%% file: main.tex
\pdfoutput=1
\documentclass[chap]{thesis}

\DeclareCaptionFont{boldmath}{\boldmath}
\usepackage{placeins}
\usepackage{flafter}
\usepackage{times}
\usepackage{latexsym}
\usepackage{graphicx}
\usepackage{adjustbox}
\usepackage{multirow}
\usepackage{multicol}
\usepackage{dirtytalk}
\IfFileExists{algorithm2e.sty}{\usepackage[ruled, vlined, linesnumbered]{algorithm2e}}{}

\usepackage{subcaption}
\usepackage{amsmath}
\usepackage{amsthm}
\usepackage{mathtools}
\usepackage{arydshln}
\usepackage{listings}
\usepackage{color, colortbl}
\usepackage{tabularx}
\usepackage{colortbl}
\usepackage{amsfonts}
\usepackage{nicefrac}       
\IfFileExists{newtxtext.sty}{\usepackage{newtxtext}}{}
\usepackage{booktabs} 
\usepackage{url}
\usepackage{hyperref}
\usepackage{xspace}
\usepackage{xcolor}         
\usepackage{wrapfig}
\usepackage{caption}
\usepackage{makecell}
\usepackage{pifont}
\usepackage{hhline}
\usepackage{array}
\usepackage{relsize}
\usepackage{amssymb}
\usepackage{graphicx}
\usepackage{calc}
\usepackage{float}
\usepackage{threeparttable}
\usepackage{cancel}

\IfFileExists{algpseudocode.sty}{\usepackage{algpseudocode}}{}
\usepackage{natbib}

\definecolor{linkdarkblue}{RGB}{0,0,139}
\hypersetup{colorlinks=true, allcolors=linkdarkblue}
\makeatletter
\newcommand\blfootnote[1]{%
  \begingroup
  \renewcommand\thefootnote{}%
  \renewcommand\@makefntext[1]{\noindent\raggedright##1}%
  \footnotetext{#1}%
  \endgroup
}
\makeatother

\usepackage[nameinlink,capitalize]{cleveref}

\Crefformat{figure}{Fig.~(#2#1#3)}
\crefformat{figure}{figure~(#2#1#3)}
\Crefformat{section}{Section~(#2#1#3)}
\crefformat{section}{section~(#2#1#3)}
\Crefformat{chapter}{Chapter~(#2#1#3)}
\crefformat{chapter}{chapter~(#2#1#3)}
\Crefformat{equation}{Eq.~(#2#1#3)}
\crefformat{equation}{equation~(#2#1#3)}
\Crefformat{table}{Table~(#2#1#3)}
\crefformat{table}{table~(#2#1#3)}
\Crefformat{appendix}{Appendix~(#2#1#3)}
\crefformat{appendix}{appendix~(#2#1#3)}

\makeatletter\@ifpackageloaded{algorithm2e}{\crefname{algocf}{alg.}{algs.}
\Crefname{algocf}{Algorithm}{Algorithms}}{}\makeatother

\definecolor{ourGreen}{RGB}{96,178,77}
\definecolor{ourRed}{RGB}{243,129,129}
\definecolor{ourBlue}{RGB}{129,190,243}
\definecolor{ourGray}{RGB}{129,129,129}
\definecolor{grey}{rgb}{0.4, 0.4, 0.4}    
\definecolor{olive}{rgb}{0.5, 0.5, 0}   

\theoremstyle{plain}

\theoremstyle{definition}

\theoremstyle{remark}

\theoremstyle{definition}

\theoremstyle{definition}

\ifdefined\algrenewcommand\algrenewcommand\algorithmicensure{\textbf{Output:}}\fi


\input{math_commands}

\makeatletter\renewcommand\@tocrmarg{2.55em plus 1fil}\makeatother  
\begin{document}

\include{front/titlepage}                 
\include{front/acknowledgment}            
\include{front/abstract}                  

\include{chapters/ch1_intro/ch1}          
\include{chapters/ch2_background/ch2}     
\include{chapters/ch3_continuous/ch3}     
\include{chapters/ch4_language/ch4}       
\include{chapters/ch5_conclusion/ch5}     

\include{references/references}           

\end{document}

%% file: math_commands.tex
\usepackage{amsmath,amsfonts,bm}

\def\eqref#1{equation~\ref{#1}}

\def\1{\bm{1}}

\newcommand{\test}{\mathcal{D_{\mathrm{test}}}}

\def\eps{{\epsilon}}

\def\rvepsilon{{\mathbf{\epsilon}}}

\def\rvf{{\mathbf{f}}}

\def\rvp{{\mathbf{p}}}

\def\rvs{{\mathbf{s}}}

\def\rvv{{\mathbf{v}}}
\def\rvw{{\mathbf{w}}}
\def\rvx{{\mathbf{x}}}
\def\rvy{{\mathbf{y}}}
\def\rvz{{\mathbf{z}}}

\DeclareMathAlphabet{\mathsfit}{\encodingdefault}{\sfdefault}{m}{sl}
\SetMathAlphabet{\mathsfit}{bold}{\encodingdefault}{\sfdefault}{bx}{n}

\newcommand{\pdata}{p_{\rm{data}}}

\newcommand{\softmax}{\mathrm{softmax}}

\DeclareMathOperator*{\argmax}{arg\,max}
\DeclareMathOperator*{\argmin}{arg\,min}

%% file: front/titlepage.tex
%
    
%
\thesistitle{
Memorization and Generalization\\in Generative Models:\\A Perspective From Energy-based\\Modeling and Associative Memory
}        
\author{Bao Pham}        
\degree{Doctor of Philosophy}        
\department{Computer Science} 
     
\signaturelines{4}     
\thadviser{Mohammed J. Zaki}
\memberone{Alex Algernon Theodore Gittens}        
\membertwo{Malik Magdon-Ismail}        
\memberthree{Dmitry Krotov}

\submitdate{[December 2026]}
\copyrightyear{2026}   

%
\titlepage     
\copyrightpage         
\tableofcontents        
{\renewcommand{\textbf}[1]{#1}\listoftables}          
{\renewcommand{\textbf}[1]{#1}\listoffigures}         

%% file: front/acknowledgment.tex
\specialhead{ACKNOWLEDGMENT}

Foremost, I would like to express my gratitude to my advisor, Dr. Mohammed J. Zaki, for his guidance, his patience, and the freedom he gave me to pursue the questions that interested me most. I also thank the members of my committee, Dr. Alex Gittens, Dr. Malik Magdon-Ismail, and Dr. Dmitry Krotov, for their time, their feedback, and the care with which they read and reviewed this thesis.

My years at RPI have shaped me as a researcher and as a person. I thank the Department of Computer Science for its support, and the students I met as a teaching assistant, who reminded me that explaining an idea clearly is the best test of understanding it. I am especially grateful to my friends at RPI, Lawrence Hiquiana, Kushal R. Bhandari, and many others, who made Troy feel like home. I would also like to thank my lab mates, Aitazaz Khan, Mohbat Tharani, and Bolun ``Namir'' Xia, for the time they spent discussing ideas with me, for their guidance, and for setting a high standard of work ethic. I wish them good fortune in their future endeavors.

I owe a special debt to IBM Research, a wonderful place where I learned how scientific research is done and made great friends who showed me the ropes during my three summers there. There I learned how to ask a good question, how to test an idea honestly, and how to keep going when an experiment does not work. I thank Dmitry Krotov for his mentorship and for the many conversations about Associative Memory that shaped this thesis, and I thank my collaborators and colleagues at IBM Research, including Benjamin Hoover and Parikshit Ram, for the discussions and the encouragement that carried this work forward. I would also like to thank the RPI-IBM Future of Computing Research Collaboration (FCRC) program for funding my research.

I am also grateful to AMD Research for the opportunity to work on 3D world modeling, which has opened a new direction of research that I hope to explore in the future. I thank my mentors, Utkarsh Saxena and Ruijing Yang, and my managers, Vikram Appia and Zhenyu Gu, for their guidance and support during my time there.

Finally, I thank my family for their support. I am very proud of my little sister, Rina, who recently started her undergraduate studies at Vanderbilt University. I am grateful for her encouragement and for her support of my curiosity and my pursuit of scientific research. I hope my journey encourages her own scientific curiosity as she begins her studies at Vanderbilt.

%% file: front/abstract.tex
 
\specialhead{ABSTRACT}

Generative models now write working code, produce realistic images, and hold long conversations, yet the mechanisms behind their \textit{intelligence} still evade our understanding. This thesis explores that puzzle through Associative Memory (AM), an idea that emerged from physics and its concept of energy, in which patterns are stored as energy minima and retrieved through attractor dynamics. Beyond their capacity, AMs form new minima known as spurious states, long regarded as a failure of memory. We argue that this failure is where creativity begins, and that memorization and generalization can coexist as outcomes of one retrieval process. We further argue that AM is broader than the energy functions of Hopfield networks, defined instead by basins of attraction.

First, we show that diffusion models behave as AMs, in which training writes data into memory and generation tries to recall it. With little data, a diffusion model forms a distinct basin of attraction around each training sample and can only reproduce what it has seen. As the data grows, these basins shrink, and spurious states emerge as attractors that blend stored samples yet match none of them. They are the first signs of generative capability and appear alongside memorized samples before the model fully generalizes. We characterize this transition by measuring the volume, energy, and curvature (or sharpness) of these basins across datasets and diffusion architectures, including Stable Diffusion.

We then turn to language, the domain of today's most prominent models, and move beyond the energy-based view of AM. We show that uniform-state discrete diffusion models are AMs capable of retrieving unseen data. Their basins of attraction arise from conditional likelihood maximization alone, which implicitly combines Hebbian storage with margin maximization. As the training set grows, basins around training examples contract while basins around unseen examples expand, until corrupted test sentences are recovered as reliably as training ones. Conditional entropy exposes this coexistence within a single sequence, separating confidently recalled tokens from uncertain ones, and cheaply diagnoses where a model lies along the transition. Larger models delay its onset.

Together, these results suggest that AM is a general principle defined by basins of attraction, and that memorization and generalization are not separate phases of generative modeling but can coexist within one model. AM thus offers a theoretical and explanatory framework for asking when a generative model recalls and when it creates.

%% file: chapters/ch1_intro/ch1.tex
\chapter{Introduction}
\label{chapter:intro}

\section{Recall or Creation?}
\label{sec:intro-motivation}

Generative models now write working code, synthesize realistic images and videos, and hold long conversations \citep{brown2020language, ho2020denoising, song2021scorebased, rombach2022high}. Their success rests on a deceptively simple principle. Given samples from an unknown data distribution, a model is trained to assign high probability to the data it has seen, whether through maximum likelihood, denoising score matching, or a conditional likelihood over parts of the data \citep{sohl-dickstein15, song2019generative, ho2020denoising, Attention2017}. Yet an objective that rewards the training data does not, by itself, specify what the model should do with data it has never seen. A model that perfectly reproduces its training set has, in a narrow sense, solved the problem it was given. That trained models nonetheless produce samples that are new, coherent, and useful remains one of the central puzzles of modern machine learning.

The puzzle becomes sharper once we look at what these models actually generate. Diffusion models trained on images can reproduce near-exact copies of their training samples \citep{somepalli2023diffusion, somepalli2023understanding, carlini2023, webster2023reproducible}, and large language models can be prompted to emit verbatim passages of their training text \citep{carlini2021extracting, carlini2023quantifying}. The same models also produce images that appear nowhere in their training data and solve tasks for which they were given few or no examples \citep{brown2020language, kojima2022large}. In practice, we want both behaviors. A useful language model should recall a fact exactly and still compose a sentence that no one has written before. This raises a question that is simple to state and difficult to answer. When a generative model produces an output, is it recalling what it has seen or creating something new?

The answer matters for several reasons. The first is practical. If generated content can be traced to memorized training data, then privacy leakage and copyright concerns may be structural consequences of how these models learn rather than occasional accidents \citep{carlini2021extracting, carlini2023, somepalli2023diffusion}. The second concerns reliability. A language model that fails to recall a fact faithfully, or that confidently produces unsupported content, generates the fluent but unfounded text often described as hallucination \citep{ji2023survey}. Distinguishing recall from creation is therefore a prerequisite for knowing when to trust a model's output. The third is scientific. Despite their empirical success, we lack an account of what these models store, where they store it, and how stored content gives rise to novel samples. A similar question sits at the heart of how the brain remembers and imagines, where recalling the past and imagining the future are thought to draw on shared memory systems \citep{schacter2007cognitive}. Explaining how memorization and generalization coexist in artificial generative models may thus contribute to a broader understanding of memory as a computational principle.

Classical learning theory has long treated memorization as the opposite of generalization, the symptom of a model that fits noise rather than structure. Modern deep networks have unsettled this view. They can fit randomly labeled data perfectly and still generalize on real data \citep{zhang2017understanding}, and memorizing rare examples can even be necessary for good performance on long-tailed distributions \citep{feldman2020does}. Recent analytical work has also shown that the memorization of facts and the generalization of rules can be achieved simultaneously by the same network \citep{farne2026rules}. For generative models, however, most studies of the memorization-to-generalization transition remain generalization-centric \citep{meehan2020non, burg2021on, kadkhodaie2023generalization, ventura2024manifolds, achilli2024losing, kamb2024analytic, ross2024geometric, biroli2024dynamical, cui2025precise, achilli2025memorization}. Memorization is typically viewed as a side effect that fades as the training set grows \citep{yoon2023diffusion, gu2023memorization}, and the intermediate regime between the two phases, where both behaviors are present, has received comparatively little attention. These studies also predominantly concern continuous data such as images, leaving the discrete setting of language comparatively less explored.

This thesis takes a different starting point. Rather than treating memorization as a flaw to be removed, we treat it as the foundation from which generalization emerges. If the ability of a model to create is built on its ability to remember, then understanding generalization requires understanding memory first. We therefore study generative models through a framework that was built to describe memory in the first place, Associative Memory.

\section{Associative Memory as a Lens on Generative Models}
\label{sec:intro-am-lens}

Associative Memory (AM) is an idea that emerged from physics and its concept of energy. In the Hopfield network \citep{Hopfield1982, AmariS.I.1972, Hopfield1984}, a set of patterns is stored as local minima of an energy function, and a corrupted or partial cue is restored by letting the state of the network descend this energy until it settles into a nearby minimum. Each stored pattern is therefore surrounded by a basin of attraction, the set of states that flow toward it, and retrieval succeeds whenever the cue lies within the correct basin. The number of patterns that a classical Hopfield network can reliably store grows only linearly with the number of neurons \citep{Amit1985StoringIN, amit1987statistical, AbuMostafa1985InformationCO}. Dense Associative Memories (DenseAMs) remove this limitation through sharper, higher-order interactions, achieving storage capacities that grow super-linearly and even exponentially with the dimension of the data \citep{DAM2016, krotov2018dense, demircigil2017model}. These modern AMs have renewed interest in the field, in part because their retrieval dynamics are closely related to the attention mechanism of Transformers \citep{Ramsauer2021, et} and to models of memory in neurobiology \citep{krotov2021large, kozachkov2025neuron}. We refer the reader to \cite{krotov2025modern} for a recent review.

A central question of AM theory is what happens when the network is asked to store more than its capacity allows. Beyond capacity, the energy landscape develops minima that do not correspond to any stored pattern. These spurious states were historically regarded as defects that corrupt retrieval \citep{Hopfield1982, Hopfield1983UnlearningHA, Amit1985StoringIN}, and loading the network further leads to a catastrophic blackout in which no meaningful attractor survives \citep{amit1987statistical}. Recent work has revealed a different outcome. When the stored patterns share an underlying structure, overloading an AM can instead create new attractors close to examples it has never seen, while the memories of its training data persist \citep{kalaj2024random}. From this perspective, the formation of new attractors beyond capacity is not only a failure of memory. It is also where generalization begins.

This observation suggests a reinterpretation of generative modeling itself. Training a generative model can be viewed as writing the training data into memory, and generation as an attempt to recall it. When recall succeeds, the model retrieves a training sample, which is memorization. When recall fails, the model settles into an attractor that is not a training sample. Such attractors include spurious states and, once enough data is available, generalized samples. For diffusion models, this correspondence can be made precise in an idealized limit. The energy implied by the optimal score of a diffusion model trained on a finite dataset takes the same log-sum-exp form as the energy of a DenseAM, with the noise level playing the role of temperature \citep{hoover2023memory, ambrogioni2023search, raya2024spontaneous}. The theory developed for AMs, including the theory of spurious states, can therefore be brought to bear on how diffusion models memorize and generalize.

\section{Associative Memory Beyond Energy}
\label{sec:intro-beyond-energy}

The energy function has long been considered the defining ingredient of an AM. It guarantees that retrieval converges, and it makes stored memories directly visible as the minima of the energy landscape. Requiring an energy function, however, also restricts the class of admissible models, since it typically demands symmetric interactions or carefully constrained architectures \citep{krotov2021hierarchical, et, kozachkov2025neuron}. Most modern networks, including the Transformers \citep{Attention2017} that underlie today's language models, satisfy no such constraint, and there is no reason to expect the computation of a generic feed-forward network to follow the gradient of an energy function.

What an AM needs in order to retrieve its memories is not the energy function itself, but the basins of attraction that the energy function produces. Basins can arise in other ways. In the classical theory of neural networks, the robustness of a stored pattern is tied to the margin with which each neuron reproduces its own state given the states of all the others \citep{gardner1988space, Gardner1988OptimalSP}. Maximizing the conditional likelihood of each variable given the rest, as in the pseudo-likelihood objective \citep{pseudo-likelihood}, has been shown to create such basins directly, even when the learned couplings are asymmetric and no energy function exists \citep{d2025pseudo}. Conditional likelihood is precisely the structure on which modern language models are trained. This thesis therefore adopts a broader view, in which a system acts as an AM whenever it forms basins of attraction around data, whether or not those basins are the minima of an energy.

This view has an important consequence for the coexistence of memorization and generalization. In an AM, a basin around a training example and a basin around an unseen example are objects of the same kind. A model that holds basins around both is at once a faithful memory and a system capable of recovering data it was never shown. The question of how memorization and generalization coexist then becomes a question about how basins of attraction form, how they grow or shrink with the amount of data and the size of the model, and how they can be measured.

\section{The Nature of This Work}
\label{sec:intro-nature}

This thesis is primarily explanatory. Its main goal is not to propose a new generative model or to improve performance on a benchmark, but to explain behaviors that widely used generative models already exhibit, and to provide tools for observing those behaviors inside a trained model. The AM perspective, however, is also constructive. As the Energy Transformer \citep{et} illustrates, energy-based modeling and AM can guide the design of new architectures that use fewer parameters than their feed-forward counterparts, remain competitive in performance, and whose computation can be read through their energy. Three principles guide the explanatory work.

First, theory serves as a lens rather than as a description of every detail. Concepts from AM theory, such as capacity, basins of attraction, spurious states, and margins, have precise meanings in idealized settings. In the continuous case, the idealized setting is a diffusion model whose score exactly matches the empirical distribution of its training data, a limit in which the connection to DenseAMs is exact. In the discrete case, it is a network of binary variables trained with pseudo-likelihood, where the link between conditional likelihood, Hebbian learning, and margins can be derived explicitly. Practical models are neither infinitely expressive nor binary. The central empirical question is therefore whether predictions derived in these idealized limits survive in finite networks trained with standard methods.

Second, memorization and generalization are also measured from the inside. The usual way to detect memorization is to compare generated samples against the training set, which becomes costly as datasets grow and says little about why a sample was produced. We still rely on such comparisons where they are needed, for instance to label generated images as memorized, spurious, or generalized, and token recovery requires reference sequences. Beyond these labels, the quantities used in this thesis are properties of the model's own dynamics. They include the size of the basin of attraction around a generated sample, the curvature of the energy landscape around it, the reliability with which the model restores a sequence after part of it has been corrupted, and the uncertainty of the model about each token it produces. Because each of these quantities has a geometric meaning in AM theory, they help explain as well as detect. They indicate not only whether a sample is memorized, but also what kind of attractor it occupies.

Third, the amount of training data is treated as a control parameter. In AM theory, the behavior of a network is governed by its load, the number of stored patterns relative to the size of the network. By training families of models on datasets of systematically increasing size, and at several model sizes, we trace the transition from memorization to generalization across its full range, rather than observing a single model at a single point.

Finally, the scope of these claims should be stated clearly. The models studied here are small compared with frontier systems, and the results establish correspondences and measurable signatures rather than causal mechanisms. Spurious states, for instance, mark the onset of generalization, but we do not prove that they cause it. The aim is to provide a coherent and testable account of how memorization and generalization can coexist at a larger scale.

\section{Research Questions}
\label{sec:intro-questions}

The perspective above leads to a central question. \textit{How can memorization and generalization coexist within the same generative model?} AM theory suggests that the answer lies in how basins of attraction form, how they change with data, and how they appear from inside the model. We address this question through three parts.

\par\addvspace{1em}
\noindent \textbf{Is generation a form of memory retrieval?}
If training writes data into memory and generation attempts to recall it, a diffusion model should behave as an AM. Below its capacity, it should hold one attractor per training sample. Beyond its capacity, it should form new attractors that belong to no single training sample, and AM theory predicts that spurious states mark the boundary between the two regimes. We ask whether finite diffusion models trained with standard methods follow this picture, and whether spurious states signal the onset of generalization.

\par\addvspace{1em}
\noindent \textbf{What makes a system an Associative Memory?}
AMs are classically defined by an energy function whose minima are the stored memories. What allows an AM to retrieve its memories, however, is the existence of basins of attraction around them, and most generative models, including language models, are neither trained with an energy nor constrained to have one. We ask whether basins of attraction, rather than an energy function, can serve as the defining property of an AM, whether conditional likelihood maximization, the objective behind these models, is sufficient to create such basins, and whether language diffusion models then display the signatures of an AM, including a memorization-to-generalization transition in which basins around training data and basins around unseen data coexist.
\par\addvspace{1em}
\noindent \textbf{Can AM theory make memorization and generalization measurable?}
Detecting memorization usually relies on comparing generated samples with the training set. Such comparisons indicate whether a sample is a copy, but they become costly at scale and say little about why the sample was produced. We ask whether quantities grounded in AM theory, such as basin volume, energy curvature, token recovery, and conditional entropy, can complement these comparisons by revealing where a model lies along the transition, how this position depends on the size of the training set and of the model, and which parts of a single output are recalled and which are created.

\section{Contributions}
\label{sec:intro-contributions}

This thesis makes two main contributions, each corresponding to one of the core chapters.

\par\addvspace{1em}
\noindent \textbf{Diffusion Models as Associative Memories.}
In Chapter~3, we interpret continuous diffusion models through the lens of DenseAMs and characterize their memorization-to-generalization transition as a function of the training set size. This chapter addresses the first research question and contributes to the third. In the small data regime, diffusion models form a distinct attractor for each training sample, akin to a DenseAM below its critical capacity. As the training set grows, the models transition to generalization. At the onset of this transition, we identify an intermediate phase, predicted by DenseAM theory, in which spurious states emerge. These samples possess sizable basins of attraction yet do not correspond to any training sample, and we argue that they are the first signs of generative capability rather than negative artifacts. We characterize memorized, spurious, and generalized samples through their abundance, the volume of their basins of attraction, and the curvature of the energy landscape around them, across a two-dimensional (toy) model, several natural image datasets, and multiple architectures. A small-scale case study of Stable Diffusion \citep{rombach2022high} further identifies candidate memorized, spurious, and generalized samples with a similar ordering of energy curvature, suggesting that these phenomena may persist in large-scale practical models. The sample types are identified by a nearest-neighbor search over the training set, but the energy curvature is computed from the trained model alone, which points toward diagnostic tools that could monitor the onset of memorization during training without such a search.

\par\addvspace{1em}
\noindent \textbf{Coexistence of Memorization and Generalization in Language.}
In Chapter~4, we extend this perspective to language, and building on \cite{d2025pseudo}, provide evidence that an explicit energy function is not required for AM behavior. This chapter addresses the second research question and contributes to the third. We study uniform-state discrete diffusion models \citep{austin2021structured, campbell2024generative, gat2024discrete, sahoo2024simple} and show, in a simplified setting of binary variables, that maximizing conditional likelihood implicitly performs Hebbian learning \citep{Hebb1949} on the training data while maximizing the classification margin of each variable, which produces basins of attraction around the data. Empirically, these models undergo a sharp memorization-to-generalization transition governed by the size of the training set. As the training set grows, basins around training sequences shrink while basins around unseen test sequences expand, until the model recovers corrupted test sequences as reliably as training ones. We introduce token recovery rates and conditional entropy as complementary probes of this transition. Conditional entropy vanishes in the memorization regime and remains finite in the generalization regime; separates recovered tokens, which have near-zero entropy, from unrecovered ones within the same sequence; and agrees with perplexity measured by an external critic model. Larger models delay the onset of the transition and narrow the entropy gap between training and generated sequences.

\section{Thesis Organization}
\label{sec:intro-organization}

The remainder of this thesis is organized as follows. Chapter~2 introduces the background on which the later chapters build, including energy-based models, continuous and discrete diffusion models, and classical and dense AMs, with an emphasis on their memory capacity and spurious states. Chapter~3 presents our study of continuous diffusion models as DenseAMs and the emergence of spurious states at the memorization-to-generalization transition. Chapter~4 extends the AM perspective to language diffusion models, where basins of attraction arise from conditional likelihood maximization without an explicit energy function. Chapter~5 summarizes our findings and discusses open questions and future directions. Chapters~3 and~4 each end with a section of supplementary material containing experimental details and additional results.

%% file: chapters/ch2_background/ch2.tex
\chapter{Background}
\label{chapter:introduction}
\section{Probabilistic Generative Modeling}
\label{sec:prob-modeling}

A generative model is an approximation of the process that produced the data, one from which new samples can be drawn, or generated. We assume that each observation $\rvx$, such as an image or a sentence, is a sample from an unknown data distribution $\pdata(\rvx)$, and we seek a parameterized distribution $p_\theta(\rvx)$ that resembles it,
\begin{equation}
    p_\theta(\rvx) \approx \pdata(\rvx).
\end{equation}
Once such a model is available, it can be used to evaluate how plausible a new observation is, to fill in missing or corrupted parts of an observation, and, most visibly, to synthesize new samples \citep{kingmaIntroVAE2019}. The difficulty is that $\pdata$ is never available directly. What we have is a finite dataset $\mathcal{D} = \{\rvx^{(1)}, \ldots, \rvx^{(K)}\}$ of $K$ independent samples, which defines the empirical distribution
\begin{equation}
    \hat{p}_\mathcal{D}(\rvx) = \frac{1}{K} \sum_{k=1}^{K} \delta\big(\rvx - \rvx^{(k)}\big),
    \label{eqn:empirical-distribution}
\end{equation}
where $\delta$ is the Dirac delta. Every learning objective in this chapter is computed from $\hat{p}_\mathcal{D}$, while the goal is to approximate $\pdata$. The gap between these two distributions is where the question of memorization and generalization arises.

\subsection{Maximum Likelihood Estimation}
\label{sec:mle}
The most common way to fit $p_\theta$ is maximum likelihood estimation (MLE). The likelihood of the parameters is the probability that the model assigns to the observed dataset,
\begin{equation}
    \mathcal{L}(\theta; \mathcal{D}) = \prod_{\rvx \in \mathcal{D}} p_\theta(\rvx),
\label{eqn:likelihood}
\end{equation}
and since products of many small probabilities are numerically unstable, one works with its logarithm,
\begin{equation}
    \log \mathcal{L}(\theta; \mathcal{D}) = \sum_{\rvx \in \mathcal{D}} \log p_\theta(\rvx).
\label{eqn:log-likelihood}
\end{equation}
The maximum likelihood estimate is the parameter that makes the observed data most probable,
\begin{equation}
    \theta^* = \arg\max_\theta \sum_{\rvx \in \mathcal{D}}\log p_\theta(\rvx).
\label{eqn:maximum-likelihood-est}
\end{equation}
Dividing \Cref{eqn:log-likelihood} by $K$ shows that MLE maximizes the average log-probability under the empirical distribution, which is the same as minimizing the Kullback-Leibler (KL) divergence \citep{kullback1951information} from $\hat{p}_\mathcal{D}$ to the model,
\begin{equation}
    \frac{1}{K} \log \mathcal{L}(\theta; \mathcal{D}) = \mathbb{E}_{\hat{p}_\mathcal{D}(\rvx)} \big[ \log p_\theta(\rvx) \big] = - \mathrm{D}_\text{KL}\big(\hat{p}_\mathcal{D}(\rvx) \,\|\, p_\theta(\rvx)\big) + \text{constant}.
    \label{eqn:mle-kl-empirical}
\end{equation}
As the number of samples grows, $\hat{p}_\mathcal{D}$ approaches $\pdata$, and MLE approaches the minimization of $\mathrm{D}_\text{KL}(\pdata \,\|\, p_\theta)$ \citep{arjovsky17a}. For any finite dataset, however, the objective is indifferent to everything outside the training set. If the model family were unrestricted, the maximizer of \Cref{eqn:mle-kl-empirical} would be $\hat{p}_\mathcal{D}$ itself, a model that places all of its mass on the training samples and reproduces nothing else. MLE, taken at face value, therefore rewards memorization. A model trained by MLE generalizes only because its parameterization, its capacity, and the way it is optimized prevent it from reaching this solution. The same observation reappears in diffusion models. A diffusion model that learns the score of its finite training set exactly, rather than the score of the underlying data distribution, can only reproduce its training samples (\Cref{sec:diffusion-ebm}).

Beyond this conceptual point, MLE poses a practical challenge. Evaluating $\log p_\theta(\rvx)$ requires $p_\theta$ to be a properly normalized distribution, and normalizing a flexible model over a high-dimensional space is generally intractable. Much of the design of deep generative models can be understood as a response to this constraint. Autoregressive models factorize the joint distribution into a product of one-dimensional conditionals, each of which is easy to normalize \citep{graves2013generating, germain15, van2016pixel, yu2022scaling, tian2024visual}. Normalizing flows build $p_\theta$ from invertible transformations of a simple base distribution, so that the change-of-variables formula gives the density exactly \citep{dinh2014nice, dinh2016density, pmlr-v37-rezende15, kingma2018glow, weng2018flow}. Variational autoencoders introduce latent variables and optimize a tractable lower bound on the likelihood \citep{kingmaVAE, rezende14}. Each of these solutions buys tractability by fixing the form of the model, whether as a sequential ordering of variables, an invertible architecture, or a particular latent-variable structure, and each constraint may limit how well the model can capture complex data \citep{song2021trainebm}.

\section{Energy-Based Models}
\label{sec:energy-based-models}
Energy-based models (EBMs) take the opposite approach. Rather than constraining the model so that it can be normalized, they give up on normalization during modeling and specify only an energy function $E_\theta(\rvx)$ that assigns low values to plausible observations and high values to implausible ones. Because the energy is not required to integrate to one, it can be represented by essentially any neural network, provided only that the corresponding distribution could in principle be normalized. This freedom has made EBMs useful for modeling images \citep{ngiam2011, du2019implicit, du2020improved}, text \citep{deng2020residual}, and graphs \citep{liu2021graphebm}. EBMs are also the natural bridge to the rest of this thesis, since the AMs discussed in \Cref{sec:AM-general} are themselves defined by energy functions.

\begin{figure}[t]
    \centering
    \includegraphics[keepaspectratio, width=1\textwidth, height=1\textheight]{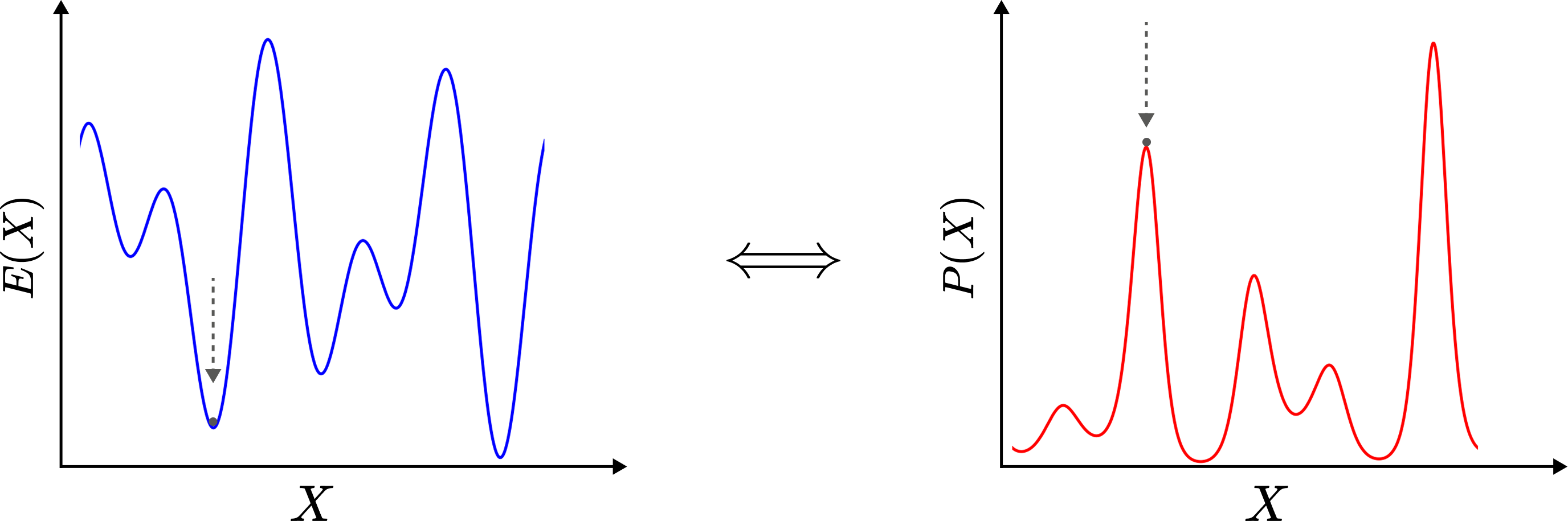}
    \caption{The relationship between {energy} and {probability}. The energy assigned to each state (left) and the probability of the same states under \Cref{eqn:boltzmann-dist-model} (right) are shown. States with low energy have high probability, as indicated by the {dashed gray arrow}.}
    \label{fig:energy-prob-relation}
\end{figure}

\subsection{Energy and Probability}
An energy function defines a probability distribution through the Boltzmann distribution,
\begin{equation}
    p_\theta(\rvx) = \frac{\exp(-E_\theta (\rvx))}{Z_\theta},
    \label{eqn:boltzmann-dist-model}
\end{equation}
where the partition function
\begin{equation}
    Z_\theta = \int \exp (-E_\theta(\rvx))\, \mathrm{d} \rvx
    \label{eqn:partition-function}
\end{equation}
ensures that the distribution integrates to one. The partition function (\ref{eqn:partition-function}) is the difficult term to compute, since it integrates over all states and changes whenever the parameters $\theta$ are updated. The exponential in \Cref{eqn:boltzmann-dist-model} translates the geometry of the energy into probability, as illustrated in \Cref{fig:energy-prob-relation}. The minima of the energy are the most probable states, and the deeper a minimum, the more probability it carries.

The price of this flexibility is paid during training. For all but the simplest energies, $Z_\theta$ is an intractable integral over the entire input space, and so is the log-likelihood $\log p_\theta(\rvx) = -E_\theta(\rvx) - \log Z_\theta$ \citep{song2021trainebm}. Two families of methods train EBMs without ever computing $Z_\theta$. The first estimates the gradient of the log-likelihood using samples from the model, which leads to contrastive divergence. The second replaces the likelihood with an objective on the gradient of the log-density with respect to the data, which leads to score matching. We review both, because each prefigures an idea used later in this thesis.

\subsection{Contrastive Divergence}
\label{sec:contrastive-div}
\begin{figure}[t]
    \centering
    \includegraphics[keepaspectratio, width=1\textwidth, height=1\textheight]{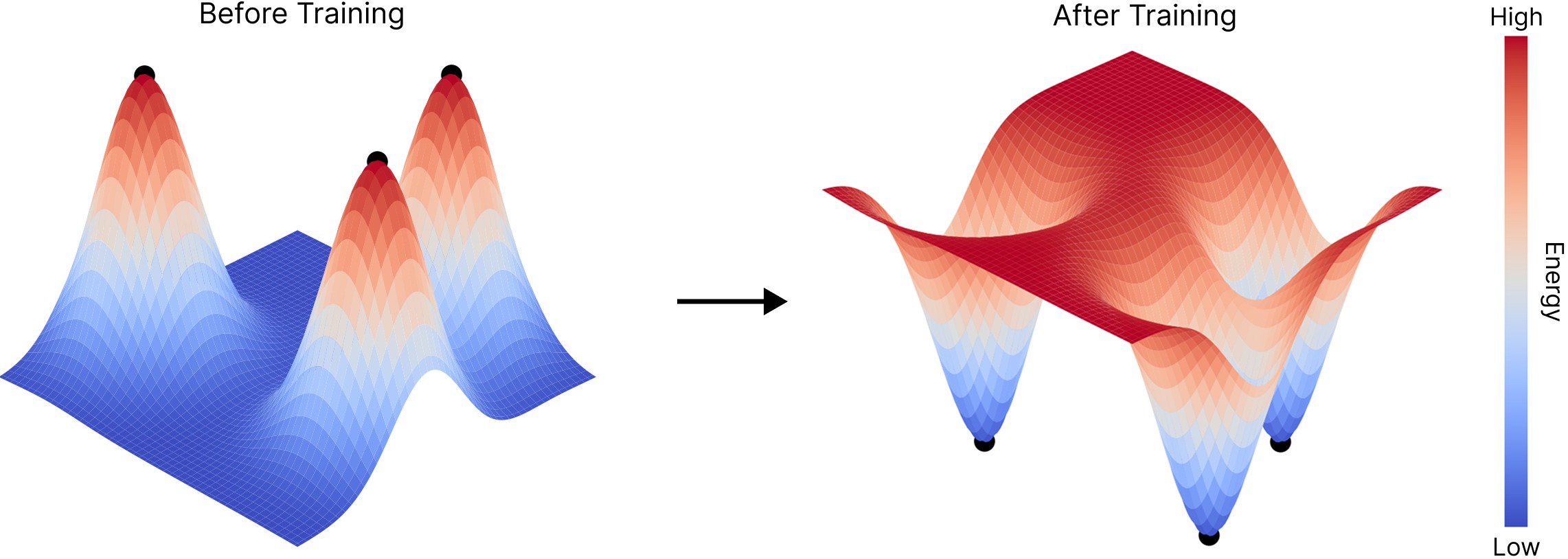}
    \caption{The effect of training an EBM on the energy landscape around the data points (shown in black). Before training, the data may lie anywhere on the landscape, including at its {highest} points. After training, the data sit at the {lowest} points, as local minima of the learned energy.}
    \label{fig:energy-manifold-up-down}
\end{figure}

In the population limit, MLE maximizes the expected log-likelihood of the data,
\begin{equation}
    \mathcal{L}(\theta) = \mathbb{E}_{\rvx \sim \pdata(\rvx)} \big [ \log p_\theta (\rvx) \big ],
    \label{eqn:expected-mle}
\end{equation}
or, equivalently, minimizes the negative log-likelihood
\begin{equation}
    -\mathcal{L}(\theta) = - \mathbb{E}_{\rvx \sim \pdata(\rvx)} \big [ \log p_\theta (\rvx) \big ],
    \label{eqn:expected-negative-mle}
\end{equation}
which differs from the KL divergence between the data and the model only by the entropy of the data, a quantity that does not depend on $\theta$,
\begin{equation}
    -\mathbb{E}_{\rvx \sim \pdata(\rvx)} \big [ \log p_\theta (\rvx) \big ] = \mathrm{D}_{KL} \big ( \pdata(\rvx) \, || \, p_\theta (\rvx) \big ) - \mathbb{E}_{\rvx \sim \pdata(\rvx)} \big [ \log \pdata (\rvx) \big ].
    \label{eqn:mle-dkl-data-model}
\end{equation}
Although the likelihood of an EBM cannot be evaluated, its gradient can be estimated, which is enough for gradient-based optimization \citep{GelfandA1990, WellingLangevin}. Differentiating \Cref{eqn:boltzmann-dist-model} gives
\begin{equation}
    \nabla_\theta \log p_\theta (\rvx) = - \nabla_\theta E_\theta(\rvx) - \nabla_\theta \log Z_\theta.
    \label{eqn:boltzmann-dist-grad}
\end{equation}
The first term is computed by automatic differentiation of the energy. The second term involves the partition function, but it has a simple form. Moving the gradient inside the integral in \Cref{eqn:partition-function} shows that it is an expectation under the model itself,
\begin{equation}
\begin{split}
    \nabla_\theta \log Z_\theta & = \frac{1}{Z_\theta}\, \nabla_\theta \int \exp \big ( -E_\theta(\rvx) \big )\, \mathrm{d} \rvx \\
    & = \int \frac{\exp \big ( -E_\theta (\rvx) \big )}{Z_\theta} \big ( - \nabla_\theta E_\theta (\rvx) \big )\, \mathrm{d} \rvx \\
    & = \mathbb{E}_{\rvx \sim p_\theta (\rvx)} \big [ - \nabla_\theta E_\theta (\rvx) \big ].
    \label{eqn:MCMC-expectation}
\end{split}
\end{equation}
This expectation can be approximated with samples $\rvx'$ drawn from the model by Markov chain Monte Carlo (MCMC). A convenient sampler for continuous data is Langevin dynamics \citep{parisiLangevin, WellingLangevin}, which iterates
\begin{equation}
    \rvx_{\tau+1} = \rvx_\tau + \frac{\eta}{2}\, \nabla_\rvx \log p_\theta(\rvx_\tau) + \sqrt{\eta}\, \rvepsilon_\tau, \qquad \rvepsilon_\tau \sim \mathcal{N}(0, \boldsymbol{I}), \qquad \tau = 0, \ldots, S - 1,
    \label{eqn:langevin}
\end{equation}
with step size $\eta$. Starting from an initial state $\rvx_0$, the sampler is run for a fixed number of steps $S$, and the final state $\rvx_S$ is taken as an approximate sample from $p_\theta$. As $\eta \to 0$ and $S \to \infty$, the distribution of $\rvx_S$ converges to $p_\theta$ under mild regularity conditions \citep{WellingLangevin}. The sampler requires only the score of the model, the gradient of the log-density with respect to the data, which for an EBM does not involve the partition function,
\begin{equation}
\begin{split}
    \nabla_\rvx \log p_\theta(\rvx) &= - \nabla_\rvx E_\theta (\rvx) - 
    \underset{ = ~ 0}{\underbrace{\nabla_\rvx \log Z_\theta}}  \\ 
    & = - \nabla_\rvx E_\theta (\rvx).
\end{split}
\end{equation}
Langevin sampling is thus a noisy descent on the energy, a picture that will reappear in the reverse process of diffusion models. Substituting a model sample $\rvx' \sim q_\theta(\rvx)$, where $q_\theta$ denotes the distribution produced by a finite run of the sampler, into \Cref{eqn:boltzmann-dist-grad} gives the stochastic gradient
\begin{equation}
     \nabla_\theta \log p_\theta (\rvx) \approx - \nabla_\theta E_\theta(\rvx) + \nabla_\theta E_\theta(\rvx'),
    \label{eqn:contrastive-loss-ebm}
\end{equation}
which is the gradient of the contrastive objective
\begin{equation}
    -\mathcal{L}_{CD}(\theta) = E_\theta(\rvx) - E_\theta(\rvx'),
    \label{eqn:contrastive-loss-ebm-simplify}
\end{equation}
when $\rvx'$ is treated as fixed. Minimizing it lowers the energy of the data and raises the energy of the states the model currently prefers. When the sampler is run for only a few steps starting from the data, the procedure is known as contrastive divergence \citep{hinton2002training}. The same two-sided principle appears much earlier in the AM literature. The Boltzmann machine is trained by contrasting statistics of the data with statistics of the model \citep{AckleyHintonSej1985}, and the unlearning procedure of \cite{Hopfield1983UnlearningHA} removes spurious memories of a Hopfield network by raising the energy of the states into which the network falls on its own. As \cite{LeCun2006ATO} emphasize, the effect of such objectives is to carve the energy landscape so that the data occupy its valleys while other states are pushed onto its slopes, as illustrated in \Cref{fig:energy-manifold-up-down}.

\subsection{Score Matching}
Contrastive divergence depends on samples from the model, and MCMC samplers can mix slowly in high dimensions, producing samples that miss parts of the distribution \citep{song2019generative, du2020improved}. Score matching avoids sampling altogether \citep{hyvarinen05a}. Instead of comparing probabilities, it compares scores, and fits the model by minimizing the Fisher divergence between the data and the model,
\begin{equation}
    \mathrm{D}_{F} \big ( \pdata(\rvx) ~ || ~ p_\theta (\rvx) \big ) = \mathbb{E}_{ \pdata(\rvx)} \bigg [ \frac{1}{2} \lVert \nabla_\rvx \log \pdata (\rvx) - \nabla_\rvx \log p_\theta(\rvx) \rVert^2_2 \bigg ].
    \label{eqn:score-matching-general}
\end{equation}
Because the scores do not involve normalization constants, $Z_\theta$ never appears. The score of the data is unknown, but \cite{hyvarinen05a} showed by integration by parts that, under mild regularity conditions, \Cref{eqn:score-matching-general} equals, up to a constant, an objective that depends only on the energy and its derivatives,
\begin{equation}
    \mathcal{L}_\text{SM} (\theta) = \mathbb{E}_{ \pdata(\rvx)} \bigg [ \sum^N_{i = 1} \bigg ( \frac{1}{2} \bigg ( \frac{\partial E_\theta(\rvx)}{\partial x_i } \bigg)^2 - \frac{\partial^2 E_\theta(\rvx)}{\partial x_i^2} \bigg ) \bigg ] + \text{constant},
    \label{eqn:score-matching-song-v1}
\end{equation}
where $N$ is the dimension of $\rvx$. Minimizing $\mathcal{L}_\text{SM}$ recovers the true parameters in the limit of infinite data \citep{hyvarinen05a}. The second derivatives are its weakness. The trace of the Hessian of the energy requires $N$ backward passes, which is prohibitive for images and other high-dimensional data. Stochastic trace estimators reduce this cost \citep{Hutchinson1990, song2019sliced}, but a simpler and more widely used solution is to add noise to the data.

\subsection{Denoising Score Matching}
Score matching also assumes that $\log \pdata$ is differentiable over the whole input space. Real data often violate this assumption. Images, for example, are quantized to integer pixel values and occupy a thin subset of their ambient space, so their score may be undefined \citep{song2021scorebased}. Both problems are addressed by perturbing the data with Gaussian noise,
\begin{equation}
    \tilde{\rvx} = \rvx + \sigma \rvepsilon, \qquad \rvepsilon \sim \mathcal{N}(0, \boldsymbol{I}),
    \label{eqn:simple-perturbation}
\end{equation}
which replaces the data distribution with the smoothed distribution
\begin{equation}
    q_\sigma(\tilde{\rvx}) = \int q_\sigma(\tilde{\rvx} \,|\, \rvx)\, \pdata(\rvx)\, \mathrm{d} \rvx,
\end{equation}
whose score is defined everywhere. Matching the score of this smoothed distribution turns out to be equivalent, up to a constant, to matching the score of the known noise kernel \citep{VincentP2011},
\begin{equation}
\begin{split}
    D_F \big ( q_\sigma(\tilde{\rvx}) ~ || ~ p_\theta (\tilde{\rvx}) \big ) &= 
    \mathbb{E}_{q_\sigma(\tilde{\rvx})} \bigg [ \frac{1}{2} \lVert \nabla_{\tilde{\rvx}} \log q_\sigma (\tilde{\rvx}) - \nabla_{\tilde{\rvx}} \log p_\theta (\tilde{\rvx}) \rVert^2_2 \bigg]\\
    &= \mathbb{E}_{q_\sigma(\rvx, \tilde{\rvx})} \bigg [ \frac{1}{2} \lVert \nabla_{\tilde{\rvx}} \log q_\sigma (\tilde{\rvx} \,|\,  \rvx) - \nabla_{\tilde{\rvx}} \log p_\theta (\tilde{\rvx}) \rVert^2_2 \bigg] + \text{constant}.
\end{split}
\label{eqn:denoising-score-matching}
\end{equation}
The second line contains neither the unknown data score nor second derivatives. For the Gaussian kernel, $\nabla_{\tilde{\rvx}} \log q_\sigma(\tilde{\rvx} \,|\, \rvx) = -(\tilde{\rvx} - \rvx)/\sigma^2 = -\rvepsilon/\sigma$, and the objective becomes the denoising score matching loss \citep{VincentP2011, song2019generative},
\begin{equation}
     \mathcal{L}_\text{DSM}(\theta) = \mathbb{E}_{\pdata(\rvx),\, \rvepsilon \sim \mathcal{N}(0, \boldsymbol{I})} \bigg [ \frac{1}{2} \Big\lVert \nabla_\rvx \log p_\theta (\rvx + \sigma \rvepsilon) + \frac{\rvepsilon}{\sigma} \Big\rVert^2_2 \bigg].
     \label{eqn:denoising-score-matching-final}
\end{equation}
The model learns to point from a noisy observation back toward the clean data, which is why the objective is called denoising. The price is that the model learns the score of $q_\sigma$ rather than that of $\pdata$. A small $\sigma$ keeps the two close, but makes the score hard to learn far from the data, where few noisy samples fall. Diffusion models resolve this trade-off by learning the score at many noise levels at once. The objective in \Cref{eqn:denoising-score-matching-final} is already the training objective of a variance-exploding diffusion model restricted to a single noise level \citep{sohl-dickstein15, ho2020denoising, song2021scorebased}, and we develop this connection in \Cref{sec:diffusion-models}.

\section{Diffusion Models}
\label{sec:diffusion-models}

The denoising score matching objective (\ref{eqn:denoising-score-matching-final}) learns the score of the data at a single, small noise level. Diffusion models extend this idea in two ways. They learn the score across a whole range of noise levels, from nearly clean data to pure noise, and they use these scores to transform noise back into data \citep{sohl-dickstein15, ho2020denoising, song2019generative, song2021scorebased}. This simple recipe now underlies state-of-the-art generators of images \citep{rombach2022high}, audio \citep{chen2020wavegrad, kong2020diffwave}, and video \citep{ho2022video, blattmann2023videoldm}. A diffusion model consists of two processes. The forward process gradually corrupts data with noise until no trace of the original sample remains. The reverse process, learned by a neural network, gradually removes the noise to produce a sample. \Cref{fig:diffusion-visual} illustrates the forward process.

\subsection{The Forward Process}
\label{sec:diffusion-forward}
\begin{figure}[t]
    \centering
    \includegraphics[width=\linewidth]{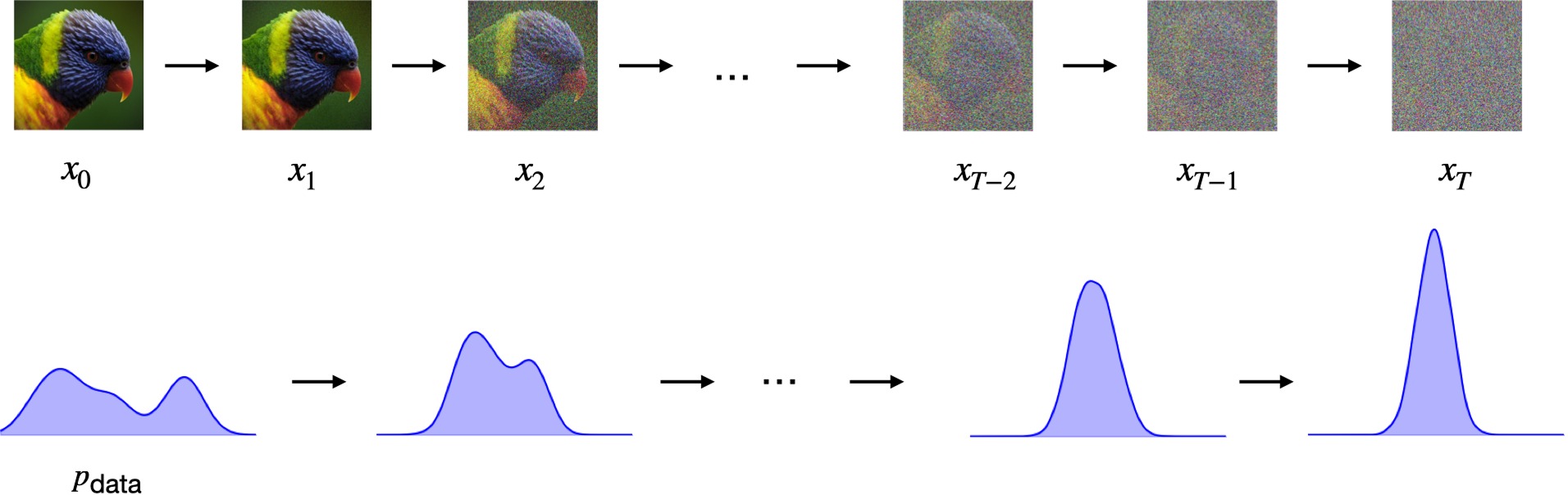}
    \caption{The forward process of a diffusion model. A clean image $\rvx_0$ is progressively corrupted by Gaussian noise over $T$ steps, until $\rvx_T$ is indistinguishable from pure noise (top row). The corresponding marginal distribution (bottom row) evolves from the complex and multimodal data distribution $\pdata$ toward a simple Gaussian. Generation reverses this process. It starts from a sample of the Gaussian and removes the noise step by step until a sample resembling the data is obtained. Reproduced from: Krotov, D., Hoover, B., Ram, P., \& Pham, B. (2025). \textit{Modern methods in associative memory}. arXiv. \url{https://doi.org/10.48550/arXiv.2507.06211}}
    \label{fig:diffusion-visual}
\end{figure}

In the discrete-time formulation of \cite{ho2020denoising}, the forward process is a Markov chain that adds a small amount of Gaussian noise at each step,
\begin{equation}
    q(\rvx_t \,|\, \rvx_{t-1}) = \mathcal{N}\big(\rvx_t;\, \sqrt{1 - \beta_t}\, \rvx_{t-1},\, \beta_t \boldsymbol{I}\big), \qquad t = 1, \ldots, T,
    \label{eqn:ddpm-forward-step}
\end{equation}
where $\rvx_0 \sim \pdata(\rvx)$ and $\beta_t \in (0, 1)$ is a prescribed variance schedule. Because a composition of Gaussian transitions is again Gaussian, the noisy sample at any step can be obtained from the clean sample in closed form,
\begin{equation}
    q(\rvx_t \,|\, \rvx_0) = \mathcal{N}\big(\rvx_t;\, \sqrt{\bar{\alpha}_t}\, \rvx_0,\, (1 - \bar{\alpha}_t) \boldsymbol{I}\big), \qquad \bar{\alpha}_t = \prod_{s=1}^{t} (1 - \beta_s),
    \label{eqn:ddpm-forward-marginal}
\end{equation}
or equivalently $\rvx_t = \sqrt{\bar{\alpha}_t}\, \rvx_0 + \sqrt{1 - \bar{\alpha}_t}\, \rvepsilon$ with $\rvepsilon \sim \mathcal{N}(0, \boldsymbol{I})$. The schedule is chosen such that $\bar{\alpha}_T \approx 0$, so that $\rvx_T$ is distributed approximately as $\mathcal{N}(0, \boldsymbol{I})$ regardless of the initial sample. The marginal distribution at step $t$,
\begin{equation}
    q_t(\rvx_t) = \int q(\rvx_t \,|\, \rvx_0)\, \pdata(\rvx_0)\, \mathrm{d}\rvx_0,
    \label{eqn:diffusion-marginal}
\end{equation}
is a smoothed version of the data distribution, and the smoothing grows with $t$. This is the progression shown in the bottom row of \Cref{fig:diffusion-visual}, in which the modes of $\pdata$ gradually merge into a single Gaussian.

\subsection{The Reverse Process and Training}
\label{sec:diffusion-reverse}
Generation runs the chain backward. Starting from $\rvx_T \sim \mathcal{N}(0, \boldsymbol{I})$, the model draws samples from learned transitions
\begin{equation}
    p_\theta(\rvx_{t-1} \,|\, \rvx_t) = \mathcal{N}\big(\rvx_{t-1};\, \boldsymbol{\mu}_\theta(\rvx_t, t),\, \sigma_t^2 \boldsymbol{I}\big),
    \label{eqn:ddpm-reverse-step}
\end{equation}
which are approximately Gaussian when each forward step adds only a small amount of noise \citep{sohl-dickstein15}. The parameters are trained by maximizing a variational lower bound on the log-likelihood $\log p_\theta(\rvx_0)$, so diffusion models remain likelihood-based models in the sense of \Cref{sec:mle}. \cite{ho2020denoising} showed that, if the mean is parameterized through a network $\rvepsilon_\theta(\rvx_t, t)$ that predicts the noise added to the sample,
\begin{equation}
    \boldsymbol{\mu}_\theta(\rvx_t, t) = \frac{1}{\sqrt{1 - \beta_t}} \Big( \rvx_t - \frac{\beta_t}{\sqrt{1 - \bar{\alpha}_t}}\, \rvepsilon_\theta(\rvx_t, t) \Big),
\end{equation}
then the bound reduces, up to a weighting of its terms, to a simple regression objective,
\begin{equation}
    \mathcal{L}_\text{simple}(\theta) = \mathbb{E}_{t,\, \rvx_0,\, \rvepsilon} \Big[ \big\lVert \rvepsilon - \rvepsilon_\theta\big(\sqrt{\bar{\alpha}_t}\, \rvx_0 + \sqrt{1 - \bar{\alpha}_t}\, \rvepsilon,\, t\big) \big\rVert_2^2 \Big],
    \label{eqn:ddpm-simple-loss}
\end{equation}
where $t$ is drawn uniformly from $\{1, \ldots, T\}$.

The noise-prediction objective is denoising score matching in disguise. From \Cref{eqn:ddpm-forward-marginal}, the score of the forward transition is $\nabla_{\rvx_t} \log q(\rvx_t \,|\, \rvx_0) = -\rvepsilon / \sqrt{1 - \bar{\alpha}_t}$. Predicting the noise is therefore equivalent to estimating the score of the noisy data, with
\begin{equation}
    \rvs_\theta(\rvx_t, t) = -\frac{\rvepsilon_\theta(\rvx_t, t)}{\sqrt{1 - \bar{\alpha}_t}} \approx \nabla_{\rvx_t} \log q_t(\rvx_t),
    \label{eqn:noise-score-relation}
\end{equation}
and \Cref{eqn:ddpm-simple-loss} is a weighted sum of the denoising score matching objective (\ref{eqn:denoising-score-matching}) over all noise levels. A diffusion model is thus a single network that has learned the score of the data at every degree of corruption.

\subsection{A Continuous-Time View}
\label{sec:diffusion-sde}
\cite{song2021scorebased} unified these models by taking the number of steps to infinity. The forward process then becomes a stochastic differential equation (SDE),
\begin{equation}
    \mathrm{d}\rvx_t = \mathbf{f}(\rvx_t, t)\, \mathrm{d}t + g(t)\, \mathrm{d}\rvw_t,
    \label{eqn:bg-forward-sde}
\end{equation}
where $\rvw_t$ is a standard Wiener process, $\mathbf{f}$ is a drift, and $g$ is a diffusion coefficient. The chain in \Cref{eqn:ddpm-forward-step} is a discretization of the variance-preserving choice $\mathbf{f}(\rvx, t) = -\frac{1}{2}\beta(t)\rvx$ and $g(t) = \sqrt{\beta(t)}$, whereas the variance-exploding choice $\mathbf{f} = 0$ recovers the multi-scale denoising score matching of \cite{song2019generative}. A classical result of \cite{anderson1982reverse} states that this process can be reversed in time by the SDE
\begin{equation}
    \mathrm{d}\rvx_t = \big[\mathbf{f}(\rvx_t, t) - g(t)^2\, \nabla_{\rvx_t} \log p_t(\rvx_t)\big]\, \mathrm{d}t + g(t)\, \mathrm{d}\bar{\rvw}_t,
    \label{eqn:bg-reverse-sde}
\end{equation}
where time flows backward and $\bar{\rvw}_t$ is a Wiener process in reversed time. The only unknown quantity is the score $\nabla_{\rvx} \log p_t(\rvx)$, which is approximated by a network $\rvs_\theta(\rvx, t)$ trained with the time-weighted denoising score matching objective
\begin{equation}
    \theta^* = \argmin_\theta\; \mathbb{E}_{t,\, \rvx_0,\, \rvx_t} \Big[ \lambda(t)\, \big\lVert \rvs_\theta(\rvx_t, t) - \nabla_{\rvx_t} \log p(\rvx_t \,|\, \rvx_0) \big\rVert_2^2 \Big],
    \label{eqn:bg-score-objective}
\end{equation}
with a positive weighting function $\lambda(t)$. The same marginal distributions are also produced by a deterministic probability flow ordinary differential equation,
\begin{equation}
    \mathrm{d}\rvx_t = \Big[\mathbf{f}(\rvx_t, t) - \frac{1}{2} g(t)^2\, \nabla_{\rvx_t} \log p_t(\rvx_t)\Big]\, \mathrm{d}t,
    \label{eqn:bg-pf-ode}
\end{equation}
which underlies fast deterministic samplers \citep{song2020denoising, song2021scorebased}.

\subsection{Diffusion Models as Energy-Based Models}
\label{sec:diffusion-ebm}
Because the score is the gradient of a log density, a diffusion model can be read through the lens of \Cref{sec:energy-based-models}. Writing the noisy marginal as $p_t(\rvx) \propto \exp(-E_t(\rvx))$, the learned score approximates $\rvs_\theta(\rvx, t) \approx -\nabla_{\rvx} E_t(\rvx)$. A diffusion model is then a family of EBMs indexed by the noise level, and the reverse process is a noisy descent on an energy landscape that changes with time. At high noise the landscape is smooth and has a single broad basin, as in the rightmost distribution of \Cref{fig:diffusion-visual}. As the noise decreases, the landscape sharpens and its minima move toward the regions where the data concentrate. By learning only the gradient of the energy, diffusion models sidestep the intractable partition function that complicates the training of EBMs. The score network is, however, a generic neural network, and it is not guaranteed to be the exact gradient of any energy function.

This reading raises a question that is central to this thesis. If the score network were perfectly accurate on a finite training set, $p_t$ would be the empirical distribution of the training data smoothed by Gaussian noise, and the reverse process would return only the training samples \citep{gu2023memorization, biroli2024dynamical}. A diffusion model that generalizes must therefore depart from this ideal in a structured way. As we discuss in Chapter~3, the energy implied by this idealized score takes the form of a DenseAM energy \citep{hoover2023memory, ambrogioni2023search}, which makes the theory of AMs reviewed in \Cref{sec:AM-general} a natural tool for understanding when diffusion models memorize and when they generalize.

\subsection{Discrete Diffusion Models}
\label{sec:discrete-diffusion}
Text is composed of discrete tokens, for which Gaussian noise is not defined. Discrete diffusion models replace it with a corruption process over a finite vocabulary \citep{austin2021structured, campbell2024generative, gat2024discrete, sahoo2024simple}. Let each token be represented by a one-hot vector $\rvx \in \mathcal{V} = \{\rvx \in \{0, 1\}^V : \sum_{i=1}^V x_i = 1\}$, where $V$ is the vocabulary size. The forward process interpolates between the clean token and a fixed prior distribution $\boldsymbol{\pi}$ over the vocabulary,
\begin{equation}
    q_t(\rvz_t \,|\, \rvx) = \text{Cat}\big(\rvz_t;\, \alpha_t \rvx + (1 - \alpha_t)\, \boldsymbol{\pi}\big), \qquad t \in [0, 1],
    \label{eqn:bg-discrete-forward}
\end{equation}
where $\text{Cat}(\cdot)$ is a categorical distribution and $\alpha_t$ decreases from $\alpha_0 \approx 1$ to $\alpha_1 \approx 0$. Two choices of prior are common. In masked diffusion, $\boldsymbol{\pi}$ places all of its mass on a special mask token, so corruption hides tokens \citep{austin2021structured, sahoo2024simple}. In uniform-state diffusion, $\boldsymbol{\pi} = \mathbf{1}/V$, so corruption replaces tokens with random ones \citep{austin2021structured, sahoo2025duality}. In the uniform case, a corrupted token looks like any other token, and the model must judge every position of the sequence, including tokens that may already be correct. Generation can therefore revise tokens repeatedly, much as the state of an AM is updated until it settles.

For a time $s < t$, the reverse transition of this forward process is available in closed form once the clean token is known \citep{austin2021structured, campbell2024generative},
\begin{equation}
    q_{s|t}(\rvz_s \,|\, \rvz_t, \rvx) = \frac{q_{t|s}(\rvz_t \,|\, \rvz_s)\, q_s(\rvz_s \,|\, \rvx)}{q_t(\rvz_t \,|\, \rvx)},
    \label{eqn:bg-discrete-posterior}
\end{equation}
which is again a categorical distribution whose probabilities are fixed by $\rvz_t$, $\rvx$, $\boldsymbol{\pi}$, and the schedule $\alpha_t$. During generation the clean token is unknown, so it is replaced by the prediction of a network $\rvx_\theta(\rvz_t, t)$,
\begin{equation}
    p_\theta(\rvz_s \,|\, \rvz_t) = q_{s|t}\big(\rvz_s \,|\, \rvz_t, \rvx = \rvx_\theta(\rvz_t, t)\big), \qquad \rvx_\theta(\rvz_t, t) = p_\theta(\rvx \,|\, \rvz_t) \in \Delta^{V},
    \label{eqn:bg-discrete-reverse}
\end{equation}
where $\Delta^{V}$ denotes the probability simplex over the vocabulary. The output of the network is therefore not a direction in which to move, but a categorical distribution over which clean token occupies each position. For a sequence of $L$ tokens, the network reads the entire corrupted sequence $\rvz_t^{1:L}$ and returns one such distribution $p_\theta(\rvx^\ell \,|\, \rvz_t^{1:L})$ for every position $\ell$.

The parameters are trained by minimizing the negative evidence lower bound (NELBO) on the log-likelihood of the data \citep{austin2021structured, sahoo2024simple},
\begin{equation}
    \begin{split}
    \mathcal{L}_\text{NELBO}(\theta) = \mathbb{E}_{q} \bigg[ &\underbrace{- \log p_\theta(\rvx \,|\, \rvz_{t_\epsilon})}_{\mathcal{L}_\text{reconstruction}} + \underbrace{\sum_{s < t} D_\text{KL}\big( q_{s|t}(\rvz_s \,|\, \rvz_t, \rvx) \,\|\, p_\theta(\rvz_s \,|\, \rvz_t) \big)}_{\mathcal{L}_\text{diffusion}} \bigg] \\
    &+ \underbrace{D_\text{KL}\big( q_1(\rvz_1 \,|\, \rvx) \,\|\, p(\rvz_1) \big)}_{\mathcal{L}_\text{prior}},
    \end{split}
    \label{eqn:bg-nelbo}
\end{equation}
where the prior term does not depend on $\theta$. Because the true and learned reverse transitions in \Cref{eqn:bg-discrete-reverse} differ only in whether the clean token is given or predicted, every term that depends on $\theta$ is a function of the predicted distribution $p_\theta(\rvx \,|\, \rvz_t)$ alone. The reconstruction term makes this explicit. For a sequence, it is the cross-entropy between the clean tokens and the predictions of the network,
\begin{equation}
    \mathcal{L}_\text{reconstruction}(\theta) = - \sum_{\ell = 1}^{L} \log p_\theta\big(\rvx^\ell \,|\, \rvz_{t_\epsilon}^{1:L}\big) = - \sum_{\ell = 1}^{L} \big\langle \rvx^\ell,\, \log \rvx_\theta^\ell(\rvz_{t_\epsilon}^{1:L}, t_\epsilon) \big\rangle,
    \label{eqn:bg-discrete-ce}
\end{equation}
where $t_\epsilon$ is a small time close to zero. Note that $\rvz_{t_\epsilon}$ is not necessarily a clean sequence itself. Since $\alpha_{t_\epsilon} \approx 1$ rather than exactly one, $\rvz_{t_\epsilon}$ is a slightly perturbed version of $\rvx$ produced by the forward process at this small time, and the reconstruction term asks the network to recover the clean tokens from an input that is nearly, but not exactly, clean. The diffusion term compares, position by position, the posterior obtained from the true token with the posterior obtained from the predicted one. In the masked case, the continuous-time limit of \Cref{eqn:bg-nelbo} reduces exactly to a time-weighted cross-entropy over the masked positions \citep{sahoo2024simple},
\begin{equation}
    \mathcal{L}_\text{NELBO}^{\infty}(\theta) = \mathbb{E}_{q} \int_0^1 \frac{\alpha_t'}{1 - \alpha_t} \sum_{\ell = 1}^{L} \log \big\langle \rvx_\theta^\ell(\rvz_t^{1:L}, t),\, \rvx^\ell \big\rangle \, \mathrm{d}t,
    \label{eqn:bg-mdlm-loss}
\end{equation}
where $\alpha_t' < 0$ is the time derivative of the schedule, so that each term is a positively weighted negative log-probability of the correct token.

These objectives differ from their continuous counterparts in an important way. A continuous diffusion model learns a score $\rvs_\theta(\rvx_t, t)$, or equivalently the noise $\rvepsilon_\theta(\rvx_t, t)$ to be removed, which tells the sampler in which direction and by how much to move the current state. Nothing of this kind is learned here. The network never estimates a gradient of a log density or the amount of corruption to undo. How far each reverse step travels is fixed entirely by the schedule $\alpha_t$ and the closed-form posterior in \Cref{eqn:bg-discrete-posterior}. What the network learns is a classifier, the conditional likelihood of each clean token given a corrupted version of the others. As we discuss in Chapter~4, this is the same structure as the pseudo-likelihood objective for AMs \citep{pseudo-likelihood, d2025pseudo}, and it is the reason basins of attraction can form in these models without an explicit energy function.

\section{Associative Memory Networks}
\label{sec:AM-general}

The previous sections described generative models whose goal is to approximate a data distribution, and showed that their training objectives, taken literally, reward memorization of the training set. We now turn to a class of models for which memorization is not a side effect but the very object of study. Associative Memory (AM) networks, also known as Hopfield networks, offer a complementary and historically rich perspective on energy-based dynamical neural systems.

These networks are designed to store and retrieve patterns by leveraging attractor dynamics on an energy landscape. Unlike generative models that aim to model the full data distribution, AM networks encode a finite set of patterns and recover them from noisy or incomplete cues. Originally inspired by physical systems such as magnets and spin glasses, and by biological principles such as Hebbian learning \citep{Hebb1949}, AM networks provide a clean and interpretable view of how information can be embedded in and recovered from a neural system.

From the beginning, the central question of AM theory has been one of capacity. How many patterns can a network of a given size store and reliably retrieve? How large is the basin of attraction around each memory, and how does it shrink as more patterns are stored? What happens to the energy landscape once the capacity is exceeded? These questions have been studied for four decades with the tools of statistical mechanics \citep{Amit1985StoringIN, amit1987statistical, gardner1988space}, and they have produced a precise vocabulary for memorization. A memory is a fixed point of the dynamics, its robustness is the size of its basin, and the limit of what a network can memorize is its capacity. More recently, the same theory has begun to describe generalization. When an AM is loaded beyond its capacity with patterns that share an underlying structure, it can form new attractors near examples it has never stored \citep{kalaj2024random, d2025pseudo}.

This combination is what makes AM essential to the questions of this thesis. Generative models are judged by whether they memorize or generalize, yet they rarely offer a direct way to see what they have stored. AM networks offer exactly this. Their memories are explicit, their retrieval dynamics are guaranteed to converge, and the transition from storing patterns to creating new ones can be located and measured. If modern generative models can be described as AMs, then this vocabulary of fixed points, basins, capacity, and spurious states becomes a language for explaining their behavior.

In this section, we provide an overview of classical and modern AM networks. We examine the core mechanisms behind their storage and retrieval, discuss their capacity and the spurious states that appear beyond it, and review Dense Associative Memories and their connection to attention. The connections between these ideas and diffusion models are developed in the later chapters.

\subsection{Classical Associative Memories}
\label{sec:classical-am}

In the midst of the second Artificial Intelligence (AI) winter, Hopfield networks were introduced as models of how biological brains might store and recall information. Developed by \cite{Hopfield1982}, while unknowingly building on the earlier work of \cite{AmariS.I.1972} and on the Lenz-Ising model introduced in 1920 \citep{Lenz-Ising-history1967}, these models became one of the first concrete examples of a dynamical neural network capable of storing binary patterns as local minima of an energy landscape and retrieving them. As a result, Hopfield networks helped revitalize interest in artificial neural networks at the time.

The inspiration behind AM networks came from two key sources, magnetism and Hebbian learning. From physics, \cite{Hopfield1982} borrowed the idea that a system can settle into a low-energy configuration, similar to how magnetic spins align with one another. Each stable pattern, or memory, corresponds to a local minimum of the energy landscape of the system. From neuroscience, the Hebbian learning rule \citep{Hebb1949, Eccles1953, Rosenblatt1958ThePA}, often summarized as ``neurons that fire together wire together,'' served as the foundation for storing patterns in the network. In classical networks, this rule strengthens the connections between co-activated neurons, allowing the system to encode patterns through their correlations and later retrieve them.

\begin{figure}[t]
    \centering
    \includegraphics[keepaspectratio, width=1.0\textwidth, height=1\textheight]{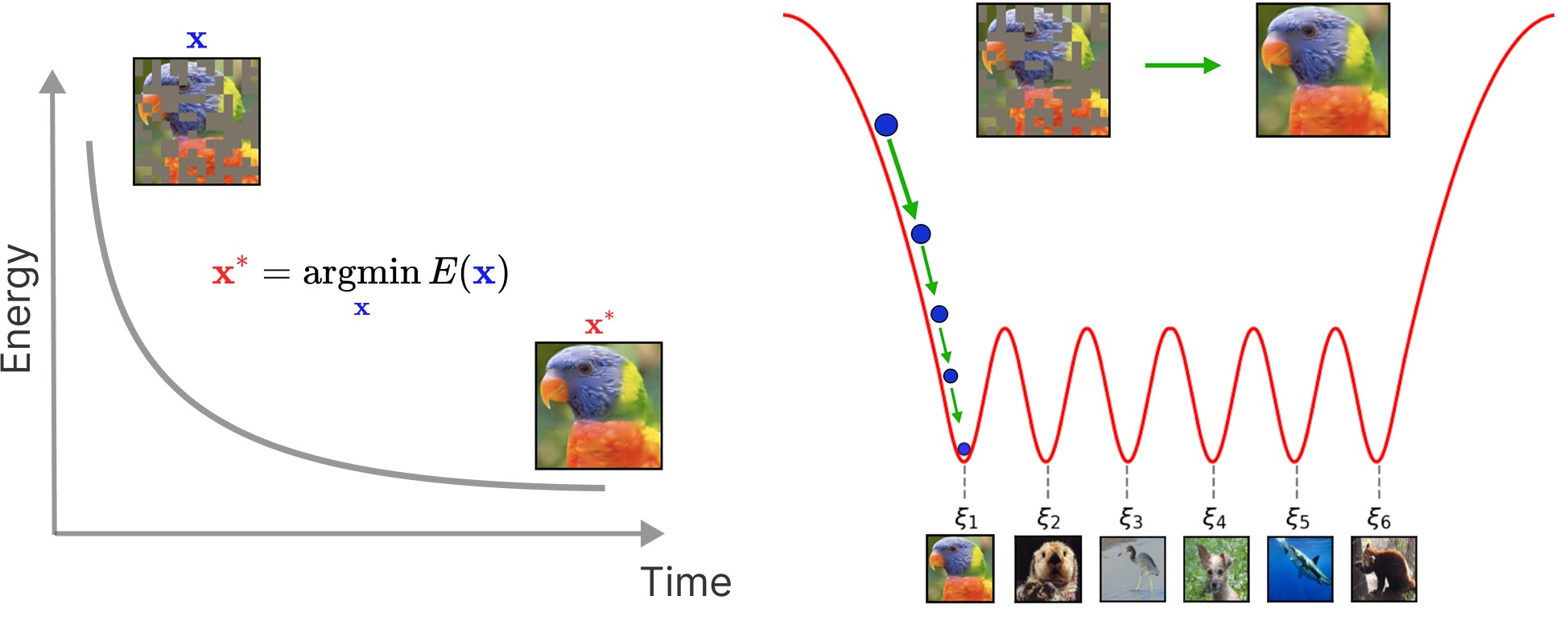}
    \caption{An illustration of the retrieval process in a high-capacity AM network. Given a query ${\rvx}$, the network performs gradient descent on its energy function with respect to the query. Because the energy is a Lyapunov function \citep{lyapunov1992general}, the dynamics are guaranteed to settle at a fixed point, here the memory ${\rvx^*}$ closest to the query. The animal images are retrieved from the ImageNet-1K dataset \citep{deng2009imagenet}.}
    \label{fig:am-memory-retrieval}
\end{figure}

Specifically, using the Hebbian learning rule, these networks form a symmetric weight matrix $\mathbf{W} \in \mathbb{R}^{N \times N}$,
\begin{equation}
    \mathbf{W} = \sum_{\mu = 1}^K  \xi^{\mu} (\xi^{\mu})^\top,  
    \label{eqn:classical-hopfield-storing}
\end{equation}
to store a set of $K$ patterns $\xi^{\mu} \in \{\pm 1\}^{N}$. The diagonal entries of $\mathbf{W}$ are set to zero to prevent self-connections. Once the matrix $\mathbf{W}$ is formed, retrieval is governed by an energy function. Given a binary query $\rvx \in \{\pm 1 \}^N$, the energy introduced by \cite{Hopfield1982} is
\begin{equation} 
    E(\rvx) = -\frac{1}{2} \rvx^\top \mathbf{W} \rvx.
    \label{eqn:classical-hopfield-energy} 
\end{equation} 
This energy determines the dynamics of the network, which iteratively updates the query $\rvx$ so as to reduce its energy. The most common scheme is the asynchronous update, in which one neuron at a time aligns with the sign of its local field, $x_i \leftarrow \text{sgn}\big(\sum_j W_{ij} x_j\big)$ \citep{Hopfield1982, Hopfield1984}. Because each such update can only decrease the energy, the dynamics are guaranteed to converge to a fixed point. Synchronous updates, in which all neurons are updated simultaneously, have also been studied, but they can lead to oscillations because the energy is no longer guaranteed to decrease monotonically \citep{Cheung87, HertzJ-Intro-NeuralComp}. If the number of stored patterns $K$ is below the capacity of the network, a noisy or incomplete version of a stored pattern is guided by the dynamics to the nearest attractor, which is the stored memory itself. This attractor-based behavior is the core mechanism behind pattern completion and retrieval in AM networks, as illustrated in \Cref{fig:am-memory-retrieval}.

A crucial property of Hopfield networks is that their energy function is a Lyapunov function, a function that decreases monotonically along the trajectories of the system and is bounded from below \citep{lyapunov1992general}. This property ensures that the dynamics are stable and converge to a local minimum of the energy landscape. The set of initial states that flow to a given minimum is its basin of attraction, and its size measures how much corruption a memory can tolerate before it is lost. Thus, unlike standard recurrent neural networks \citep{ElmanRNN, LSTM, JordanRNN}, which process sequences of inputs and learn temporal dependencies, AM networks perform fixed-point retrieval by evolving an input state toward a local energy minimum through gradient descent-like dynamics.

\subsection{Memory Capacity and Spurious States}
\label{sec:limited-capacity}

The capacity of the classical AM network was the first question asked of it, and its answer exposed the most important limitation of the model. \cite{Hopfield1982} observed experimentally that a network of $N$ neurons can store roughly $0.15N$ random, uncorrelated patterns. Using the replica method from the statistical mechanics of spin glasses, \cite{Amit1985StoringIN} later derived the critical capacity analytically, $K^\text{max} \approx 0.14N$. The analysis revealed a sharp transition governed by the load $\alpha = K/N$. Below the critical load, the stored patterns are stable attractors with finite basins. Above it, interference between the patterns destroys them all at once, and the network enters a phase in which no stored pattern can be retrieved \citep{amit1987statistical}. As more patterns are added, the attractor landscape thus becomes increasingly crowded, recall becomes unstable, and the dynamics converge to incorrect states \citep{AbuMostafa1985InformationCO}.

The capacity of the Hebbian rule is not the capacity of the network itself. \cite{gardner1988space} asked how many patterns could be stored by any choice of couplings, and showed that the answer reaches $K^\text{max} = 2N$ for uncorrelated patterns. Her analysis tied the stability of each memory to a margin, the amount by which each neuron agrees with the local field produced by all the others, and showed that larger margins produce larger basins of attraction \citep{gardner1988space, Gardner1988OptimalSP}. This view of memory as a set of classification problems, one per neuron, reappears in Chapter~4, where it explains how basins can form without an energy function.

The crowding of patterns as their number $K$ increases leads to a second phenomenon, the appearance of spurious states. These are stable states of the network that do not correspond to any of the stored patterns, as illustrated in \Cref{fig:dam-am-contrast}. They emerge from crosstalk between memories, particularly as the number of stored patterns approaches the capacity of the network, and in the classical model many of them are mixtures of a few stored patterns \citep{Amit1985StoringIN}. Hopfield later proposed an ``unlearning'' mechanism that suppresses spurious attractors by raising their energy without disrupting the true memories \citep{Hopfield1983UnlearningHA}. Interestingly, this unlearning process is akin to the contrastive divergence objective (\ref{eqn:contrastive-loss-ebm-simplify}) discussed in \Cref{sec:contrastive-div}. It is, however, computationally expensive and does not fully eliminate the issue in practice. Classical AM networks are also highly sensitive to correlations between patterns. Hebbian storage works well only when the patterns are nearly uncorrelated. When patterns overlap, interference increases and retrieval deteriorates, which makes classical Hopfield networks poorly suited to natural data, where correlations are the rule.

For most of the history of the field, spurious states were therefore treated as defects to be removed. Recent work suggests a different reading. When the stored patterns are not random but are samples from a structured distribution, overloading the network need not lead to a blackout. Instead, new attractors can form near examples drawn from the same distribution that were never stored, while the stored patterns remain retrievable \citep{kalaj2024random, d2025pseudo}. Beyond capacity, an AM can therefore begin to generalize. This observation turns the classical question of capacity into a question about the transition from memorization to generalization, which is the question this thesis asks of diffusion models.

\subsection{Dense Associative Memories}
\label{sec:DenseAM}

Following their initial popularity, AM networks gradually fell out of favor because of their limited storage capacity. For decades, they remained largely of theoretical interest, overshadowed by deep learning architectures that prioritized end-to-end optimization and continuous representations. This changed with the introduction of Dense Associative Memories (DenseAMs), also known as modern Hopfield networks, which reformulated the AM energy function to achieve much larger storage capacities.

As demonstrated by \cite{DAM2016}, AM models can achieve a capacity that grows super-linearly with the number of neurons $N$ by applying a rapidly growing interaction function to the overlap between the query and each memory. Specifically, the energy in \Cref{eqn:classical-hopfield-energy} can be generalized to
\begin{equation}
    E(\rvx) = - \sum^K_{\mu = 1} F  \big ( (\xi^{\mu})^\top \rvx \big ),
    \label{eqn:general-dam-energy}
\end{equation}
where $F(\cdot)$ is an interaction function and $\xi^{\mu}$ is one of the $K$ stored memories. When $F$ is quadratic, this energy reduces to the classical one,
\begin{equation}
    \begin{split}
        E(\rvx) &= - \sum^K_{\mu = 1} \big ( (\xi^{\mu})^\top \rvx \big )^2 \\ 
        & = - \rvx^\top \bigg( \sum^{K}_{\mu = 1} \xi^\mu  (\xi^\mu)^\top  \bigg) \rvx \\ 
        & = - \rvx^\top \mathbf{W} \rvx,
    \end{split}
\end{equation}
which is the energy (\ref{eqn:classical-hopfield-energy}) of \cite{Hopfield1982} up to a constant factor.

\begin{figure}[t]
    \centering
    \includegraphics[keepaspectratio, width=1\textwidth, height=1\textheight]{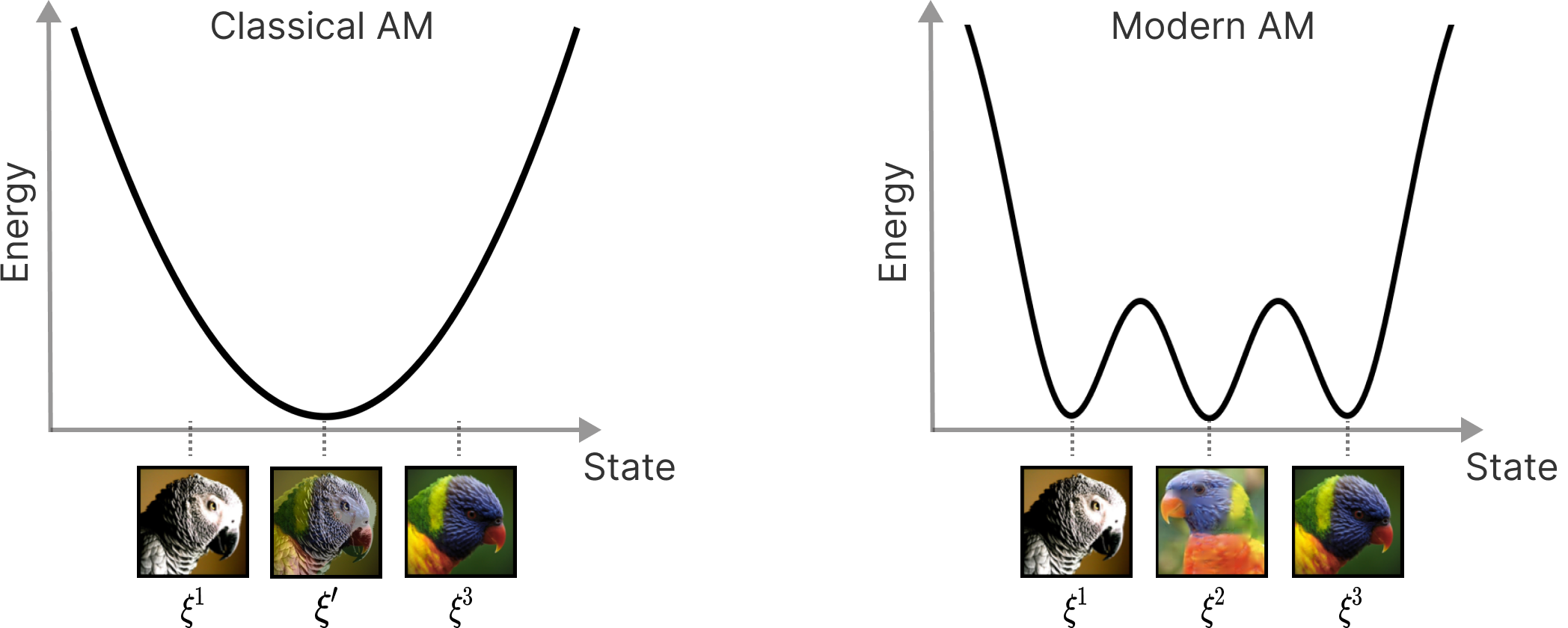}
    \caption{The energy landscape of a classical AM network (left) and of a DenseAM (right). Memories in classical AM networks interfere with one another much more strongly than in DenseAMs. This interference produces spurious patterns (shown as $\xi'$), which can be combinations of stored patterns or random-like states. In the classical network (left), they are shown as an interpolation between two stored patterns, and they hamper retrieval. DenseAMs instead use higher-order interaction functions, which reduce the interference among memories. The animal images are retrieved from the ImageNet-1K dataset \citep{deng2009imagenet}.}
    \label{fig:dam-am-contrast}
\end{figure}

\begin{figure}[t]
    \centering
    \includegraphics[keepaspectratio, width=1\textwidth, height=1.\textheight]{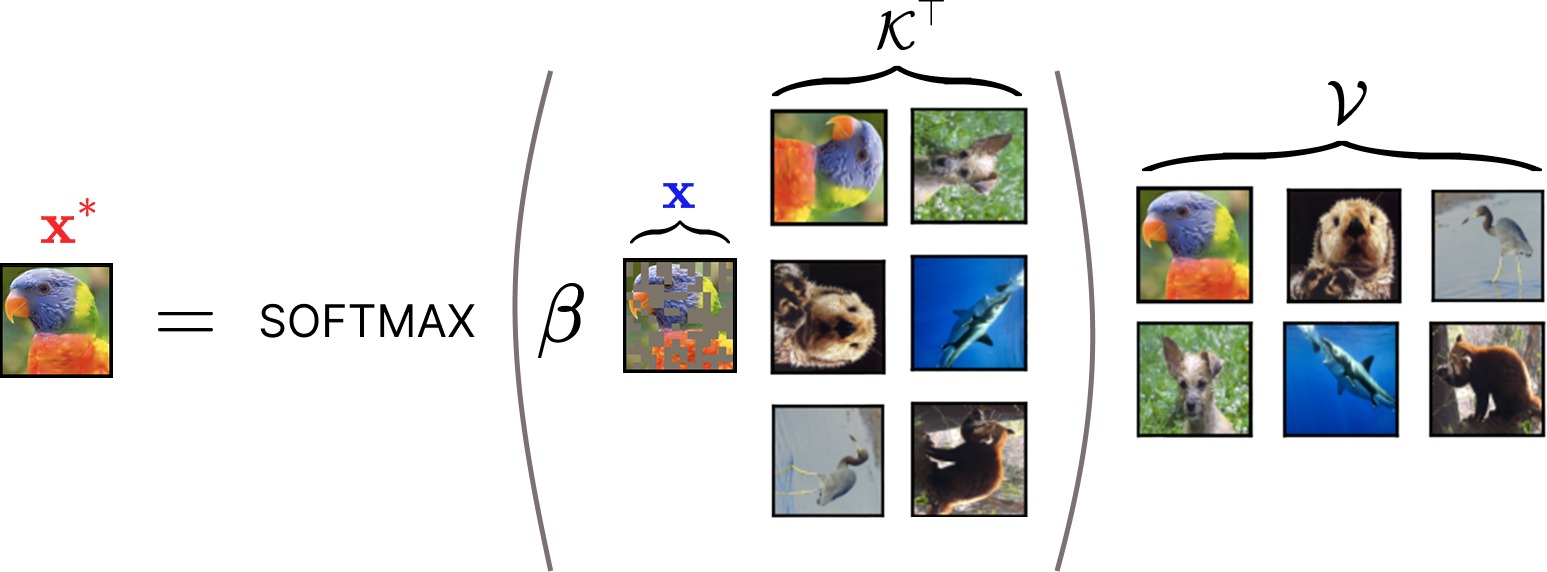}
    \caption{An illustration of the DenseAM update rule (\ref{eqn:attention-dam}). The query ${\rvx}$ is compared with the memories $\xi$, which serve as both the key matrix $\mathcal{K}$ and the value matrix $\mathcal{V}$. The $\softmax$ assigns a weight to each memory according to its similarity to the query ${\rvx}$, and the weighted sum of the values retrieves the closest memory ${\rvx^*}$. The animal images are retrieved from the ImageNet-1K dataset \citep{deng2009imagenet}.}
    \label{fig:dam-softmax}
\end{figure}

The choice of $F(\cdot)$ can be interpreted as defining a similarity measure between the query $\rvx$ and the stored memories $\xi^\mu$ \citep{millidge2022universal}. By increasing the steepness of $F(\cdot)$, DenseAMs shape the energy landscape to have deeper and more distinct minima around each stored memory \citep{krotov2023new}. This reduces interference, improves the reliability of retrieval, and allows the network to store a significantly larger number of patterns, as shown in \Cref{fig:dam-am-contrast}. With a polynomial interaction $F(x) = x^n$, \cite{DAM2016} showed that a DenseAM can store $K^\text{max} \propto N^{n - 1}$ patterns. With an exponential interaction $F(x) = \exp(x)$, \cite{demircigil2017model} showed that the capacity becomes exponential in $N$. Capacity thus becomes a design choice rather than a fixed limitation, and the steepness of the interaction controls where the boundary between memorization and its failure lies. In addition to their larger capacity, DenseAMs offer more flexibility in how patterns are represented and retrieved. Unlike classical AM networks, which operate on binary states, DenseAMs naturally support continuous queries and memories. This allows for more expressive memory representations and makes the model amenable to learning its memories with stochastic gradient descent \citep{DAM2016}.

One of the most striking insights from recent work on DenseAMs is their strong connection to the attention mechanism of Transformers \citep{Attention2017}. \cite{Ramsauer2021} introduced an alternative form of the DenseAM energy,
\begin{equation}
    E(\rvx) = -\text{LSE} (\beta, \xi, \rvx) + \frac{1}{2} \rvx^\top \rvx + \text{constant},
    \label{eqn:dam-softmax-energy-v1}
\end{equation}
where 
\begin{equation}
    \text{LSE} (\beta, \xi, \rvx) = \frac{1}{\beta} \log \sum^{K}_{\mu = 1} \exp \bigg  ( \beta (\xi^\mu)^\top \rvx \bigg ), 
\end{equation}
$\xi \in \mathbb{R}^{K \times N}$ is the memory matrix, and the inverse temperature $\beta$ controls the separation between memories. The constant in \Cref{eqn:dam-softmax-energy-v1} does not depend on $\rvx$ and can be ignored. Using the concave-convex procedure \citep{CCP-Yuilee}, \cite{Ramsauer2021} showed that the update rule for the query $\rvx$ reduces to
\begin{equation}
    \rvx^\text{new} = \softmax \big ( \beta \, \rvx \, \xi^\top  \big) \, \xi,
    \label{eqn:attention-dam}
\end{equation}
which is equivalent to the attention mechanism. Specifically, letting the key be $\mathcal{K} = \xi$, the value be $\mathcal{V} = \xi$, and the query be $\mathcal{Q} = \rvx$, \Cref{eqn:attention-dam} becomes
\begin{equation}
    \rvx^\text{new} = \softmax \big ( \beta \mathcal{Q} \, \mathcal{K}^\top \big) \, \mathcal{V},
\end{equation}
which is the attention formulation of \cite{Attention2017}.

Unlike standard attention, however, the key and value are shared as the memory matrix $\xi$, while the query is a dynamical variable that evolves according to the DenseAM energy, as shown in \Cref{fig:dam-softmax}. Depending on the inverse temperature $\beta$, a minimum of \Cref{eqn:dam-softmax-energy-v1} can be (1) a global fixed point that averages all stored patterns, (2) a metastable state that averages several similar patterns, or (3) a single stored pattern when $\beta$ is large \citep{Ramsauer2021}. The inverse temperature therefore moves the network between retrieving individual memories and producing averages of them, a behavior that will reappear in diffusion models, where the noise level plays the role of temperature. This formulation also offers a powerful interpretation of attention as a step of gradient descent on a principled energy function, connecting two paradigms that had been studied in isolation.

\section{Conclusion}
In this chapter, we introduced the concepts on which the rest of this thesis builds. Probabilistic generative models are trained on finite datasets with objectives that, taken literally, reward memorization. EBMs offer a flexible way to represent such models, and diffusion models learn their scores across all levels of noise, in both continuous and discrete settings. AM networks, in turn, have always been about memorization. Their theory explains what a network can store, how robustly it stores it, and what happens beyond its capacity, where spurious states and, with structured data, generalization emerge. The following chapters connect these paradigms. Chapter~3 views continuous diffusion models as DenseAMs whose memories are their training samples, and Chapter~4 shows that the conditional likelihood objective of discrete diffusion models forms basins of attraction without an energy function. Together, they study how memorization and generalization coexist in diffusion models for images and for language.

%% file: chapters/ch3_continuous/ch3.tex
\providecommand{\cfig}[2]{\hyperref[#1]{\ref*{#1}#2}}

\chapter{Emergence of Diffusion Models from Associative Memory}
\label{chapter:dam-and-diffusion}
\blfootnote{Portions of this chapter have previously appeared as:\newline Pham, B.\textsuperscript{*}, Raya, G.\textsuperscript{*}, Negri, M., Zaki, M. J., Ambrogioni, L., \& Krotov, D. (2026). Emergence of diffusion models from associative memory through memorization and generalization. \textit{Nature Communications}. \url{https://doi.org/10.1038/s41467-026-78727-y}\newline Pham, B.\textsuperscript{*}, Raya, G.\textsuperscript{*}, Negri, M., Zaki, M. J., Ambrogioni, L., \& Krotov, D. (2025). \textit{Memorization to generalization: Emergence of diffusion models from associative memory}. arXiv. \url{https://doi.org/10.48550/arXiv.2505.21777}\newline \textsuperscript{*}Equal contribution as first author.}

\section{Introduction}
\label{sec:ch3-intro}

Hopfield networks are energy-based Associative Memory (AM)  models, which conceptualize memories as attractor states corresponding to local minima of their energy function, and the memory retrieval process as the dynamical convergence towards such attractors \citep{Hopfield1982, Hopfield1984, AmariS.I.1972}. Modern variants of such networks, called Dense Associative Memories (DenseAMs) \citep{DAM2016, krotov2018dense}, have helped revitalize  scientific interest in these ideas and paved the way for more sophisticated AM systems \citep{demircigil2017model, agliari2020neural, agliari2020tolerance, albanese2022replica, millidge2022universal, sahaCLAM, krotov2023new, et}, driven primarily by their connection to the attention mechanism in Transformers \citep{Ramsauer2021, et} and neurobiology \citep{krotov2021large, kozachkov2025neuron}.

Simultaneously, diffusion models (DMs) \citep{sohl-dickstein15} have recently gained a large popularity, due to their flexibility and accuracy in modeling various high-dimensional distributions \citep{ho2020denoising, song2019generative, song2021scorebased, rombach2022high}. However, despite their effectiveness, DMs pose challenges related to data privacy and security, as concerns grow about their tendency to replicate training data alongside a lack of understanding of how generalization arises in these models \citep{somepalli2023diffusion, somepalli2023understanding, carlini2023, wen2024detecting, jeon2024understanding, webster2023reproducible}. Hence, such matters emphasize the need for further understanding of memorization and generalization behaviors of DMs.

Currently, much of the recent works on the memorization-to-generalization transition of DMs tackle it in a generalization-centric fashion \citep{meehan2020non, burg2021on, somepalli2023diffusion, somepalli2023understanding, kadkhodaie2023generalization, ventura2024manifolds, achilli2024losing, kamb2024analytic, ross2024geometric, biroli2024dynamical, cui2025precise, achilli2025memorization}. Namely, the memorization phenomenon is typically viewed as a ``small side effect,'' alleviated by drastically increasing the training data size \citep{yoon2023diffusion, gu2023memorization}. Although these studies shed some light on memorization and generalization, they do not fully explore the \textit{intermediate transition regime} between these two phases. Our work adopts a complimentary approach which further ties memorization and generalization together. From the beginning, we cast the diffusion modeling pipeline into the AM framework. The training phase of diffusion modeling is conceptualized as the operation of {\em writing the training data into the memory}. The generation phase is viewed as {\em the attempt of memory recall}. This recall can be successful, resulting in the retrieval of training samples (memorization), or unsuccessful, resulting in the generation of new previously unseen samples (generalization). Consequently, this point of view allows us to apply the theory developed for AMs to the memorization-to-generalization transition in DMs, particularly when DMs operate in the memorization regime.  

Recent works \citep{hoover2023memory, ambrogioni2023search, raya2024spontaneous} have also begun establishing theoretical connections between DenseAMs and DMs, illustrating that the score function of typical DMs can be interpreted as the gradient of a DenseAM's energy function. The focus of our work is the peculiar phenomenon commonly observed in AMs, called spurious states. Although spurious states is a cornerstone phenomenon in AM literature, they have not been predicted or found in DMs. Thus, the demonstration of  \textit{the emergence of spurious states} can further bridge DenseAMs and DMs. Historically considered as detrimental to reliable memory recall \citep{Hopfield1982, Hopfield1983UnlearningHA, Hopfield1984, Amit1985StoringIN, AbuMostafa1985InformationCO}, spurious states were viewed as negative artifacts of AM networks. In contrast, for generative modeling these same spurious states serve as the first signatures of generalization capabilities, thus becoming a desirable emergent phenomenon \citep{kalaj2024random}.

\begin{figure}[t]
    \centering
    \includegraphics[keepaspectratio, width=1\textwidth, height=0.5\textheight]{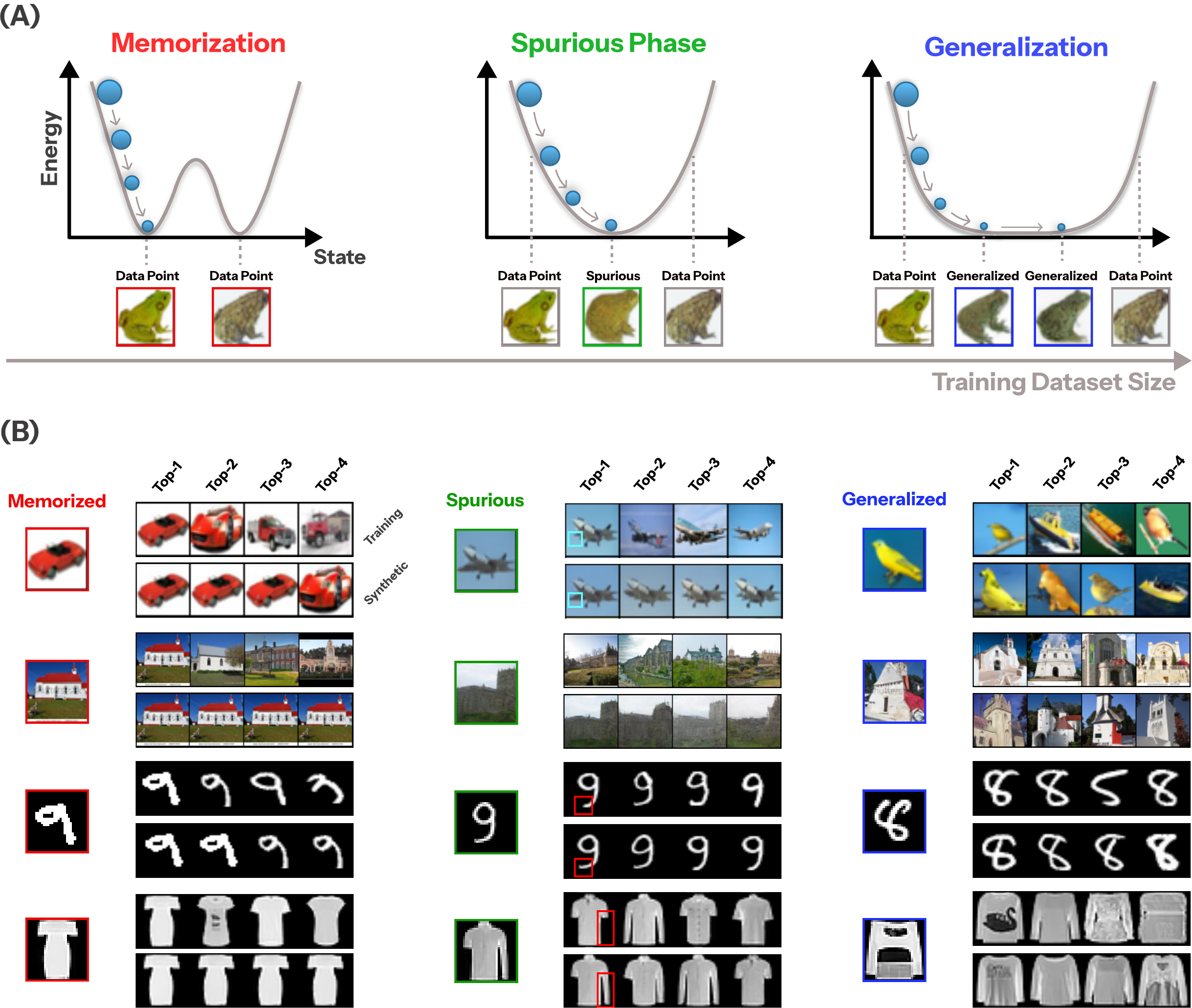}
    \caption{Schematic illustration of the energy landscape and the sample type at each phase of the transition. (A) Behavior of the energy landscape around generated samples as the size of the training dataset is increased. In the small data regime, the model stores the training data as local minima of the energy. When the amount of training data exceeds the model's memory capacity, spurious patterns are formed and training data points are no longer energy minima. Subsequent increase of the training data size leads to the generalization phase, defined by the formation of continuous manifold of the low energy states.(B) Examples of {memorized}, {spurious}, and {generalized} samples in their respective columns for four datasets, MNIST \citep{mnist}, FASHION-MNIST \citep{fmnist}, CIFAR-10 \citep{cifar10}, and LSUN-CHURCH \citep{lsun}. See \Cref{sec:ch3-detection} for our definitions of these three sample types. For each target image (shown on the left), its top-4 nearest neighbors from the training set (top row) and the synthetic set (bottom row) are shown to highlight the novelty and commonality of the target image with respect to its training and synthetic sets. To help highlight the novelty of spurious patterns, we provide rectangular markers guiding where the features differ in their corresponding nearest neighbors. 
    }
    \label{fig:general-energy-transition}
\end{figure}

In accord with previous literature, we empirically find that DMs undergo a transition from memorization to generalization as the number of training points increases on realistic datasets. Unlike previous works, at the onset of generalization we detect the emergence of spurious patterns, which have sizable basins of attraction, but do not correspond to memorized data points. No longer negative artifacts, spurious states are now observed to be positive indicators of early generalization in DMs and they may also exist even during the full generalization stage of DMs, including Stable Diffusion \citep{rombach2022high}. Our work does not prove that these spurious states are the causal driver of generalization. However, the confluence of our empirical results suggests that this is a plausible interpretation of the observed phenomena. Lastly, we provide theoretical descriptions distinguishing these peculiar states from memorized and generalized patterns, in terms of energy landscapes. Our findings suggest a close parallel between two seemingly different systems, DMs and DenseAMs, studied by two disjoint communities. Specifically, the DenseAM's perspective on DMs can illuminate some of the computational properties of DMs at the boundary of the memorization-generalization transition and deep inside the memorization phase. 

The remainder of this chapter is organized as follows. \Cref{sec:ch3-dm-am} makes the connection between diffusion models and DenseAMs precise and explains why spurious states should be expected. \Cref{sec:ch3-detection} introduces the detection metrics that classify each generated sample as memorized, spurious, or generalized. \Cref{sec:ch3-results} presents our results on natural image datasets, on a two-dimensional toy model, and on Stable Diffusion, characterizing the three sample types through their abundance, the volume of their basins of attraction, and the curvature of the energy around them. \Cref{sec:ch3-discussion} discusses the implications and limitations of these findings, and the experimental details and additional results are collected in \Cref{sec:ch3-supp}.

\section{Diffusion Models and Dense Associative Memories}
\label{sec:ch3-dm-am}
Motivated by \cite{hoover2023memory}, \cite{ambrogioni2023search}, and \cite{raya2024spontaneous}, we first provide a theoretical link between DMs and DenseAMs through the perspective of energy. We build on the continuous-time formulation of diffusion models reviewed in \Cref{sec:diffusion-sde}. Let $\rvy \in \mathbb{R}^N$ denote a sample drawn from a target data distribution $p(\rvy)$, with $\rvx_0 = \rvy$. The forward and reverse processes are given by Eqs.~(\ref{eqn:bg-forward-sde}) and (\ref{eqn:bg-reverse-sde}), and the score $\nabla_{\rvx_t} \log p_t(\rvx_t) = -\nabla_{\rvx_t} E(\rvx_t, t)$ is estimated by a neural network $s_\theta(\rvx_t, t)$, where $\theta$ denotes its set of trainable parameters, trained with denoising score matching across multiple time steps \citep{ho2020denoising, song2021scorebased, hyvarinen05a, VincentP2011}.

Consider the training data distribution in the variance-exploding setting of $f(\rvx_t, t) = 0$ and $g(t) = \sigma$. In this case, the marginal probability distribution of noisy samples at time $t$ can be computed exactly as
\begin{equation}
    p(\rvx_t, t) = \underset{\rvy \sim \text{data}}{\mathbb{E}} \Bigg[\frac{1}{(2\pi\sigma^2 t)^{\frac{N}{2}}}\exp\Big(- \frac{\lVert \rvx_t - \mathbf{y} \rVert^2_2}{2 \sigma^2 t}\Big) \Bigg] .
\end{equation}
Assuming the empirical distribution of the data $p(\mathbf{y}) = \frac{1}{K}\sum\limits_{\mu=1}^K \delta^{(N)}(\mathbf{y} - \boldsymbol{\xi}^\mu)$, where $\boldsymbol{\xi}^\mu$ represents an individual data point (belonging to a dataset with training data size $K$), this marginal distribution can be written as 
\begin{equation}
    p(\rvx_t, t) = \frac{1}{K} \sum \limits_{\mu=1}^K \frac{1}{(2\pi\sigma^2 t)^{\frac{N}{2}}}  \exp\Big({ - \frac{\lVert \rvx_t - \boldsymbol{\xi}^\mu \rVert^2_2}{2 \sigma^2 t}}\Big)
    \overset{\text{def}}{\equiv} \exp\Bigg (- {\frac{E^\text{DM}(\rvx_t, t)}{2 \sigma^2 t}}\Bigg ),
    \label{eq:marginal}
\end{equation}
where we also defined the energy $E^\mathrm{DM}$ of the DM, up to $\rvx$--independent terms, to be 
\begin{equation}
    E^\text{DM}(\rvx_t, t) = -2 \sigma^2 t \log\bigg[\sum\limits_{\mu=1}^K \exp \Big(- \frac{\lVert \rvx_t - \boldsymbol{\xi}^\mu \rVert^2_2}{2 \sigma^2 t}\Big) \bigg] .
    \label{eq:energy-diffusion}
\end{equation}
As already observed in \cite{ambrogioni2023search}, Eq.~(\ref{eq:energy-diffusion}) is closely related to a commonly studied DenseAM's model, see for example \cite{sahaCLAM}:
\begin{align}
    E^\text{AM}(\rvx) = -\beta^{-1} \log\bigg[\sum\limits_{\mu=1}^K \exp\Big(- \beta \lVert \rvx - \boldsymbol{\xi}^\mu \rVert^2_2\Big) \bigg].
    \label{eq:DenseAM}
\end{align}
A direct comparison of Eqs.~(\ref{eq:energy-diffusion}) and (\ref{eq:DenseAM}) highlights a close connection between these two seemingly different frameworks. Specifically, the data points $\boldsymbol{\xi}^\mu$ in the DM framework play the role of memories in the AM formulation. The variance of the noise added during the forward process (\ref{eqn:bg-forward-sde}) plays the role of the effective temperature $\beta^{-1}$ (which controls the separation of memories), while the reverse process (\ref{eqn:bg-reverse-sde}) corresponds to memory retrieval dynamics \citep{hoover2023memory, ambrogioni2023search}. 

For a single data point $\boldsymbol{\xi}^1$ or $K = 1$, Eq. (\ref{eq:DenseAM}) has a single local minimum at that stored pattern, where the shape of this energy landscape is independent of the value of $\beta$. This is an example of a memorized state. In contrast, for $K = 2$, there are several possibilities. The energy landscape can have two local minima when $\beta\rightarrow\infty$ is large. But, for finite values of $\beta$, there exist configurations of patterns such that Eq. (\ref{eq:DenseAM}) has only one local minimum:
\begin{equation}
    \boldsymbol{\eta} = \argmin_{\rvx} E^\text{AM}(\rvx), 
\end{equation}
such that $\boldsymbol{\eta}\neq\boldsymbol{\xi}^1$ and $\boldsymbol{\eta}\neq\boldsymbol{\xi}^2$. This is an example of a \textit{spurious state}. Such states are ubiquitous in AM models, when the amount of training data (memories) crosses the critical memory storage capacity of the model (which depends on the correlations between the training data points). Thus, the following question can be posed: \textit{Is it possible to observe spurious states in DMs trained using score matching objective when the energy landscape is modeled by the score-predicting neural network?} 

\section{Detecting Memorized, Spurious, and Generalized Samples}
\label{sec:ch3-detection}
 
The main goal of our work is to establish the existence of spurious states in DMs trained on natural high-dimensional datasets. This section explains the design of detection metrics that can classify any generated sample into one of three categories -- \textit{memorized}, \textit{spurious}, and \textit{generalized} samples -- to aid in our analyses. Fundamentally, these metrics defined below rely on two datasets: the training dataset $\mathsf{S}$ used to train the DM, and the synthetic dataset $\mathsf{S}'$ generated from the fully trained DM. The size of $\mathsf{S}'$ is assumed to be much bigger than $\mathsf{S}$. 

Our \textbf{core intuition} is rooted in the energy landscape, where its topography dictates the nature of the generated samples, as shown in Fig.~(\cfig{fig:general-energy-transition}{A}). Memorized samples originate from the basins of attraction the model creates around individual data points from the training set $\mathsf{S}$. Because these samples are attractors, the generation process frequently converges to exact memorized samples at small data sizes, producing outputs in the synthetic set $\mathsf{S}'$ that are perfect duplicates of training data. In contrast, spurious samples arise from new emergent basins of attraction, formed during training, which do not correspond to any sample in $\mathsf{S}$. These emergent minima are also stable attractors, which explains why they appear as duplicates within $\mathsf{S}'$ but are absent from $\mathsf{S}$. Finally, generalized samples are drawn from flatter regions of the energy landscape. Since these areas support a continuum of low-energy states rather than a single sharp minimum, the probability of landing on the exact same point twice is negligible, resulting in unique novel creations.

\begin{figure}[t]
    \centering
    \includegraphics[keepaspectratio, width=1.0\textwidth, height=0.5\textheight]{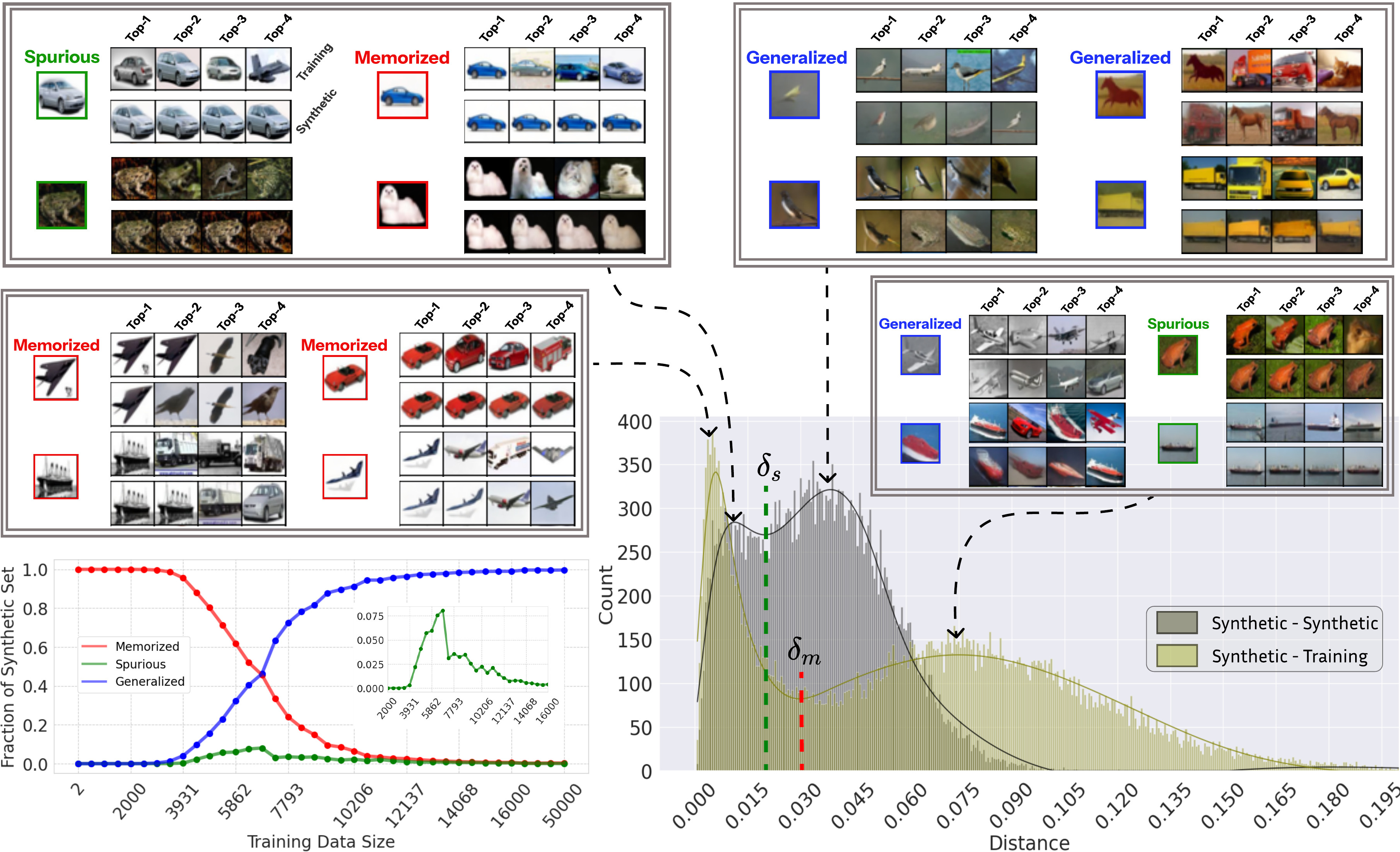}
    \caption{Different sample types identified in accordance to the bi-modality in distance histograms for CIFAR-10 \citep{cifar10}. The {grey histogram} shows the distances between synthetic samples and their nearest neighbors from the synthetic set $\mathsf{S}'$ (which is $4$x the size of the training set). The threshold $\delta_s$ defines a boundary between the two peaks. The {olive histogram} depicts the distances from the synthetic samples to their closest neighbor from the training set $\mathsf{S}$, with threshold $\delta_m$ separating the two peaks. {Memorized} samples are located in the left peak of the olive histogram, below $\delta_m$. In contrast, {generalized} and {spurious} samples appear to the right of $\delta_m$ in the olive histogram. Examples of the generated samples forming each of the four peaks of the histograms are shown in the inset frames. For each generated sample, top-4 nearest neighbors from the training set are shown in the top row, and those from the synthetic set are shown in the bottom row. Training set size $K=7310$ is used in this figure, but the discussed phenomena are general and largely independent of this specific value. The fraction of the memorized, spurious, and generalized samples in the pool of all generated samples is shown in the bottom left panel as a function of the training set size. The inset shows amplified spurious fraction ({green curve}). 
    }
    \label{fig:cifar10-hist}
\end{figure}

With this intuition in mind we examine two histograms, see Fig.~(\ref{fig:cifar10-hist}). First, for every element of  $\mathsf{S}'$, their distance to their closest nearest neighbor from $\mathsf{S}$ is shown by the {olive histogram}. Second, for every element of $\mathsf{S'}$, their distance to their closest nearest neighbor in $\mathsf{S}'$ (excluding itself) is shown by  the {gray histogram}. Both histograms exhibit a clear bimodal shape, indicating that at least \textit{two groups} of samples are present in $\mathsf{S}'$. This bimodality makes it possible to select the memorized  $\delta_m$ and spurious $\delta_s$ thresholds to separate the samples forming the two peaks in each histogram. The left peak of the olive histogram identifies memorized samples by their small distance to the training set $\mathsf{S}$, leaving spurious and generalized samples in the right peak. The gray histogram distinguishes memorized and spurious samples from generalized samples: its left peak contains high-frequency duplicates (both memorized and spurious), while its right peak consists of unique generalized samples. This bimodality provides an empirical basis for defining our detection metrics below.

Following \cite{yoon2023diffusion}, we define the memorization detection metric $\mathcal{M}$. A sample $\hat{\rvx} \in \mathsf{S}'$ is considered memorized if its distance to its nearest neighbor $\rvx_1$, from the training set $\mathsf{S}$, falls below a threshold $\delta_m \in \mathbb{R}$:
\begin{equation}
    \mathcal{M}(\hat{\rvx}, \mathsf{S}) = \mathbb{I} \bigg{(} d(\hat{\rvx}, \rvx_1) \leq \delta_m \bigg{)} , 
    \label{eqn:mem-detection}
\end{equation}
where $\mathbb{I}$ represents the indicator function and $\delta_m$ is a threshold derived from the olive distance histograms, see Fig.~(\ref{fig:cifar10-hist}). Second, a sample $\hat{\rvx} \in \mathsf{S}'$ is spurious if it is not a memorized state but is a duplicate of the synthetic set $\mathsf{S}'$. This is detected by the metric $\mathcal{S}$, which identifies non-memorized samples that are close to their respective nearest neighbor $\rvx'_1$ within the synthetic set $\mathsf{S}'$:
\begin{equation}
    \mathcal{S} (\hat{\rvx}, \mathsf{S}, \mathsf{S}') = \mathbb{I} \bigg{(} d(\hat{\rvx}, \rvx'_1) \leq \delta_s \bigg{)} \land \neg \mathcal{M}(\hat{\rvx}, \mathsf{S}), 
    \label{eqn:spur-detection}
\end{equation}
where $\delta_s \in \mathbb{R}$ is a threshold value chosen based on the gray distance histogram. Finally, a sample $\hat{\rvx}$ is generalized if it is neither memorized nor spurious:
\begin{equation}
    \mathcal{G}(\hat{\rvx}, \mathsf{S}, \mathsf{S}') = \neg \mathcal{M}(\hat{\rvx}, \mathsf{S}) \land \neg\mathcal{S}(\hat{\rvx}, \mathsf{S}, \mathsf{S}') . 
    \label{eqn:gen-detection}
\end{equation}

With these metrics, we computed the fractions of memorized, spurious, and generalized samples in the synthetic sets for various training data sizes for different datasets, see Fig.~(\ref{fig:transition-a}). These datasets include MNIST \citep{mnist}, FASHION-MNIST \citep{fmnist}, CIFAR-10 \citep{cifar10}, and down-sampled LSUN-CHURCH \citep{lsun}. For each dataset, we trained DDPM-based DMs \citep{ho2020denoising} for $M=38$ different data sizes. For each model $\alpha=1,...,M$ trained on a training set $\mathsf{S}_\alpha$, a synthetic set $\mathsf{S}'_\alpha$ was generated. To account for duplication, we ensured that each $\mathsf{S}'_\alpha$ is $4$ times the size of its corresponding $\mathsf{S}_\alpha$. We experimented with up to $8$x ratio to the training set size and did not notice qualitative differences in our conclusions, see \Cref{sec:transition-details}. Each sample $\hat{\rvx} \in \mathsf{S}'_\alpha$ is classified as either memorized, spurious, or generalized using Eqs.~(\ref{eqn:mem-detection})-(\ref{eqn:gen-detection}). The fractions of these three sample types with respect to $|\mathsf{S}'_\alpha|$ are plotted in Fig.~(\ref{fig:transition-a}). For each dataset, our smallest model was trained on the training set of $|\mathsf{S}_1|=2$ data points while the largest model was trained on the entire original training set $\mathsf{S}_M = \mathsf{S}$. For the MNIST and FASHION-MNIST datasets, we used $\ell_2$-distance, and the FID metric \citep{heusel2017gans} was used for the two remaining datasets. Notably, we did not observe significant differences between either using $\ell_2$-distance or FID metric for MNIST and FASHION-MNIST, and thus, we opted for a more efficient option ($\ell_2$-distance) for these two datasets. 

\FloatBarrier
\section{Results}
\label{sec:ch3-results}

\par\addvspace{1em}
\noindent \textbf{The Memorization-to-Generalization Transition}. We argue that the answer to this question is positive, and the spurious states indeed emerge, as DMs progress from memorization to generalization as the training data size $K$ increases. When $K$ is small, DMs initially memorize their training data points and store them as local minima of their energy function. In the large data regime, a different phase appears where a sufficient increase of $K$ fosters the creation of attractor states corresponding to the manifold of the generated samples. Spurious states appear at the boundary of this transition and correspond to emergent attractor states of the energy function, absent in the training set, but at the same time, having distinct basins of attraction around them. This sequence of transitions is illustrated in Fig.~(\cfig{fig:general-energy-transition}{A}), which provides a schematic illustration of the possible locations of memorized and spurious patterns on the energy landscape as the training set size is increased. 

In order to empirically establish this sequence of transitions, we have trained a set of conventional DMs on datasets of gradually increasing sizes. Each of the trained models is characterized by its training set $\mathsf{S}$, and a synthetic set of generated samples drawn from this model $\mathsf{S}'$ (the synthetic set is assumed to be very large). A sample $\hat{\rvx} \in \mathsf{S}'$ is memorized if it has duplicates in both training and synthetic sets. This is clear from the three universality classes shown in Fig.~(\cfig{fig:general-energy-transition}{A}). Since memorized samples have a basin of attraction around them, the same sample should appear in the synthetic set more than once, and that generated sample should be a copy of the training data point. A sample $\hat{\rvx} \in \mathsf{S}'$ is spurious if it has duplicates in the synthetic set, but not in the training set. Again, this is clear from Fig.~(\cfig{fig:general-energy-transition}{A}): spurious samples have a well-defined basin of attraction around them, but the bottom of that basin corresponds to a newly generated sample -- not the training example. Finally, the generalized sample $\hat{\rvx} \in \mathsf{S}'$ has no duplicates in either the training or synthetic sets. Generalized samples occupy flat manifolds in the configuration space. For this reason, every generation produces a slightly different sample resulting in absence of duplicates in $\mathsf{S}'$. These criteria correspond to the detection metrics defined in \Cref{sec:ch3-detection}.   

Examples of generated samples belonging to these three categories are shown in Fig.~(\cfig{fig:general-energy-transition}{B}). For each sample, the top-4 nearest neighbors from the training set (top row) and synthetic set (bottom row) are shown. It is instructive to visually examine some of the samples classified as spurious. For instance, the first spurious sample in Fig.~(\cfig{fig:general-energy-transition}{B}) (generated from the model trained on CIFAR-10 dataset \citep{cifar10}) is a plane that has a large wing (highlighted by the cyan frame), which is absent in the training set, but has an exact duplicate in the synthetic set. Another interesting example is shown in the fourth row (generated from the model trained on FASHION-MNIST dataset \citep{fmnist}). There are many examples of shirts in the training set with either two sleeves, or no sleeves. The generated sample shows only one sleeve. This sample has exact duplicates in the synthetic set, indicating the existence of a well-defined basin of attraction around it, but does not appear in the training set. These examples illustrate the general phenomenon, which is predicted by the AM perspective and is clearly observed empirically for conventional diffusion models trained with the score matching objective. 

\begin{figure}[t]
    \centering  
    \includegraphics[keepaspectratio, width=1.0\textwidth, height=0.5\textheight]{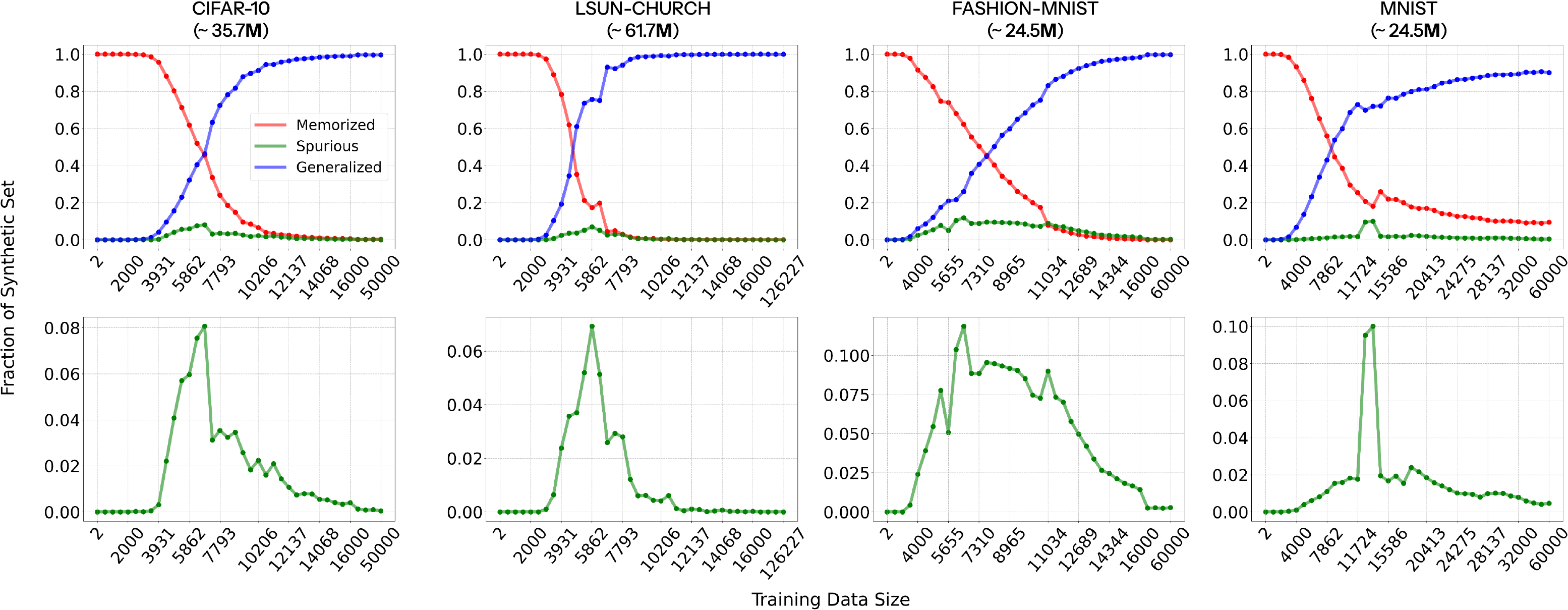}
    \caption{Fractions of {memorized}, {spurious}, and {generalized} samples in synthetic sets across training set sizes and various datasets. As the training data size $K$ increases, memorization decreases and the fraction of generalized samples steadily increases (top row). The fraction of spurious patterns rises and decreases at the boundary between the memorization and generalization phases (bottom row). The synthetic set size at each point is $4$x its respective training set size.
}
\label{fig:transition-a}
\end{figure}

The abundance of memorized, spurious, and generalized samples is quantified as a function of training data set size in Fig.~(\ref{fig:transition-a}). Here, a family of DM models has been trained for a gradually increasing size of the training set $K$. For small $K$, the generated samples are dominated by memorized data points. As the training set gets larger, the amount of memorized samples decreases, while the amount of generalized samples increases. The spurious states appear as a peak at the boundary of this transition. 

\par
\begin{figure}[t]
    \centering  
    \includegraphics[keepaspectratio, width=1.0\textwidth, height=0.5\textheight]{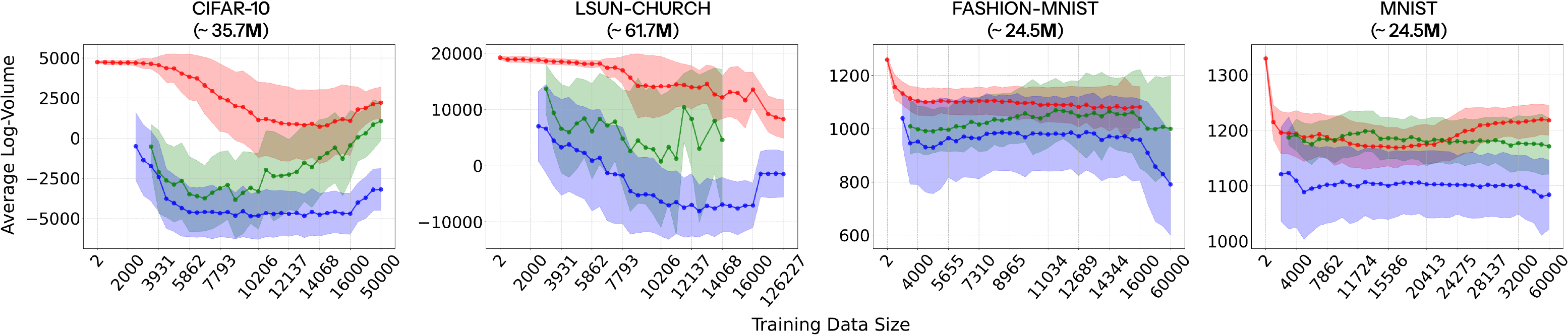}
    \caption{Volume of the basins of attraction for three distinct sample types and various datasets as the training set size increases. The average log-volume is computed using Eq.~(\ref{eqn:volume-hypersphere}). In general, memorized samples have the largest basin volume at the beginning of the transition. However, as the training set size increases, their basin volume decreases and approaches that of the spurious samples' volume. Statistically, memorized and spurious samples have larger basin volume than that of the generalized samples across all datasets. The shaded regions indicate the standard deviation of the log-volume.
}
\label{fig:transition-b}
\end{figure}

\addvspace{1em}
\noindent \textbf{Volume of the Basins of Attraction}. To characterize the volume of the basins of attraction, motivated by \cite{biroli2024dynamical}, \cite{raya2024spontaneous}, and \cite{li2024critical}, we measured the critical time $t_c$, which is the maximal time when the reverse process (\ref{eqn:bg-reverse-sde}) can still recover a target pattern $\hat{\rvx}$ from its perturbation $\rvx_{t_c}$ within an error margin. The time $t_c$ defines the radius $R(\hat{\rvx}, \rvx_{t_c}) = \lVert \hat{\rvx} - \rvx_{t_c} \rVert$ of the basin of the attraction of a sample $\hat{\rvx}$, which is the Euclidean norm of the difference between $\hat{\rvx}$ and its recoverable perturbation $\rvx_{t_c}$. The volume of this basin is simply defined by the volume of the $N$-dimensional sphere of that radius:
\begin{equation}
    V(\hat{\rvx}, \rvx_{t_c}) = \frac{\pi^\frac{N}{2}}{\Gamma(\frac{N}{2} + 1)} R(\hat{\rvx}, \rvx_{t_c})^N,
    \label{eqn:volume-hypersphere}
\end{equation}
where $\Gamma(\cdot)$ is the Gamma function. The results of the average logarithm of Eq.~(\ref{eqn:volume-hypersphere}) are shown in Fig.~(\ref{fig:transition-b}) for samples belonging to each distinct category.

The distinct nature of these three different states is further substantiated by the hierarchy of the log-volume of their basins of attraction in Fig.~(\ref{fig:transition-b}). Memorized samples consistently exhibit the largest log-volume, followed by spurious samples, while generalized samples have near-zero volume. This is in accord with the intuitive expectation for the energy profiles in Fig.~(\cfig{fig:general-energy-transition}{A}). Memorized and spurious samples should have substantial basins of attraction around them, while generalized samples should have almost vanishing basins, since they occupy an almost flat energy manifold. 

\par
\begin{figure}[t]
    \centering  
    \centering
    \includegraphics[keepaspectratio, width=1.0\textwidth, height=0.5\textheight]{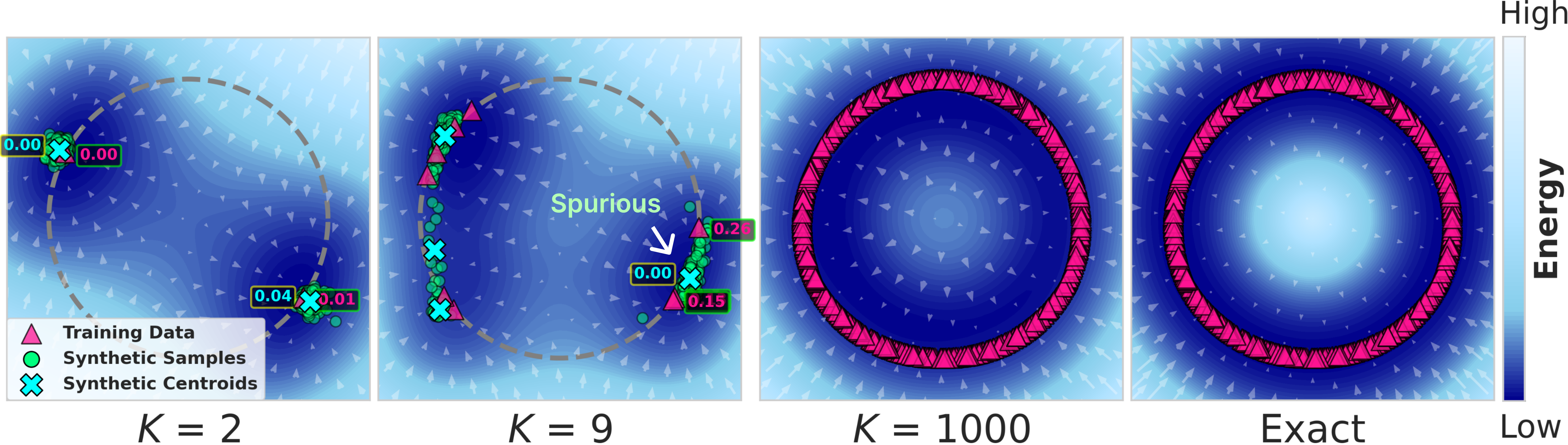} 
    \caption{~Energy landscape evolution for the 2D toy model as training data size $K$ increases. Models trained at $K \in \{2, 9, 1000\}$, using the VE-SDE based diffusion pipeline from \cite{song2021scorebased}, with training data (represented as pink triangles) sampled from the unit circle (shown in gray). Generated samples are shown alongside the learned score field or neural network, aligned with the negative gradient of the energy (\ref{eq:energy-diffusion}). Hierarchical clustering identifies structure within the generations, with the generated samples' centroids (represented as cyan crosses) are visualized by their numerical value of the energy at the location of that centroid. The energy values of the training points are also shown for memorized samples and data points located close to spurious samples. The right-most panel shows the exact solution as $K \rightarrow \infty$ derived in Eq.~(\ref{eqn:toy-model-energy}). As $K$ grows, the model initially memorizes individual data points, forming isolated basins. Around $K = 9$, spurious patterns, distinct low-energy attractors not present in the data, emerge and signal the onset of generalization. At large $K$, the model enters a fully generalized regime, where low-energy states lie on a flat continuous manifold. The visualization is shown at $t = 10^{-5}$.}
    \label{fig:toy-example}
\end{figure}

\addvspace{1em}
\noindent \textbf{Curvature of the Energy}. While the changing fractions of memorized, spurious, and generalized samples constitute a good signature of the memorization-to-generalization transition, the underlying mechanism is also rooted in the geometry of the model's learned energy landscape, as shown in Figs.~(\cfig{fig:general-energy-transition}{A}) and (\ref{fig:toy-example}). To elucidate this mechanism, we quantify the energy curvature at the minima corresponding to each sample type. 

\begin{figure}[t]
    \centering
    \includegraphics[keepaspectratio, width=1.0\textwidth, height=0.5\textheight]{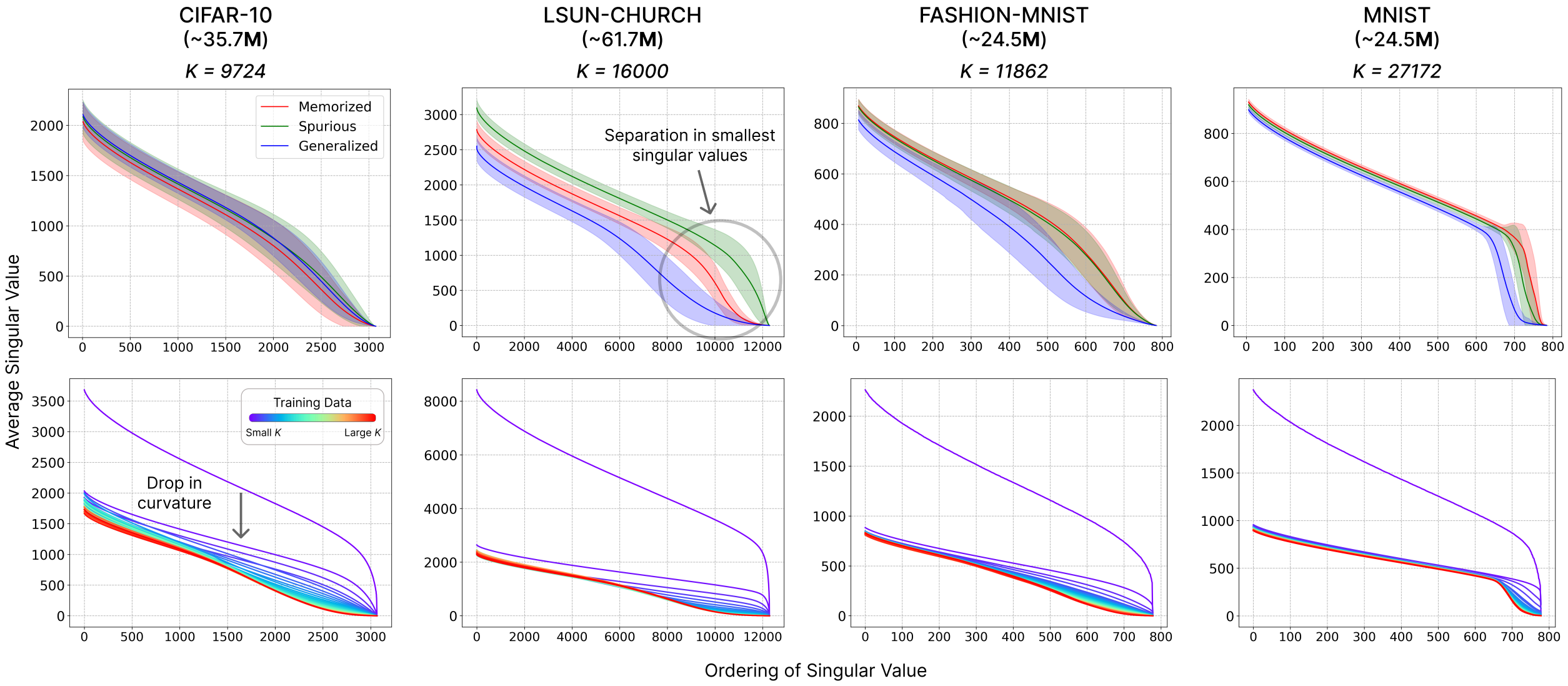}
    \caption{Spectral analysis of the energy curvature for different sample types and various datasets.
    The shaded region is the standard deviation of the singular values. The average singular values computed for the three distinct sample types are shown in the top row. Memorized and spurious samples generally exhibit higher energy curvature, characterized by fewer near-zero singular values than generalized samples. The average singular values computed for the training data points are shown in the bottom row, where the changing color scheme denotes {small} to large data sizes, suggesting a drop in curvature (indicated by the decrease in singular values) as $K$ increases.
    }
    \label{fig:energy-curvature-a}
\end{figure}

To measure this curvature, we rely on the relationship $\nabla_{\rvx_t} \log p_t (\rvx_t) = - \nabla_{\rvx_t} E_t(\rvx_t)$ to analyze the Hessian of the energy function or the Jacobian of the score, evaluated at the location of each generated sample $\hat{\rvx} \in \mathbb{R}^N$. Specifically, following \cite{stanczuk2022your}, \cite{achilli2024losing}, and \cite{ventura2024manifolds}, we numerically approximate the spectrum of the energy Hessian by exploiting the tangent space of the model's learned manifold at a given pattern $\hat{\rvx}$. First, we generate a set of $4N$ perturbations around $\hat{\rvx}$ using a small time $t_0$ in Eq.~(\ref{eqn:bg-forward-sde}). The learned score is then evaluated at each perturbed point $\hat{\rvx}_{t_0}$ and the resulting $4N$ score vectors form a matrix $\mathsf{M} \in \mathbb{R}^{N \times 4N}$. Singular value decomposition is then performed on $\mathsf{M}$ to extract its singular values, serving as our measure of the local curvature of the energy around $\hat{\rvx}$. The curvature spectra for the three sample types for different datasets are shown in the top row of Fig.~(\ref{fig:energy-curvature-a}), see \Cref{sec:energy-curvature} for additional details.

The distinct universality classes of the energy behavior around generated samples, illustrated in Fig.~(\cfig{fig:general-energy-transition}{A}), suggests that the three classes of generated samples should have different behaviors of the curvature of the energy around them. The curvature of the energy profile is characterized the Hessian of the energy (or equivalently the Jacobian of the score function). We have performed the singular value decomposition on the resulting estimated Jacobian matrix, see details below. The ranked singular values are shown in Fig.~(\ref{fig:energy-curvature-a}). This analysis reveals that the curvature of the energy is largely consistent with the intuitive expectations from Fig.~(\cfig{fig:general-energy-transition}{A}). For instance, for MNIST and LSUN-CHURCH datasets there is a clear separation of spectral curves for memorized, spurious, and generalized samples in the region of smallest singular values. These small singular values correspond to flat directions in the ambient space, which do not result in energy increase around corresponding samples. Additionally, we performed a similar analysis for training data. The results reveal that as the training set size increases, the curvature around these training samples on average decreases. All these results are consistent with intuitive expectation predicted by Fig.~(\cfig{fig:general-energy-transition}{A}): as the training data size $K$ increases, the memorized and spurious samples tend to have higher curvature (indicated by their fewer near-zero singular values) compared to the generalized samples. 

While a consistent separation in the curvature spectra of memorized, spurious, and generalized samples is not readily apparent for all training data sizes $K$, likely due to the geometric complexity of the energy landscapes near transitions, we were able to observe the separation of the curvature spectra between memorized, spurious, and generalized samples at many sizes of the training dataset (prominently at or after the peak of the spurious states). In addition, we repeated a similar analysis for training data points. There is a consistent decrease in the singular values measured at the locations of the training data points as their number $K$ increases, see bottom row of Fig.~(\ref{fig:energy-curvature-a}). This provides a direct geometric evidence of the mechanism behind generalization: the model transitions from creating sharp, high-curvature minima for individual data points in the memorization regime to learning a smoother, lower-curvature manifold that encompasses the entire dataset in the generalization regime. 

\par\addvspace{1em}
\noindent \textbf{A Two-Dimensional Toy Model}. Natural datasets considered above are high-dimensional, which makes it difficult to directly visualize the energy behavior around generated samples. To address this issue, we designed a very simple synthetic model that makes it possible to evaluate the energy landscape of DM directly. 

Imagine that the training data points live in two dimensions and lie on a unit circle:
\begin{equation}
    y_1^2 + y_2^2 =1 \label{eq: 2Dcircle}
\end{equation}
where $y_1$ and $y_2$ are coordinates in this 2D space. We are interested in exploring how the shape of $E^\text{DM}$ (\ref{eq:energy-diffusion}) changes as the number of training data points $K$ increases. 
To explore this aspect, we trained a set of DMs in this precise setting. Each model is trained on a dataset of size $K$. The resulting energy landscapes are shown in Fig.~(\ref{fig:toy-example}). For a small data size of $K = 2$, the DM exhibits memorization behavior -- the local minima of the energy correspond to the training data points. At $K=9$ we observed the first signs of \textit{spurious states}. At this stage, the model begins to learn emergent (different from the training data) local minima of the energy. Subsequent increase of the size of the training set leads to fully generalized behavior, illustrated at $K=1000$. At this stage, all of the samples from the model live in close proximity of the exact data manifold. The right panel shows the analytical expression of the energy landscape defined by Eq.~(\ref{eqn:toy-model-energy}), which can be obtained in the limit $K\rightarrow\infty$.

Below, we derive the exact form of the energy landscape $E^\text{AM}$ (\ref{eq:DenseAM}) in the limit when the number of training data is infinite or $K \rightarrow \infty$. In this case, the distribution of data points can be described by a continuous density of states: 
\begin{equation}
    p(\rvy) = \frac{1}{\pi} \delta\big(y_1^2+y_2^2-1\big),
    \label{2D density}
\end{equation}
where $\delta(\cdot)$ is the Dirac delta function. The probability of the generated data is proportional (up to terms independent of the state $\rvx$) to 
\begin{equation}
    p(\rvx) \sim \int\limits_{-\infty}^{+\infty} dy_1 dy_2\ p(\rvy)\ e^{- \beta \lVert \rvx - \rvy \rVert^2_2} 
    \label{eq: continous circle}
\end{equation}
In order to evaluate this integral, it is easiest to introduce polar coordinates for both the state vector $\rvx$ and the training data $\rvy$:
\begin{equation}
\begin{cases}
    x_1 = R \cos(\phi)\\
    x_2 = R \sin(\phi)
\end{cases} \ \ \ \ \ \ \ \ \ \ \begin{cases}
    y_1 = r \cos(\varphi)\\
    y_2 = r \sin(\varphi)
\end{cases}
\label{eq:polar-coords}
\end{equation}
The integral (\ref{eq: continous circle}) can then be written as
\begin{equation}
p(\rvx) \sim \int\limits_{0}^{2\pi} d\varphi\int\limits_0^\infty r dr \frac{1}{\pi} \delta(r^2-1) e^{-\beta[R^2+r^2-2Rr\cos(\varphi-\phi)]}
= e^{-\beta (R^2+1)} I_0(2\beta R)
\label{eq:circle-integral}
\end{equation}
and explicitly computed using the definition of the modified Bessel functions \citep{gradshteyn2014table}.

Using the definition of $E^\text{AM}$, see Eq. (\ref{eq:DenseAM}), the energy of this toy model can be written as 
\begin{equation}
\begin{split}
     E^\text{AM}(R, \phi) & =  R^2+1 - \frac{1}{\beta} \log\big[I_0(2\beta R)\big] \underset{\beta\rightarrow\infty}{\approx} (R - 1)^2,
\end{split}
    \label{eqn:toy-model-energy}
\end{equation}
In this formula, $R$ is the radius of the unit circle, $\phi$ is the polar angle, and $I_0(\cdot)$ is a modified Bessel function of the first kind. The dependence on $\phi$ in Eq.~(\ref{eqn:toy-model-energy}) completely disappears from the final result, since the local minima of the resulting energy function form a {\em flat manifold}. In the limit of inverse temperature $\beta \rightarrow \infty$, the shape of the energy landscape is described by a parabola centered around $R=1$. The limit of $K \rightarrow \infty$ describes the fully generalized phase of this toy model.

The results presented in Fig.~(\ref{fig:toy-example}) perfectly align with the intuition suggested by possible energy behaviors around generated samples illustrated in Fig.~(\cfig{fig:general-energy-transition}{A}). At small $K$ the model memorizes the training data. Around $K=9$ the model begins to form spurious states, i.e., the generated samples have lower energy than the energy of the nearby training data points. As $K$ continues to increase, the model learns a flat energy landscape around the manifold of the data. That learned energy landscape $E^\text{DM}$ (\ref{eq:energy-diffusion}) is very similar to the empirical energy $E^\text{AM}$ of DenseAM (\ref{eq:DenseAM}) in the limits of large $K$ in the regions close to the data manifold (the two energies may deviate from each other outside the manifold of the data, since there are no samples presented to DM outside the data manifold). Thus, the results of this simple two-dimensional model support the claims of the empirical evaluations on the natural datasets. 

\par
\begin{figure}[t]
    \centering
    \includegraphics[keepaspectratio, width=1.0\textwidth, height=0.5\textheight]{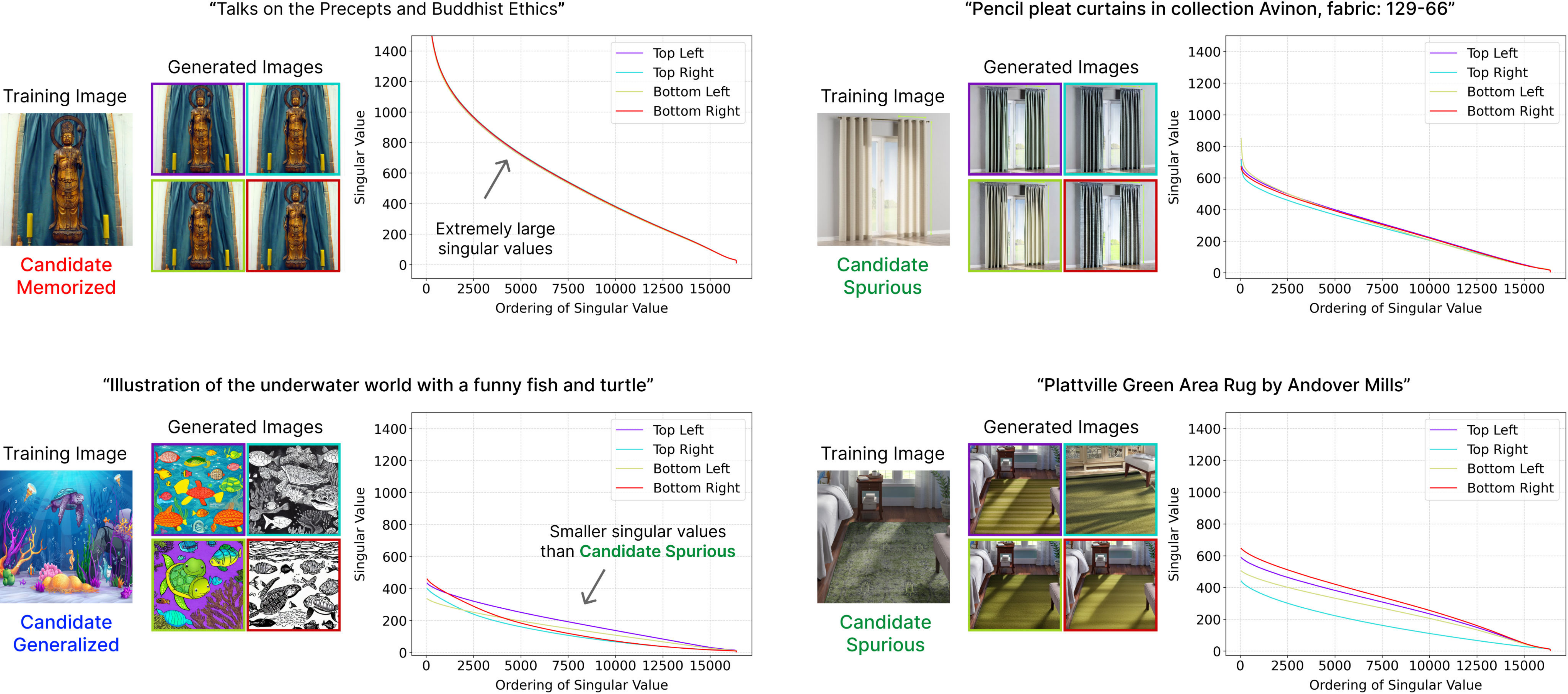}
    \caption{Spectral analysis of energy curvature for different candidate examples of the three distinct sample types in Stable Diffusion model trained on LAION \citep{schuhmann2022laion}. Here, each candidate exhibits different spectra signatures. The candidate memorized sample has much larger singular values, while the candidate generalized sample has much smaller ones. The selected and possibly spurious candidates have larger singular values than the candidate generalized sample, but smaller than that of the candidate memorized sample. The common trait between candidate memorized and spurious samples is that both are stable attractors, as demonstrated by their repeated generations given 4 different initial points (or noise vectors) and the associated text-prompt. The y-axis is clipped at the value of 1500 to better contrast the shown examples' spectra.}
    \label{fig:energy-curvature-b}
\end{figure}

\addvspace{1em}
\noindent \textbf{Stable Diffusion}. So far our studies focused on relatively small scale datasets: MNIST \citep{mnist}, FASHION-MNIST \citep{fmnist}, CIFAR-10 \citep{cifar10}, and LSUN-CHURCH \citep{lsun}. Focusing on these settings allowed us to train a dense grid of models with varying size of the training set and carefully examine the existence of duplicates in the training and synthetic sets. Such a careful analysis would be intractable for large-scale real world scenarios such as Stable Diffusion \citep{rombach2022high}: training these models requires substantial compute resources,  the training set is large, and the synthetic set needs to be even larger in order to reliably perform the analysis similar to what was described above. At the same time, the analysis based on the curvature of the energy landscape around generated samples is more feasible (see Fig.~\ref{fig:energy-curvature-b}). 

With this in mind, we have identified several candidate samples from Stable Diffusion that fall into three categories: sharp energy curvature, intermediate energy curvature, and almost flat energy. Examples of these samples together with their singular values of the score function Jacobian are shown in Fig.~(\ref{fig:energy-curvature-b}). For each sample a language prompt was given to the model and the generated samples produced by that model was recorded. The prompt used were taken from the training set and the generated samples are shown together with the training image corresponding to that prompt. The results of this analysis reveal that the high curvature samples (candidate memorized states) are exact duplicates of the training image. Intermediate curvature samples (candidate spurious states) are somewhat similar to the training image, but deviate from it in certain aspects (color and pattern of the rug, color of the curtains). Finally, small curvature samples (candidate generalized states) demonstrate a greater degree of variability and larger deviations from the training image. Given the computational complexity of large models, this analysis has only been performed at small scale (small number of prompts and generated samples). Qualitatively, these results align with our primary findings on the carefully studied small dataset models.  

To validate these geometric findings on a large-scale model, we performed a case study using Stable Diffusion \citep{rombach2022high}, trained on the LAION dataset \citep{schuhmann2022laion}, shown in Fig.~(\ref{fig:energy-curvature-b}). The analysis of handpicked candidate examples of memorized, spurious, and generalized samples from LAION reveals a similar hierarchy of energy curvatures. As expected, heavily memorized samples occupy high-curvature minima, evidenced by large singular values. In contrast, generalized samples exhibit a significantly lower curvature signature, while spurious samples occupy an intermediate range of curvatures. Notably, the spectra for all generated samples remain relatively sharp, with few near-zero singular values, as similarly found in \cite{jeon2024understanding}. 

\FloatBarrier
\section{Discussion}
\label{sec:ch3-discussion}
In conclusion, we established a novel connection between DMs and DenseAMs by identifying the emergence of creativity (generalization) in DMs with the failure of successful memory recalls via an energy-based Dense Associative Memory. Using this perspective, we empirically identified a novel intermediate phase, {\em spurious states}, previously overlooked in related DM literature, which marks the onset of generalization in these models as the training data size increases.  Additional analyses of the volume of the basins of attraction on samples generated from DMs, helps validate our hypothesis delineated in Fig. (\cfig{fig:general-energy-transition}{A}). Our findings in Stable Diffusion suggest that similar phenomena persist even in large models trained on massive datasets. Thus, our framework provides a new tool for understanding the computational mechanisms of DMs, opening new avenues for characterizing their behaviors.

The main result of our work is the empirical demonstration that DMs, trained with standard methods, generate spurious samples at the onset of memorization-to-generalization transition. The existence of these samples is a cornerstone of energy-based AM models. By casting DMs' generation step into DenseAM setting, one can theoretically expect the emergence of these states at the transition boundary. We have characterized computational properties of memorized, spurious, and generalized samples by studying their abundance, volume of the basins of attraction, and curvature of the energy landscape as a function of the DM training set size. Spurious samples in DMs have well-defined basins of attraction around them and have not been discussed in prior literature.  

Our theoretical arguments pertain to the idealistic setting in which the neural network $s_\theta (\rvx_t, t)$ is capable of learning the empirical distribution obtained by adding noise to the training data. This limiting case is often adopted in theoretical studies of DMs, see \cite{biroli2024dynamical}, \cite{biroli2023generative}, \cite{achilli2025memorization}, and \cite{vastola2025generalization}. From a practical perspective, this is achieved if $s_\theta (\rvx_t, t)$ is over-parametrized and trained for a long time, and the generation process is terminated at a small but non-zero time $t_0$, see Eq. (\ref{eq:energy-diffusion}). This simplification allows us to make a theoretical prediction that DMs must generate spurious states. In practical settings, the size of the score modeling neural network is finite. Additionally, the score function itself is modeled by a general feed-forward neural network and, for this reason, is not guaranteed to represent the gradient of a scalar energy function. It is non-trivial that the theoretical prediction obtained in the regime of an infinitely large neural network can be extrapolated to the regime in which the size of $s_\theta (\rvx_t, t)$ is finite and the amount of training data is large. A possible perspective on our contribution is the empirical demonstration that this extrapolation is valid, and spurious states indeed exist in practically used settings. 

In the AM literature, specifically on Hopfield networks, the concept of spurious states is often entangled with the phenomenon of mixture states, which are linear combinations of a certain number of memory vectors. While in certain kinds of AM models these two notions can be closely related, it is important to emphasize that they are different concepts. Spurious states are defined by the existence of a basin of attraction around them. They may or may not be mixture states. The main message of our work is to establish that spurious states exist in DMs. We do not claim that these spurious states are necessarily mixture states. It is up to the model, defined by the score modeling neural network, to decide how the spurious states are constructed from the training data. 

Careful characterization of the memorization-generalization transition and the computational aspects of spurious states that define the onset of this transition may offer future insights for guiding the development of methods that can mitigate DMs' memorization. For instance, one can imagine diagnostic tools that monitor the prevalence of spurious states among generated samples (for example by leveraging spectral analysis method, see Fig. \ref{fig:energy-curvature-a}) during training process. If the spurious states become abundant, one could stop training or inject additional data into the training set at the onset of this regime. Thus, emergence of spurious states can serve as an early warning signal of potential memorization. The benefit of using this diagnostic method is that it does not require searching for the memorized samples in the training set, which can be unpractical due to its large size. One can also imagine incorporating regularizers into the architecture of the score-predicting neural network or the loss function that could penalize high curvature of the energy landscape around generated samples. These are some of the examples of promising directions of future work that may utilize the properties of spurious states for monitoring and reducing effects of memorization in practical large scale models. 

Additionally, recent works underscore training time as a central factor in balancing generalization against memorization. For instance, \cite{bonnaire2025diffusion} identifies two distinct training timescales: a period of high-quality generation (independent of the number of data points) and a time where memorization scales linearly with the training dataset size. Furthermore, while increased model capacity typically enhances memorization, as shown by other works \cite{yoon2023diffusion} and \cite{li2024generalization}, in contrast \cite{favero2025bigger} suggests that larger architectures may paradoxically shorten the optimal generalization window, making a shorten training time (early stopping) a viable strategy for alleviating memorization. Finally, \cite{kamb2024analytic} provides a structural basis for this emergent creativity through a `patch-mixing' mechanism, where the model composes novel images by recombining localized features from the training samples. 

Overall, these perspectives complement our interpretation of DMs through the lens of AM. While this work does not focus on aspects of training time and early stopping, these recent studies also suggest that the memorization-to-generalization transition is dependent on the model capacity, and most importantly, the training dataset size -- both attributes explored in our work. Thus, it leaves the exploration of the training time effects on the emergence of spurious states in DMs for future works. Moreover, the `patch-mixing' mechanism offers a good explanation for how the model constructs the emergent attractor states we observe, where novel patterns are formed by recombining features of stored memories into stable, non-memorized energy minima. By confirming the existence and distinct computational properties of spurious states in DMs, we help highlight the complexity of the memorization and generalization problems, positioning these emergent attractors not as detrimental artifacts of AM, but as a distinct computation state that possess a basin of attraction (unlike generalized states) akin to memorized states. 
%

\FloatBarrier
\input{chapters/ch3_continuous/ch3_supp}

\clearpage

%% file: chapters/ch3_continuous/ch3_supp.tex
\section{Supplementary Material}
\label{sec:ch3-supp}
This section collects the experimental details and additional results that support the findings of this chapter.\footnote{Code is available at \mbox{\url{https://github.com/Lemon-cmd/Diffusion-Models-and-Associative-Memory}}.}

\subsection{The Two-Dimensional Toy Model}
\label{sec:toy-model-energy-details}
\label{sec:toy-model-experiment}
\par\addvspace{1em}
\noindent \textbf{Dataset}. To construct the dataset, we uniformly sampled points on the unit circle. Concretely, we had 
$\rvy \sim p(\rvy) = \frac{1}{2\pi}\delta(r-1)$, where $r = \sqrt{y_1^2 + y_2^2}$ is the radius and $\delta(r - 1)$ is the delta function ensures that all probability mass lies on radius $r = 1$. The factor $\frac{1}{2\pi}$ guarantees that the angles are sampled uniformly, relying on the identity:
\begin{equation}
    \delta(\sqrt{y_1^2 + y_2^2}-1) = 2\delta(y_1^2 + y_2^2-1),
\end{equation}
where $p(\rvy) = \frac{1}{\pi} \delta(y^2_1 + y_2^2 - 1)$. In practice, we sampled polar coordinates $(r, \varphi)$ with $r = 1$ and uniformly sampled the angular coordinate $\varphi \in [0, 2\pi]$. We then converted $(r, \varphi)$ to Cartesian coordinates ($y_1, y_2$) using $y_1 =r \cos \varphi$ and $y_2=r \sin \varphi$.

Moreover, we created a data set of $60000$ points using a fixed random seed, and then built smaller subsets by progressively adding distinct samples without replacement, ensuring that each subset is a strict subset of the next. This approach allowed us to systematically examine how the model behaves as we vary the size of the training set.

\par\addvspace{1em}
\noindent \textbf{Exact Score}. Following Eq.~(\ref{eqn:toy-model-energy}), the energy can be written in closed form, whose local minima form the continuous data manifold at $R = 1$\footnote{Throughout this chapter, we use $r$ for the polar radius of the training data and $R$ for the polar radius of the generated samples.}. Furthermore, in the case of many data points or $K \rightarrow \infty$, we have $p(\rvy) = \frac{1}{\pi}\delta(y_1^2 + y_2^2-1)$. Since $p(R) \propto \exp^{-E^\text{AM}(R)}$, the score can be described as 
\begin{equation}
    \begin{split}
        -\nabla_R \log p(R) &= \nabla_R E^\text{AM}(R) \\
        & = 2R - \frac{2\beta I_1(2\beta R)}{\beta I_0(2\beta R)} \\
        &=  2R - 2\frac{I_1(2\beta R)}{I_0(2\beta R)} ,
    \end{split}
\end{equation}
where $E^\text{AM}(R)$ is Eq. (\ref{eqn:toy-model-energy}), and $I_1$ and $I_0$ are the first-order and zero-order modified Bessel functions of the first kind, respectively. With further expansion, the score becomes
\begin{equation}
    \nabla_R \log p(R) = 2 \left(\frac{I_1(2\beta R)}{I_0(2\beta R)} - R \right) \quad \Rightarrow \quad \nabla_x \log p(x) =   2 \left(\frac{I_1(2\beta R)}{I_0(2\beta R)} - R \right) \frac{x}{R} .
    \label{eq:toy-model-score}
\end{equation}
We implemented Eq. (\ref{eqn:toy-model-energy}) on a $20 \times 20$  grid to obtain the exact energy and score of the model, overlaying 1000 samples from the training set. For visualization purposes, we set $\beta=20$, ensuring a well-defined energy landscape. 
Additionally, we normalized the energy by subtracting its minimum value to ensure that the lowest energy point remains at zero.

To align with the theoretical framework, we employed the variance-exploding (VE) SDE of the form:
\begin{equation}
    \mathrm{d}\rvx_t = \sigma \,\mathrm{d}\mathbf{w}_t
    \label{eq:toymodel-sde}
\end{equation}
where the diffusion coefficient $g(t) = \sigma$. Assuming $t\in (0, 1]$, the variance of the diffusion kernel is given by $ \int_0^t g^2(s) \,\mathrm{d}s = \sigma^2 t$, enabling us to construct the corresponding Gaussian kernel: 
\begin{equation}
    p(\rvx_t | \rvy) = \mathcal{N}(\rvx_t; \rvy, \sigma^2 t \boldsymbol{I})
    \label{eq:kernel-toyexample}
\end{equation}
However, in practice we used $t \in [\epsilon, 1]$ with $\epsilon = 10^{-5}$ for numerical stability. The generative dynamics are given by   
\begin{equation}
    \mathrm{d}\mathbf{x}_{t} = \big[-\sigma^2 \nabla_{\mathbf{x}_{t}} \log p_t(\mathbf{x}_{t})\big] \mathrm{d}t + \sigma^2\,\mathrm{d}\mathbf{w}_{t}.
    \label{eq:toymodel-backward-sde}
\end{equation}
where this runs backwards in time to match the description of \cite{song2021scorebased}. The deterministic Probability Flow ODE is
\begin{equation}
    \frac{\mathrm{d}\mathbf{x}_{t}}{\mathrm{d}t } = -\frac{1}{2}\sigma^2 \nabla_{\mathbf{x}_{t}} \log p_t(\mathbf{x}_{t})
    \label{eq:toymodel-pf-ode}
\end{equation}
which shares the same marginal distributions as the SDE (\ref{eq:toymodel-backward-sde}) and is useful for likelihood estimation.

\par\addvspace{1em}
\noindent \textbf{Empirical distribution}. We represented the data as an empirical distribution 
 \begin{equation}
      p(\mathbf{y})=\frac{1}{K}\sum_{\mu=1}^K \delta^{(N)}(\mathbf{y}-\boldsymbol{\xi}^\mu)
 \end{equation}
 where $N$ is the dimensionality of the data point $\rvy$. Since $\rvx_t$ is drawn from the forward process distribution $p(\rvx_t | \rvy)$ that is conditioned on the data point $\rvy$, which can be expressed as a Gaussian kernel $\mathcal{N}(\rvx_t; \mathbf{y}, \sigma^2 t \boldsymbol{I})$. We obtained the distribution $p(\mathbf{x}_t, t)$ as
\begin{equation}
    \begin{split}
    p(\mathbf{x}_t, t) &= \int p(\mathbf{x}_t \mid \mathbf{y}) \,p(\mathbf{y}) \,\mathrm{d}\mathbf{y} \\
    &=\frac{1}{K}\sum_{\mu=1}^K 
    \int \mathcal{N}\!\bigl(\mathbf{x}_t; \mathbf{y}, \sigma^2 t \boldsymbol{I} \bigr)\,
    \delta^{(N)}\!\bigl(\mathbf{y}-\boldsymbol{\xi}^\mu\bigr)\,\mathrm{d}\mathbf{y}\\
    & = \frac{1}{K}\sum_{\mu=1}^K
\mathcal{N}\!\bigl(
   \mathbf{x}_t;\,\boldsymbol{\xi}^\mu,\,\sigma^2 t\, \boldsymbol{I}
\bigr)\\
    &=\frac{1}{K}\sum_{\mu=1}^K \frac{1}{(2\pi\,\sigma^2t)^{\frac{N}{2}}}\exp\!\Bigl(
-\,\frac{\|\mathbf{x}_t - \boldsymbol{\xi}^\mu\|_2^2
    }{2\,\sigma^2 t}\Bigr) 
    \end{split}
\end{equation}
which is the marginal distribution underlying $E^\text{DM}$ in \Cref{eq:energy-diffusion}, when the variance of the Gaussian kernel is $\sigma^2 t$.

\par\addvspace{1em}
\noindent \textbf{Energy Computation}. Following \cite{song2021scorebased} we computed the log-likelihood $\log p_\theta(\rvx_0)$ given by a diffusion model, with the instantaneous change of variable formula \citep{NeuralODE}, where $\rvx_0$ are the generated samples from the model. Replacing Eq. (\ref{eq:toymodel-backward-sde}) into the log-likelihood equation, 
\begin{align}  
    \label{eqn:loglikelihood-dm}
    \log p_0(\rvx_0;\mathbf{\theta}) = \log p_T(\rvx_T) + \int_0^T \nabla \cdot \tilde{\rvf}(\rvx_t, t)  \mathrm{d}t
\end{align}
with $\tilde{\rvf}(\rvx_t, t) =-\frac{1}{2}\sigma^2 \nabla_{\mathbf{x}_{t}} \log p_t(\mathbf{x}_{t})$. The function $\tilde{\rvf}(\rvx_t, t)$ comes from the above Probability Flow ODE (\ref{eq:toymodel-pf-ode}), and $\nabla \cdot \tilde{\rvf}(\rvx_t, t)$ denotes the divergence of the function (or the trace of its Jacobian). 

To estimate the likelihood of the model, we integrated $\nabla \cdot \tilde{\rvf}(\rvx_t, t)$, from a small time $\eps=10^{-5}$ to $T=1$, using a numerical integrator and added the prior logarithmic likelihood to it following Eq. (\ref{eqn:loglikelihood-dm}). The divergence term $\nabla \cdot \tilde{\rvf}(\rvx_t, t)$ is computed using the Laplacian, which is computationally feasible for this toy example, instead of using the Hutchinson trace estimator \citep{Hutchinson1990} done in \cite{song2021scorebased}. We use the RK45 method \citep{DORMAND198019} implemented in the ODE solver \textit{scipy.integrate.solve\_ivp} from Scipy \citep{SciPy} to solve the above integral. 

To compute the energy, we used the equation of $E^\text{DM}$ in \Cref{eq:energy-diffusion}, which can be also computed using Eq. (\ref{eqn:loglikelihood-dm}). From the Boltzmann distribution $p(\rvx) = \frac{1}{Z}\exp\bigl[-\beta E(\mathbf{x})\bigl]$, we obtained $\log p(\rvx) \propto -\beta E(\mathbf{x})$. In our case, the inverse temperature is $\beta = \frac{1}{2 \sigma^2 t}$. The equation for our VE-SDE thus becomes 
\begin{equation}
    E^\text{DM}(\rvx_t, t) = -2 \sigma^2 t \log p(\rvx_t, t)  
\end{equation}
To visualize the energy landscape, we observed the energy at $t=0.15$ which corresponds to $\beta=3.3$.

\par\addvspace{1em}
\noindent \textbf{Model and Training Details}. For each training data size $K$, we used an MLP-based network to estimate the score $s_\theta(\rvx_t, t)$. The general architecture includes \textbf{(1)} a Fourier random feature (FRF) timestep embedding layer and a linear layer which projects the concatenation of FRF timestep embedding and input into a latent dimension of 256; \textbf{(2)} an encoder which consists of four non-convolutional residual blocks (using the same latent dimension) with Swish \citep{Swish} activation in between; \textbf{(3)} a decoder which consists of the same number of residual blocks; and \textbf{(4)} a linear block which projects the latent variable back to the 2D space.

 The network was trained in continuous time with $\sigma=1$ using the objective function from \cite{ho2020denoising} and \cite{song2021scorebased}. Optimization is performed using the Adam \citep{Adam} optimizer with a learning rate $lr = 10^{-4}$. The batch size is set to $\textbf{min}(K,500)$. All models were trained for $800000$ iterations with the maximum batch size of $500$. Meanwhile, we used Euler-Maruyama discretization method \citep{kloeden1992stochastic}, with 1000 discretized steps, to solve the reverse SDE (\ref{eq:toymodel-backward-sde}) to generate samples.

\par\addvspace{1em}
\noindent \textbf{Clustering of Generated Points}. To cluster the generated samples (as part of visualizing their concentrations), we used the \texttt{AgglomerativeClustering} algorithm from scikit-learn, a part of SciPy \citep{SciPy}.

\par\addvspace{1em}
\noindent \textbf{Log-Likelihood of Diffusion Models}.\label{sec:likelihood-dm-details} For concreteness, we provide the likelihood computation of Eq. (\ref{eqn:loglikelihood-dm}) within this section to highlight the formulation of Eq. (\ref{eqn:loglikelihood-dm}) done by \cite{song2021scorebased} and \cite{NeuralODE}.

\par\addvspace{1em}
\noindent \textbf{Probability Flow ODE}. Suppose we have the following forward process:
\begin{equation}
    \mathrm{d}\rvx_t = \mathbf{f}(\rvx_t, t) \mathrm{d}t + \mathbf{g}(\rvx_t, t) \mathrm{d} \mathbf{w}_t
\end{equation}
where $\mathbf{f}(\cdot, t) : \mathbb{R}^{N} \rightarrow \mathbb{R}^N$, $\mathbf{g}(\cdot, t): \mathbb{R}^{N} \rightarrow \mathbb{R}^{N \times N}$ and $N$ denotes the dimensionality of $\rvx_t$. Using the derivations done by \cite{song2021scorebased} in their Appendix D, the evolution of the marginal probability density $p_t(\rvx_t)$ is
\begin{equation}
    \frac{\partial p_t(\rvx_t)}{\partial t} = - \sum^N_{i = 1} \frac{\partial}{\partial x_i} \big [  f_i (\rvx_t, t) p_t (\rvx_t)\big ] + \frac{1}{2} \sum^N_{i = 1} \sum^N_{j = 1} \frac{\partial^2}{\partial x_i \partial x_j} \bigg [\sum^N_{k = 1} g_{ij} (\rvx_t, t) g_{jk} (\rvx_t, t) p_t(\rvx_t) \bigg] 
\end{equation}
which corresponds to the Fokker-Planck equation \citep{OksendalSDE}: 
\begin{equation}
\begin{split}
    \frac{\partial p_t(\rvx_t)}{\partial t} & = - \sum^N_{i = 1} \frac{\partial}{\partial x_i} \big [  f_i (\rvx_t, t) p_t (\rvx_t)\big ] \\
    & \quad \,\, + \frac{1}{2} \sum^N_{i = 1} \frac{\partial}{\partial x_i} \bigg [ \sum^N_{j = 1} \frac{\partial }{\partial x_j} \bigg [\sum^N_{k = 1} g_{ij} (\rvx_t, t) g_{jk} (\rvx_t, t) p_t(\rvx_t) \bigg] \bigg ] \\
    & = - \sum^N_{i = 1} \frac{\partial}{\partial x_i} \big [  f_i (\rvx_t, t) p_t (\rvx_t)\big ] \\
    & \quad \,\, + \frac{1}{2} \sum^N_{i = 1} \frac{\partial}{\partial x_i} \bigg [ p_t (\rvx_t) \, \nabla \cdot \big[\mathbf{g} (\rvx_t, t) \mathbf{g} (\rvx_t, t)^\top \big] \\
    & \qquad \qquad \qquad + p_t(\rvx_t) \big [ \mathbf{g} (\rvx_t, t) \mathbf{g} (\rvx_t, t)^\top \big ] \nabla_{\rvx_t} \log p_t(\rvx_t) \bigg] \\ 
    & = - \sum^N_{i = 1} \frac{\partial}{\partial x_i} \bigg \{  f_i (\rvx_t, t) p_t (\rvx_t) \\
    &\quad \,\, - \frac{1}{2} \bigg [ \nabla \cdot \big[\mathbf{g} (\rvx_t, t) \mathbf{g} (\rvx_t, t)^\top \big] + \big [ \mathbf{g} (\rvx_t, t) \mathbf{g} (\rvx_t, t)^\top \big ] \nabla_{\rvx_t} \log p_t(\rvx_t) \bigg ] p_t(\rvx_t) \bigg \} \\ 
    & = - \sum^N_{i = 1} \frac{\partial}{\partial x_i} \big [ \tilde{f}_i(\rvx_t, t) \big ] p_t(\rvx_t)
\end{split}    
\label{eqn:diffusion-fokker-placnk}
\end{equation}
where $\tilde{\rvf}(\rvx_t, t) = \rvf (\rvx_t, t) - \frac{1}{2} \nabla \cdot \big[\mathbf{g} (\rvx_t, t) \mathbf{g} (\rvx_t, t)^\top \big] - \frac{1}{2}  \big [ \mathbf{g} (\rvx_t, t) \mathbf{g} (\rvx_t, t)^\top \big ] \nabla_{\rvx_t} \log p_t(\rvx_t) $. With careful inspection of Eq. (\ref{eqn:diffusion-fokker-placnk}), it is equal to the Liouville equation if the diffusion term $\tilde{\mathbf{g}}(\rvx, t) = 0$ and essentially, it is the probability flow ODE where
\begin{equation}
\begin{split}
\mathrm{d} \rvx_t & = \tilde{\rvf}(\rvx_t, t)\mathrm{d} t +\underset{0}{\underbrace{\tilde{\mathbf{g}} (\rvx_t, t) \mathrm{d} \rvw_t}} \\ 
& = \bigg \{\rvf (\rvx_t, t) - \frac{1}{2} \nabla \cdot \big[\mathbf{g} (\rvx_t, t) \mathbf{g} (\rvx_t, t)^\top \big] -\frac{1}{2} \big [ \mathbf{g} (\rvx_t, t) \mathbf{g} (\rvx_t, t)^\top \big ] \nabla_{\rvx_t} \log p_t(\rvx_t) \bigg \} \mathrm{d} t
\end{split}
\label{eqn:general-probability-flow-ode}
\end{equation}
Using the ODE (\ref{eqn:general-probability-flow-ode}), we can derive an appropriate probability flow ODE from the forward process detailed in \Cref{sec:ch3-dm-am}. For example, we have the following equation
\begin{equation}
    \mathrm{d} \rvx_t = \underset{:= \, \tilde{\mathbf{f}}(\rvx_t, t)}{\underbrace{\bigg \{ \mathbf{f} (\rvx_t, t) - \frac{1}{2} g(t)^2 \nabla_{\rvx_t} \log p_t (\rvx_t)\bigg \}}} \mathrm{d} t
\end{equation}
for the toy model where $\mathbf{f} (\rvx_t, t) = 0$.

\par\addvspace{1em}
\noindent \textbf{Log-likelihood}. Furthermore, if we first take the logarithm of Eq. (\ref{eqn:diffusion-fokker-placnk}) where 
\begin{equation}
    \begin{split}
        \frac{\partial \log p_t(\rvx_t)}{\partial t} = \frac{1}{p_t(\rvx_t)} \frac{\partial p_t(\rvx_t)}{\partial t} = - \sum^N_{i = 1} \frac{\partial \tilde{f}_i(\rvx_t, t)}{\partial x_i} = \nabla \cdot \tilde{\rvf} (\rvx_t, t)
    \end{split}
\end{equation}
we can compute the log-likelihood of $p_0(\rvx_0)$ using the following equation
\begin{align} 
    \log p_0(\rvx_0) = \log p_T(\rvx_T) + \int_0^T \nabla \cdot \tilde{\rvf}(\rvx_t, t)  \mathrm{d}t
    \label{eqn:loglikelihood-dm-general}
\end{align}
where $\nabla \cdot \tilde{\rvf}(\rvx_t, t)$ is parameterized as $\nabla \cdot \tilde{\rvf}_\theta (\rvx_t, t)$ since $s_\theta(\rvx_t, t) = \nabla_{\rvx_t} \log p_t(\rvx_t; \theta)$. Moreover, in general the term $\nabla \cdot \tilde{\rvf}_\theta (\rvx_t, t)$ is computed via the Hutchinson trace estimator \citep{Hutchinson1990}, 
\begin{equation}
    \nabla \cdot \tilde{\rvf}_\theta (\rvx_t, t) = \mathbb{E}_{p(\epsilon)} \big [ \epsilon^\top \nabla_{\rvx} \tilde{\rvf}_\theta (\rvx_t, t) \epsilon\big] 
\end{equation}
where $\nabla_{\rvx} \tilde{\rvf}_\theta (\rvx_t, t)$ denotes the Jacobian of $\tilde{\rvf}_\theta (\rvx_t, t)$ and the random variable $\epsilon$ satisfies $\mathbb{E}_{p(\epsilon)} [\epsilon]= 0$ and $\text{Cov}_{p(\epsilon)} [\epsilon] = \boldsymbol{I}$. 

However, for the toy example, due to its low dimensionality we found that it is possible to not utilize the Hutchinson trace estimator at all, and instead opted to compute the trace of the Jacobian without using the noise estimators.

\subsection{Memorization-to-Generalization Transition}
\label{sec:transition-details}
\par\addvspace{1em}
\noindent \textbf{Selection of Points}. For the computation of \Cref{fig:transition-a} and \Cref{fig:transition-b}, we followed the experiment setup of \cite{yoon2023diffusion} and conducted an initial sparse search of the transition, starting at data size $500$ and doubling it all the way to the total data size $|\mathsf{S}|$, i.e., $K \in \{500, 1000, 2000, \dots, |\mathsf{S}|\}$. We then identified two transitional critical points, $A$ and $B$, see Table (\ref{tab:transitions}), to conduct a more thorough search to elucidate the memorization-to-generalization transition. 

Here, point $A$ indicates the initial drop in memorization while $B$ signals the plateauing of memorization. To capture the finer details of this transition, we performed a search of $30$ inclusive linearly spaced points, from point $A$ to point $B$. For regions outside of the transition, using linear spacing, we sampled $5$ points from $|\mathsf{S}| = 2$ to point $A$, and another $5$ points from $B$ to the total dataset size $|\mathsf{S}|$, inclusively. In other words, we trained a separate model for each selected data size and generated the corresponding evaluation and synthetic sets while keeping the same configuration for all models, see Table (\ref{tab:models}). Then, we computed the memorization, spurious, and generalization fractions using the detection metrics in \Cref{sec:ch3-detection}. The selection process for the values $\delta_m$ and $\delta_s$ is explained below.

\begin{table}[t]
\centering
\caption{A table showing the critical points $A$ and $B$ of the memorization-generalization transition for each dataset. We use 30 linearly spaced points between $A$ and $B$ to finely characterize the memorization-to-generalization transition.}
\label{tab:transitions}
\begin{tabular*}{\textwidth}{@{\extracolsep\fill}lccc}
\toprule%
& \multicolumn{3}{@{}c@{}}{Training Data Size} \\ \cmidrule{2-4}%
Dataset & Point $A$ & Point $B$ & Total \\
\midrule
MNIST & 4000 & 32000 & 60000 \\ 
CIFAR-10 & 2000 & 16000 & 50000 \\ 
LSUN-CHURCH & 2000 & 16000 & 126227 \\
FASHION-MNIST&  4000 & 16000 & 60000 \\ 
\bottomrule
\end{tabular*}

\vspace{1.5em}
\caption{Table displaying the range of the memorized and spurious threshold values in the computation of the transition plots in \Cref{fig:transition-a}. The full set of threshold values is provided alongside the code.}
\label{tab:thresholds}
\begin{tabular*}{\textwidth}{@{\extracolsep\fill}lccc}
\toprule%
& & \multicolumn{2}{c}{Range [min, max]} \\ 
\cmidrule{3-4}
Dataset & Initial Latent Dim. & $\delta_m$ & $\delta_s$ \\ 
\midrule
MNIST & 128 & $[2, 4.5]$ & $[2, 4]$ \\ 
FASHION-MNIST & 128 & $[2, 2.5]$ & $[2.5, 3]$ \\ 
\hline 
& 64 & $[0.01, 0.04]$ & $[0.015, 0.025]$ \\
CIFAR10 & 96 & $[0.015, 0.04]$ & $[0.015, 0.025]$ \\
& 128 & $[0.015, 0.035]$ & $[0.015, 0.025]$ \\
\hline 
& 64 & $[0.06, 0.15]$ & $[0.01, 0.1]$ \\
LSUN-CHURCH & 96 & $[0.1, 0.125]$ & $[0.05, 0.06]$ \\
& 128 & $[0.1, 0.12]$ & $[0.04, 0.06]$ \\
\bottomrule
\end{tabular*}

\vspace{1.5em}
\caption{A table displaying both model and training configurations for each dataset.}
\label{tab:models}
\resizebox{\textwidth}{!}{\begin{tabular}{lccccc}
\toprule%
& \multicolumn{5}{c}{Configurations} \\ 
\cmidrule{2-6}
Dataset & Initial Latent Dim. & Channel Multipliers & Num. of Param. & Batch Size & Training Iterations \\
\midrule
 MNIST & 128 & (1, 2, 2) & 24.5M & 128 & 400,000 \\ 
FASHION-MNIST & 128 & (1, 2, 2) & 24.5M & 128 & 400,000 \\ 
\hline 
& 64 & (1, 2, 2, 2) & 8.9M & 128 & 500,000 \\
CIFAR-10 & 96 & (1, 2, 2, 2) & 20.1M & 128 & 500,000 \\
& 128 & (1, 2, 2, 2) & 35.7M & 128 & 500,000 \\
\hline 
& 64 & (1, 1, 2, 2, 4, 4) & 27.4M & 64 & 800,000 \\
LSUN-CHURCH & 96 & (1, 1, 2, 2, 4, 4) & 61.7M & 64 & 800,000 \\
& 128 & (1, 1, 2, 2, 4, 4) & 109.7M & 64 & 800,000\\
\bottomrule
\end{tabular}}
\end{table}

\par\addvspace{1em}
\noindent \textbf{Selection of Distance Metric}. For high-dimensional datasets, CIFAR-10 \citep{cifar10} and LSUN-CHURCH \citep{lsun}, we utilized LPIPS \citep{lpips} with the AlexNet \citep{AlexNet} backbone, as the function $d(\cdot, \cdot)$ for both memorization and spurious detection metrics. We selected this approach since it is a commonly used perceptual metric that compares the similarity between two images based on their feature representations. Moreover, it has been shown to better align with human judgment of visual similarity, making it ideal for assessing the quality and diversity of generated samples in these high-dimensional image datasets \citep{lpips}. For simpler datasets like MNIST \citep{mnist} and FASHION-MNIST \citep{fmnist}, where images are single-channel and less complex, we found that $L_2$-distance suffices for both memorization and spurious detection metrics.

\par
\begin{figure}[t]
    \centering
    \includegraphics[keepaspectratio, width=1\textwidth, height=0.7\textheight]{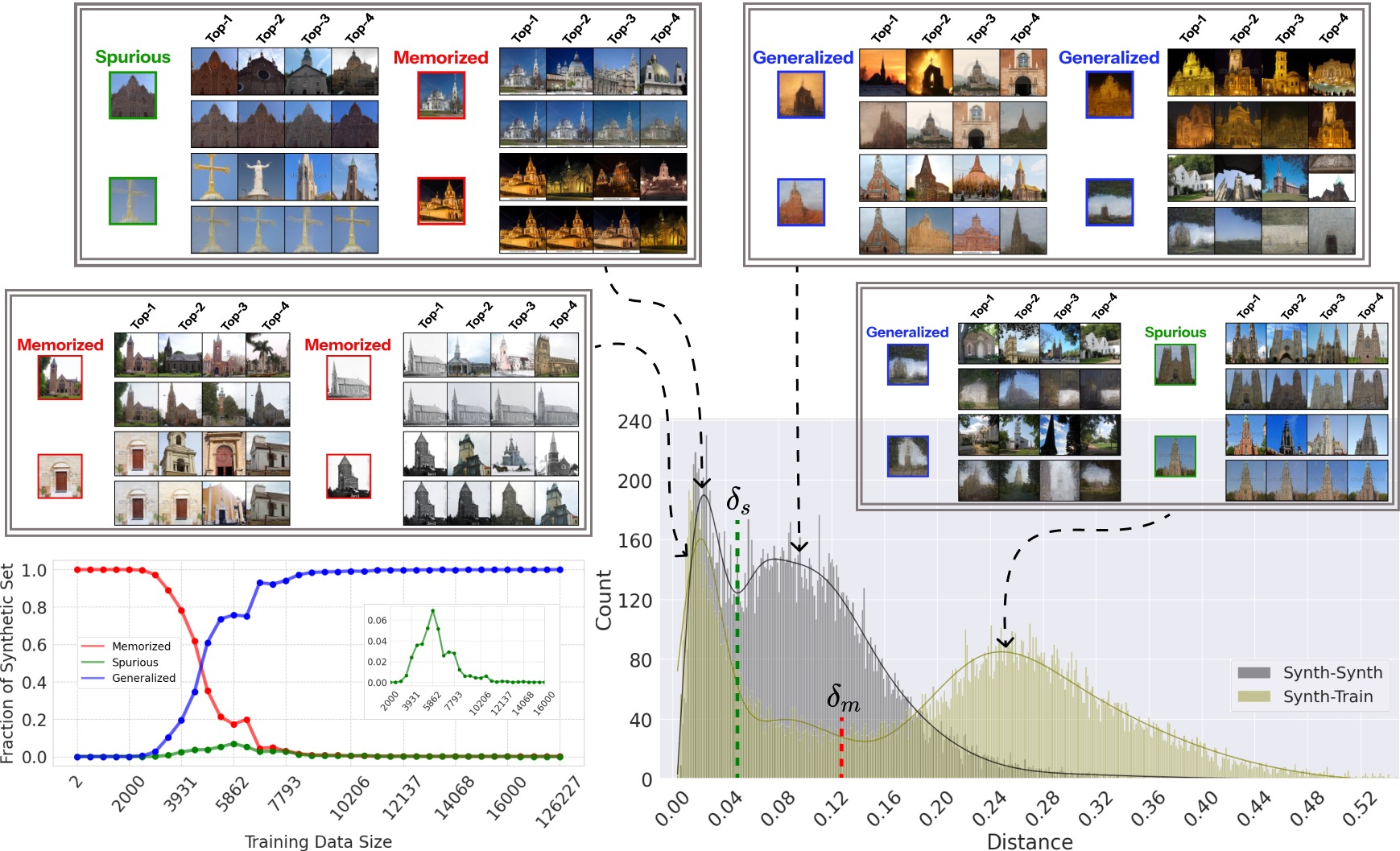}
    \caption{Different sample types across the memorization-to-generalization transition for LSUN-CHURCH \citep{lsun}, defined by the spurious and memorized thresholds, $\delta_s$ and $\delta_m$. The {grey histogram} shows the distances between synthetic samples and their nearest neighbors from the synthetic set $\mathsf{S}'$. The threshold $\delta_s$ is defined as a boundary between the two peaks. The {olive histogram} depicts the distances from the synthetic samples to their closest neighbor from the training set $\mathsf{S}$, with threshold $\delta_m$ separating the two peaks. The threshold $\delta_m$ is chosen much stricter here such that it works well for all training dataset sizes via visual inspection. {Memorized} samples are located in the left peak of the olive histogram, below $\delta_m$. In contrast, {generalized} and {spurious} samples appear to the right of $\delta_m$ (olive histogram). Examples of the generated samples forming each of the four peaks of the histograms are shown in the inset frames. For each generated sample top-4 nearest neighbor images from the training set are shown in the top row, and top-4 nearest neighbors from the synthetic set are shown in the bottom row. Training set size $K = 4896$ was used in this figure (at the peak of the frequency of spurious states), but phenomena discussed are generic and largely independent of this specific value. The fraction of the memorized, spurious, and generalized samples in the pool of all generated samples is shown in the bottom left panel as a function of the training set size. The inset shows amplified spurious fraction ({green curve}). 
    }
    \label{fig:church-hist-figure}
\end{figure}

\begin{figure}[t]
    \centering
    \includegraphics[keepaspectratio, width=1\textwidth, height=0.7\textheight]{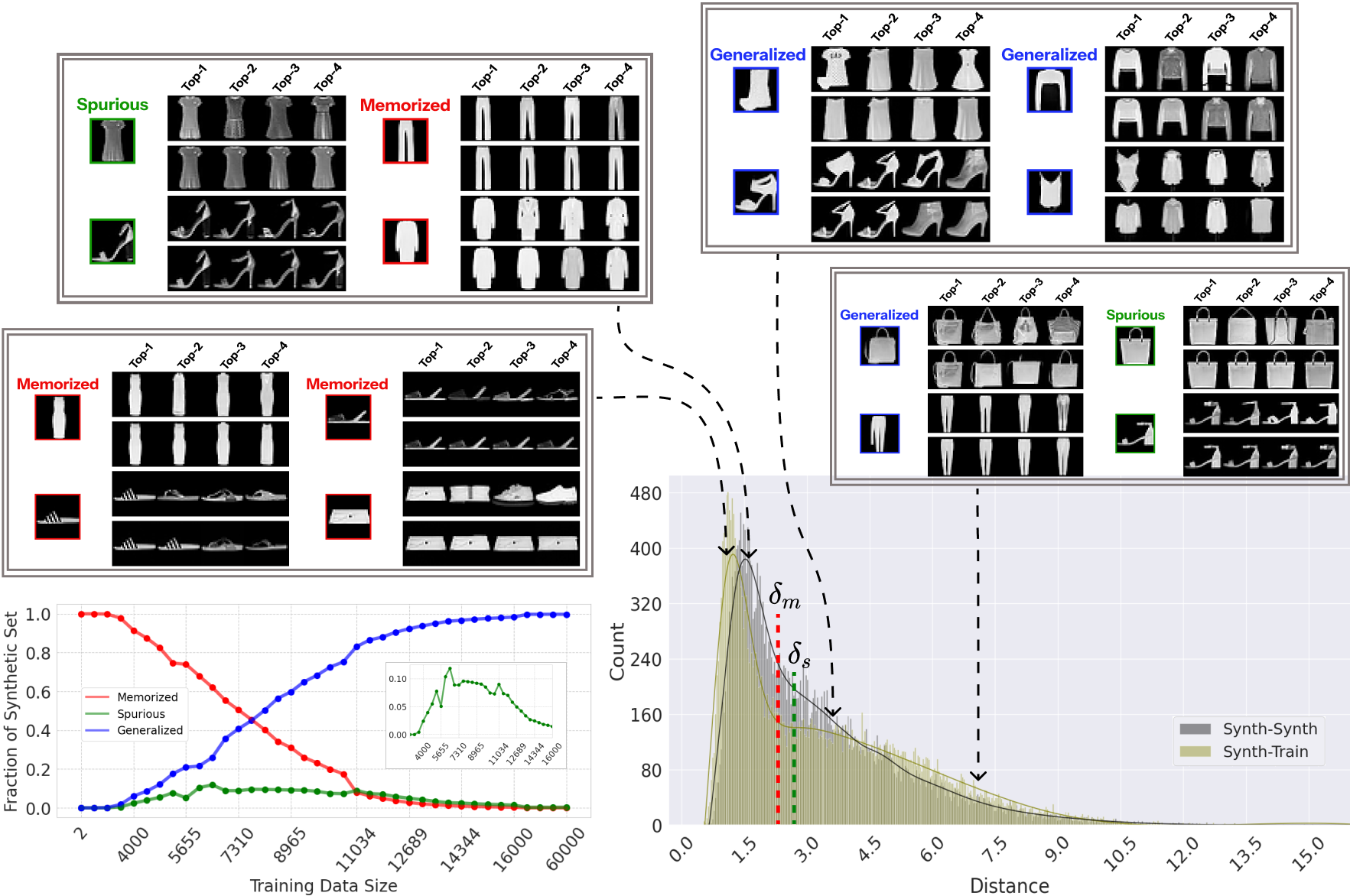}
    \caption{Different sample types across the memorization-to-generalization transition for FASHION-MNIST \citep{fmnist}, defined by the spurious and memorized thresholds, $\delta_s$ and $\delta_m$. The {grey histogram} shows the distances between synthetic samples and their nearest neighbors from the synthetic set $\mathsf{S}'$. The threshold $\delta_s$ is defined as a boundary between the two peaks. The {olive histogram} depicts the distances from the synthetic samples to their closest neighbor from the training set $\mathsf{S}$, with threshold $\delta_m$ separating the two peaks. {Memorized} samples are located in the left peak of the olive histogram, below $\delta_m$. In contrast, {generalized} and {spurious} samples appear to the right of $\delta_m$ (olive histogram). Examples of the generated samples forming each of the four peaks of the histograms are shown in the inset frames. For each generated sample top-4 nearest neighbor images from the training set are shown in the top row, and top-4 nearest neighbors from the synthetic set are shown in the bottom row. Training set size $K = 7724$ was used in this figure (at the peak of the frequency of spurious states), but phenomena discussed are generic and largely independent of this specific value. The fraction of the memorized, spurious, and generalized samples in the pool of all generated samples is shown in the bottom left panel as a function of the training set size. The inset shows amplified spurious fraction ({green curve}). 
    }
    \label{fig:fmnist-hist-figure}
\end{figure}

\begin{figure}[t]
    \centering
    \includegraphics[keepaspectratio, width=1\textwidth, height=0.7\textheight]{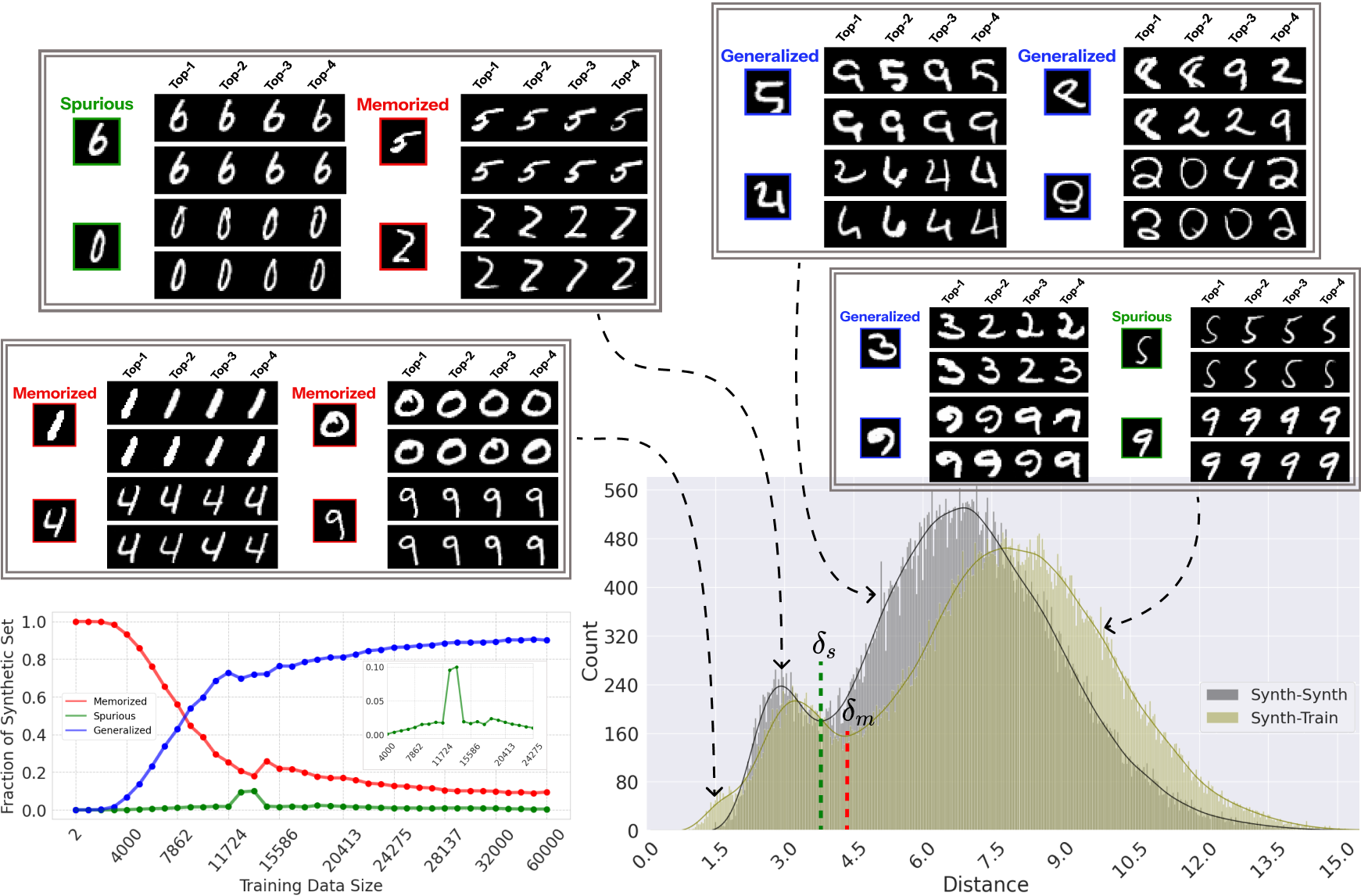}
    \caption{Different sample types across the memorization-to-generalization transition for MNIST \citep{mnist}, defined by the spurious and memorized thresholds, $\delta_s$ and $\delta_m$. The {grey histogram} shows the distances between synthetic samples and their nearest neighbors from the synthetic set $\mathsf{S}'$. The threshold $\delta_s$ is defined as a boundary between the two peaks. The {olive histogram} depicts the distances from the synthetic samples to their closest neighbor from the training set $\mathsf{S}$, with threshold $\delta_m$ separating the two peaks. {Memorized} samples are located in the left peak of the olive histogram, below $\delta_m$. In contrast, {generalized} and {spurious} samples appear to the right of $\delta_m$ (olive histogram). Examples of the generated samples forming each of the four peaks of the histograms are shown in the inset frames. For each generated sample top-4 nearest neighbor images from the training set are shown in the top row, and top-4 nearest neighbors from the synthetic set are shown in the bottom row. Training set size $K = 21379$ was used in this figure, which is slightly past the peak of the frequency of spurious states.  The phenomena discussed are generic and largely independent of this specific value. The fraction of the memorized, spurious, and generalized samples in the pool of all generated samples is shown in the bottom left panel as a function of the training set size. The inset shows amplified spurious fraction ({green curve}). 
    }
    \label{fig:mnist-hist-figure}
\end{figure}

\addvspace{1em}
\noindent \textbf{Selection of Threshold Values}. The detection thresholds, $\delta_m$ and $\delta_s$, were set based on the chosen distance metric $d(\cdot, \cdot)$ and visual inspection of the distance histograms for bimodality as described in \Cref{sec:ch3-detection}. Moreover, these thresholds were chosen to reflect the varying visual complexity and feature richness of the datasets, and their values for each dataset are shown in Table (\ref{tab:thresholds}). Specifically, for each training data size $K$ of each dataset, we manually inspected the bimodality of the olive and gray histograms to select $\delta_m$ and $\delta_s$. Moreover, to better tune these thresholds at each $K$, we manually inspected the least $5$ spurious samples, ensuring that they do not resemble memorized samples, while doing the same for memorized samples. If the resemblance of one set to another is too strong, we decreased the corresponding threshold. Please refer to Figs.~(\ref{fig:church-hist-figure})-(\ref{fig:mnist-hist-figure}) below and \Cref{fig:cifar10-hist} for examples of the histograms. Please note that the bimodality of the gray histograms in FASHION-MNIST might not be significant, but it is enough to select their respective $\delta_s$.


\par\addvspace{1em}
\noindent \textbf{Model and Training Details}. For each point in \Cref{fig:transition-a} and \Cref{fig:transition-b}, we trained a DDPM-based diffusion model, where the score model is a PixelCNN++ based U-Net \citep{van2016conditional, salimans2017pixelcnn++}. We kept the variances, $\beta_\text{min} = 10^{-4}$ and $\beta_\text{max}$ = $2 \times 10^{-2}$, timesteps $T = 1000$, and learning rate $lr =$ $2 \times 10^{-4}$ for all models and datasets. Each model has $2$ residual blocks \citep{he2016deep} for each down-sampling and up-sampling layer, while an attention block is placed at $16\times$ resolution. We only modified the channel multipliers for each model based on the complexity of the dataset, see Table (\ref{tab:models}). If the training data size $K$ is smaller than the specified batch size, the batch size is set to be equal to $K$. For generation or inference, we used the exponential moving average of each trained model, as delineated in \cite{ho2020denoising}, which is obtained with the decay value set as $0.9999$ during training. Please note that we did not use random flipping in the training of our models, since we want our measurements to reflect the training data size at best as random flipping implicitly increases the number of patterns that the models see during training. However, we did use dropout (of value $0.1$) for the training of CIFAR-10 \citep{cifar10}, MNIST \citep{mnist}, and FASHION-MNIST \citep{fmnist} models. For LSUN-CHURCH dataset \citep{lsun}, the images were center-cropped and down-sampled to $64 \times 64$ resolution. Lastly, for each of the training set $\mathsf{S}_\alpha$, where $\alpha = 1, \dots, M$, they are split from the original dataset given a specific size, using the same random seed value of $3407$.

\par
\begin{figure}[t]
    \centering
    \includegraphics[keepaspectratio, width=1.0\linewidth]{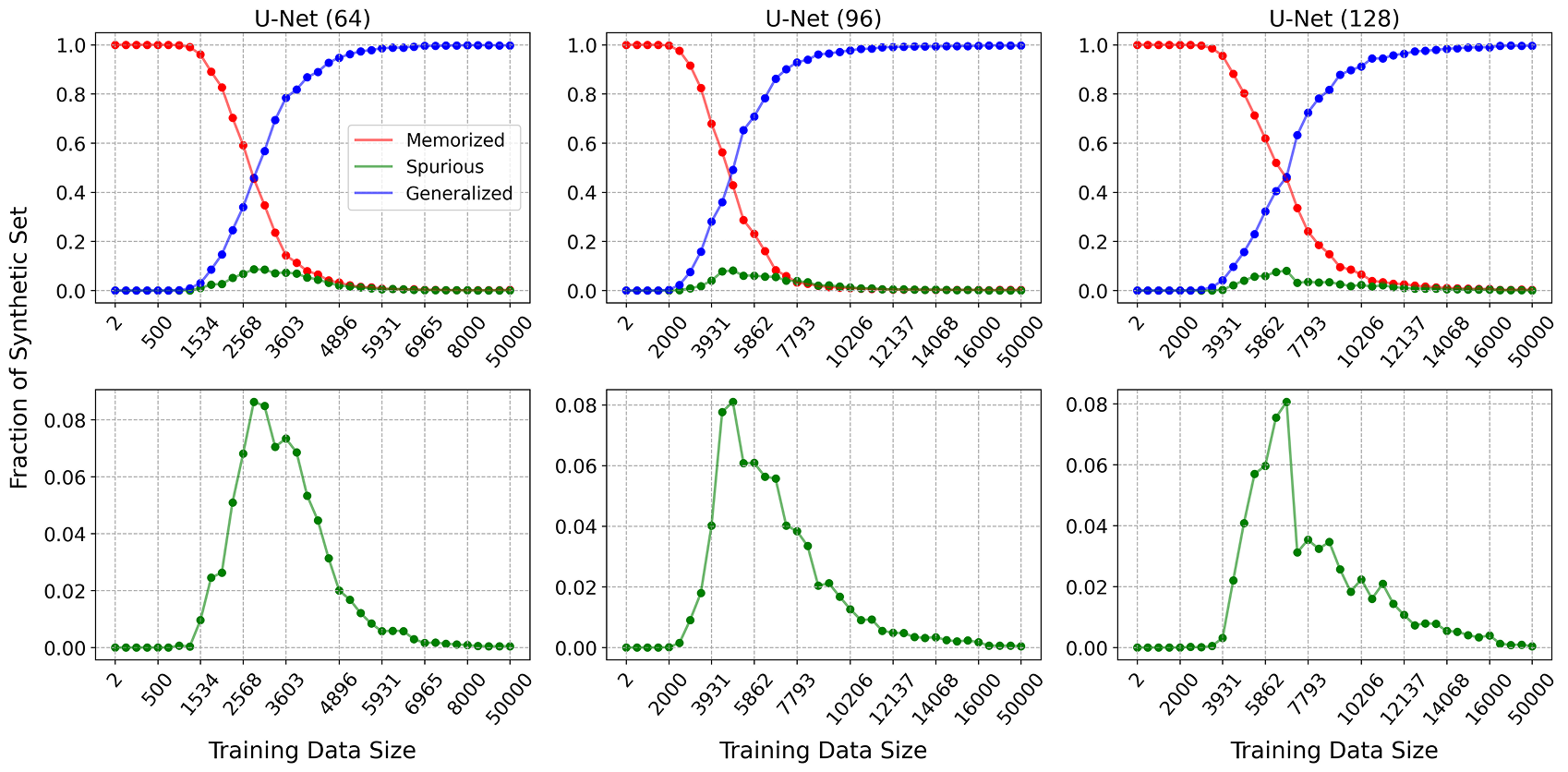}
    \caption{
    The fractions of {memorized}, {spurious}, and {generalized} samples in synthetic sets across different training sizes for CIFAR-10 \citep{cifar10} and various model's widths. As the training data size $K$ increases, memorization decreases while the emergence of spurious patterns follows and the fraction of generalized patterns steadily increase (top row). The fraction of spurious patterns rises and decreases at the boundary between the memorization and generalization phases (bottom row). Additionally, with a smaller number of parameters, the memorization-generalization transition happens at an earlier stage (or smaller $K$), since the model has a memorization capacity. In contrast, as the model's size increases, the transition happens at an later stage (or larger $K$), where it is able to memorize more training data. The numbering, e.g., $64$, $96$, and $128$, denotes the initial latent dimension of the U-Net, see Table~(\ref{tab:models}).
    }
    \label{fig:cifar10-transitions-extra}
\end{figure}

\begin{figure}[t]
    \centering
    \includegraphics[keepaspectratio, width=1.0\linewidth]{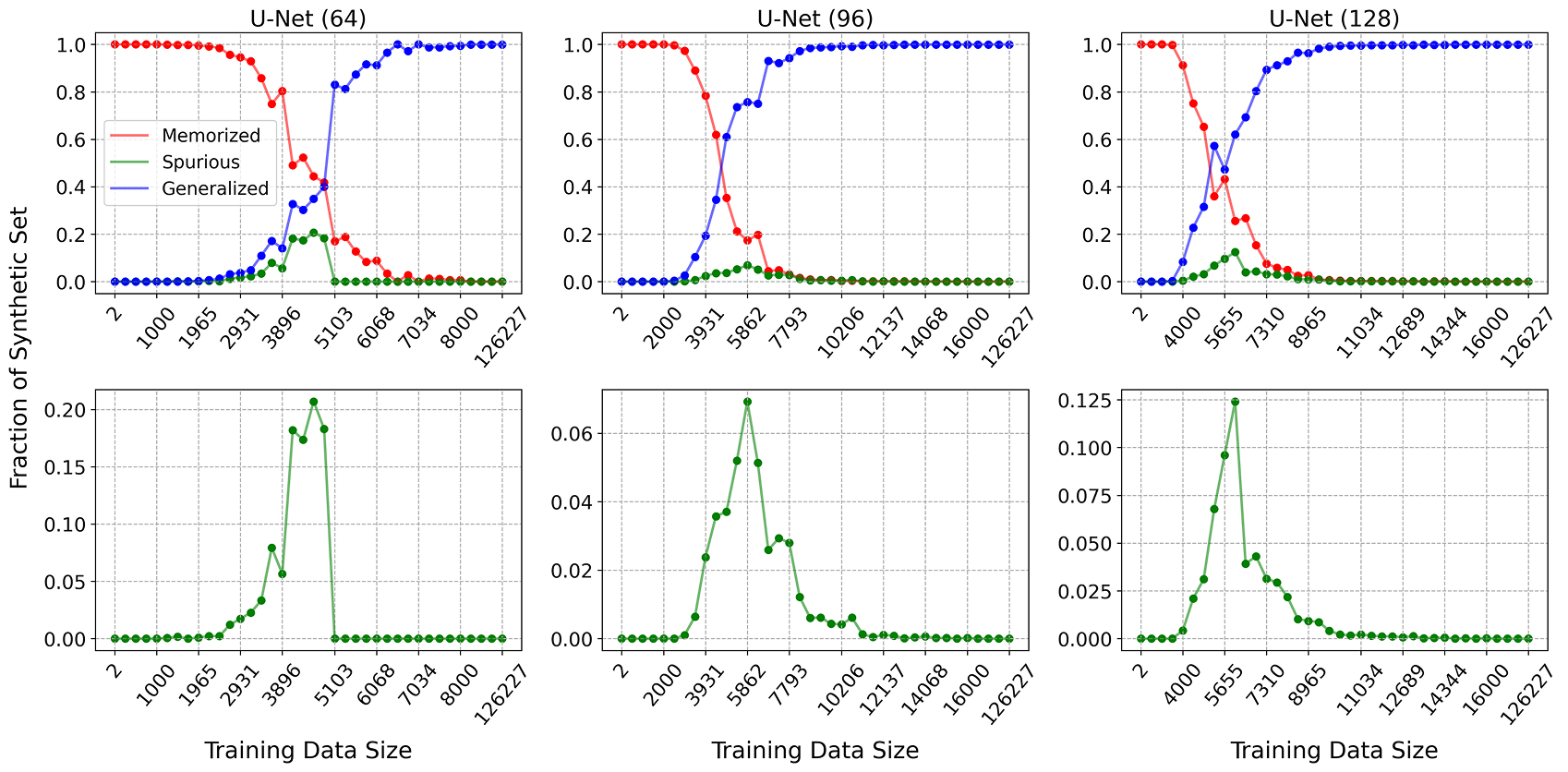}
    \caption{
    The fractions of {memorized}, {spurious}, and {generalized} samples in synthetic sets across different training sizes for LSUN-CHURCH \citep{lsun} and various model's widths. As the training data size $K$ increases, memorization decreases while the emergence of spurious patterns follows and the fraction of generalized patterns steadily increase (top row). The fraction of spurious patterns rises and decreases at the boundary between the memorization and generalization phases (bottom row). Additionally, with a smaller number of parameters, the memorization-generalization transition happens at an earlier stage (or smaller $K$), since the model has a memorization capacity. In contrast, as the model's size increases, the transition happens at an later stage (or larger $K$), where it is able to memorize more training data. The numbering, e.g., $64$, $96$, and $128$, denotes the initial latent dimension of the U-Net, see Table~(\ref{tab:models}).
    }
    \label{fig:church-transitions-extra}
\end{figure}

\addvspace{1em}
\noindent \textbf{Additional Results}. Inspired by \cite{yoon2023diffusion} and \cite{gu2023memorization}, we extend our investigation on the memorization-generalization transition by studying the effects of varying the parameter size of the diffusion model for CIFAR-10 \citep{cifar10} and LSUN-CHURCH \citep{lsun}. The results of this experiment are recorded in Figs.~(\ref{fig:cifar10-transitions-extra})-(\ref{fig:church-transitions-extra}) for these two datasets respectively. Specifically, we trained two additional sets of 38 diffusion models, with the initial latent dimensions of 64 and 96 respectively, and repeated the same experimental setup delineated above, see also Table~(\ref{tab:models}).

For CIFAR-10 and the U-Net (64) models, we identified the critical points, $A$ and $B$, to be $500$ and $8000$ respectively. Meanwhile, for the U-Net (96) models, these critical points are $2000$ and $16000$. To select $\delta_m$ and $\delta_s$, we followed the same process detailed above by observing the distance histograms to see where bimodality occurs with respect to the distance value, and manually inspecting the images to ensure there is a minimal amount of false positives in memorized and spurious samples. Similarly, for the LSUN-CHURCH models, we identified the critical points, $A$ and $B$, to be $1000$ and $8000$ for the U-Net (64) models; and for the larger U-Net (128) models, $4000$ and $16000$ as the critical points. Please refer to Table (\ref{tab:thresholds}) for the threshold values. 



With the smaller U-Net (64) models, we observe the memorization-generalization transition happens at an earlier stage with respect to the larger U-Net (128) models, where the peak of spurious samples appears at $K = 2568$ instead of $K = 5862$ for CIFAR-10, and $K = 4862$ instead of $K = 6086$ for LSUN-CHURCH.
Overall, these results indicate that as the parameter size of the diffusion model increases, the memorization capacity of the model is increased and thus, generalization is delayed since the model can memorize more of its training data denoted by the appearance of spurious patterns happening at a larger training data size $K$. In contrast, when the memorization capacity is reduced, generalization is initiated at an earlier stage (or smaller training data size) given a larger parameterized model. Most importantly, the emergence of spurious patterns remains consistent despite the variations in the parameter size of the diffusion model: these patterns quickly rise and fall at the boundaries of the memorization-to-generalization transition.

\begin{figure}[t]
    \centering
    \includegraphics[keepaspectratio, width=0.65\linewidth]{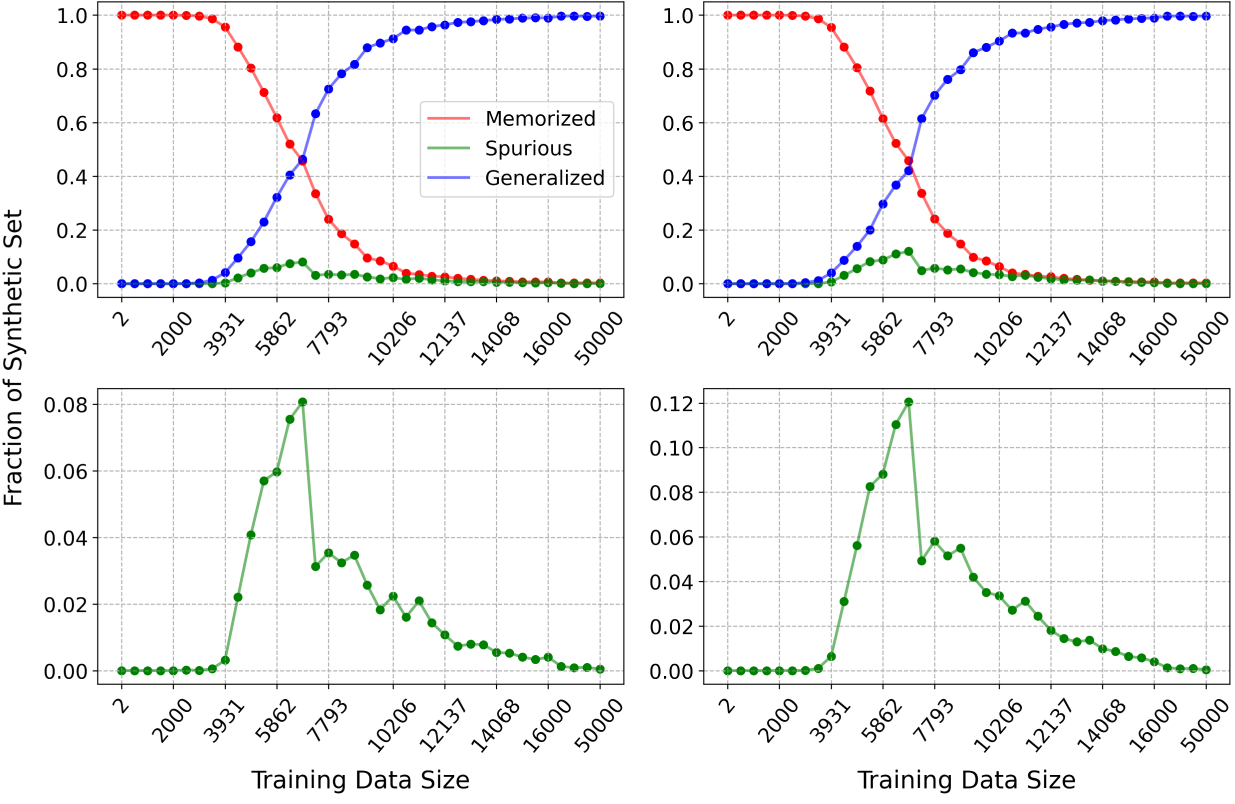}
    \caption{
    Memorization-Generalization transition plots for the CIFAR-10 dataset comparing $\boldsymbol{4\times}$ and $\boldsymbol{8\times}$ synthetic set sizes. The memorization-generalization transition is computed with $4\times$ synthetic sets (left) and with $8\times$ synthetic sets (right) for the same training data sizes. With the larger synthetic sets, more spurious patterns are detected around the peak of the transition. However, the overall trend is maintained, as these patterns quickly rise during the co-existence of memorization and generalization and fall as generalization becomes the dominant phase. Both figures use the same memorized and spurious threshold values detailed in Table~(\ref{tab:thresholds}).
    }
    \label{fig:cifar10-4x-8x}
\end{figure}

Additionally, to verify the consistency of the memorization-generalization transition, we explored the influence of larger synthetic sets and recorded our results in Fig.~(\ref{fig:cifar10-4x-8x}). Specifically, we repeated our experiment done in \Cref{fig:transition-a} for CIFAR-10 using $8\times$ synthetic sets, which was originally done using $4 \times$ synthetic sets. Based on these results, we observe a slight increase in the fraction of spurious samples when using the larger $8\times$ synthetic sets. Specifically, at $K = 5862$, the fraction increases from $0.04$ to roughly $0.06$ with respect to the synthetic set size. Nonetheless, the trend of memorization-generalization, with the emergence of spurious samples, is still maintained regardless of the multiplier for the synthetic size. This aspect highlights the consistency of the spurious phenomenon in diffusion models in relation to the varying of training data size.

\par\addvspace{1em}
\noindent \textbf{Additional Visualizations of Distance Histogram}. Figs.~(\ref{fig:church-hist-figure})-(\ref{fig:mnist-hist-figure}) show the distance histograms used to select the thresholds for LSUN-CHURCH, FASHION-MNIST, and MNIST, complementing \Cref{fig:cifar10-hist} for CIFAR-10.

\clearpage 

\subsection{The Volume of the Basin of Attraction}
\label{sec:basin_of_attraction_extra_details}
\par\addvspace{1em}
\noindent \textbf{Algorithm Details}. To obtain the average log-volume results recorded in \Cref{fig:transition-b}, we first computed the critical time $t_c$ given an image $\hat{\rvx} = \rvx_0$. Specifically, this is done by using Alg.~(\ref{alg:critical_time}), where $t_c$ is obtained using the deterministic DDIM \citep{song2020denoising} sampler to denoise the perturbed version $\rvx_t$ of $\rvx_0$ for each step in $t = 1, \dots, T$, given $N'$ number of denoising steps starting backward from $t$ to $0$. To save computational time, the algorithm utilizes a stride constant $s$ to skip over some of the timesteps $t =1, 1 + s, 1 + 2s, \dots, T - s - 1$. For each timestep $t$, the variance preserving forward process is applied onto the targeted image $\rvx_0$ yielding $\rvx_t$. To account for the variations in noise, the image $\rvx_0$ is perturbed for $M$ number of times at each step $t$. 

To check whether if the critical time $t_c$ is reached, the algorithm utilizes a distance metric $d(\cdot, \cdot)$, which is LPIPS \citep{lpips}, equipped with the \texttt{vgg-16} backbone \citep{VGGsimonyan2014very} for all of the datasets. The distance metric $d(\cdot, \cdot)$ checks whether if a recovered image $\hat{\rvx}_0$, obtained from the DDIM sampler \citep{song2020denoising} with $N'$ denoising steps, is similar to the original pattern $\rvx_0$ within an error threshold $\delta_d \in (0, 1)$. Unlike the experiment on the memorization-generalization transition detailed in Sec.~(\ref{sec:transition-details}), we can afford to trade computational time for better accuracy in this experiment by using a larger backbone for LPIPS. To keep track of the critical time $t_c$ for all perturbations, we initialized it as a vector of integers $t_c \in \mathbb{N}^{M}$, where $M$ is the number of perturbations or trials for the image $\rvx_0$. To ensure that $t_c$ is correct, we utilized a binary stopping vector $\hat{S} \in \{0, 1\}^M$, which keeps track of whether each perturbation of $\rvx_0$ is no longer recoverable, such that we only update an entry of the vector $t_c$ if and only if the corresponding entry in $\hat{S}$ is zero or not stopped. 

Since the entire process of finding $t_c$ is a rough search over a set of timesteps, we require a `breaking condition' for the search. In Alg.~(\ref{alg:critical_time}), this condition is determined by the value $p \in [0, 1]$, which is a probability corresponding to the recovery probability of the image $\rvx_0$ with respect to its $M$ perturbations. To compute $p$, we calculated the distance vector $\mathbf{d} = d(\hat{\rvx}_0, \rvx_0)$ where $\hat{\rvx}_0$ is the predicted version of $\rvx_0$ obtained from denoising $\rvx_t$. Then, we computed the recovery binary vector $\mathbf{r} \in [0, 1]^{M}$ by applying the indicator function onto $\mathbf{d}$ using the threshold $\delta_d$. Lastly, we calculated the probability value $p = \frac{1}{M} \sum^M_{i = 1} \mathbf{r}_i$. If this 
value $p$ is below the probability threshold $\delta_p$, we stop the search for critical time across all trials.

\begin{algorithm}[H]
\caption{Compute critical time $t_c$ for a pattern}
\label{alg:critical_time}
\DontPrintSemicolon
\SetKwInOut{Input}{Input}
\SetKwInOut{Output}{Output}
\SetArgSty{textnormal}
\small
\Input{Pattern $\hat{\rvx}$, score model $s_\theta$, DDIM sampler $\Phi$ with $N'$ denoising steps, forward process $\mathcal{F}$, total diffusion time $T$, time stride $s$, number of trials $M$, distance metric $d$, error threshold $\delta_d$, probability threshold $\delta_p$}
\Output{Critical time $t_c \in \mathbb{N}^M$}
$\rvx_0 \gets \text{duplicate}(\hat{\rvx}, M)$ \tcp*{Duplicate the pattern $M$ times}
$t_c \gets \mathbf{0}$, \ $\hat{S} \gets \mathbf{0}$ \tcp*{Critical times and stopped trials}
\For{$t = 1, 1+s, 1+2s, \dots, T - s - 1$}{
    $\epsilon \sim \mathcal{N}(0, \boldsymbol{I})$, \ $\rvx_t \gets \mathcal{F}(\rvx_0, \epsilon, t)$ \tcp*{Forward process}
    $\hat{\rvx}_0 \gets \Phi(s_\theta, \rvx_t, t, N')$ \tcp*{DDIM denoising}
    $\mathbf{r}_i \gets \mathbb{I}\big(d(\hat{\rvx}_{0,i}, \rvx_{0,i}) \leq \delta_d\big)$ for $i = 1, \dots, M$ \tcp*{Recovery of each trial}
    $p \gets \frac{1}{M} \sum_{i=1}^M \mathbf{r}_i$ \tcp*{Recovery probability}
    \lIf{$p < \delta_p$}{\Return $t_c$}
    $\hat{S} \gets \text{update}(\hat{S}, \mathbf{r})$, \ $t_c \gets \text{update}(t, t_c, \hat{S})$\;
}
\Return $t_c$\;
\end{algorithm}

\par\addvspace{1em}
\noindent \textbf{Experimental Details}. For each training data size $K$, we used a maximum of the top 512 samples belonging to each of the sample types: training data, memorized, spurious, and generalized samples. Note, with the exception of the training data points, we sorted samples belonging to each set, identified via our detection metrics, by their distance from least to greatest. We computed the logarithm of the $N$-ball volume in \Cref{eqn:volume-hypersphere} using the $t_c$ we computed from Alg.~(\ref{alg:critical_time}) where $\Gamma(\cdot)$ is the Euler's gamma function \citep{sebah2002introductionGammaFunction} and $N$ is the dimensionality of the image. The radius $R(\hat{\rvx}, \rvx_{t_c})$ is the Euclidean norm of the difference between a given image $\hat{\rvx} \in \mathbb{R}^N$ and its recoverable perturbation $\rvx_{t_c}$ computed from the forward process. \Cref{fig:transition-b} shows the average log-volume calculated for each training data size $K$ per dataset. 

Meanwhile, for Alg.~(\ref{alg:critical_time}), we fixed the stopping probability $\delta_p$ as $0.8$, the number of perturbations $M$ as $20$, the DDIM steps $N'$ as $10$ and the stride constant $s$ as $10$ for all datasets. At some training data size $K$, in which there are very few samples in either memorized, spurious or generalized set, we opted to not include the statistics of such a set, partly due to its large variance. We remove the statistics of such a set if the number of its samples is less than $0.1\%$ of its corresponding synthetic set for a training data size $K$.

For CIFAR-10 and LSUN-CHURCH, we utilized the threshold $\delta_d$ as $0.03$ and $0.12$ respectively. We kept these threshold values similar to the memorized threshold values $\delta_m$ detailed in Table~(\ref{tab:thresholds}), since they have proven to be effective at separating the memorized samples from non-memorized ones in the previous experiment. However, since it is difficult to select an appropriate threshold value for the $L_2$-distance metric (due to its range being between $0$ and $\infty$), we opted to use LPIPS to experiment with MNIST and FASHION-MNIST using the same threshold $\delta_d = 0.1$ for both. To find this optimal threshold value, we selected three data sizes, where each belongs to one of the three phases: memorization, spurious, or generalization. We then observed which of the several threshold values, starting from $0.01$ to $0.15$ with the increment of $0.01$, reflects best of the performance given by LPIPS using \texttt{vgg-16} \citep{VGGsimonyan2014very} as its backbone. Specifically, we evaluated these thresholds using $K \in \{ 4000, 11724, 60000 \}$ for MNIST, and $K \in \{ 4000, 8965, 60000\}$ for FASHION-MNIST. We simply performed Alg.~(\ref{alg:critical_time}) for those three data sizes on a small batch of $32$ samples for the three sample types, and manually checked whether the recovered samples (using the obtained critical time $t_c$ to denoise from) are similar to the original images and if this observation is consistent across those three data sizes. 

\subsection{The Curvature of the Energy}
\label{sec:energy-curvature}
\par\addvspace{1em}
\noindent \textbf{Overview}. Following \cite{achilli2024losing} and \cite{ventura2024manifolds}, we utilized their stable version of algorithm for computing the singular values of the Jacobian of the score function, detailed in Alg.~(\ref{alg:central-diff}). The motivations for the algorithm are first detailed in \cite{stanczuk2022your} and later expanded by \cite{achilli2024losing} and \cite{ventura2024manifolds}. Here we recite the motivations laid out in \cite{ventura2024manifolds}.

Consider a generative diffusion model with $p_0(\rvx)$ defined on a $d$-dimensional manifold $\mathcal{M}$, where $\mathcal{M} = \mathcal{M}_0$ denotes the manifold at $t = 0$. We can define a time-dependent set of points:
\begin{equation}
    \mathcal{M}_t = \{ \rvx^* | \tilde{s}_\mathcal{M} (\rvx^*, t) = 0, \,\,\text{with} \,\, J_\mathcal{M} (\rvx^*, t) \,\,\text{n.s.d.} \}
    \label{eqn:stable-latent-set}
\end{equation}
which \cite{ventura2024manifolds} name the \textbf{stable latent set} of the diffusion process. The negative semi-definiteness (n.s.d.) is a stability condition of the Jacobian matrix $J_\mathcal{M} (\rvx^*, t)$ of the support score $\tilde{s}_\mathcal{M} (\rvx^*, t)$, defined as the score function obtained from the uniform data distribution $\tilde{p}_0 (\rvx) = \frac{1}{|\mathcal{M}|} \delta_\mathcal{M} (\rvx)$. 

\begin{figure}[t]
    \centering
    \includegraphics[keepaspectratio, width=1\linewidth]{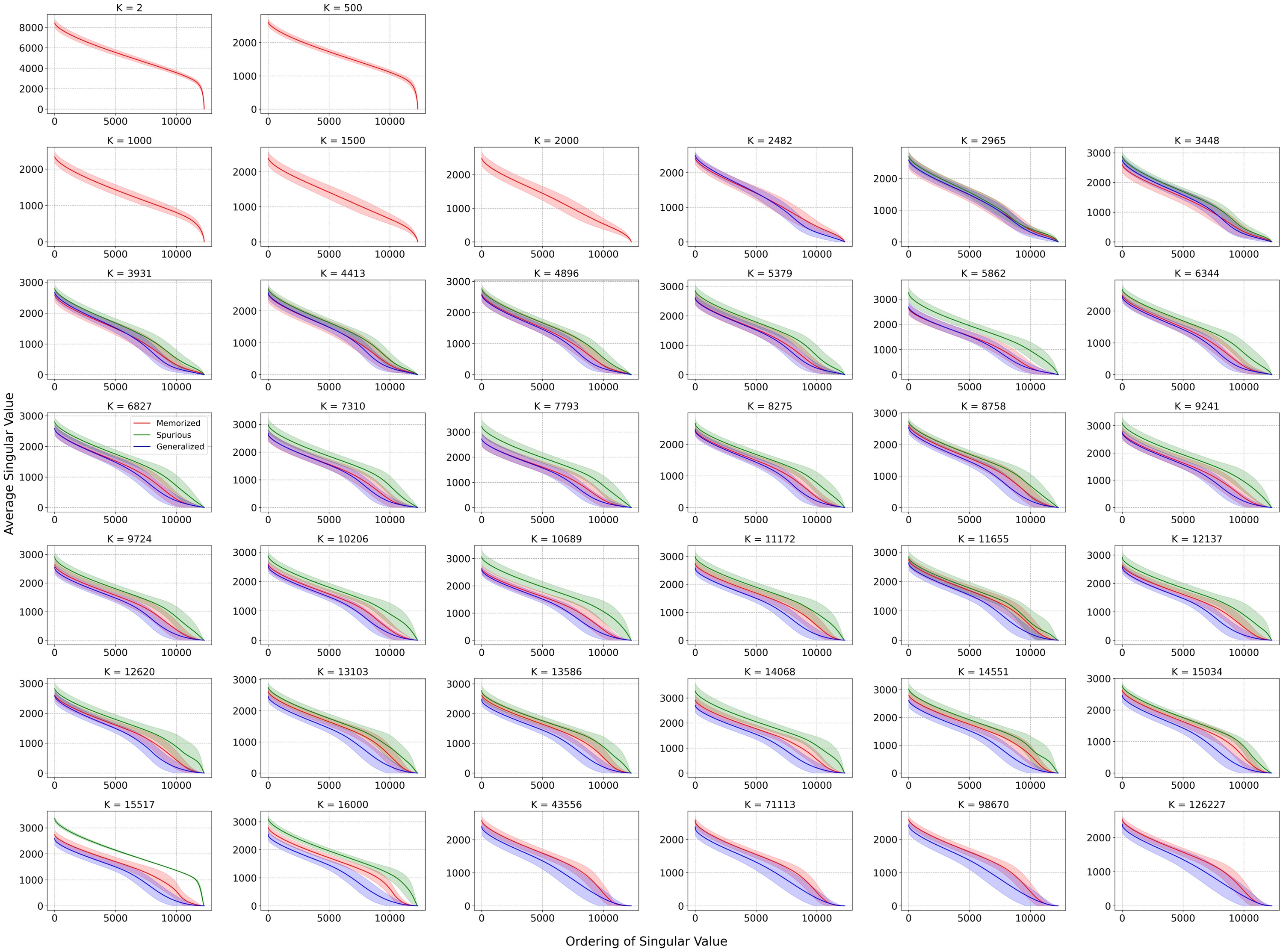}
    \caption{
    Average singular value spectra of the approximated score's Jacobian for memorized, spurious, and generalized samples from diffusion models trained on the LSUN-CHURCH \citep{lsun} dataset, shown across 38 increasing training data sizes $K$. The plots reveal a clear geometric hierarchy in the intermediate training regime, where memorized samples consistently exhibit the highest curvature (largest singular values), followed by spurious, and finally generalized samples. As the model enters the full generalization phase at large $K$, the average spectra for all three sample types become very distinguishable. Shaded region is the standard deviation of the singular values of different samples of each category.
    }
    \label{fig:church-all-svs}
\end{figure}

\begin{figure}[t]
    \centering
    \includegraphics[keepaspectratio, width=1\linewidth]{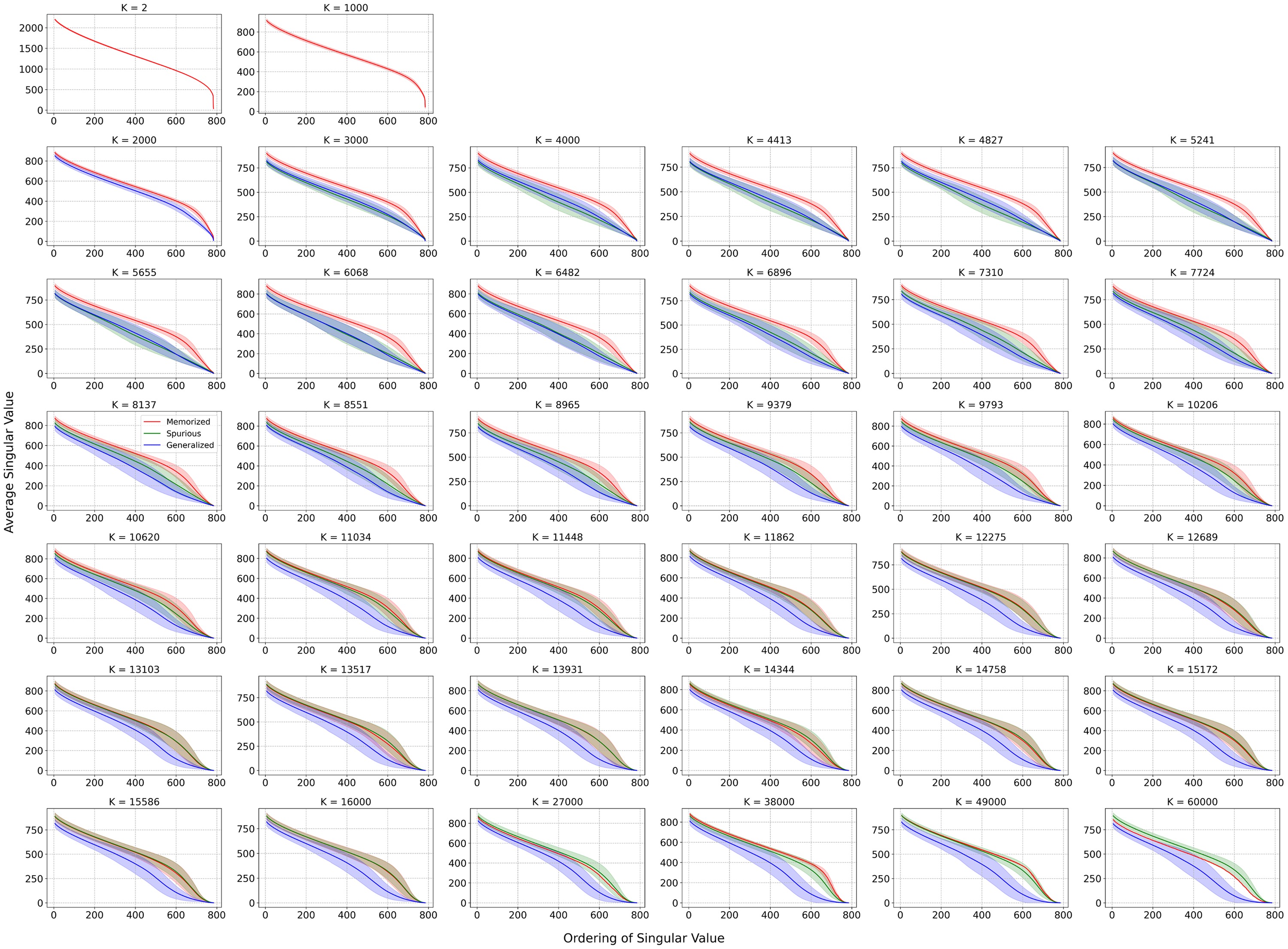}
    \caption{
    Average singular value spectra of the approximated score's Jacobian for memorized, spurious, and generalized samples from diffusion models trained on the FASHION-MNIST \citep{fmnist} dataset, shown across 38 increasing training data sizes $K$. The plots reveal a clear geometric hierarchy in the intermediate training regime, where memorized samples consistently exhibit the highest curvature (largest singular values), followed by spurious, and finally generalized samples. As the model enters the full generalization phase at large $K$, the average spectra for all three sample types become very distinguishable. Shaded region is the standard deviation of the singular values of different samples of each category.
    }
    \label{fig:fmnist-all-svs}
\end{figure}

\begin{figure}[t]
    \centering
    \includegraphics[keepaspectratio, width=1\linewidth]{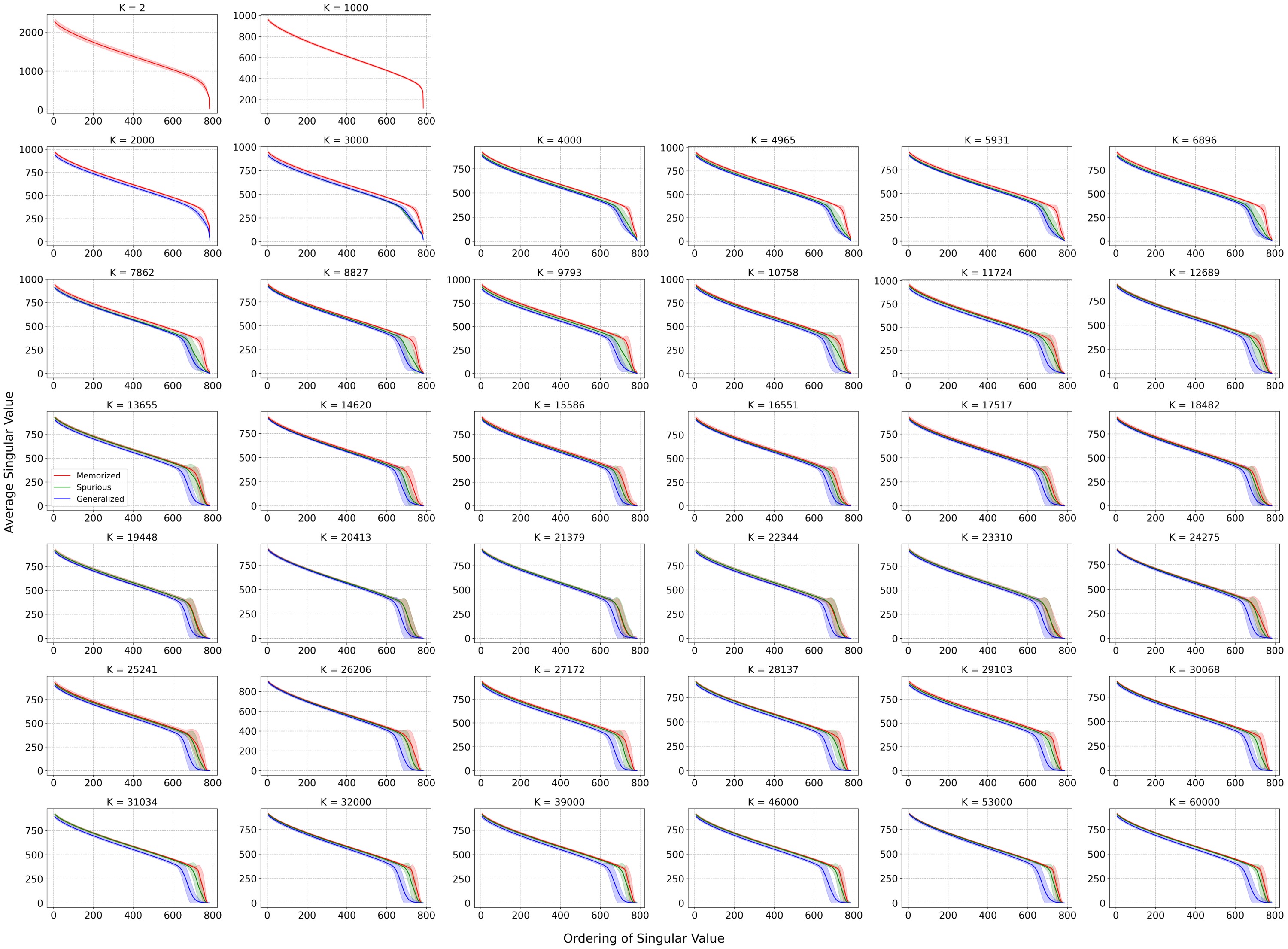}
    \caption{
    Average singular value spectra of the approximated score's Jacobian for memorized, spurious, and generalized samples from diffusion models trained on the MNIST \citep{mnist} dataset, shown across 38 increasing training data sizes $K$. The plots reveal a clear geometric hierarchy in the intermediate training regime, where memorized samples consistently exhibit the highest curvature (largest singular values), followed by spurious, and finally generalized samples. As the model enters the full generalization phase at large $K$, the average spectra for all three sample types become very distinguishable. Shaded region is the standard deviation of the singular values of different samples of each category.
    }
    \label{fig:mnist-all-svs}
\end{figure}

\begin{figure}[t]
    \centering
    \includegraphics[keepaspectratio, width=1\linewidth]{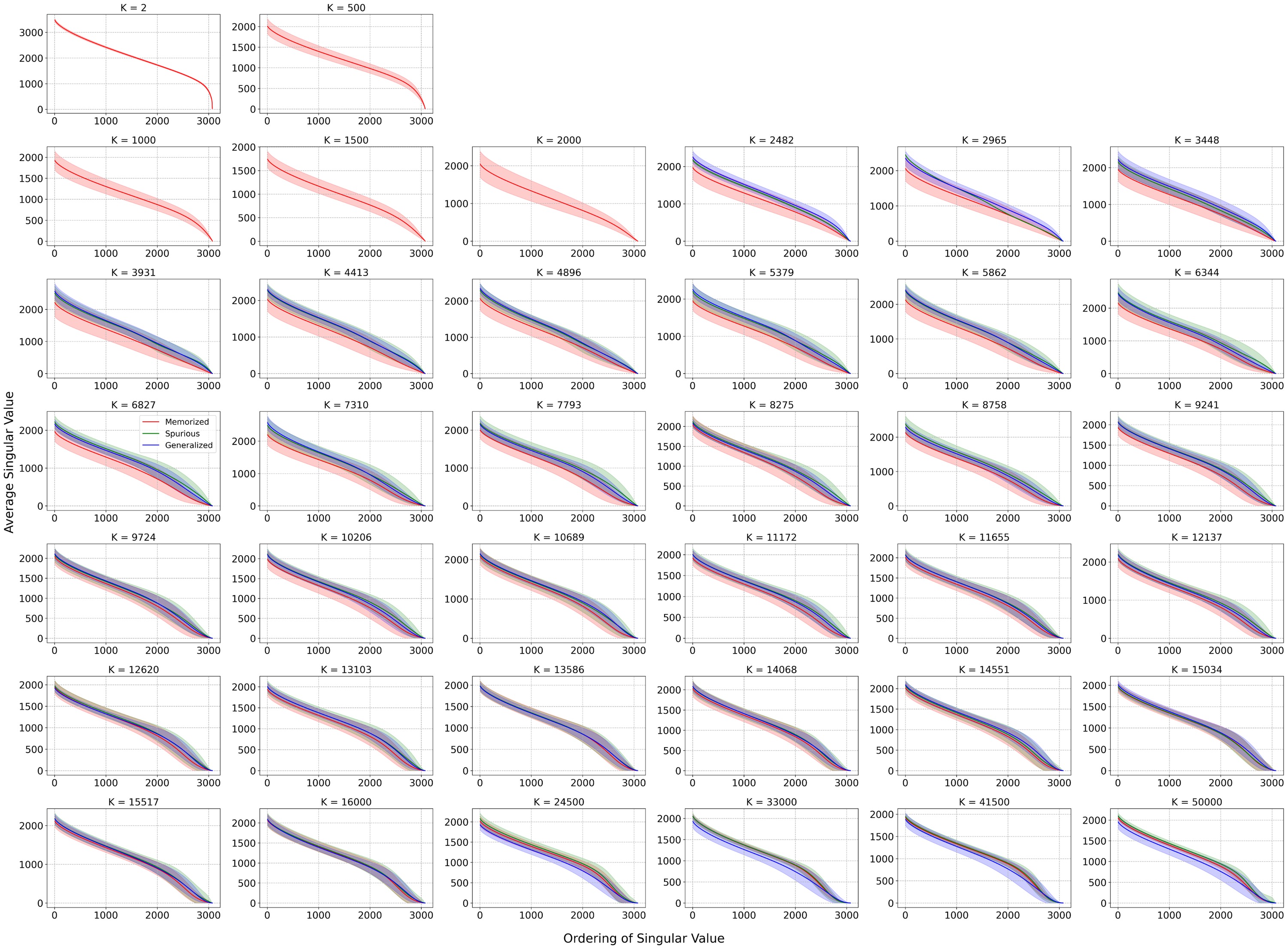}
    \caption{
    Average singular value spectra of the approximated score's Jacobian for memorized, spurious, and generalized samples from diffusion models trained on the CIFAR-10 \citep{cifar10} dataset, shown across 38 increasing training data sizes $K$. In contrast to the clear separation as seen in other datasets, the spectra for all three sample types remain heavily overlapped throughout the entire transition from memorization to generalization. However, as the model enters the full generalization phase at large $K$, the average spectra for all three sample types become less indistinguishable. Shaded region is the standard deviation of the singular values of different samples of each category.
    }
    \label{fig:cifar10-all-svs}
\end{figure}

Using the relationship $s_\theta(\rvx, t) = -\frac{\epsilon_\theta(\rvx, t)}{\sigma_t}$, the diffusing particles typically explore shells of a radius that concentrates on $\sigma_t$. Then, for a small perturbation $\rvp$ around a state $\rvx^*$ on the latent manifold $\mathcal{M}_t$, the score function is well approximated by linearization:
\begin{equation}
    s(\rvp, t) \approx J(\rvx^*, t) \rvp = \sum_j (\rvv_j \cdot \rvp) \lambda_j(\rvx^*, t) \rvv_j
\end{equation}
where $\rvv_j$ and $\lambda_j$ are the respective $j$-th eigenvector and eigenvalue of the negative Jacobian $-J(\rvx^*, t)$. Small perturbations, which aligned with the tangent space of $\mathcal{M}_t$, correspond to small eigenvalues; while orthogonal perturbations correspond to high eigenvalues, as the score tends to push the stochastic dynamics towards its fixed-points. Hence, based on these relationships, the spectrum of eigenvalues can provide detailed information regarding the local geometry of the stable latent set (\ref{eqn:stable-latent-set}). Moreover, as already indicated by \cite{stanczuk2022your}, one can estimate the dimensionality of the manifold from the location of a drop, or a sharp change, in the sorted spectrum of eigenvalues. The full range of the spectra results is shown in Figs.~(\ref{fig:church-all-svs})-(\ref{fig:cifar10-all-svs}). 

\par\addvspace{1em}
\noindent \textbf{Experimental Details}. Using Alg.~(\ref{alg:central-diff}), we computed the singular values of the Jacobian of the learned score function $s_\theta$ given a small time $t_0$ for all datasets shown in \Cref{fig:energy-curvature-a}. For each sample type, we utilized a maximum of 512 samples and computed the singular values at $t_0 = 2$ for all datasets and training data sizes. The singular values are sorted from greatest to least, and the $top-5$ values are ignored to avoid the case of very large singular values skewing the visualizations. Meanwhile, for the spectra results of the training data points shown in the bottom row of \Cref{fig:energy-curvature-a}, we utilized $\min(K, 2048)$ number of training samples for all datasets and training data sizes. In this visualization, we utilized the rainbow color scheme and plotted the singular values for small training data sizes, starting from the purple color, to larger ones, ending at the red color.

\begin{figure}[t]
    \centering
    \includegraphics[keepaspectratio, width=1\linewidth]{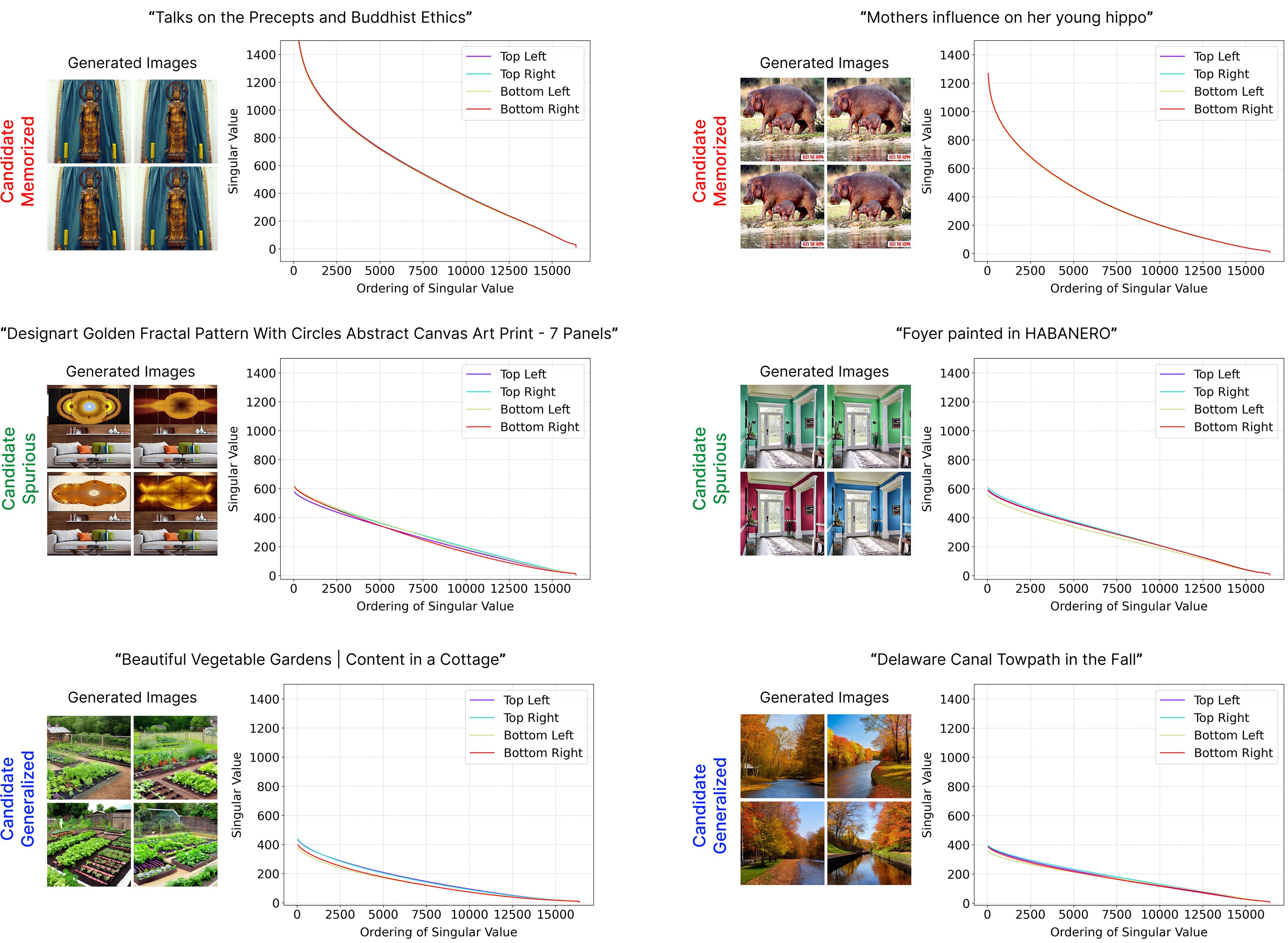}
    \caption{
    Additional examples of candidate memorized, spurious, and generalized samples from Stable Diffusion trained on LAION dataset \citep{schuhmann2022laion}, each corresponding to a distinct curvature signature. The candidate memorized sample has much larger singular values, while the candidate generalized sample has much smaller ones. The selected possible spurious samples have larger singular values than the candidate generalized sample, but much less in contrast to the candidate memorized sample. The common trait between candidate memorized and spurious samples is that both are stable attractors, as demonstrated by their repeated similar generations given different initial points (or noise vectors) and the conditioning on text-prompt. The y-axis is clipped at the value of 1500 to better contrast the shown examples' spectra. The associated training image for each prompt can be found in \cite{carlini2023}, \cite{webster2023reproducible}, and \cite{wen2024detecting} and on our code \href{https://github.com/Lemon-cmd/Diffusion-Models-and-Associative-Memory/blob/main/laion_candidates.json}{repository}. For copyright reasons, we omitted the inclusion of original training images in this figure.
    }
    \label{fig:stable-extra}
\end{figure}

Finally, regarding our experiment with Stable Diffusion \citep{rombach2022high} for \Cref{fig:energy-curvature-b}, following \cite{wen2024detecting} and \cite{jeon2024understanding} we utilized its version 1.4 from the package Diffusers \citep{von-platen-etal-2022-diffusers}. The candidate memorized and spurious prompts we utilized are originally found by \cite{webster2023reproducible} and \cite{webster2023duplication}. Additional examples are shown in Fig.~(\ref{fig:stable-extra}). Meanwhile, for candidate generalized samples, their prompts are obtained from LAION2B-Aesthetic \citep{schuhmann2022laion}. Instead of using Alg.~(\ref{alg:central-diff}), we opted for Alg.~(\ref{alg:non-central-diff}) and computed the singular values of the Jacobian of the score using $t_0 = 5$, due to the lack of computational resources. Please note that, according to \cite{ventura2024manifolds}, Alg.~(\ref{alg:central-diff}) produces a much smaller number of extreme singular values than the later algorithm. For generation of the samples, we utilized DDIM \citep{song2020denoising} sampler with different noise vectors initialized from a fixed seed. For the visualizations, we skipped the $top-50$ singular values and set the limit for the y-axis as 1500. 

\vspace{0.2in}

\begin{algorithm}[H]
\caption{Compute singular values of the Hessian of the energy using central difference for a pattern}
\label{alg:central-diff}
\DontPrintSemicolon
\SetKwInOut{Input}{Input}
\SetKwInOut{Output}{Output}
\SetArgSty{textnormal}
\small
\Input{Target sample $\rvx_0 \in \mathbb{R}^N$, time $t_0$, score model $s_\theta$, forward process $\mathcal{F}$}
\Output{Singular values $\{ s_i \}_{i=1}^{N}$}
$\mathsf{M} \gets \text{Empty}()$ \tcp*{Create an empty matrix}
\For{$i = 1, 2, \dots, 4N$}{
    $\epsilon \sim \mathcal{N}(0, \boldsymbol{I})$ \tcp*{Sample noise}
    $\rvx^+ \gets \mathcal{F}(\rvx_0, \epsilon, t_0)$, \ $\rvx^- \gets \mathcal{F}(\rvx_0, -\epsilon, t_0)$ \tcp*{Paired perturbations}
    $\mathsf{M}^{(i)} \gets \frac{1}{2}\big(s_\theta(\rvx^+, t_0) - s_\theta(\rvx^-, t_0)\big)$ \tcp*{Add a column to $\mathsf{M}$}
}
$\{ s_i \}_{i=1}^{N} \gets \text{SVD}(\mathsf{M})$ \tcp*{Singular values}
\Return $\{ s_i \}_{i=1}^{N}$\;
\end{algorithm}

\vspace{1.5em}
\begin{algorithm}[H]
\caption{Compute singular values of the Hessian of the energy for a pattern}
\label{alg:non-central-diff}
\DontPrintSemicolon
\SetKwInOut{Input}{Input}
\SetKwInOut{Output}{Output}
\SetArgSty{textnormal}
\small
\Input{Target sample $\rvx_0 \in \mathbb{R}^N$, time $t_0$, score model $s_\theta$, forward process $\mathcal{F}$}
\Output{Singular values $\{ s_i \}_{i=1}^{N}$}
$\mathsf{M} \gets \text{Empty}()$ \tcp*{Create an empty matrix}
\For{$i = 1, 2, \dots, 4N$}{
    $\epsilon \sim \mathcal{N}(0, \boldsymbol{I})$, \ $\rvx^+ \gets \mathcal{F}(\rvx_0, \epsilon, t_0)$ \tcp*{Perturbation}
    $\mathsf{M}^{(i)} \gets s_\theta(\rvx^+, t_0)$ \tcp*{Add a column to $\mathsf{M}$}
}
$\{ s_i \}_{i=1}^{N} \gets \text{SVD}(\mathsf{M})$ \tcp*{Singular values}
\Return $\{ s_i \}_{i=1}^{N}$\;
\end{algorithm}

\par\addvspace{1em}
\noindent \textbf{Curvature Results}. Figs.~(\ref{fig:church-all-svs})-(\ref{fig:cifar10-all-svs}) show the full curvature spectra for every training data size $K$ and dataset.

\par\addvspace{1em}
\noindent \textbf{Additional Examples of Stable Diffusion}. \Cref{fig:stable-extra} shows additional candidate memorized, spurious, and generalized samples from Stable Diffusion.

\clearpage 
\subsection{Relative Energy Gap}
\label{sec:relative_energy_details}

\par\addvspace{1em}
\noindent \textbf{Overview}. In this section, we provide an additional experiment on the relative energy gap, which measures the gap between the energy depth of training data points and each of the three distinct sample types, using the method for approximating the relative potential from \cite{raya2024spontaneous}.

Consider the reverse SDE in \Cref{eqn:bg-reverse-sde},
\begin{align}
    \mathrm{d} \rvx_t = [ \rvf(\rvx_t, t) - g(t)^2 \nabla_{\rvx_t} \log p_t (\rvx_t) ] \mathrm{d}t + g(t) \mathrm{d} \bar{\rvw}_t .
\end{align}
We can re-express the above equation with the potential function $u(x, t)$:
\begin{equation}
    -\nabla_{\rvx_t} u(\rvx_t, t) = \mathbf{f}(\rvx_t, t) - g(t)^2  \nabla_{\rvx_{t}}  \log p_t(\rvx_t) 
    \label{eqn:reverse-sde-reexpress}
\end{equation}
by using the drift term. By moving the minus sign and taking the integration of Eq. (\ref{eqn:reverse-sde-reexpress}) with respect to $\rvx_t$, we have
\begin{equation}
    u(\rvx_t, t) = g(t)^2 \log p_t(\rvx_t) + \int^0_{\rvx_t} f(\rvz, t) d\rvz,
    \label{eqn:potential_energy_function_appendix}
\end{equation}
which is the potential function $u(\rvx_t, t)$ of the generative process. Substituting this potential function (\ref{eqn:potential_energy_function_appendix}) into the reverse SDE, we have the following generative process:
\begin{equation}
    \mathrm{d} \rvx_t = -\nabla_{\rvx_t} u(\rvx_t, t) \mathrm{d} t + g(t) \mathrm{d}\bar{\rvw}_t .
    \label{eqn:generative-process-potential}
\end{equation}

For our experimental setting, where we used the DDPM \citep{ho2020denoising} setting, we have the following variance preserving forward SDE:
\begin{equation}
    \mathrm{d} \rvx_t = -\frac{1}{2} \beta(t) \rvx_t \mathrm{d} t + \sqrt{\beta(t)} \mathrm{d} \rvw_t ,
\end{equation}
and the corresponding reverse (or generative) process SDE
\begin{equation}
    \mathrm{d} \rvx_t = \bigg [ -\frac{1}{2} \beta(t) \rvx_t - \beta(t)  \nabla_{\rvx_t}  \log p_t(\rvx_t) \bigg ] \mathrm{d} t + \sqrt{\beta(t)} \mathrm{d} \bar{\rvw}_t .
    \label{eqn:vp-reverse-sde}
\end{equation}
With Eq.~(\ref{eqn:generative-process-potential}), we have 
\begin{equation}
    \mathrm{d} \rvx_t = -\nabla_{\rvx_t} u(\rvx_t, t) \mathrm{d} t + \sqrt{\beta(t)} \mathrm{d} \bar{\rvw}_t    ,
\end{equation}
and by using Eq.~(\ref{eqn:reverse-sde-reexpress}), the potential for our experimental setting equates to
\begin{equation}
\begin{split}
    u(\rvx_t, t) &= g(t)^2 \log p_t(\rvx_t) + \int_{\rvx_t}^0 f(\rvz, t) d\rvz \\ 
    & = \beta(t) \log p_t (\rvx_t) + \frac{1}{2} \beta(t) \int_{\rvx_t}^0 \rvz_t \mathrm{d} \rvz \\ 
    &= \beta(t) \log p_t (\rvx_t) - \frac{1}{4} \beta(t) \rvx_t^2 . 
\end{split}  
\label{eqn:potential_function_vp}
\end{equation}

\begin{figure}[t]
\centering
\includegraphics[keepaspectratio, width=0.9\textwidth, height=0.7\textheight]{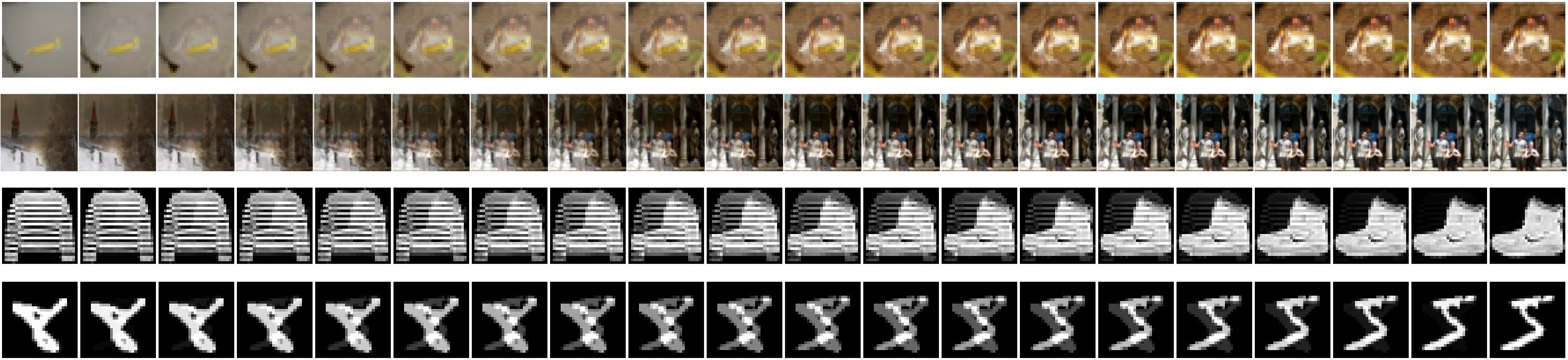}
\caption{Circular interpolation of a target image $\rvx_1$ and a reference image $\rvx_2$ over $\alpha \in \{0, \dots, \frac{\pi}{2}\}$ for different datasets.}

\label{fig:relative-energy-interpolation}

\end{figure}

Assume we have two images, $\rvx_1 (t)$ and $\rvx_2 (t)$, where $\rvx_1$ is a target image and $\rvx_2 (t)$ is a reference image at time $t$. Since we are interested in the energy of unperturbed images, let $\rvx_1 = \rvx_1(0)$ and $\rvx_2 = \rvx_2(0)$. As discussed in \cite{raya2024spontaneous}, the $N$-dimensional potential of a diffusion model can be projected down into 1-dimension via a generative trajectory between $\rvx_1$ and $\rvx_2$. We can obtain such a trajectory via circular interpolation between $\rvx_1$ and $\rvx_2$ following
\begin{equation}
     \tilde{\rvx}(\alpha) = \cos (\alpha) \rvx_1 + \sin (\alpha) \rvx_2 ,
     \label{eqn:circular_interpolation}
\end{equation}
and compute the relative potential (up to a constant) as a function of $\alpha$ by using 
\begin{equation}
    \tilde{u}(\alpha) = u(\tilde{\rvx} (\alpha)) = \int^{\pi / 2}_{0} \nabla u (\tilde{\rvx} (\alpha)) \cdot \rvv \, \mathrm{d} \alpha ,
    \label{eqn:relative-potential-int}
\end{equation}
where $\rvv = - \sin(a) \rvx_1 + \cos(a) \rvx_2$. See Fig.~(\ref{fig:relative-energy-interpolation}) for a visualization of this circular interpolation between two samples. For discrete times, we instead estimate the relative energy as 
\begin{equation}
    \tilde{u}(\bar{\alpha}_L) \approx \sum^{L}_{i = 1} \nabla u (\tilde{\rvx} (\bar{\alpha}_i)) \cdot \rvv \Delta \alpha 
    \label{eqn:relative-potential-discrete}
\end{equation}
given that $\bar{\alpha} = \{0, \dots, \frac{\pi}{2} \}$ and the potential $u(\cdot)$ is obtained from the above Eq.~(\ref{eqn:potential_function_vp}).

\par\addvspace{1em}
\noindent \textbf{Experimental Details}. In our experimentation, we utilized a maximum of the top 512 samples for each of three sample types, memorized, spurious, and generalized, to compute the relative energy following Eq.~(\ref{eqn:relative-potential-discrete}), at each training data size $K$. For training data points, we instead utilized $\min (K, 2048)$ number of samples. Note, with the exception of the training data set, the samples in the three sets are sorted from least to greatest, based on their distance corresponding to their respective detection metric (\Cref{sec:ch3-detection}). We computed the relative energy of each set following Alg.~(\ref{alg:relative_energy}), which is strictly following the formulations detailed above. Note, we randomly selected and utilized an image from the entire training set as part of the relative energy calculation. 

\begin{figure}[t]
\centering  
\includegraphics[keepaspectratio, width=0.905\textwidth, height=0.7\textheight]{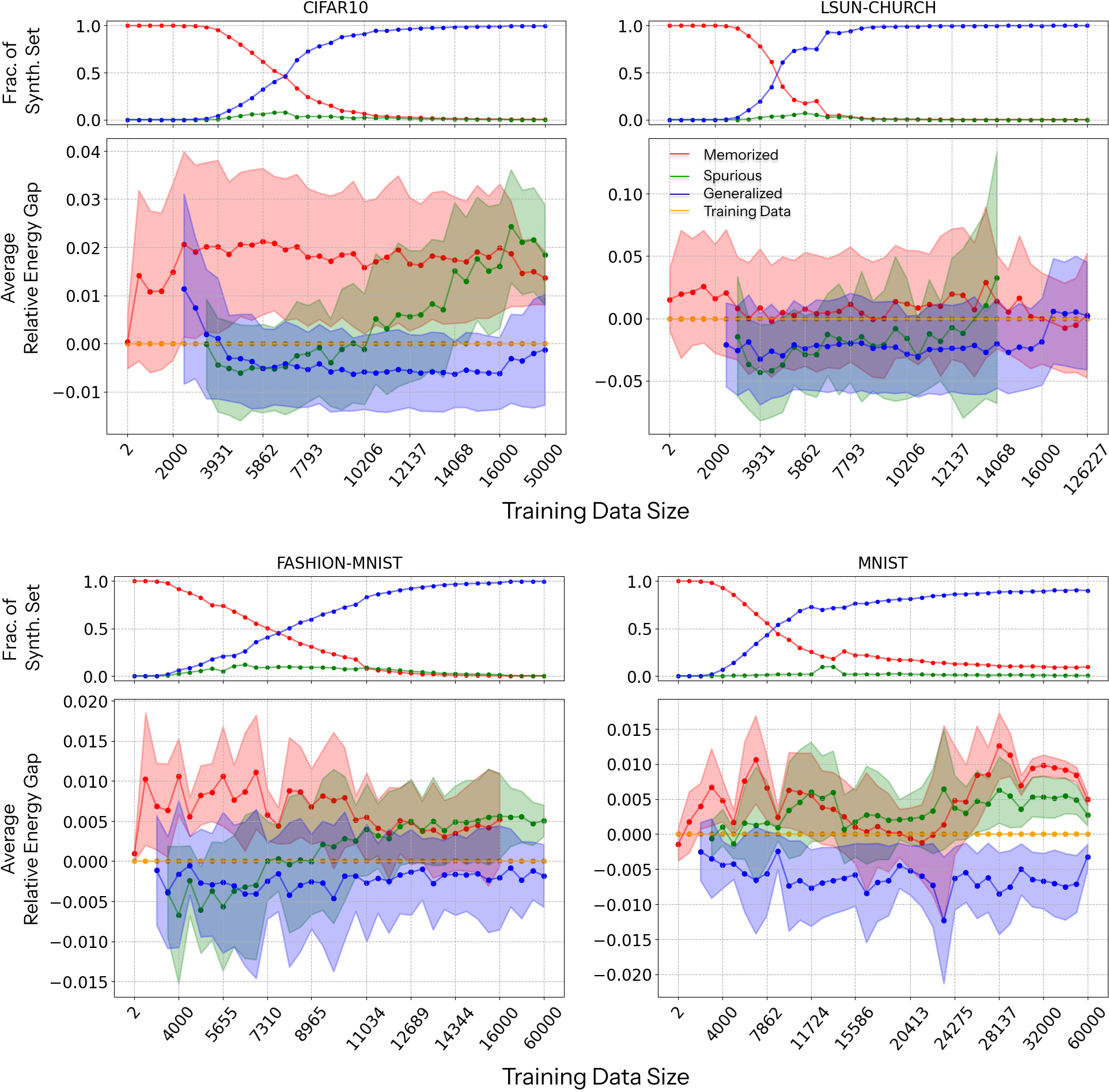}
\caption{
Average energy gap of each sample type and the training samples for the four datasets as the training data size $K$ grows, computed using Eq.~(\ref{eqn:relative-potential-discrete}). Gaps are measured relative to training samples, whose energy is set to zero since there is no gap to itself. A maximum of 512 samples per type is used at each $K$, or the full set is used instead if there are fewer than 512 points. Shaded regions show standard deviation of the gap values. With the exception of LSUN-CHURCH, the relative energy of the memorized samples matches that of the training data samples at $K = 2$, but quickly diverges as $K$ increases. Similarly, spurious samples' relative potential resembles that of the training data samples during the intermediate phase of the memorization-generalization phase. However, the potential quickly diverges as the later stages of generalization are initiated and consequently, the gap between generalized and training data samples' relative energy significantly decreases. 
}
\label{fig:all-relative-energy}
\end{figure}

To define the average relative energy gap, let $\tilde{u}_\text{target} \in \mathbb{R}^{M}$ be a set of $M$ relative potentials computed from using either a memorized, spurious, or generalized set and following Eq.~(\ref{eqn:relative-potential-discrete}). Please note that $M$ is capped at $512$ for the three sample types. Meanwhile, let $\tilde{u}_\text{data} \in \mathbb{R}^{M'}$ be the set of relative potentials computed from the training dataset at size $K$ where $M' = \min(K, 2048)$. The average relative energy gap is defined as 
\begin{equation}
\Delta (\tilde{u}_\text{target}, \tilde{u}_\text{data}) = \frac{1}{M} \sum^M_{i = 1} \big ( \tilde{u}_\text{target} \big )_i - \frac{1}{M'} \sum^{M'}_{j = 1} \big ( \tilde{u}_\text{data} \big )_j = \mu(\tilde{u}_\text{target}) - \mu(\tilde{u}_\text{data})
 \label{eqn:relative-energy-gap}
\end{equation}
Using Eq.~(\ref{eqn:relative-energy-gap}), we computed the average relative energy gap and recorded our results in Fig.~(\ref{fig:all-relative-energy}). The shaded regions reported in this figure represent the standard deviations computed from using $\tilde{u}_\text{target}$, and the average relative energy gap as the mean value. Since at some training data size $K$, there might be very few samples in one of the three sets, we opted not to include the relative energy gap of such sets, partly due to their large variance. Similar to the basin of attraction experiment, we determine whether a set is too small if the number of its samples is less than $0.1\%$ of its corresponding synthetic set. Finally, for our interpolation values $\bar{\alpha} = \{0, \dots, \frac{\pi}{2}\}$ used for Eq.~(\ref{eqn:relative-potential-discrete}), we generate $20$ points starting from 0 to $\frac{\pi}{2}$ using linear spacing. 

\par\addvspace{1em}
\noindent \textbf{Discussion}. The results for the relative energy gap across the memorization-generalization transition for all datasets are shown in Fig.~(\ref{fig:all-relative-energy}). As the training data size $K$ increases, the gap for memorized samples is initially close to zero and quickly diverges. Spurious samples' relative potential resembles that of the training data samples during the intermediate phase of the memorization-generalization phase, where their potential is the lowest among the three sample types. However, the potential of spurious states quickly diverges as the later stages of generalization are initiated, and consequently, the gap between generalized and training data samples' relative energies became the most narrowed among the three distinct sample types -- further emphasizing the changing in the energy landscape as the DM undergoes its transition from memorization to generalization as $K$ increases.

\vspace{1em}
\begin{algorithm}[H]
\caption{Compute the discretized relative energy between a target and a reference sample}
\label{alg:relative_energy}
\DontPrintSemicolon
\SetKwInOut{Input}{Input}
\SetKwInOut{Output}{Output}
\SetArgSty{textnormal}
\small
\Input{Target pattern $\rvx_1 \in \mathbb{R}^N$ and reference pattern $\rvx_2 \in \mathbb{R}^N$ (both samples at $t = 0$), score model $s_\theta$, interpolation values $\bar{\alpha}$, initial variance $\beta_0$}
\Output{Relative energy $\tilde{u}(\bar{\alpha}_L)$}
$\bar{\alpha} \gets \{ 0, \dots, \frac{\pi}{2} \}$ \tcp*{Linear spacing from $0$ to $\frac{\pi}{2}$}
$\Delta \alpha \gets \bar{\alpha}_1 - \bar{\alpha}_0$, \ $\tilde{u}(\bar{\alpha}_L) \gets 0$\;
\For{$\alpha \in \bar{\alpha}$}{
    $\tilde{\rvx} \gets \cos (\alpha) \rvx_1 + \sin (\alpha) \rvx_2$, \ $\rvv \gets - \sin(\alpha) \rvx_1 + \cos(\alpha) \rvx_2$ \tcp*{Interpolate}
    $\nabla_{\tilde{\rvx}} \log p_0(\tilde{\rvx}) \gets -s_\theta(\tilde{\rvx}, 0) / \sqrt{\beta_0}$\;
    $\mathbf{g} \gets \big( -\frac{1}{2} \beta_0 \tilde{\rvx} - \beta_0 \nabla_{\tilde{\rvx}} \log p_0(\tilde{\rvx}) \big) \odot \rvv \, \Delta \alpha$ \tcp*{Eq.~(\ref{eqn:vp-reverse-sde})}
    $\tilde{u}(\bar{\alpha}_L) \gets \tilde{u}(\bar{\alpha}_L) + \sum_{j=1}^{N} g_j$ \tcp*{Eq.~(\ref{eqn:relative-potential-discrete})}
}
\Return $\tilde{u}(\bar{\alpha}_L)$\;
\end{algorithm}

\par\addvspace{1em}
\noindent \textbf{Average Relative Energy Gap Results}. \Cref{fig:all-relative-energy} reports the average relative energy gap of each sample type across the memorization-to-generalization transition.

\clearpage

\subsection{Hardware Details}
The training of the models detailed in Table~(\ref{tab:models}) was done using NVIDIA Tesla V100 GPUs, where a single GPU was used to train each model. Each GPU has 32GB of memory and is linked with Power9 processors, clocking at 3.15 GHz maximum. Each CIFAR-10 model took roughly 18, 20, and 22 hours to train for different initial latent dimensions, e.g., 64, 96, 128, respectively. Meanwhile, each LSUN-CHURCH model requires roughly 48, 56, and 103 hours of training with respect to the model's initial latent dimension. In contrast, FASHION-MNIST and MNIST models require approximately 17 hours to be fully trained (per model). Please note that these durations are with respect to the required training iterations for each model and their corresponding dataset detailed in Table~(\ref{tab:models}). Also, note that we utilized the same hardware for our analyses.

%% file: chapters/ch4_language/ch4.tex
\providecommand{\cfig}[2]{\hyperref[#1]{\ref*{#1}#2}}
\providecommand{\Cat}{\text{Cat}}
\providecommand{\boldpi}{\boldsymbol{\pi}}
\providecommand{\dd}{\mathrm{d}}

\chapter{Coexistence of Memorization and Generalization\\in Language Diffusion Models}
\label{chapter:language-dm-am}
\blfootnote{Portions of this chapter have previously appeared as: Pham, B., Zaki, M. J., Ambrogioni, L., Krotov, D., \& Negri, M. (2026). Language diffusion models are associative memories capable of retrieving unseen data. In H.-Y. Lee, S. Dev, J. J. Li, \& D. Elliott (Eds.), \textit{Proceedings of the 2026 conference on empirical methods in natural language processing} (pp. 19377--19404). Association for Computational Linguistics. \url{https://doi.org/10.18653/2026.emnlp-main.835}}

\section{Introduction}
\label{sec:ch4-intro}
Generative diffusion models \citep{sohl-dickstein15} have set new standards for image and video generation \citep{ho2020denoising, song2019generative, song2021scorebased, rombach2022high}. Yet, alongside their remarkable generative power, these models exhibit a well-documented tendency to reproduce their training data \citep{somepalli2023diffusion, somepalli2023understanding, carlini2023, webster2023reproducible}. Although recent studies have addressed the interplay between memorization and generalization in diffusion models (DMs), they predominantly focus on the continuous image domain \citep{yoon2023diffusion, kadkhodaie2023generalization, biroli2024dynamical, kamb2024analytic, achilli2024losing, wen2024detecting, jeon2024understanding, pham2025memorization, achilli2025memorization}. Their mechanics in the discrete domain, particularly for language modeling, remain poorly investigated.\footnote{Code is available at \mbox{\url{https://github.com/Lemon-cmd/Associative-Memory-and-Language-Diffusion}}.} 

A particularly compelling instance of these questions arises in the context of large language models. For instance, \citet{brown2020language} and \citet{kojima2022large} have found that these models are capable of few-shot and even zero-shot capabilities, referring to the model's ability to perform novel tasks with minimal or even no task-specific examples, or alternatively the possibility of retrieving appropriate responses never encountered during training. The reasons behind the emergence of these capabilities remain unclear, and the push for models capable of both factual recall and creative behaviors makes it difficult to define and assess what \textit{generalization} means in this regime.

\begin{figure}[t]
     \centering
     \includegraphics[width=1.0\textwidth]{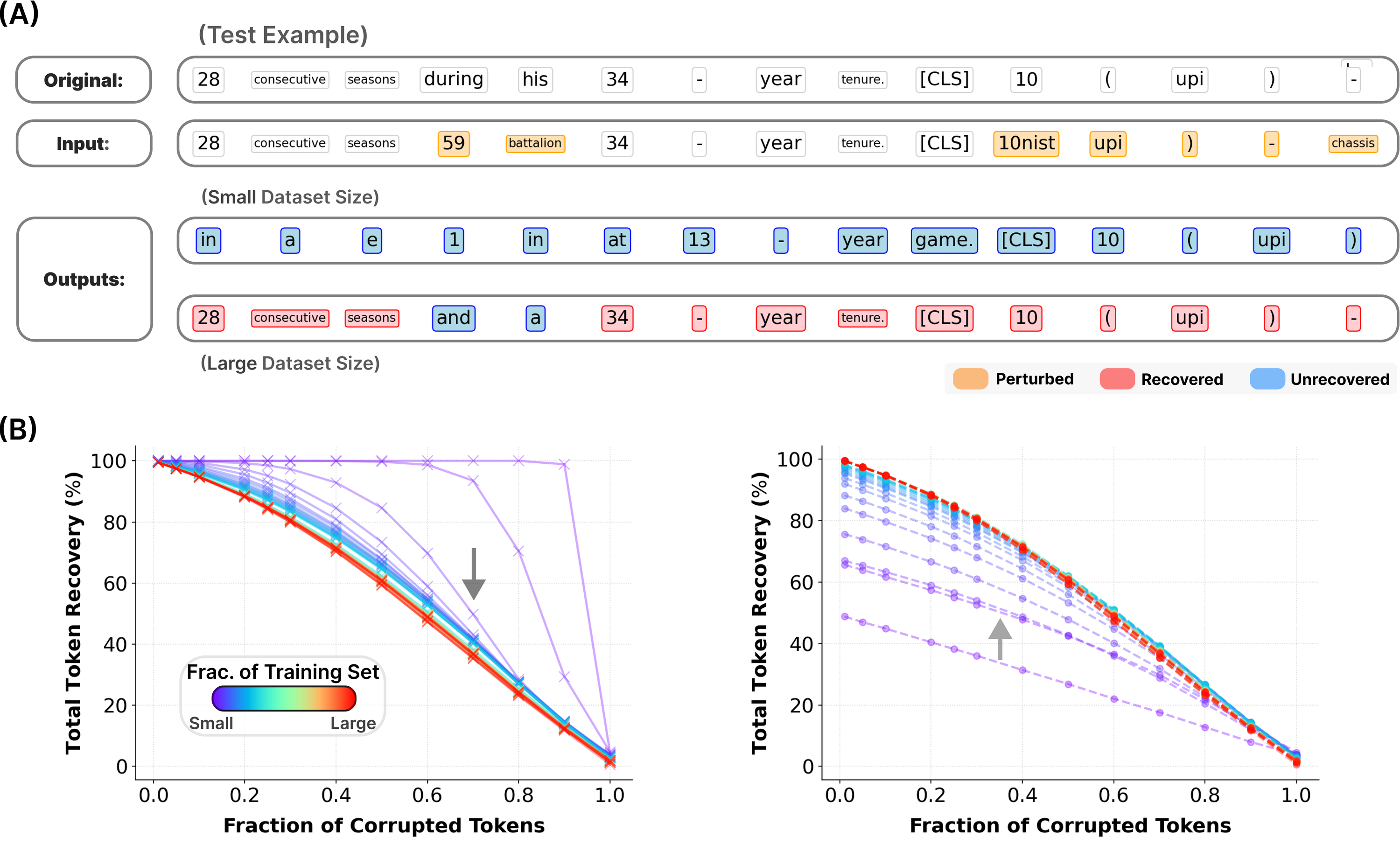}
     \caption{Basins around training examples shrink and basins around test examples expand as the training dataset size increases.
     (A) Textual examples showing two \textit{Tiny} ($\sim$24M) UDDMs' token recovery at noise level $t = 0.2$, where each is trained on two different training dataset sizes. With a small training dataset, the model fails to recognize unseen test sequences and alters all tokens. With a larger training set, the tokens of unseen sequences, however, become stable and recoverable by the sampling process. (B) Average total token recovery rates (\%), including both non-corrupt and corrupted tokens, for training (left) and test (right) sequences across varying corruption levels. Line colors indicate the fractions of the training dataset used (ranging from small to large). As data scales, the model’s ability to flawlessly recover training examples drops (indicating shrinking basins), while its recovery rate of unseen test examples improves (indicating expanding basins). The convergence of these rates at large training set sizes (red curves) marks the transition from memorization to generalization.
     Deterministic (greedy) sampling was used across these experiments to isolate from stochastic noise.
     }
    \label{fig:basin_shrinkage}
\end{figure}

Motivated by these phenomena, this chapter studies Uniform-based Discrete Diffusion Models (UDDMs) \citep{austin2021structured, campbell2024generative, gat2024discrete, sahoo2024simple} through the theoretical lens of Associative Memories (AMs) \citep{krotov2025modern}. Fundamentally, a generative system functions as an AM by reliably retrieving stored data points, by producing distinct basins around them \citep{Hopfield1982, gardner1988space}. Historically, attempting to overload these systems with too many data points leads to a catastrophic \textit{memory blackout}, where all meaningful attractors are destroyed \citep{amit1987statistical}. However, \citet{kalaj2024random} has recently revealed a counterintuitive regime: overloading an AM can instead trigger a generalization phase where new attractors spontaneously form near unseen examples from the underlying data distribution, while the memorization of the training data points persists. The desired capability of language modeling of having both factual recall and creative behaviors  resembles the co-existence of memory attractors and novel attractors in AMs. For this reason, it is useful to describe the generalization phase of UDDMs from the perspective of AMs, by focusing on the retrieval of seen and unseen data points. Notably, the co-existence of these two seemingly contradicting capabilities has been studied in an analytically tractable teacher-student setting \citep{farne2026rules}.

\begin{figure}[t]
    \centering
    \includegraphics[width=1.0\linewidth]{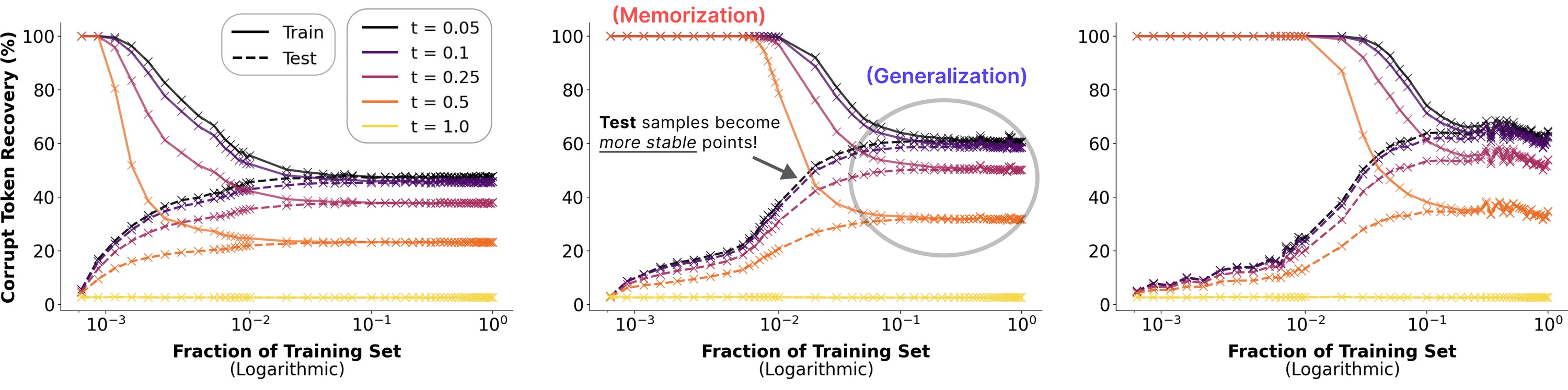}
    \caption{The convergence of corrupted training and test token recovery rates marks the phase transition from memorization to generalization. The plots display the average token recovery rate (\%) as the fraction of the training dataset grows, comparing the model's ability to denoise perturbed tokens (using stochastic sampling) from the training set (solid lines) against unseen test sequences (dashed lines) across various perturbation levels based on time $t$ on the LM1B dataset \citep{chelba2013one} for the \textit{Tiny} ($\sim$24M, left), \textit{Small} ($\sim$135M, middle), and \textit{Medium} ($\sim$384M, right) models. In the limited-data regime, the model exhibits memorization, perfectly recovering training samples while failing on test data. As the dataset expands, the recovery rates for training and test sequences converge to an identical recovery rate, demonstrating that unseen samples have become stable attractors. Notably, the model's size dictates the timing of this shift, where the \textit{Medium} model requires a significantly larger fraction of data to trigger the transition, effectively prolonging the memorization phase.}
    \label{fig:token_recovery_rate_combined}
\end{figure}

Typically, AMs are characterized with an explicit and well-defined energy function, as seen in Hopfield networks \citep{Hopfield1982, AmariS.I.1972} and Dense Associative Memories \citep{DAM2016, krotov2021large}. However, this reliance on an energy function is not strictly necessary to guarantee attractor dynamics. As shown by \cite{d2025pseudo}, conditional likelihood maximization alone produces basins around the data points. Importantly, dropping the energy requirement is conceptually essential for extending the AM framework to deep feed-forward architectures. There is no reason to expect that a generic feed-forward network can be written as an energy-based model, since this interpretation restricts the class of admissible architectures due to its constraints on the symmetry of the network's weights \citep{krotov2021hierarchical, et, kozachkov2025neuron}. On the contrary, the conditional likelihood structure already exists in many widely used architectures (such as the Transformer \citep{Attention2017}), making it a natural connection with AMs. 

\par\addvspace{1em}
\noindent \textbf{Contributions}. Leveraging conditional likelihood dynamics, this chapter establishes a connection between UDDMs and AMs by showing that attractor basins can be formed through conditional likelihood maximization alone (via pseudo-likelihood \citep{pseudo-likelihood}) and provide a benefit for AM. Specifically, in a simplified setting detailed in \Cref{sec:am_pseudo}, we establish that maximizing conditional likelihood implicitly enforces Hebbian learning \citep{Hebb1949} on data points while maximizing their classification margins, providing a formal bridge between AM and generative models (like UDDMs) which rely on conditional likelihood. Moreover, unlike previous works which connect continuous DMs and AMs \citep{hoover2023memory, ambrogioni2023search, pham2025memorization}, we extend their results to the discrete setting of language modeling. 

This theoretical link of conditional likelihood maximization and AM allows us to interpret UDDMs as AM systems and reveals a sharp memorization-to-generalization transition governed by the training dataset size. As the training set gets bigger, basins around training examples shrink while (partial) basins around unseen test examples expand, eventually converging to a regime where novel samples become stable attractors (see Figs.~(\ref{fig:basin_shrinkage}) and (\ref{fig:token_recovery_rate_combined})). During memorization, the UDDM fails to recognize tokens of unseen test sequences as stable points, frequently altering them during the reverse process. Once generalized, however, the model is likely to maintain these tokens, preserving them if unperturbed and successfully recovering them from partially corrupted sequences via the reverse process (see Figs.~(\cfig{fig:basin_shrinkage}{A}) and (\cfig{fig:unrecovered_recovered}{A}) for visual examples of test and training points' token recovery). 

Empirically, we detect this transition using \textit{token recovery rates} and \textit{conditional entropy} as complementary probes for token stability and the memorization-to-generalization transition. Crucially, token-level conditional entropy distinguishes between different token recovery behaviors, with successfully recovered tokens exhibiting near-zero entropy (see Fig.~(\ref{fig:unrecovered_recovered})). Meanwhile, sequence-level conditional entropy serves as a practical metric for detecting generalization, as the entropy distributions of training and synthetic sequences align perfectly once the transition is reached (see Fig.~(\ref{fig:seq_cond})). Notably, conditional entropy has also been found to be crucial for scaling laws in autoregressive language models \citep{cagnetta2026deriving}, suggesting the versatility and usefulness of this metric in language modeling. 

Finally, we validate our proposed metrics via the typical measurement of perplexity using an external critic model as commonly done in language modeling \citep{radford2019language}. We observe a strong alignment between these metrics, not just in the detection of memorization-to-generalization transition point, but also in the estimation of sample quality (see Fig.~(\ref{fig:perplexity})). The success of this validation indicates that our metrics are effective for assessing the generative capabilities of UDDMs, while also providing a principled geometric interpretation of such capabilities.

The remainder of this chapter is organized as follows. \Cref{sec:ch4-prelim} recalls the uniform-state discrete diffusion models we study. \Cref{sec:am_pseudo} establishes the connection between conditional likelihood maximization and AMs in a simplified setting. \Cref{sec:ch4-methods} describes our three experiments and the two metrics, token recovery rate and conditional entropy, used to probe the memorization-to-generalization transition. \Cref{sec:ch4-results} presents our results, and \Cref{sec:ch4-conclusion} concludes with a discussion of limitations. The experimental details and additional results are collected in \Cref{sec:ch4-supp}.

\section{Uniform-State Discrete Diffusion}
\label{sec:ch4-prelim}
We study the uniform-state discrete diffusion models (UDDMs) reviewed in \Cref{sec:discrete-diffusion}, specifically the formulation of \cite{sahoo2025duality}. A clean token $\rvx \in \mathcal{V}$ is corrupted by the forward process in \Cref{eqn:bg-discrete-forward} with the uniform prior $\boldsymbol{\pi} = \mathbf{1}/V$ over a vocabulary of size $V$, and the reverse process replaces the unknown clean token in the closed-form posterior of \Cref{eqn:bg-discrete-posterior} with the prediction of a network, $p_\theta(\rvx \,|\, \rvz_t) = \rvx_\theta(\rvz_t, t)$, as in \Cref{eqn:bg-discrete-reverse}. The parameters are trained with the NELBO objective in \Cref{eqn:bg-nelbo}.

By framing the network's prediction as a categorical probability distribution, the cross-entropy terms within $\mathcal{L}_\text{NELBO}$ naturally enforce a finite classification margin and therefore store training data points. As we will explore next, this reliance on cross-entropy provides a mathematical link between discrete diffusion dynamics and AM systems. Please see \Cref{sec:duo} for more details on this form of UDDMs, the reverse process (\ref{eqn:bg-discrete-reverse}), and the NELBO (\ref{eqn:bg-nelbo}).

\section{Associative Memories from Conditional Sampling}
\label{sec:am_pseudo}

To establish a connection between AMs and UDDMs, we rely on a specific aspect that emerges from modeling sequences of discrete variables in DMs: the \textit{core assumption}
is that the denoising process of a sequence $\rvz^{1:L}$ of length $L$ factorizes for each token $\rvz^\ell$ as 
\begin{equation}
    p^\theta_{s|t}(\rvz_s^{1:L} \mid \rvz^{1:L}_t) = \prod _{\ell=1}^L \psi^{\theta}_{s|t}(\rvz_s^\ell \mid \rvz^{1:L}_t), 
\end{equation}
where $\psi^\theta_{s|t}$ denotes the conditional probability:
\begin{equation}
\psi^\theta_{s|t} (\rvz^\ell_s \mid \rvz^{1:L}_t) = \text{Cat}\big(\rvz^\ell_s;\,f^\ell_\theta(\rvz^{1:L}_t) \big).
\label{eq:categorical_conditional_sampling}
\end{equation}
Here, $f_\theta^\ell (\mathbf{z}^{1:L}_t) = \mathrm{softmax}_V \big[ \beta(t) \mathbf{u}_\theta^\ell(\mathbf{z}_t^{1:L}) \big ]$, where $\mathbf{u}^\ell(\mathbf{z}_t^{1:L})$ maps the sequence $\rvz_t^{1:L}$ to raw logits via a diffusion transformer \citep{peebles2023scalable}, and the softmax is applied over the $V$ categories per position $\ell$. The inverse temperature $\beta(t)$ depends on the diffusion schedule $\alpha(t)$ and typically increases as $t\to0$. These conditional probabilities enter the cross-entropy terms in NELBO (\ref{eqn:bg-nelbo}). This reliance on cross-entropy establishes the connection to AMs, where minimizing this loss implicitly creates basins of attraction around data points during conditional sampling \citep{d2025pseudo}, and translates with minimal changes to UDDMs.

\subsection{Simpler Setting: Binary Variables and Linear Logits}
Consider an AM of $L$ binary neurons $\rvs^\ell \in \{\pm 1\}$ with a non-symmetric coupling matrix $\mathbf{W} \in \mathbb{R}^{L \times L}$, where its \textit{diagonal entries are zero}, we have the following deterministic update rule: 
\begin{equation}
\rvs^\ell_\tau  = \mathrm{sgn} \bigg ( \sum^L_{m = 1} \mathbf{W}^{\ell m } \; \rvs^m_{\tau + 1} \bigg ),
\label{eq:deterministic_update}
\end{equation}
where we adopted the notation of the discrete time $\tau$ running backwards to highlight the connection with the backward dynamics of DMs. 
Consider also a set of $P$ examples $\boldsymbol{\Xi} \in \{\pm 1\}^{P \times L}$, where $\rvx^{1:L} \in \boldsymbol{\Xi}$ is a binary vector of length $L$.

To build an AM, it is not sufficient to find an optimal coupling matrix $\mathbf{W}^*$ such that $\forall \rvx^{1:L} \in \boldsymbol{\Xi}$ is a fixed point of \Cref{eq:deterministic_update}, because it would not guarantee finite basins of attraction around the examples. To create such basins, a stronger condition is instead needed:
\begin{equation}
\rvx^\ell = \mathrm{sgn} \bigg ( \sum^L_{m = 1} \mathbf{W}^{\ell m} \rvx^m + \kappa \bigg ), \quad \, \forall \ell = 1, \dots, L
\label{eq:perceptrons_margin}
\end{equation}
where the classification margin $\kappa \in \mathbb{R}^+$ ensures that each fixed point is robust to a finite amount of variable flips from the deterministic update rule \citep{gardner1988space, Gardner1988OptimalSP, forrest1988content, benedetti2022supervised}. Larger $\kappa$ implies larger basins. Given the load $\gamma=P/L$, there exists a maximum margin $\kappa_{\mathrm{max}}(\gamma)$. As shown by \citet{Soudry2018} and \citet{Montanari2024NegativePerceptron}, training a Perceptron with the cross-entropy loss in the separable regime implicitly solves \Cref{eq:perceptrons_margin} with $\kappa=\kappa_{\mathrm{max}}(\gamma)$, suggesting how to produce large basins of attraction around the training examples.

\par\addvspace{1em}
\noindent \textbf{Conditional sampling}. To connect the cross-entropy loss with conditional sampling, it is useful to interpret this deterministic update (\ref{eq:deterministic_update}) as:
\begin{equation}
\rvs^\ell_\tau = 
\argmax_{\rvs^\ell} 
\; \psi_{\tau | \tau + 1} \big ( \rvs^\ell_\tau  \mid \rvs_{\tau + 1}^{1:L} ;  \mathbf{W}^\ell \big )
\end{equation}
where we select the most probable state under the conditional distribution (between position $\ell$ and its \textit{neighborhood} or $1 \dots L$ positions in the binary vector excluding $\ell$\footnote{Since the diagonal entries of our coupling matrix $\mathbf{W}$ are zeroed, position $\ell$ \textit{does not attend to itself}.}):
\begin{equation}
\psi_{\tau | \tau + 1} ( \rvs^\ell_\tau \; | \; \rvs_{\tau + 1}^{1:L};  \mathbf{W}^\ell) = \frac{\exp \Big ( \rvs^\ell_\tau f^\ell_{\mathbf{W}} (\rvs_{\tau + 1}^{1:L}) \Big )}{2 \cosh \Big ( f^\ell_{\mathbf{W}} (\rvs_{\tau + 1}^{1:L} ) \Big )},
\label{eq:conditional_probability_ising}
\end{equation}
where $f^\ell_{\mathbf{W}}(\rvs_{\tau + 1}^{1 : L}) = \beta \sum_{m = 1}^L  \mathbf{W}^{\ell m} \, \rvs^m_{\tau + 1}$ with a fixed inverse temperature $\beta$\footnote{\Cref{eq:conditional_probability_ising} is obtained using $\psi(\rvs^\ell | \rvs^{1:L}) \propto \exp (\beta \rvs^\ell \sum_{ m} \mathbf{W}^{\ell m} \rvs^m)$.}.
For binary variables, \Cref{eq:conditional_probability_ising} yields a logistic form. However, for generic categorical variables, $f_\ell(\rvs^{1:L})$ are logits inside $\mathrm{softmax}(\cdot)$ like that of   \Cref{eq:categorical_conditional_sampling}. 

\par\addvspace{1em}
\noindent \textbf{Conditional-likelihood maximizes classification margins}. \Cref{eq:conditional_probability_ising} suggests a way to train the model: by minimizing the following loss function, the negative logarithm of the conditional-likelihood that is also called pseudo-likelihood \citep{pseudo-likelihood}:
\begin{equation}
\begin{split}
\mathcal{L}(W) &= -\frac{1}{P} \sum_{\rvx \in \Xi} \log \prod_{l=1}^{L} \psi(\rvx^{l} | \rvx^{1:L}; \mathbf{W}^{l}) \\
&= -\frac{1}{P} \sum_{\rvx \in \Xi} \sum_{l=1}^{L} \Big[\rvx^{l} f_{\mathbf{W}}^{l}(\rvx^{1:L}) - \log 2 \cosh\big(f_{\mathbf{W}}^{l}(\rvx^{1:L})\big) \Big]
\end{split}
\label{eqn:pseudo-likelihood}
\end{equation}
It has been shown by \citet{d2025pseudo} that the above objective (\ref{eqn:pseudo-likelihood}) produces \textit{basins of attraction around the training data points} in the dynamics of the conditional sampling. 
Unlike classical AM, the couplings $\mathbf{W}$ induced by conditional likelihood do not need to be symmetric, and therefore \textit{no explicit global energy function is required}. The existence of attractor-like behavior follows directly from the structure of the conditional probabilities.

To have an intuition on why the loss (\ref{eqn:pseudo-likelihood}) promotes classification margins, we can derive it with respect to the coupling matrix $\mathbf{W}$:
\begin{equation}
\begin{split}
\frac{\dd \mathcal{L}(\mathbf{W})}{\dd \mathbf{W}^{lm}} \propto -\frac{1}{P} \sum_{\mathbf{x}\in\Xi} \; \underbrace{\rvx^l \rvx^m}_{\text{Hebbian}} \Big [\underset{\text{Penalty}}{\underbrace{1 - \tanh(M^\ell (\rvx^{1:L}))}} \Big ].
\end{split}
\label{eqn:grad_pseudo}
\end{equation}
Here, we denote the local classification margin as $M^\ell(\rvx^{1:L}) = \rvx^\ell f^\ell_{\mathbf{W}}(\rvx^{1:L})$. The Hebbian term is factored out via identity $\tanh (f^\ell_{\mathbf{W}}(\rvx^{1:L})) = \tanh(\rvx^\ell M^\ell(\rvx^{1:L})) = \rvx^\ell \tanh (M^\ell(\rvx^{1:L}))$, given $\rvx^\ell \rvx^\ell = 1$. This yields two distinct gradient terms: a classical Hebbian learning \citep{Hebb1949} component for pattern storage and a margin-modifying component. Because the gradient penalty $1 - \tanh(M^\ell (\rvx^{1:L})) \approx 2 e^{- 2 M^\ell(\rvx^{1:L})}$ decays exponentially for correctly classified patterns with wide margins, the objective concentrates learning on samples with the smallest margins. The loss is thus minimized when all margins are maximized, driving weight magnitudes to diverge in the separable regime \citep{Soudry2018}. 

Furthermore, we verify numerically that the optimization of loss (\ref{eqn:pseudo-likelihood}) leads to basins around training examples in \Cref{sec:numerical_validation_of_PL}. Specifically, we confirm the results of \cite{Soudry2018} by reporting the lower bound computed from the idealized $\kappa_{\max}(\gamma)$ and the one computed from the empirically achieved $\kappa_{\text{achieved}}(\gamma)$ in \Cref{fig:basin_verification}. Notably, the shrinking of basins demonstrated in \Cref{fig:token_recovery_rate_combined} for UDDMs is also exhibited within the simple model trained using pseudo-likelihood (\ref{eqn:pseudo-likelihood}), supporting that the fundamental AM properties in UDDMs do not depend on the NELBO objective but on the underlying conditional-likelihood structure.

\par\addvspace{1em}
\noindent \textbf{Main differences with UDDMs.}~The above objective (\ref{eqn:pseudo-likelihood}) is directly related only to the term $\mathcal{L}_\text{reconstruction}$ in \Cref{eqn:bg-nelbo}, since AMs traditionally rely on a fixed temperature rather than annealing the temperature during its dynamics. By following this analogy, the reverse diffusion process can be interpreted as a stochastic AM retrieval dynamics for categorical variables where we also anneal the temperature, similarly to the continuous setting \citep{ambrogioni2023search, pham2025memorization}. From this perspective, we suspect that the additional terms in \Cref{eqn:bg-nelbo} are useful to enlarge basins of attraction when conditional probabilities are parametrized with deep architectures (like a transformer), but we leave this study for future work.

\section{Methods}
\label{sec:ch4-methods}

As shown in \citet{d2025pseudo} and \citet{kalaj2024random},
the AM produced by conditional-likelihood (\ref{eqn:pseudo-likelihood}) is capable of generalization: when trained on a sufficiently large dataset, the system creates new attractors that are strongly correlated with both training and test examples. In this section, taking inspirations from these results, we can ask whether similar regimes of memorization and generalization, found in AMs, can appear in UDDMs. Consequently, we design three experiments of increasing realism from the UDDM perspective, progressively bridging the gap between the AM retrieval setting and the standard generative process of a UDDM. Across these experiments, we explore the memorization-to-generalization transition as a function of the training dataset size and model scale.

\par\addvspace{1em}
\noindent \textbf{Experiment 1: Deterministic retrieval from reference examples}. To emulate the zero-temperature retrieval dynamics of an AM without the confounding effects of stochastic generation noise, we establish a specialized baseline setting. As reported in Fig.~(\ref{fig:basin_shrinkage}), we initialize the reverse process from corrupted training and test examples, and replace the standard stochastic sampling with deterministic greedy dynamics. At each step, we take the $\argmax$ of the conditional probabilities rather than sampling from them. In this setting, the perturbation level is decoupled from the diffusion time, which is always set to $t=1$ while the fraction of corrupted input tokens is varied, allowing us to cleanly measure the token recovery rate as a function of the input noise level. 

\par\addvspace{1em}
\noindent \textbf{Experiment 2: Stochastic retrieval from reference examples}. To understand how the standard diffusion dynamics alter the stability and structural boundaries of these basins under realistic operating conditions, we introduce stochasticity back into the UDDMs. In Figs.~(\ref{fig:token_recovery_rate_combined}) and (\ref{fig:unrecovered_recovered}) we retain the initialization from training and test examples, but restore the standard stochastic reverse dynamics. The perturbation level is now tied to the diffusion time $t$ as in the typical UDDM scheme, where each value of $t$ determines both the fraction of corrupted tokens and the time step at which the reverse process is initialized, providing a dynamic look at token recovery across varying levels of input noise.

\par\addvspace{1em}
\noindent \textbf{Experiment 3: Standard generative process}. Finally, we run the full standard reverse dynamics starting from \textit{noise}, independent of any reference sequences, e.g., training or test sequences, in Figs.~(\ref{fig:seq_cond}) and (\ref{fig:perplexity}). Because ground-truth samples are unavailable, token recovery rate is no longer a viable metric. Instead, we use conditional entropy (\ref{eqn:cond-ent}) as our metric. In Fig.~(\ref{fig:perplexity}) we validate the performance of UDDM across the memorization-to-generalization transition by measuring the perplexity of the generated sequences with GPT-2 Large \citep{radford2019language} used as the critic model.

We analyze these experiments on the LM1B dataset \citep{chelba2013one} via two metrics, as a function of the training dataset size: \textit{token recovery rate} and \textit{conditional entropy}, which serve as proxies for the stability and geometry of the attractors, respectively. For more details and results, please refer to \Cref{sec:experimental-details} and \Cref{sec:add-results}, respectively.

\par\addvspace{1em}
\noindent \textbf{Corrupt Token Recovery}. We define the corrupt token recovery rate as the accuracy with which the model recovers a target or original sequence of length $L$, either a training or a test example, via applying the reverse process on a noisy sequence defined at time $t \in (0, 1]$: 
\begin{equation}
    R(\rvx^{1 : L}, \hat{\rvx}^{1:L}) = \frac{1}{|\mathcal{M}|} \sum_{j \in \mathcal{M}} \delta (\rvx^j, \hat{\rvx}^j \big ),
\label{eqn:token-recovery-rate}
\end{equation}
where $\rvx^{1 : L}$ is the original sequence, $\hat{\rvx}^{1 : L}$ is the recovered sequence after running a denoising process (either greedy or stochastic), $\mathcal{M} = \{ j : \delta(\rvx^j, \rvz_t^j) = 0\}$ is the set of indices in the input sequence $\rvz_t^{1: L}$ denoting the positions where the tokens of $\rvx^{1:L}$ have been changed after applying the forward process (\ref{eqn:bg-discrete-forward}), and $\delta(\cdot, \cdot)$ denotes the Kronecker delta function. 

\par\addvspace{1em}
\noindent \textbf{Conditional Entropy}. 
While token recovery rates indicate whether the system successfully returns to an attractor, UDDMs also provide direct access to conditional likelihood, allowing us to probe the local geometry and sharpness of these basins. Here, the conditional entropy of an individual token $\rvx^\ell$ in a sequence, given its associated perturbed sequence $\rvz_t^{1 : L}$ at some time $t$ is defined as:
\begin{equation}
\begin{split}
\mathcal{H}(\rvx^\ell | \rvz_t^{1 : L}) = -\sum^V_{k = 1} \big [\rvx_\theta^\ell(\rvz_t^{1 : L}, t) \big ]_k \log \big [ \rvx_\theta^\ell(\rvz_t^{1 : L}, t) \big ]_k,
\end{split}
\label{eqn:cond-ent}
\end{equation}
where $\rvx_\theta^\ell(\rvz_t^{1: L}, t)$ is the output of the diffusion transformer for position $\ell$ of the sequence over a vocabulary of size $V$. Meanwhile, the conditional entropy of a sequence is the sum of each token's conditional entropy for all positions from 1 to $L$, i.e., $\mathcal{H}(\rvx^{1: L} \; | \; \rvz_t^{1: L}) = \sum^L_{\ell = 1} \mathcal{H}(\rvx^\ell \; | \; \rvz_t^{1 : L})$. Please refer to \Cref{sec:entropy_curvature} for our discussion on the relationship between conditional entropy and energy curvature (or sharpness).

\section{Results}
\label{sec:ch4-results}

Our empirical analysis demonstrates that, as the training set size increases, basins around training examples shrink and basins around test examples expand (see Fig.~(\ref{fig:basin_shrinkage})). Changes in the structure of basins are reflected in token recovery rates: for larger datasets the token recovery rate for training examples drops (see Fig.~(\ref{fig:token_recovery_rate_combined}), \textit{solid} lines), while the recovery rate for unseen test samples increases (\textit{dashed} lines in Fig.~(\ref{fig:token_recovery_rate_combined})). For a certain size onward, the retrieval behaviors of both seen and unseen sequences converge to an identical recovery rate, mirroring the convergence of training and test basins and completing the transition from memorization to generalization. Scaling up the model size delays this transition, consistently with continuous DMs \citep{yoon2023diffusion, pham2025memorization}.  

\begin{figure}[t]
     \centering
     \includegraphics[width=1.\textwidth]{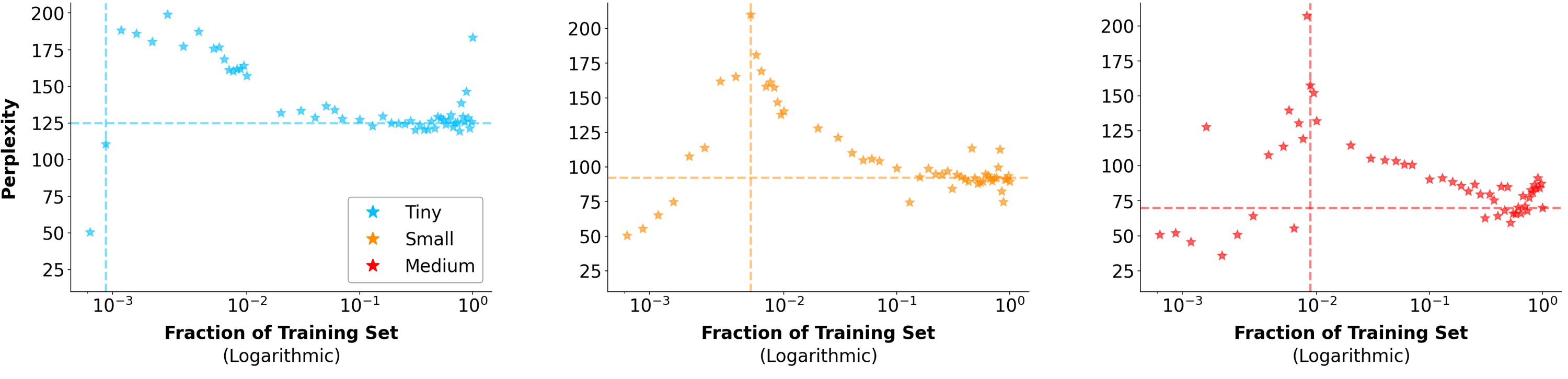}
     \caption{Perplexity trends during the memorization-to-generalization transition. Evaluation of the \textit{Tiny} (left), \textit{Small} (middle), and \textit{Medium} (right) UDDMs as the fraction of the training set increases. 
     The vertical dashed lines roughly illustrate the transition point where the model switches from memorization to generalization. Perplexity starts at a very low value during the memorization regime, increases as the training dataset size grows, and it then decreases to a lower value upon reaching generalization.
     The horizontal dashed lines highlight the converging perplexity value, showing that scaling up the model leads to improved perplexity at the end of this transition.
     The precise position of each vertical line is the last point where the recovery rate is $100\%$ for training tokens in \Cref{fig:token_recovery_rate_combined}. 
     Perplexity is computed using GPT-2 Large \citep{radford2019language} as the critic model.}
     \label{fig:perplexity}
\end{figure}

\begin{figure}[t]
    \centering
    \includegraphics[width=1.0\linewidth]{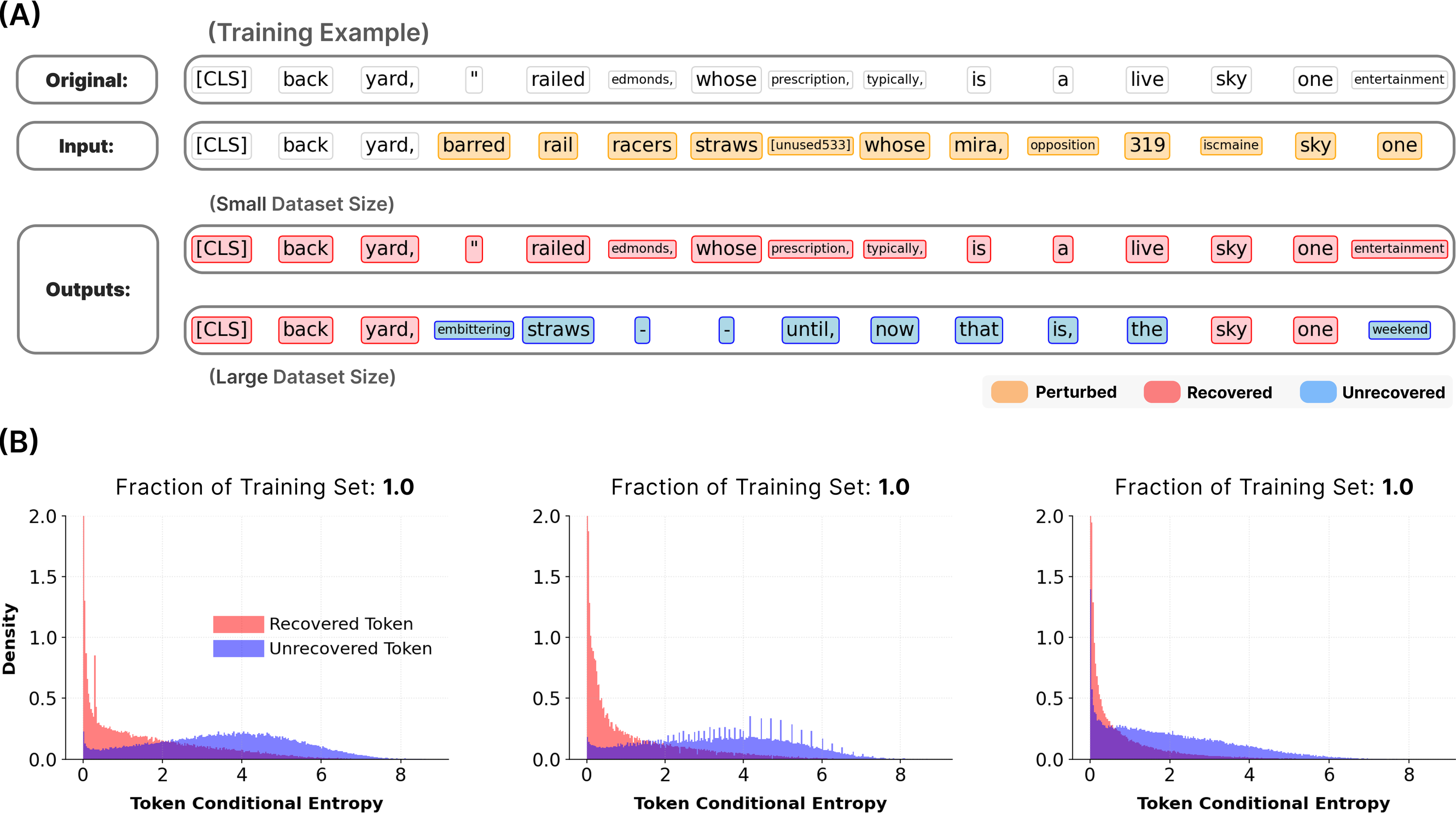}
    \caption{
    Token-level conditional entropy highlights different token recovery behaviors. (A) Textual examples showing how two \textit{Medium} UDDMs, trained on two different training dataset sizes, recover corrupted training sequences at noise level $t = 0.5$. The model trained on a small training set perfectly memorizes and restores the original text. In contrast, the model trained on a large training set recovers some original words but actively alters others to synthesize a novel sentence. (B) Density histograms of conditional entropy for individual tokens from \Cref{fig:token_recovery_rate_combined} for the \textit{Tiny} (left), \textit{Small} (middle), and \textit{Medium} (right) models. Successfully recovered tokens consistently demonstrate much lower entropy, indicating higher model confidence. While this gap between recovered and unrecovered tokens narrows in larger models, a surprising number of highly stable, low-entropy tokens persist even during the generalization phase. The $y$-axis is clipped to better contrast these two token types.
    }
    \label{fig:unrecovered_recovered}
\end{figure}

\begin{figure}[t]
    \centering
    \includegraphics[width=1.\linewidth]{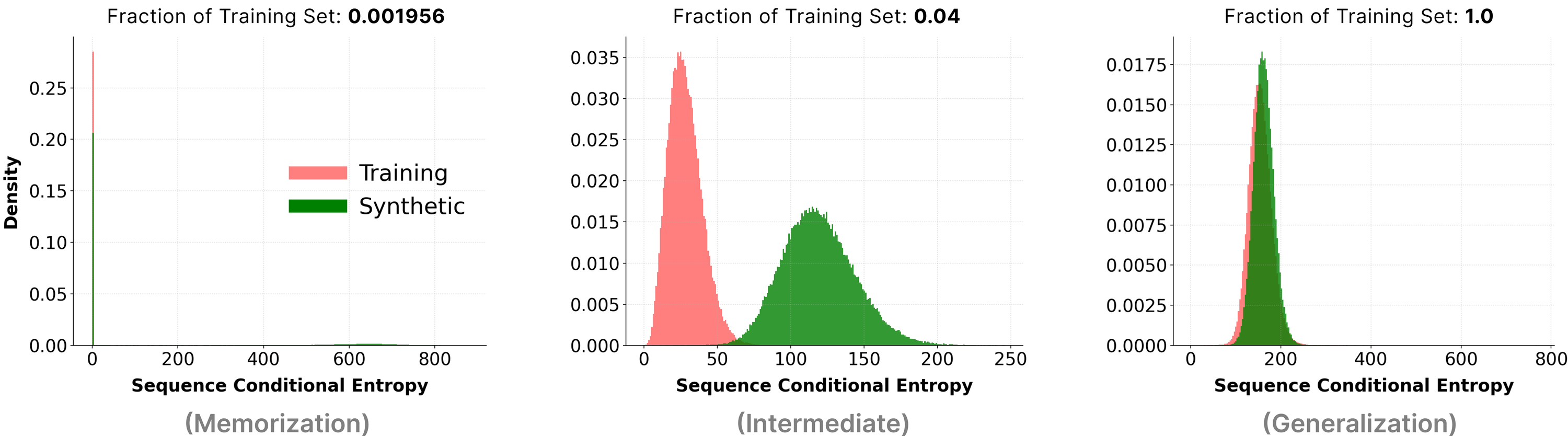}
    \caption{Sequence conditional entropy highlights the memorization to generalization transition. The histograms compare the sequence entropy distributions of training and synthetic (or generated) sequences respectively, for three selected fractions of the training set for the \textit{Medium} UDDMs. At a very low fraction of the training set, the conditional entropy of training sequences is near zero, aligning very well alongside the conditional entropy of many generated sequences, indicating the regime of memorization. However, as the fraction of the training set reaches the total, the synthetic and training distributions of the sequence conditional entropy converge, signifying generalization. See \Cref{sec:add-results} for the full range of histograms across various model sizes.
    }
    \label{fig:seq_cond}
\end{figure}

For these results, please note that a high recovery rate on test sequences does not imply that a basin has formed around them, since it could also be explained just by a correct conditional prediction. We keep the naming of `basin' for two reasons. First, the denoising is still non-trivial: the rate remains well above the injected noise, so the dynamics returns close to the original sequence rather than to an unrelated one. Second, a sentence admits many valid completions, and a rate below $100\%$ is what we should expect, since valid alternatives are counted as unrecovered tokens. Both are compatible with the attractor being flat, with the flat directions corresponding to the equivalent formulations of a given sentence, which also accounts for the dependence on the noise level: without corruption the greedy reverse process initialized on a test sequence stays on it, giving $100\%$ recovery, while any finite corruption drops it below $100\%$ (see Fig.~(\cfig{fig:basin_shrinkage}{B})), because the state can be moved to a different point of the same basin. Consistently, the unrecovered tokens in the generalization phase are usually grammatically and semantically valid (see Figs.~(\cfig{fig:basin_shrinkage}{A}), (\ref{fig:train_recovery_examples}), and (\ref{fig:test_recovery_examples})).

Moreover, as shown in \Cref{fig:unrecovered_recovered}, successful token recovery is characterized by near-zero conditional entropy, whereas unrecovered tokens exhibit higher and non-zero conditional entropy value. Interestingly, a distinct fraction of highly stable and low-entropy tokens persists even at the later generalization phase, suggesting that there are tokens which are very stable and less likely to change through the generative process. Using the average \textit{sequence-level} conditional entropy of training and generated sequences, we observe a conditional \textit{entropy gap} at the intermediate regime of the transition (see Fig.~(\ref{fig:seq_cond})): at very low fractions of the training set, the conditional entropy distributions of training and generated sequences coincide and peak near zero. In the intermediate regime, these distributions however separate from each other until the training set size is large enough, and once again overlap each other (but at \textit{non-zero} values of conditional entropy). 

Perplexity benchmarks using GPT-2 Large validate these phase transitions (see Fig.~(\ref{fig:perplexity})): we observe that the region where perplexity increases (to the \textit{left} of the transition) corresponds to the memorization phase, which suggest that the perplexity is initially low due to the model's duplication of training sequences. Conversely, the region where perplexity decreases (to the \textit{right} of the transition) corresponds to the generalization phase where basins around test sequences start forming, suggesting that these novel basins are related to the generative capabilities of the model. 

The correspondence between token recovery rate in \Cref{fig:token_recovery_rate_combined} and perplexity in \Cref{fig:perplexity} also holds for the rightmost region where the perplexity plateaus: the length of the plateaus is similar in both figure. The plateaus of lower perplexity correspond to the converging token recovery rates, see \Cref{fig:ppl_combined} in \Cref{sec:add-results} for a combined version of Figs.~(\ref{fig:token_recovery_rate_combined}) and (\ref{fig:perplexity}). This alignment suggests that token recovery rate captures the same underlying phenomenon: it identifies the generative regime without an external model and offers a geometrical interpretation, where generalization is defined by the evolution of basins around training and test sequences.

Note that recovery under denoising is not the same as training-data extraction \citep{carlini2023, webster2023reproducible}. Our metrics probe the stability of the dynamics near a given sequence, and carry no information on the probability of reaching a specific training or test sequence starting from noise. The non-monotonic perplexity of Fig.~(\ref{fig:perplexity}) is only indirect evidence in this direction: the low perplexity at small fractions is compatible with the model duplicating training sequences, while the plateau after the transition we can only interpret as good in-distribution generation.

\section{Conclusion}
\label{sec:ch4-conclusion}
By framing UDDMs as AMs, we show that an explicit energy function is not required to guarantee attractor dynamics. Instead, basins can form through conditional likelihood maximization alone, which reveals a sharp memorization-to-generalization transition governed by the training set size. As training dataset scales up, basins around training examples shrink while basins around unseen test sequences expand until they become indistinguishable. This evolution allows the model to recover both seen and unseen sequences from perturbations, showing a form of co-existence of memory and creativity. 

Furthermore, token recovery rate and conditional entropy serve as complementary geometric probes of this transition: conditional entropy vanishes during memorization but remains finite during generalization, providing a measurable entropy gap between training and synthetic sequences that narrows with model scale. Larger models delay this transition, requiring more data to generalize before both metrics stabilize past the phase boundary. In this picture the attractors in the generalization phase are flat, where the flat directions correspond to equivalent formulations of a sentence. We leave the study of these directions, which requires assessing semantic equivalence, to future work.

Finally, a distinct fraction of highly stable, low-entropy tokens persists even during the generalization phase, when novel text is being synthesized. We hypothesize that these tokens correspond to rigid syntactic structures, predictable function words, or common idioms that must remain stable to preserve linguistic fluency, and that act therefore as micro-attractors within an otherwise novel sequence. This leaves an interesting question for us and the community to explore.

\par\addvspace{1em}
\noindent \textbf{Limitations}. This chapter opens some directions for future investigation. First, how good the new proposed metrics are in practically relevant cases remains an open question: while we showed a promising consistent picture with the perplexity measurement, an extensive analysis of how token recovery rate and conditional entropy correlate with standard evaluation metrics would be needed to consolidate their validity as proxies for generalization. Second, UDDMs provide a convenient and tractable setting for this study, but extending these ideas to regimes where factual recall appears in practice would require significantly larger models, up to and including large language models, which would introduce conceptual and practical challenges beyond the scope of this chapter.

\FloatBarrier
\input{chapters/ch4_language/ch4_supp}

%% file: chapters/ch4_language/ch4_supp.tex
\section{Supplementary Material}
\label{sec:ch4-supp}
This section collects the experimental details and additional results that support the findings of this chapter.

\subsection{Additional Details on Memorization to Generalization}
\label{sec:experimental-details}

\begin{table}[t]
\centering
\caption{A table showing hyperparameters of the UDDMs for the \textit{Tiny}, \textit{Small}, and \textit{Medium} sets. These variables are obtained from \citet{sahoo2025duality}. Scale by sigma indicates that the model takes the inverse temperature or the appropriate diffusion scheduling parameter at time $t$ instead of the typical approach of conditioning on $t$.}
\label{tab:model_configs}
\begin{tabular*}{\textwidth}{@{\extracolsep\fill}lccc}
\toprule
& \multicolumn{3}{@{}c@{}}{\textbf{Model Type}} \\ \cmidrule{2-4}
\textbf{Hyperparameters} & Tiny & Small & Medium \\
\midrule
Hidden Size & 256 & 768 & 1024 \\ 
Conditioning Size & 128 & 128 & 128\\ 
Length & 128 & 128 & 128 \\ 
Num. of Blocks & 8 & 12 & 24 \\ 
Num. of Heads & 8 & 12 & 16 \\
Scale by Sigma & True & True & True \\ 
Dropout & 0.1 & 0.1 & 0.1 \\
\bottomrule
\end{tabular*}
\end{table}

\par\addvspace{1em}
\noindent \textbf{Setup}. For the experiments in \Cref{sec:ch4-results}, we trained three sets of UDDMs, labeled as \textit{Tiny}, \textit{Small}, and \textit{Medium}, utilizing the code base and approach of \citet{sahoo2025duality}\footnote{The code-base used in our experiments is publicly available at \href{https://github.com/s-sahoo/duo}{https://github.com/s-sahoo/duo} under the Apache 2.0 license, and we use it solely for research purposes consistent with its intended use.
}. For more details on this variant of UDDMs, please refer to the discussion in \Cref{sec:duo} below. Meanwhile, the backbone of our trained UDDMs is the diffusion transformer architecture from \cite{peebles2023scalable}. The configurations of our three variations of UDDMs are described in \Cref{tab:model_configs}\footnote{The LM1B dataset \citep{chelba2013one} is publicly available at \href{https://www.statmt.org/lm-benchmark/}{https://www.statmt.org/lm-benchmark/} and is intended for research purposes in statistical language modeling, which is consistent with our use.}.

Meanwhile, there are a total of $162$ models, or $54$ models for each of the three UDDM sets we have trained. All models are trained up to 1 million training iterations following the procedures detailed in \citet{sahoo2025duality}. For the selection of the fraction of the training dataset sizes, we initially start with the fraction $n = 10^{-2}$ and increment it by $ \Delta n = 0.03$ all the way to the full dataset. However, to further magnify the memorization phase, we train more points using linear spacing (of $17$ points, inclusively) starting at $10^{-4}$ to $10^{-2}$, and another set of points using linear spacing (of $7$ points, inclusively) starting from $10^{-2}$ to $0.07$. Lastly, our models are trained on the LM1B dataset \citep{chelba2013one}, using \texttt{GPT-2} tokenizer \citep{radford2019language}, where our model handles the block size (or sequence length) of $128$, and all of them are initialized from the same random seed.

\par\addvspace{1em}
\noindent \textbf{Token Recovery Rate}. To obtain the results in \Cref{fig:token_recovery_rate_combined}, we utilized our trained models, from the three sets corresponding to \textit{Tiny}, \textit{Small}, and \textit{Medium} UDDMs. Here, we computed the analysis of randomly chosen $5 \times 10^{3}$ samples belonging to their respective training set (in accordance with their fraction of training dataset size) and also $5 \times 10^{3}$ sequences from the test set. We performed the perturbation of these sequences using the forward process (\ref{eqn:forward-duo}) at time $t$, and run the reverse process (\ref{eqn:bg-discrete-reverse}) starting at or near the same time $t$ back to a small terminal time $\epsilon = 10^{-5}$. Finally, to measure our recovery rate, we applied \Cref{eqn:token-recovery-rate}, which measures the rate of perturbed tokens being recovered. 

Meanwhile, for \Cref{fig:basin_shrinkage}, we are interested in the shrinkage of the basins of our training and unseen test sequences. To obtain a more comprehensive view of this shrinkage without worrying about the stochastic noise added during each reverse process (\ref{eqn:bg-discrete-reverse}) step as $t \to 0$, we utilized the greedy or deterministic sampling dynamics instead. Specifically, in \Cref{eqn:bg-discrete-reverse}, to obtain $\rvz_s$ as we traverse back in time, we are effectively sampling from a categorical distribution $\text{Cat}(\cdot)$. To convert this stochastic process to a deterministic one, we simply replaced $\text{Cat}(\cdot)$ with the $\argmax (\cdot)$ operation. Similarly to the previous experiment, we applied perturbation using the forward process (\ref{eqn:forward-duo}), and utilized the greedy process to denoise our sequences. The timesteps we used are $\{0.1, 0.2, 0.3, \dots, 1.0\}$ and the same values of $t$ we used in \Cref{fig:token_recovery_rate_combined}.

\par\addvspace{1em}
\noindent \textbf{Conditional Entropy}. 
In \Cref{fig:unrecovered_recovered}, we computed the token conditional entropy following \Cref{eqn:cond-ent} for unrecovered and recovered tokens at $t = 10^{-5}$ identified in \Cref{fig:token_recovery_rate_combined} using $5 \times 10^{3}$ samples (per training and per test sequences), respectively. Meanwhile, for the results in \Cref{fig:seq_cond} and \Cref{fig:entropy_over_time}, we generated a synthetic set of $10^{5}$ samples for each of the trained models and performed our analyses -- alongside using at most $10^{5}$ sequences belonging to the training set that each model was trained with. It is important to note that the results in \Cref{fig:entropy_over_time} are the average conditional entropy computed across sequences and their respective tokens, while \Cref{fig:seq_cond} showcases the sequence conditional entropy, which is the sum of all of the tokens' conditional entropy in a given sequence. Please see Figs.~(\ref{fig:additional_train_synth_hist_tiny})-(\ref{fig:additional_train_synth_hist_medium}) for the full histograms of conditional entropy on training versus generated samples, and Figs.~(\ref{fig:additional_recovered_unrecovered_hist_tiny})-(\ref{fig:additional_recovered_unrecovered_hist_medium}) for more conditional entropy histograms of the unrecovered versus recovered tokens shown in \Cref{fig:token_recovery_rate_combined}.

\par\addvspace{1em}
\noindent \textbf{Textual Examples}. In Figs.~(\ref{fig:basin_shrinkage}) and (\ref{fig:basin_shrinkage_extra}), we are interested in studying the stability of unseen test tokens across different perturbation levels (although we are much more interested in the case where there is little noise for these experiments). For the textual examples illustrated in Figs.~(\cfig{fig:basin_shrinkage}{A}) and (\cfig{fig:unrecovered_recovered}{A}), we used the models that have been trained with the fractions of the training dataset: $0.000719$ and $1.0$. Although these two figures utilize different sampling dynamics (i.e., greedy versues stochastic), in general we do not spot much differences between the greedy and stochastic processes for our token recovery experiments. Please see Figs.~(\cfig{fig:unrecovered_recovered}{A}), (\ref{fig:train_recovery_examples}), and (\ref{fig:test_recovery_examples}) for the textual examples that collected using the typical stochastic process.

\par\addvspace{1em}
\noindent \textbf{Perplexity}. Similarly to our results on conditional entropy, we utilized the same sets of $10^{5}$ generated sequences belonging to each trained UDDM, respectively. The code to compute perplexity is also from \citet{sahoo2025duality}. Keep in mind, the reported conditional entropy results are computed from our UDDMs, unlike perplexity, which requires the reference or critic model of the pre-trained GPT2-Large \citep{radford2019language} as mentioned previously.

\par\addvspace{1em}
\noindent \textbf{Hardware}. The training of the UDDMs are done using NVIDIA Tesla V100 GPUs. Each GPU has 32GB of memory and is linked with Power9 processors, clocking at 3.15 GHz maximum. For each model, we used 4 GPUs and an effective total (or global) batch size of $512$ samples. For each GPU, the local batch size is set as $64$, requiring 2 gradient accumulation steps.

\subsection{Uniform-State Discrete Diffusion and Duality with Gaussian}
\label{sec:duo}

\textbf{Further Details on UDDMs}. As mentioned in \Cref{sec:ch4-prelim}, the true reverse posterior of a previous timestep $s < t$ corresponding to the forward process (\ref{eqn:bg-discrete-forward}) is
\begin{equation}
\begin{split}
     \rvz_s \sim q_{s | t}(\rvz_s | \rvz_t, \rvx) &= \text{Cat} \Big (\rvz_s; g(\rvz_t, \rvx) \Big ) \\
     &= \text{Cat} \Bigg (\rvz_s ; \frac{V \alpha_t \rvz_t \odot \rvx + (\alpha_{t|s} -\alpha_t)\rvz_t}{V \alpha_t \langle \rvz_t, \rvx \rangle + (1 - \alpha_t)} + \frac{(\alpha_s - \alpha_t)\rvx + (1 - \alpha_{t | s}) \frac{\mathbf{1}}{V}}{V \alpha_t \langle \rvz_t, \rvx \rangle + (1 - \alpha_t)} \Bigg ), 
\end{split}
\label{eqn:discrete-reverse-process-full}
\end{equation}
where $\langle \cdot, \cdot \rangle$ denotes the dot product, $\odot$ denotes Hadamard product, the relative diffusion parameter is $\alpha_{t|s} =\frac{\alpha_t}{\alpha_s}$, and we have a uniform prior over $\mathcal{V} (\boldpi = \mathbf{1} / {V})$ \citep{sahoo2024simple, sahoo2025duality}.

Since the true clean token $\rvx$ is unknown during the generative process (\ref{eqn:discrete-reverse-process-full}), the approximate reverse posterior is defined as $p^\theta_{s|t}(\rvz_s | \rvz_t) = q_{s | t} (\rvz_s | \rvz_t, \rvx = \rvx_\theta(\rvz_t, t))$. The neural network $\rvx_\theta(\rvz_t, t) \approx \rvx$ is trained to predict the clean token $\rvx$ at any time $t$ according to \Cref{eqn:bg-discrete-reverse}, where $p_\theta (\rvx | \rvz_t) = \rvx_\theta (\rvz_t, t)$. The network's parameters $\theta$ are optimized via the Negative Evidence Lower Bound (NELBO) objective \citep{austin2021structured, sahoo2024simple}: 
\begin{equation}
\mathcal{L}_\text{\tiny NELBO}=
    \mathbb{E}_q \Big [
    \underset{\mathcal{L}_\text{reconstruction}}{\underbrace{- \log p_\theta (\rvx | \rvz_0)}} + \underset{\mathcal{L}_\text{diffusion}}{\underbrace{\sum_{s < t} D_\text{KL} \big [ q(\rvz_s | \rvz_t , \rvx) \, || \, p_\theta (\rvz_s | \rvz_t ) \big ]}}
    \Big ] + \underset{\mathcal{L}_\text{prior}}{\underbrace{D_\text{KL} \big [ q(\rvz_1 | \rvx) \, || \,  p_\theta (\rvz_1) \big ]}}.
    \label{eqn:nelbo-full}
\end{equation}
As discussed in \Cref{sec:am_pseudo}, the term $\mathcal{L}_\text{reconstruction}$ is more related to our simple setting in that section due to the fact that it only deals with a single finite temperature value (or it does not rely on temperature annealing). However, we suspect that the other terms in NELBO (\ref{eqn:nelbo-full}) further help the maximization of classification margin (or basin) of the data points.

\par\addvspace{1em}
\noindent \textbf{Duality of Uniform and Gaussian}. In continuous diffusion modeling, we typically rely on the diffusion mapping of a data distribution $q_\text{data}$ to a simple prior distribution that is often the standard Gaussian distribution $\mathcal{N}(0, \boldsymbol{I}_V)$. The marginal distribution of the noisy latent variable $\rvw_t \sim \tilde{q}_t(\cdot | \rvx)$ at time $t$ is defined as:
\begin{equation}
    \rvw_t \sim \tilde{q}_t(\rvw_t | \rvx; \tilde{\alpha}_t) = \mathcal{N} (\rvw_t ;  \tilde{\alpha}_t \rvx , (1 -\tilde{\alpha}_t^2) \boldsymbol{I}_V),
\end{equation}
where $\tilde{\alpha}_t \in [0, 1]$ is the diffusion parameter that is a monotonically decreasing function in $t$. The boundary conditions are $\tilde{q}_{t = 0} \approx q_\text{data}$ and $\tilde{q}_{t = 1} = \mathcal{N}(0, \boldsymbol{I}_V)$.

However, as shown in \citet{sahoo2025duality}, there exists a connection between the Gaussian and Uniform diffusion processes for the discrete setting. Specifically, we can utilize the operator, $\argmax : \mathbb{R}^V \rightarrow \mathcal{V}$, to map a continuous vector $\rvw \in \mathbb{R}^V$ to the one-hot vector corresponding to $\argmax (\rvw) = \underset{\rvz \in \mathcal{V}}{\argmax} \,\, \rvz^\top \rvw$. Then, we can define the discrete marginals to be 
\begin{equation}
\rvz_t = \argmax ( \rvw_t ),
\end{equation}
and the conditional probability mass function $p_t(\rvz_t | \rvx)$ marginalized over $\rvw_t \sim \tilde{q}_t (\rvw_t | \rvx; \tilde{\alpha}_t )$ such that 
\begin{equation}
    \rvz_t \sim P_t \big(\rvz_t | \rvx; \mathcal{T} (\tilde{\alpha}_t) \big ) = \Cat \bigg ( \rvz_t ; \mathcal{T}(\tilde{\alpha}_t) \rvx + (1 - \mathcal{T}(\tilde{\alpha}_t) ) \frac{\mathbf{1}}{V} \bigg ) ,
\label{eqn:discrete-marginal-dual}
\end{equation}
where $\mathcal{T} : [0, 1] \rightarrow [0, 1]$ is the Gaussian Diffusion Transformation operator. This operator is defined as the following Gaussian integral:
\begin{equation}
    \alpha_t = \mathcal{T}(\tilde{\alpha}_t) = \frac{V}{V - 1} \bigg [ \int^\infty_{-\infty} \phi \bigg ( z - \frac{\tilde{\alpha}_t}{\sqrt{1 - \tilde{\alpha}^2_t}} \bigg ) \Phi^{V - 1} (z) \dd z - \frac{1}{V} \bigg ]
\end{equation}
where $\phi(z) = \frac{\exp (-z^2)}{\sqrt{2 \pi}}$ is the standard Normal distribution and $\Phi(z) = \int^z_{-\infty} \phi(t) \dd t$ is the respective cumulative distribution.

Overall, there exists a fundamental connection between Uniform-state discrete and Gaussian diffusion processes, shown in \citet{sahoo2025duality}. Specifically, this formal connection is expressed as 
\begin{equation}
    \rvz_t \sim q_t(\rvz_t | \rvx; \mathcal{T}(\tilde{\alpha}_t)) = [\argmax]_{\boldsymbol{*}} \,\tilde{q}_t (\rvw_t | \rvx ; \tilde{\alpha}_t)
    \label{eqn:forward-duo}
\end{equation}
where $\boldsymbol{*}$ denotes the push-forward of the $V$-dimensional Gaussian density $\tilde{q}_t$ under $\argmax$ which yields a categorical distribution of $V$ categories. For full details, please refer to \citep{sahoo2025duality}.

\subsection{Conditional Entropy and Curvature}
\label{sec:entropy_curvature}
In the more common setting, where AM have an energy landscape, the sharpness of the basin dictates retrieval dynamics \citep{DAM2016, krotov2023new, krotov2025modern}. Low entropy implies more deterministic attractors (\textit{memorization}), while high entropy signals a flatter landscape with distributed probability mass \citep{biroli2024dynamical}. This flatness facilitates \textit{generalization}, where basins widen and merge to capture the underlying data manifold, enabling the synthesis of novel patterns \citep{pham2025memorization}.

In this section, inspired by \citet{biroli2024dynamical} and \citet{d2025pseudo}, we relate entropy and the curvature (or sharpness) of the energy in the continuous setting, using local approximation, to show the connection between these two ideas. Here, assume that the clean data $\rvx \in \mathbb{R}^d$ and its perturbed version $\rvz_t \in \mathbb{R}^d$ at time $t$.

\par\addvspace{1em}
\noindent \textbf{Proof Sketch}. Consider the conditional distribution $p(\rvx | \rvz_t)$ defined by an energy function $E(\rvx ; \rvz_t)$:
\begin{equation}
    p(\rvx | \rvz_t) = \frac{1}{Z(\rvz_t)} e^{-E(\rvx ; \; \rvz_t)},
\end{equation}
where $Z(\rvz_t) = \int e^{-E(\rvy, \rvz_t)} \dd \rvy$ is the partition function. We assume the distribution is peaked around a mode $\rvx^*$, representing the most likely clean data point given the noisy observation $\rvz_t$.

To analyze the local geometry, we perform a second-order Taylor expansion of the energy $E(\rvx; \rvz_t)$ with respect to $\rvx$, centered around the mode $\rvx^*(\rvz_t)$:
\begin{equation}
    E(\rvx; \rvz_t) \approx E(\rvx^*; \rvz_t) + (\rvx - \rvx^*)^\top \nabla_{\rvx} E(\rvx^*; \rvz_t) + \frac{1}{2} (\rvx - \rvx^*)^\top \mathbf{H}(\rvz_t) (\rvx - \rvx^*).
    \label{eqn:taylor-expan-cond-energy-min}
\end{equation}
Since $\rvx^*$ is a local minimum of the energy surface defined by $\rvz_t$, the gradient $\nabla_{\rvx} E(\rvx^*; \rvz_t)$ vanishes. The matrix $\mathbf{H}(\rvz_t) = \nabla_{\rvx}^2 E(\rvx^*; \rvz_t)$ is the Hessian of the energy, representing the local curvature or \textit{sharpness} of the energy basin conditioned on $\rvz_t$.

Using \Cref{eqn:taylor-expan-cond-energy-min} and assuming that we are at the minimum where $\nabla_\rvx E(\rvx^*; \rvz) = \mathbf{0}$, we can perform Laplace approximation for the partition function $Z(\rvz_t)$: 
\begin{equation}
    \begin{split}
    Z(\rvz_t) \approx & \int e^{- \big( E(\rvx^*; \rvz_t) + \frac{1}{2} (\rvx - \rvx^*)^\top \mathbf{H}(\rvz_t) (\rvx - \rvx^*) \big )} \dd \rvx\\
    &= e^{-E(\rvx^*; \rvz_t )}  \int e^{-\frac{1}{2} (\rvx - \rvx^*)^\top \mathbf{H}(\rvz_t) (\rvx - \rvx^*) } \dd \rvx \\ 
    &= (2 \pi)^\frac{d}{2} \, e^{-E(\rvx^*; \rvz_t )} \, \mathrm{det} \big (\mathbf{H}(\rvz_t) \big )^{-\frac{1}{2}}
    \end{split}
\end{equation}
The conditional entropy $\mathcal{H} (\rvx | \rvz_t)$ can now be defined. Using the relationship $\log p = -E - \log Z$ and substitute it into $\mathcal{H} (\rvx | \rvz_t)$, we have 
\begin{equation}
\begin{split}
\mathcal{H} (\rvx | \rvz_t) & = \underset{\rvx \sim p(\rvx | \rvz_t)} {\mathbb{E}} \bigg [ E (\rvx ; \rvz_t) + \log Z(\rvz_t) \bigg ] \\ 
& = \underset{\rvx \sim p(\rvx | \rvz_t)} {\mathbb{E}} \bigg [ E(\rvx; \rvz_t) + \frac{d}{2} \log (2 \pi) - E(\rvx^* ; \rvz_t ) - \frac{1}{2} \log \big [ \mathrm{det}\big(\mathbf{H} (\rvz_t) \big) \big ]  \bigg ]
\end{split}
\label{eqn:entropy-pf1}
\end{equation}
If we substitute \Cref{eqn:taylor-expan-cond-energy-min} into the term $E(\rvx; \rvz_t)$ in \Cref{eqn:entropy-pf1}, we then have 
\begin{equation}
\begin{split}
 \mathcal{H} (\rvx | \rvz_t) & \approx \underset{\rvx \sim p(\rvx | \rvz_t)} {\mathbb{E}} \bigg [ \cancel{E(\rvx^* ; \rvz_t)} +   \frac{d}{2} \log (2 \pi) \cancel{- E(\rvx^* ; \rvz_t )} - \frac{1}{2} \log \big [\mathrm{det}\big(\mathbf{H} (\rvz_t) \big) \big ] + C \bigg ] \\ 
 & \approx \underset{\rvx \sim p(\rvx | \rvz_t)} {\mathbb{E}} \bigg [ - \frac{1}{2} \log \big [\mathrm{det}\big(\mathbf{H} (\rvz_t) \big) \big ] +C \bigg ] 
\end{split} 
\end{equation}
where $C$ is a constant involving the omitted terms from our substitution of \Cref{eqn:taylor-expan-cond-energy-min}.

\par\addvspace{1em}
\noindent \textbf{Discussion}. Overall, this derivation highlights that the conditional entropy is inversely proportional to the log-determinant of the Hessian at the mode, and aligns well to the findings of \citet{biroli2024dynamical} where a collapse in entropy corresponds to the system getting trapped in small-disjoint regions of the configuration space. However, in this chapter, we study UDDMs, which are not continuous DMs. Thus, we lack the formulations that attempt to link up their conditional entropy with the sharpness in the discrete setting of language or text modeling. But we suspect that the connection between Uniform and Gaussian distributions in the discrete setting, laid out by \citet{sahoo2025duality}, provides some clues to further this link between conditional entropy and energy in a future work.  

\subsection{Numerical Validation of the Simplified Setting}
\label{sec:numerical_validation_of_PL}

In this section we propose a numerical verification of the binary or linear-logit analysis of \Cref{sec:am_pseudo}, specifically that optimizing the conditional likelihood in \Cref{eqn:pseudo-likelihood} implicitly leads to a maximum-margin solution, which in turn leads to basins of attraction around training data points. Please note that these results are all already established \citep{Gardner1988OptimalSP, Soudry2018, d2025pseudo}, and that we reproduce them here for convenience.

\par
\begin{figure}[t]
    \centering
    \includegraphics[width=0.49\linewidth]{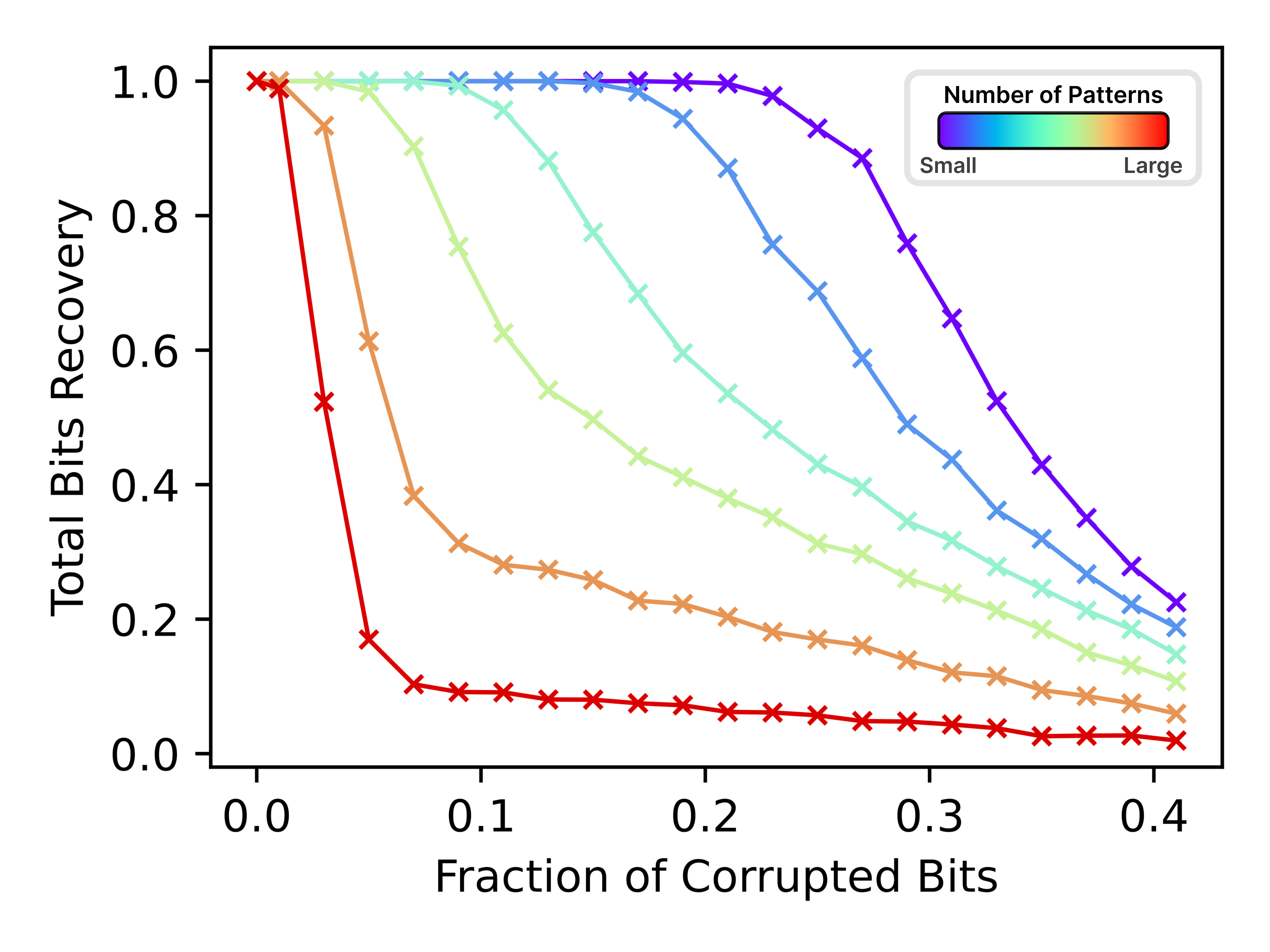}
    \includegraphics[width=0.49\linewidth]{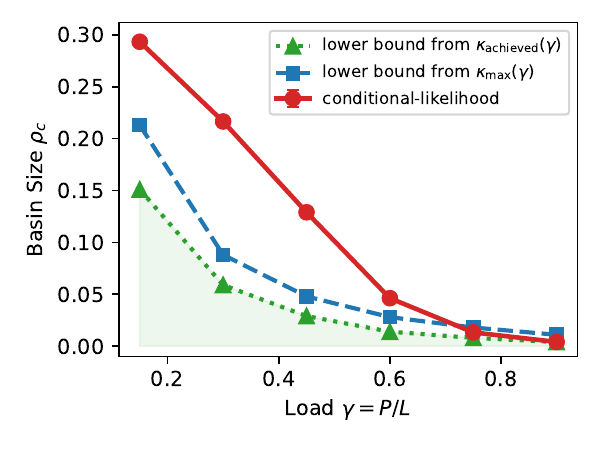}
    \caption{
    The implicit maximization of the margin guarantees large basins of attraction. Both panels refer to an associative memory model trained with conditional likelihood on random bit sequences. Average total bit recovery rates for training sequences across varying corruption levels are shown in the left panel. Line colors indicate the sizes of the training datasets used. This panel captures the same qualitative memorization phenomenon of \Cref{fig:basin_shrinkage}. The right panel shows the basin size $\rho_c(\gamma)$ of the trained model (red curve) compared to the margin-derived lower bound, evaluated both at the idealized $\kappa_{\max}(\gamma)$ and at the empirically achieved $\kappa_{\text{achieved}}(\gamma)$ (blue and green curves, respectively). The measured basin size remains above the achieved-margin bound at every load, confirming margin maximization is sufficient to produce basins of the predicted size. The red points correspond to the critical noise level for which the token recovery rate drops below 0.99. Each point is the average of five samples, and the error bars are too small to be visible.}
    \label{fig:basin_verification}
\end{figure}

\addvspace{1em}
\noindent \textbf{Setup.}
We generate $P = \gamma L$ random i.i.d.\ binary sequences (or patterns), where each sequence has a length of $L = 250$, and train the recurrent coupling matrix $\mathbf{W}$ by minimizing the pseudo-likelihood loss of \Cref{eqn:pseudo-likelihood} for each row independently (unregularized logistic regression), which is known to converge in direction to the maximum-margin solution \citep{Soudry2018}. Here, $\gamma = \frac{P}{L}$ is the load or the ratio between the number of patterns $P$ and their dimension $L$. For each load $\gamma$, we then measure the network's empirical basin of attraction: starting from a stored pattern, we flip a fraction $\rho$ of its bits, run the recurrent dynamics to a fixed point, and record the mean overlap between the recovered state and the original pattern, averaged over multiple patterns and corruption realizations. Basin size $\rho_c(\gamma)$ is defined as the largest corruption level for which this overlap remains above $0.99$, following the same criterion used in \citet{d2025pseudo}. The results are shown in \Cref{fig:basin_verification}.

\par\addvspace{1em}
\noindent \textbf{Theoretical comparison.}
We compare $\rho_c(\gamma)$ against the noise tolerance implied by the Gardner capacity \citep{Gardner1988OptimalSP}:
\begin{equation}
\frac{1}{\gamma_c(\kappa)} = (1+\kappa^2)\,\Phi(\kappa) + \kappa\,\phi(\kappa),
\label{eq:gardner_capacity}
\end{equation}
with $\Phi$ the cumulative distribution function and $\phi$ the probability density function of the standard normal distribution, inverted numerically to obtain $\kappa_{\max}(\gamma)$.
Here $\kappa$ is normalized so that a neuron's pre-activation has unit variance under random patterns, i.e.\ $\kappa$ is a signal-to-noise ratio (a $z$-score). Corrupting a fraction $\rho$ of the $N$ inputs perturbs the pre-activation by a sum of $\rho N$ independent random contributions, which by the central limit theorem has standard deviation $\sim 2\sqrt{\rho}$. A stored pattern with margin $\kappa$ therefore remains a fixed point as long as the injected noise does not exceed the margin, i.e.\ $\kappa > 2\sqrt{\rho}$. This gives a closed-form, margin-derived lower bound on basin size, which we plot alongside the measured $\rho_c(\gamma)$.

\par\addvspace{1em}
\noindent \textbf{Slow convergence to $\kappa_{\max}$.}
In practice, the margin actually achieved by finite-time training, $\kappa_{\text{achieved}}(\gamma)$, undershoots the asymptotic optimum $\kappa_{\max}(\gamma)$, since convergence to the max-margin direction under logistic loss is logarithmic in training time \citep{Soudry2018}, a phenomenon also reported in \citet{d2025pseudo}. We therefore report both curves: the lower bound computed from the idealized $\kappa_{\max}(\gamma)$, and the one computed from the empirically achieved $\kappa_{\text{achieved}}(\gamma)$, which is the fair comparison for a network trained for a finite number of steps.

\par\addvspace{1em}
\noindent \textbf{Result.}
\Cref{fig:basin_verification} shows that the measured basin size lies consistently above the bound derived from $\kappa_{\text{achieved}}(\gamma)$ at every load tested, confirming that the margin does provide a quantitative guarantee on basin size. This directly supports the mechanism claimed in \Cref{sec:am_pseudo}: maximizing the margin is sufficient to produce basins of attraction of predictable size.

\clearpage
\subsection{Additional Results}
\label{sec:add-results}
\begin{figure}[t]
    \centering
    \includegraphics[width=1.0\linewidth]{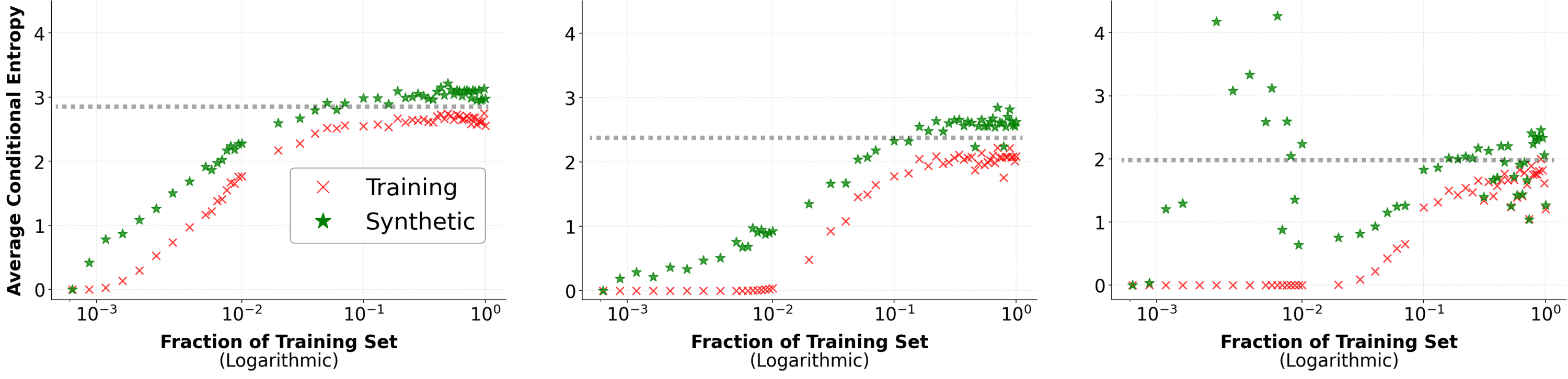}
    \caption{Average conditional entropy for training versus synthetic sequences. As the training dataset size increases, the average (over sequences and tokens) conditional entropy for both training and synthetic (or generated) sequences naturally rises for the \textit{Tiny} (left), \textit{Small} (middle), and \textit{Medium} (right) models. Here, we plot the dashed horizontal line in each plot to show the location of the separation between the average conditional entropies of training and synthetic sequences. Initially, there exists an ``entropy gap'' which separates these two sample types given small fractions of the training set, depicting the UDDM's uncertainty during generation. However, scaling up the model's size narrows this entropy gap, demonstrating that higher parameter counts increase the model's confidence in its own generated text during the generalization regime.}
    \label{fig:entropy_over_time}
\end{figure}

This section collects additional results supporting \Cref{sec:ch4-results}. \Cref{fig:entropy_over_time} shows the average conditional entropy of training and synthetic sequences as a function of the training set size.

\subsubsection{Shrinkage and Expansion of Basins}
\begin{figure}[t]
    \centering
    \begin{minipage}{0.8\textwidth}
        \centering
        \begin{subfigure}[b]{0.465\textwidth}
             \centering
             \includegraphics[width=\textwidth]{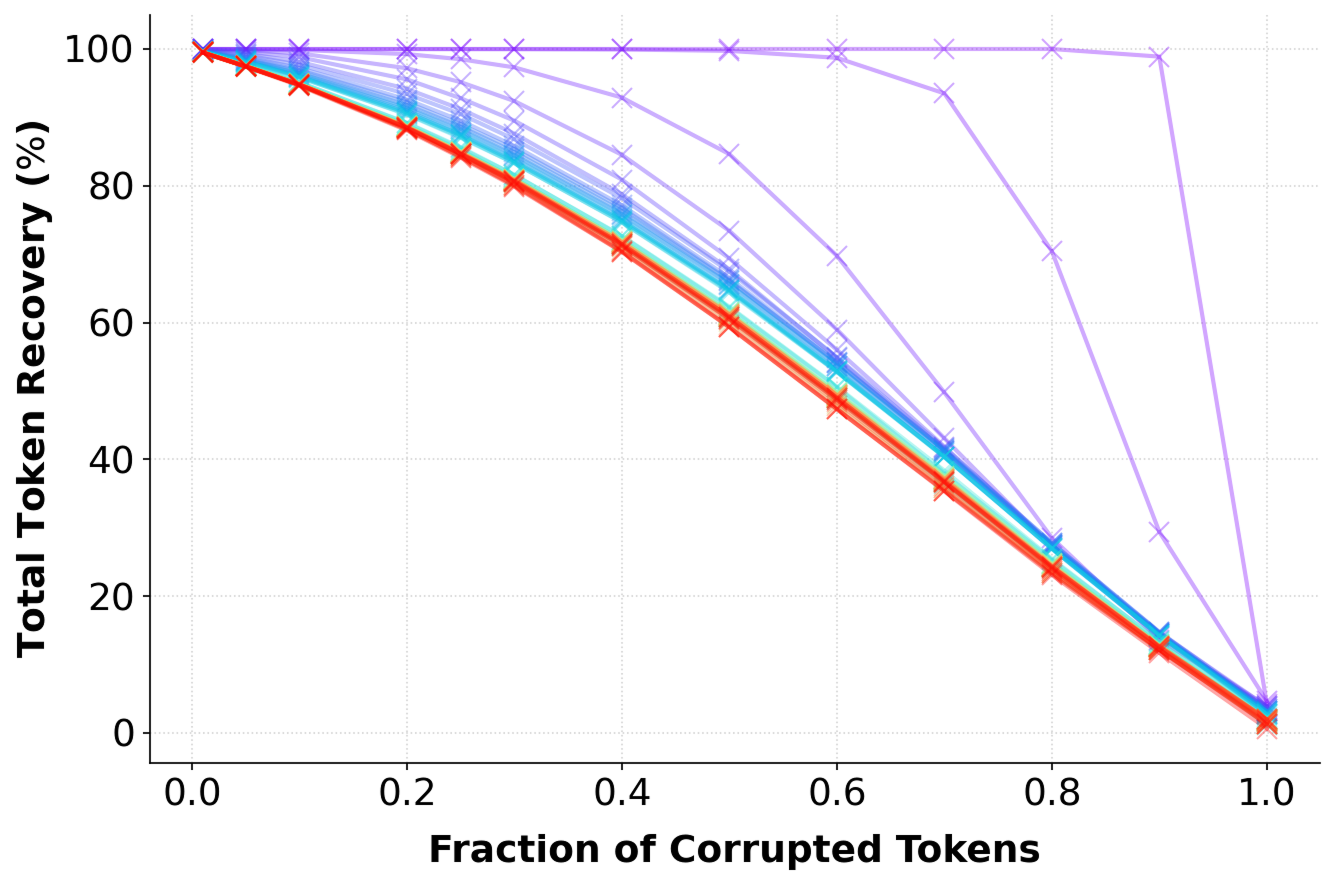}
                     \end{subfigure}
        \hfill
        \begin{subfigure}[b]{0.465\textwidth}
             \centering
             \includegraphics[width=\textwidth]{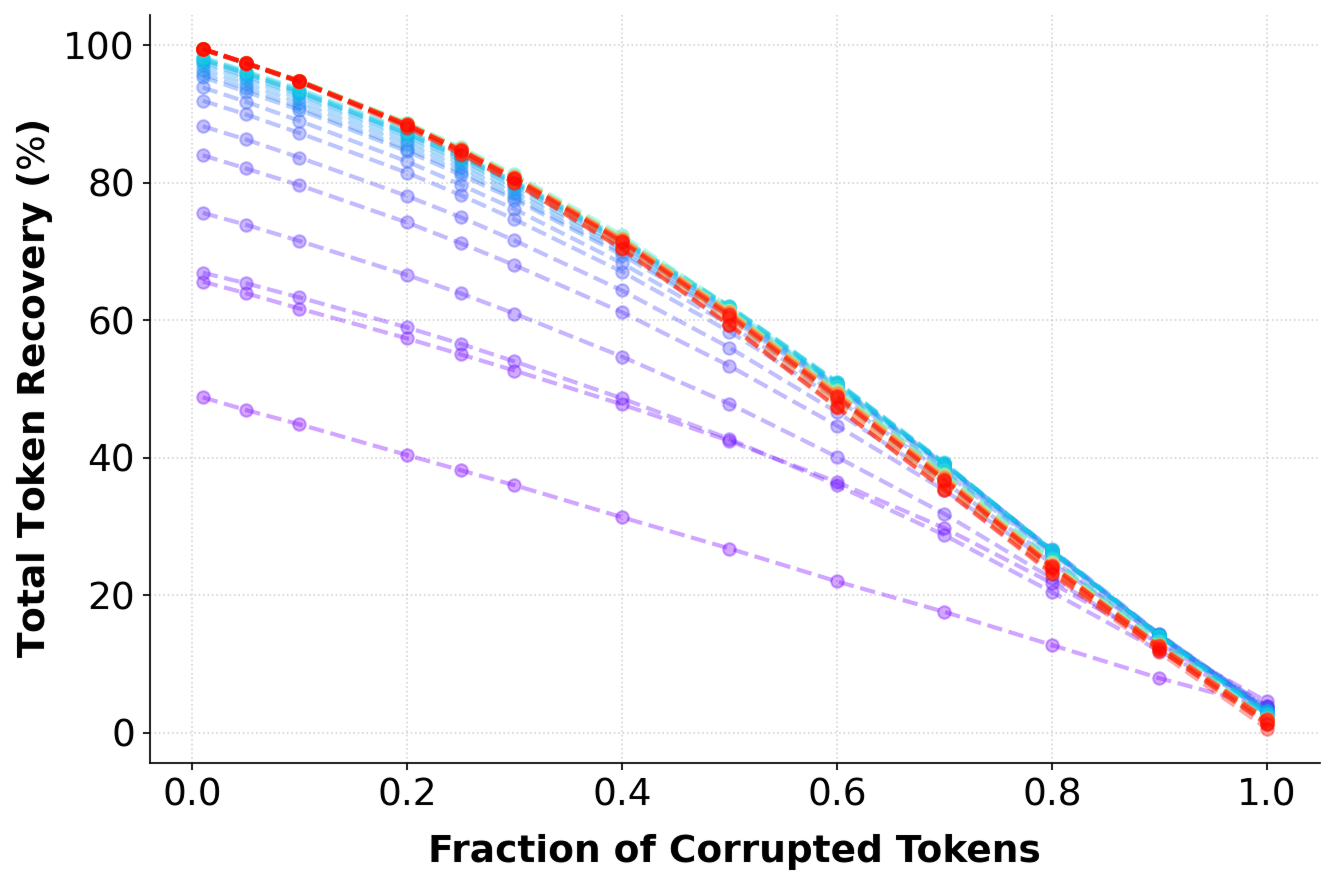}
                     \end{subfigure}
        \vspace{3.5mm}
    \end{minipage}
    \begin{minipage}{0.8\textwidth}
        \centering
                \begin{subfigure}[b]{0.465\textwidth}
             \centering
             \includegraphics[width=\textwidth]{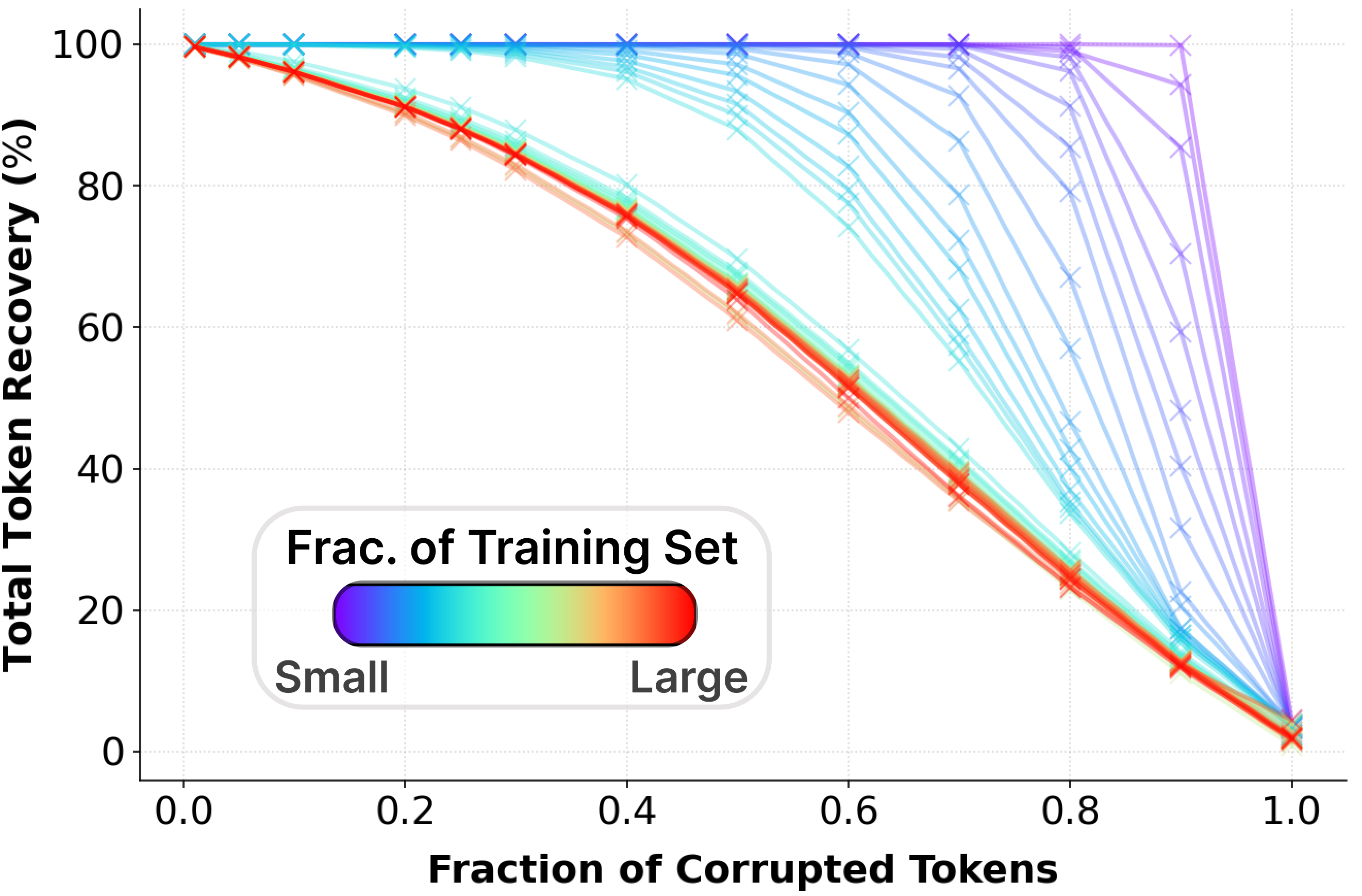}
                     \end{subfigure}
        \hfill
        \begin{subfigure}[b]{0.465\textwidth}
             \centering
             \includegraphics[width=\textwidth]{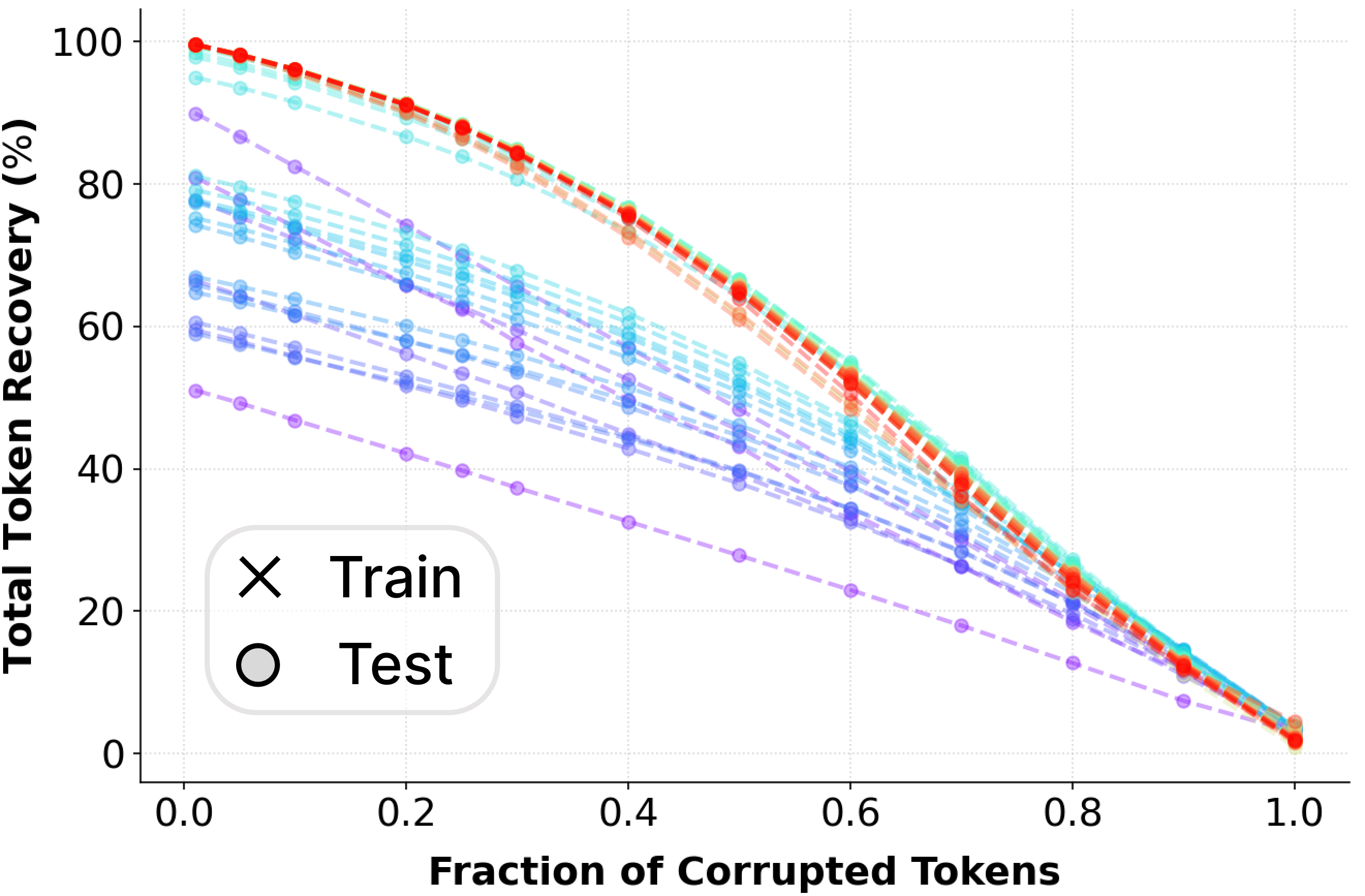}
                     \end{subfigure}
        \vspace{3.5mm}
    \end{minipage}
    \begin{minipage}{0.8\textwidth}
        \centering
                \begin{subfigure}[b]{0.465\textwidth}
             \centering
             \includegraphics[width=\textwidth]{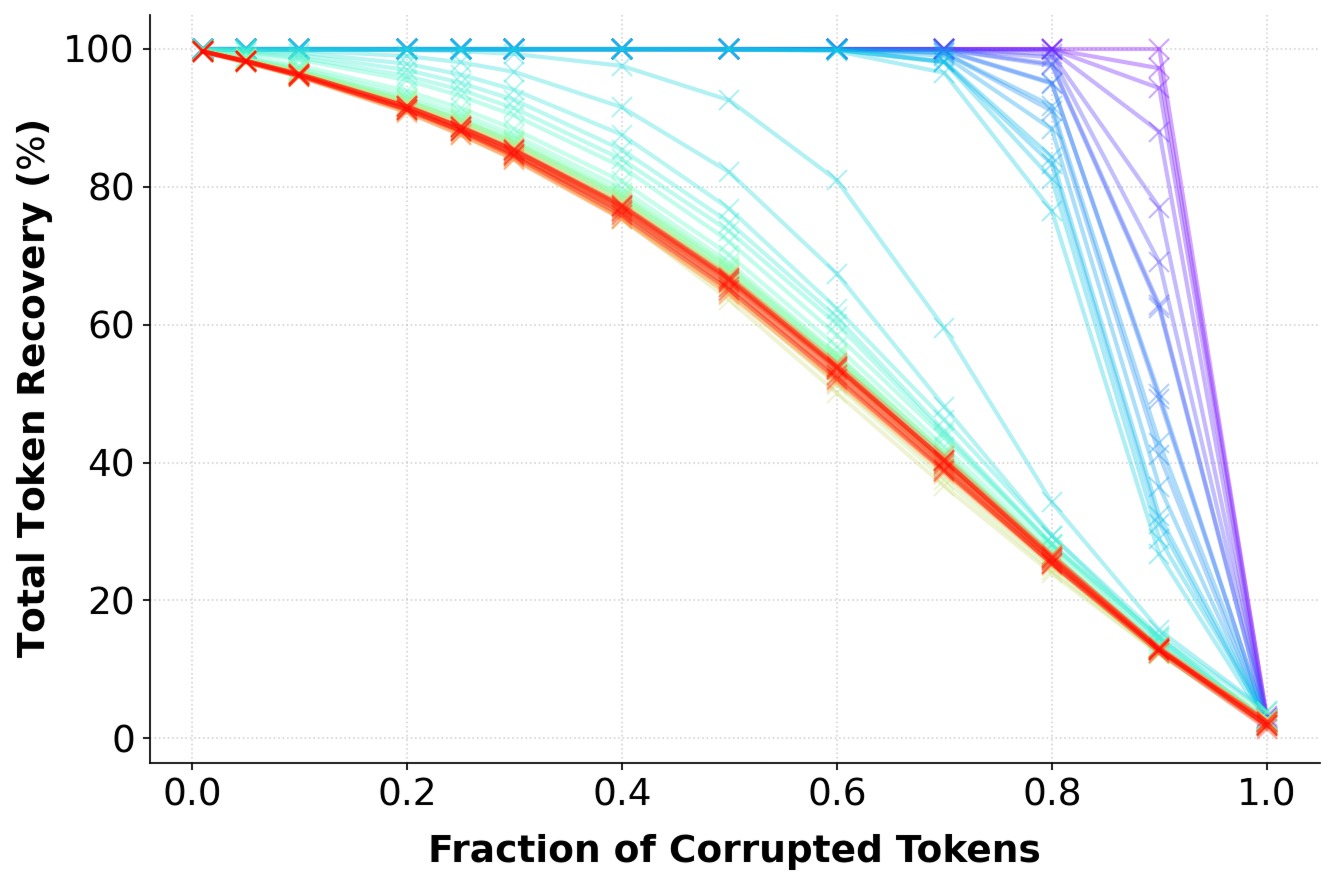}
                     \end{subfigure}
        \hfill
        \begin{subfigure}[b]{0.465\textwidth}
             \centering
             \includegraphics[width=\textwidth]{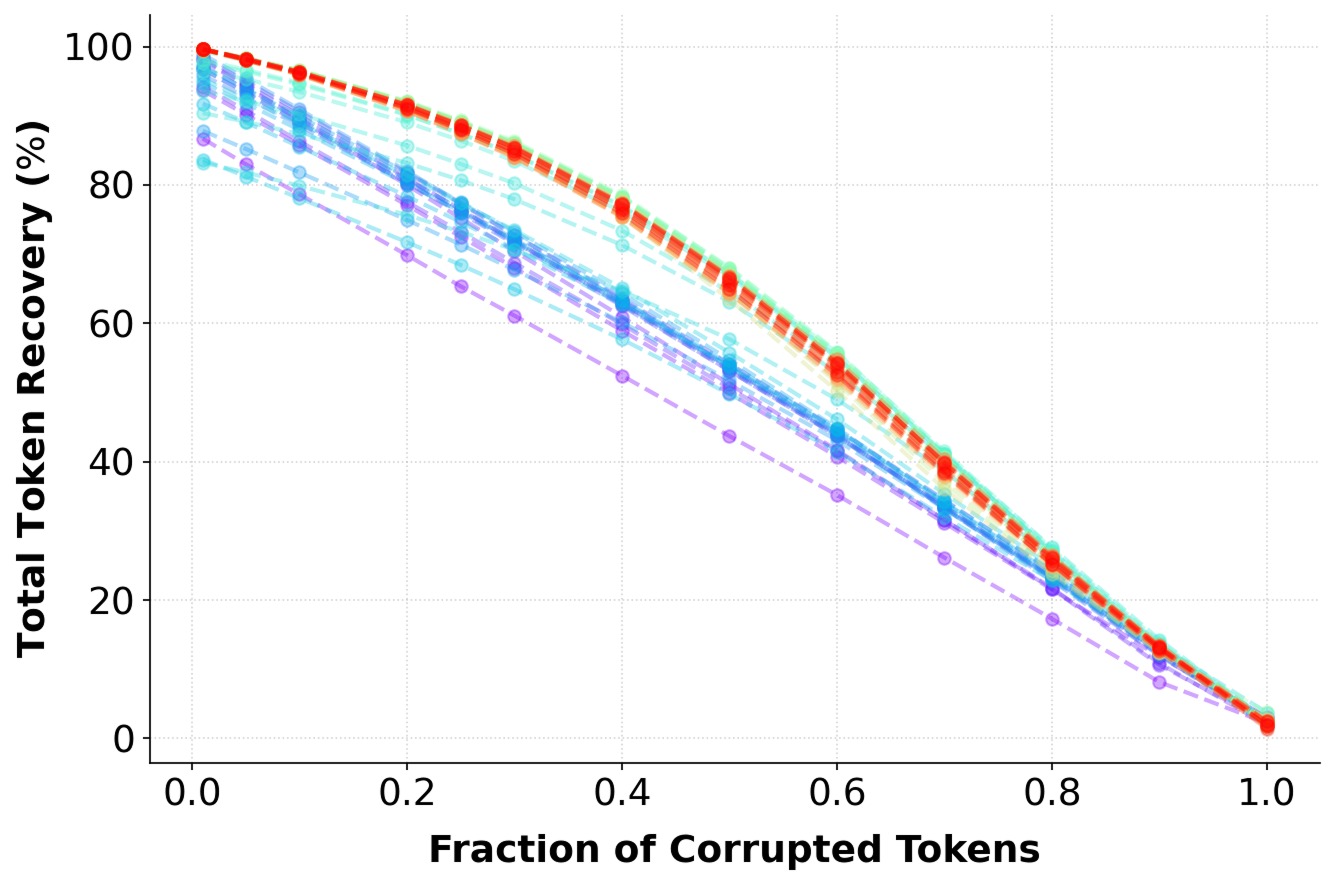}
                     \end{subfigure}
    \end{minipage}
    \caption{Shrinkage and expansion of training and test samples' basins of attraction during the memorization-to-generalization transition for various UDDM sizes. Total token recovery rates (\%), including non-corrupt and corrupted, are shown across varying levels of corruption for training sequences (left column) and unseen test samples (right column), with the \textit{Tiny} ($\sim$24M), \textit{Small} ($\sim$135M), and \textit{Medium} ($\sim$384M) models in the top, middle, and bottom rows, respectively. The color gradient represents the fraction of the training dataset used, ranging from small to large. As the training dataset grows, the recovery rate for training examples diminishes, reflecting a shrinkage in the basins of attraction around explicitly memorized points. Simultaneously, the recovery rate for unseen test samples improves, demonstrating an expansion of the basins of attraction around novel examples within the broader data distribution. The convergence of these recovery rates at large dataset sizes (red curves) signifies that unseen test samples have effectively become stable attractors, marking the shift from pure memorization to generalization.}
    \label{fig:basin_shrinkage_extra}
\end{figure}

\Cref{fig:basin_shrinkage_extra} extends \Cref{fig:basin_shrinkage} to all three model sizes, showing the total token recovery rates of training and test sequences across corruption levels and training set sizes.

\subsubsection{Visualizations of Text Recovery Examples}
\label{sec:text_recovery_examples}
\begin{figure}[t]
    \centering
    \begin{minipage}{0.88\textwidth}
        \begin{subfigure}[b]{1.0\textwidth}
            \centering
            \includegraphics[width=\linewidth]{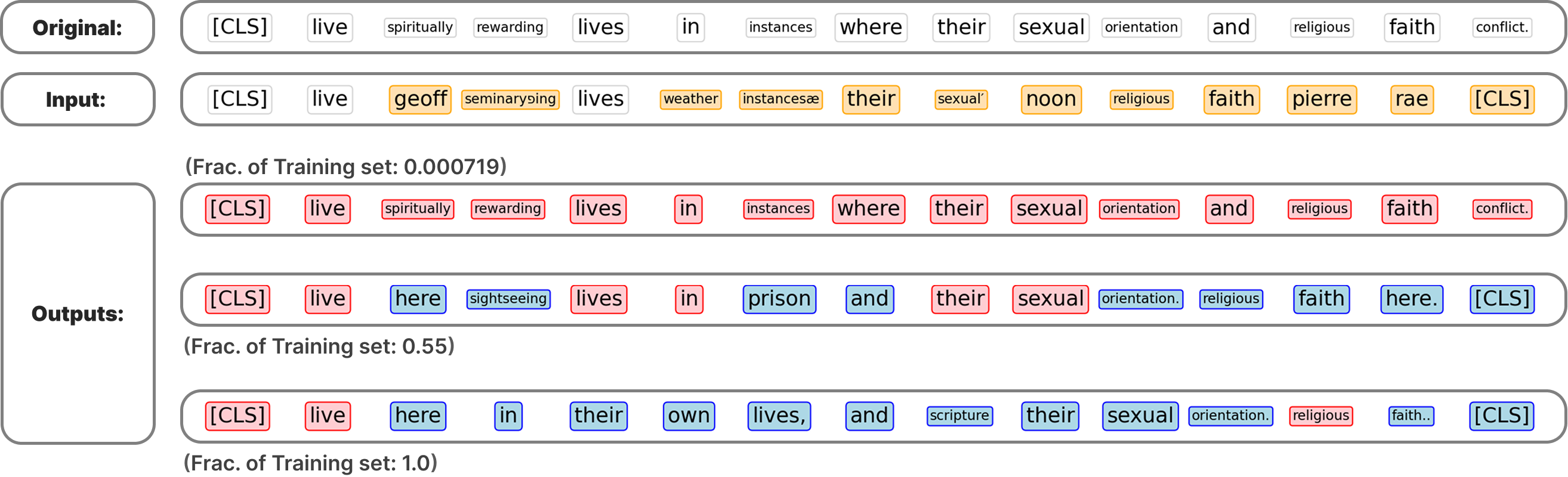}
                    \end{subfigure}
    \end{minipage}
    \vskip 0.35cm
    \begin{minipage}{0.88\textwidth}
        \begin{subfigure}[b]{1.0\textwidth}
            \centering
            \includegraphics[width=\linewidth]{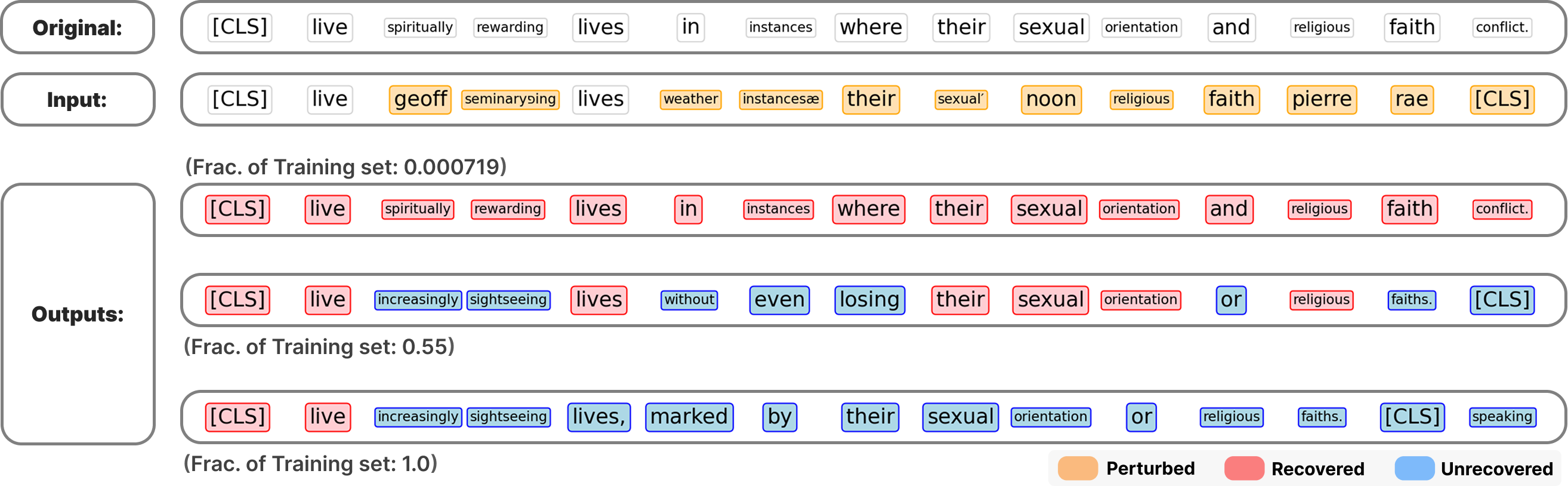}
                    \end{subfigure}
    \end{minipage}
    \vskip 0.35cm
    \begin{minipage}{0.88\textwidth}
        \begin{subfigure}[b]{1.0\textwidth}
            \centering
            \includegraphics[width=\linewidth]{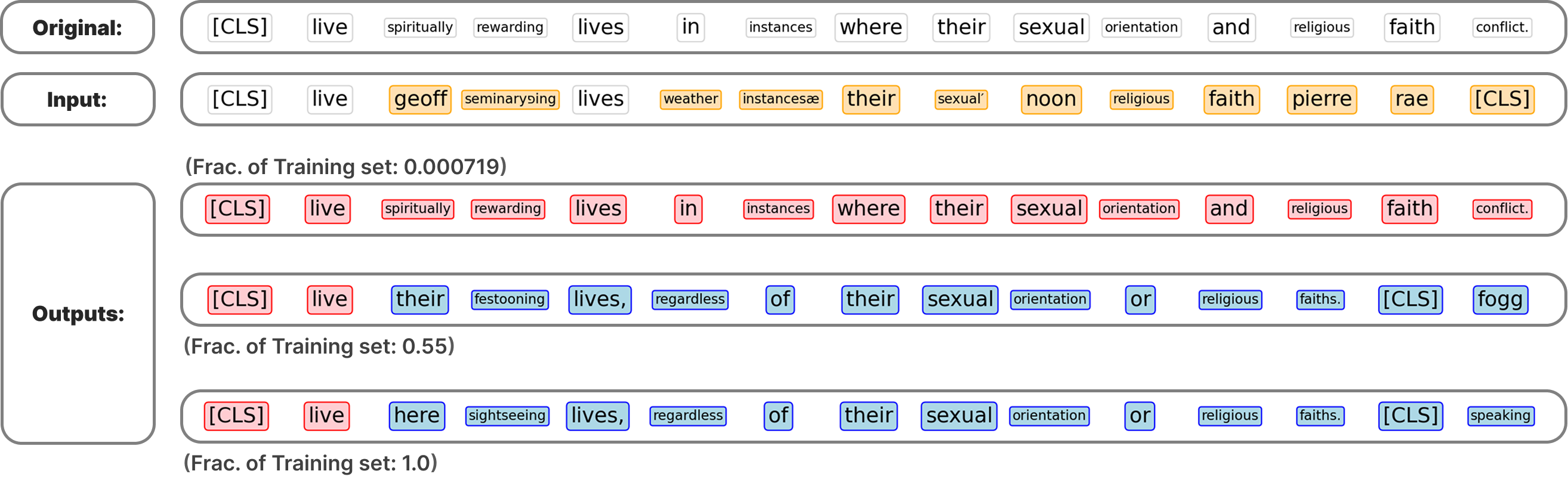}
                    \end{subfigure}
    \end{minipage}
    \caption{An illustration of the model's ability to recover tokens from perturbed sequences on training examples at three different fractions of the training dataset and sizes of the UDDMs. Results are shown for the \textit{Tiny} ($\sim$24M, top), \textit{Small} ($\sim$135M, middle), and \textit{Medium} ($\sim$384M, bottom) models. Perturbation is computed at $t = 0.5$ and the typical stochastic reverse process is performed afterwards. As the training dataset size increases, the model's ability to recover perturbed tokens becomes worse and in contrast, its generative ability improves. Meanwhile, there is no distinction among the various UDDMs' sizes, where the overall trend of memorization and generalization in relation to the training dataset size persists. }
    \label{fig:train_recovery_examples}
\end{figure}

\begin{figure}[t]
    \centering
    \begin{minipage}{0.88\textwidth}
        \begin{subfigure}[b]{1.0\textwidth}
            \centering
            \includegraphics[width=\linewidth]{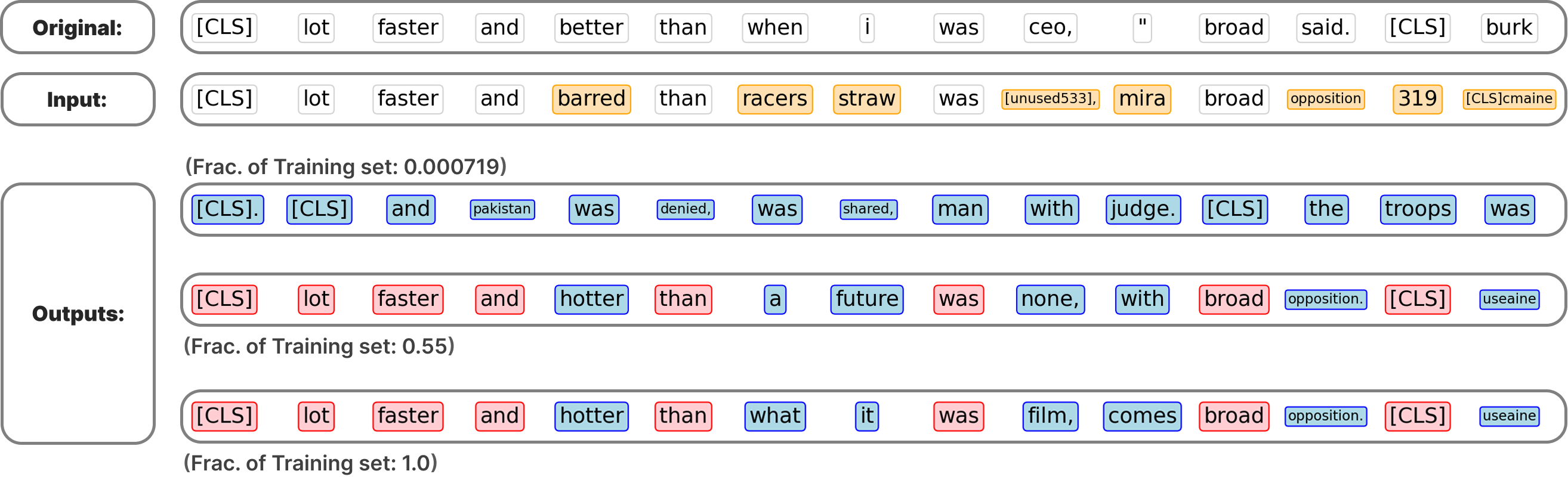}
                    \end{subfigure}
    \end{minipage}
    \vskip 0.35cm
    \begin{minipage}{0.88\textwidth}
        \begin{subfigure}[b]{1.0\textwidth}
            \centering
            \includegraphics[width=\linewidth]{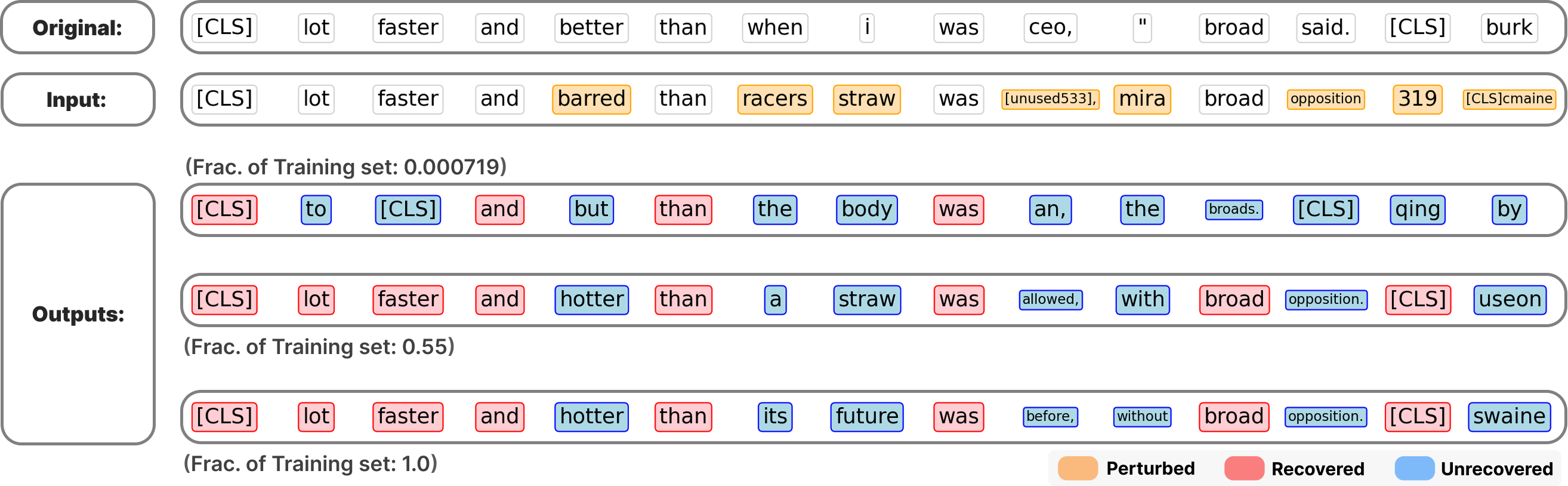}
                    \end{subfigure}
    \end{minipage}
    \vskip 0.35cm
    \begin{minipage}{0.88\textwidth}
        \begin{subfigure}[b]{1.0\textwidth}
            \centering
            \includegraphics[width=\linewidth]{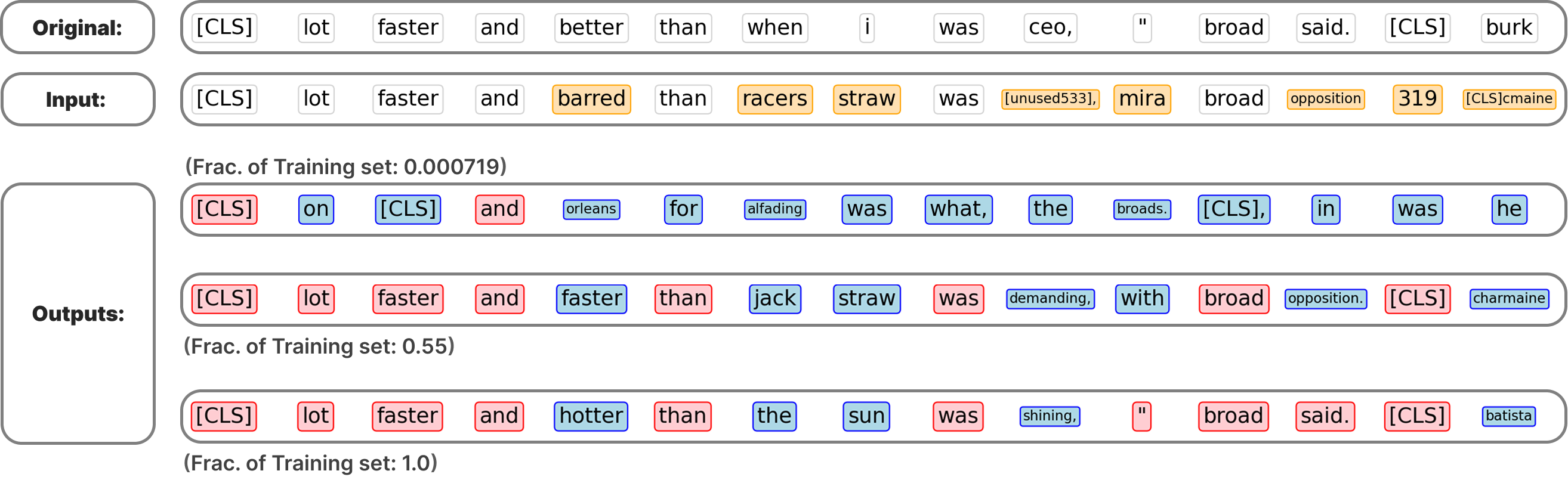}
                    \end{subfigure}
    \end{minipage}
    \caption{An illustration of the model's ability to recover tokens from perturbed sequences on test examples at different fractions of the training dataset and sizes of UDDMs. Results are shown for the \textit{Tiny} ($\sim$24M, top), \textit{Small} ($\sim$135M, middle), and \textit{Medium} ($\sim$384M, bottom) models. Perturbation is computed at $t = 0.5$ and the stochastic reverse process is performed afterwards. In the beginning, the model is unable to recognize unperturbed unseen test tokens, where it often changes them to other tokens. However, as the training dataset size increases, the UDDM is more likely to maintain the unperturbed test tokens rather than `flipping' them. 
    }
    \label{fig:test_recovery_examples}
\end{figure}

Figs.~(\ref{fig:train_recovery_examples}) and (\ref{fig:test_recovery_examples}) show textual examples of token recovery from perturbed training and test sequences, respectively, using the stochastic reverse process.

\subsubsection{Conditional Entropy Histograms}
\label{sec:additional-hists}
\begin{figure}[t]
    \centering
    \includegraphics[width=1\linewidth]{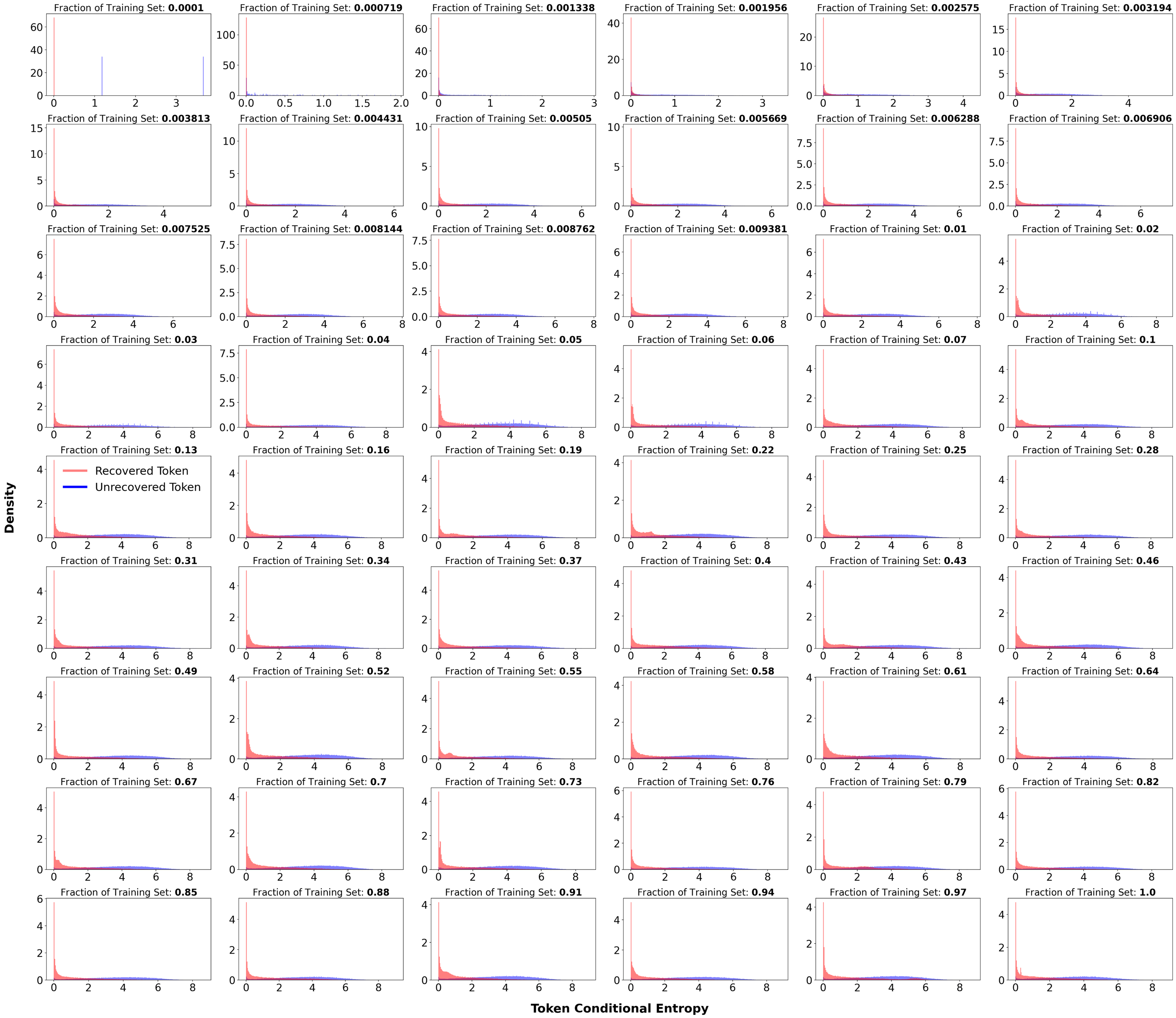}
    \caption{An illustration of the density of conditional entropy for two categories of tokens, recovered and unrecovered, computed at $t = 0.25$ for the \textit{Tiny} model. The subplots are ordered by the fraction of training dataset, ranging from $10^{-4}$ (top-left) to $1.0$ (bottom-right). As the fraction of training data increases, recovered tokens concentrate near zero entropy (high confidence), while unrecovered tokens exhibit a broad distribution at higher entropy.
    }
    \label{fig:additional_recovered_unrecovered_hist_tiny}
\end{figure}

\begin{figure}[t]
    \centering
    \includegraphics[width=1.0\linewidth]{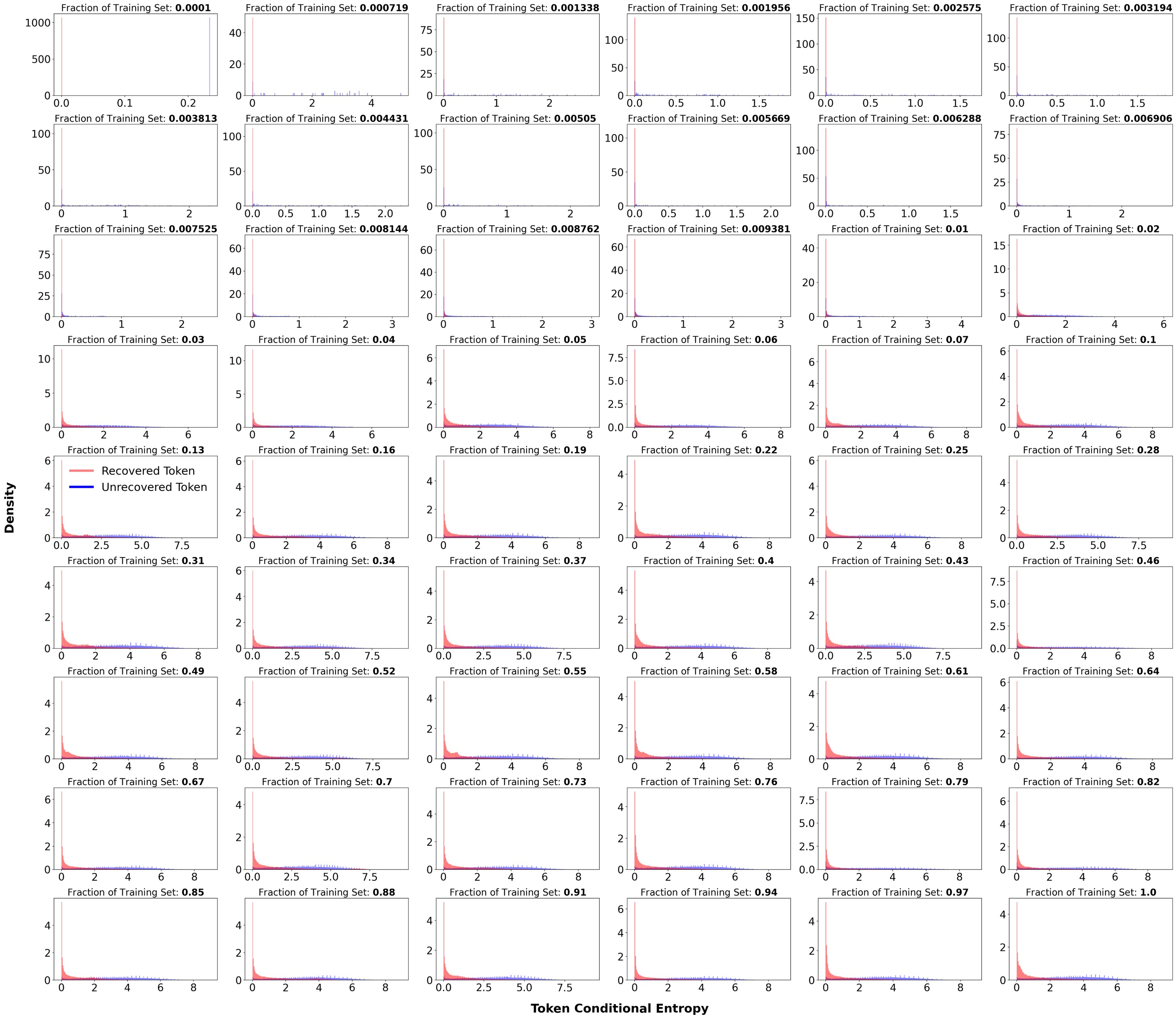}
    \caption{An illustration of the density of conditional entropy for two categories of tokens, recovered and unrecovered, computed at $t = 0.25$ for the \textit{Small} model. The subplots are ordered by the fraction of training dataset, ranging from $10^{-4}$ (top-left) to $1.0$ (bottom-right). As the fraction of training data increases, recovered tokens concentrate near zero entropy (high confidence), while unrecovered tokens exhibit a broad distribution at higher entropy.
    }
    \label{fig:additional_recovered_unrecovered_hist_small}
\end{figure}

\begin{figure}[t]
    \centering
    \includegraphics[width=1.0\linewidth]{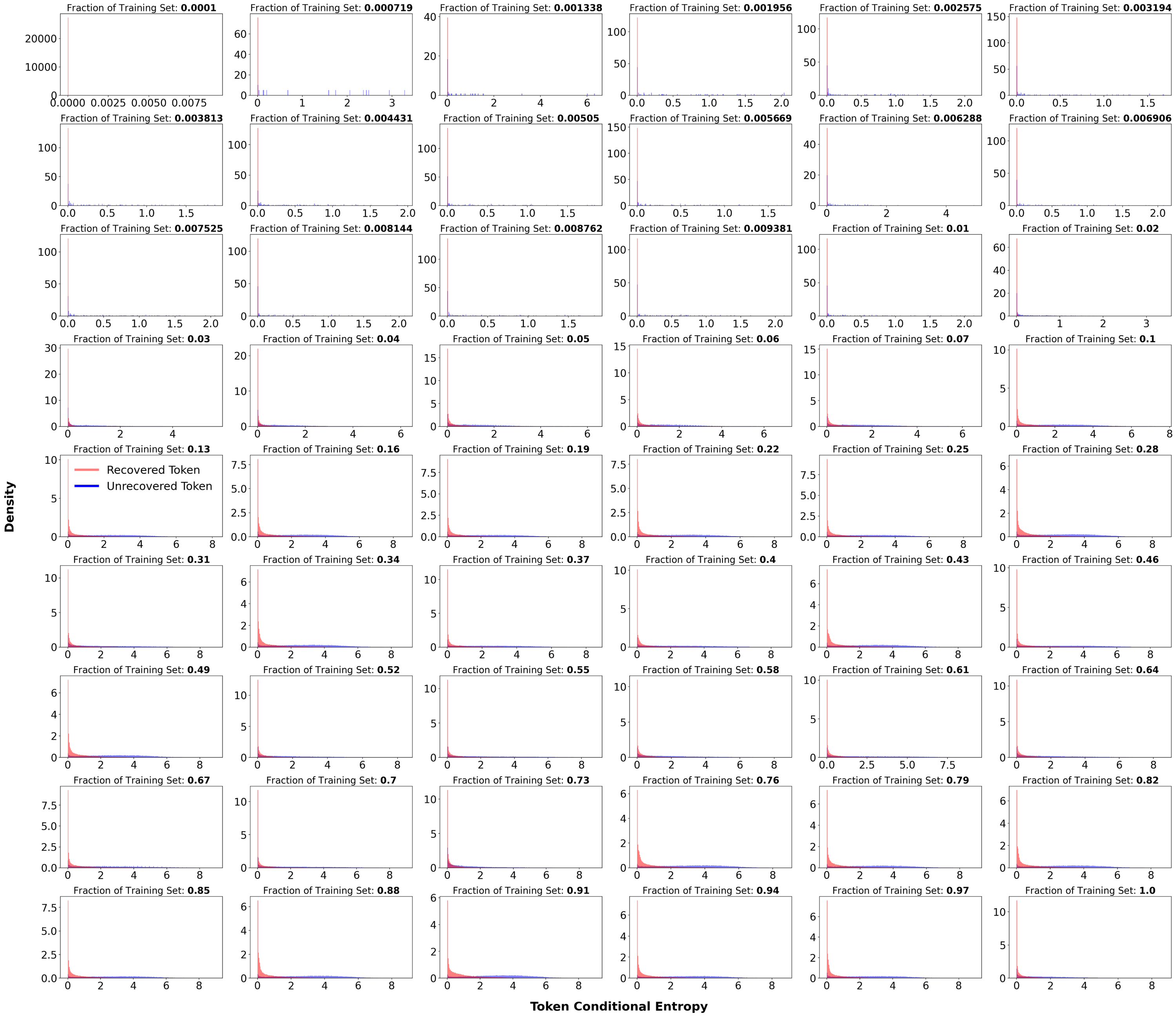}
    \caption{An illustration of the density of conditional entropy for two categories of tokens, recovered and unrecovered, computed at $t = 0.25$ for the \textit{Medium} model. The subplots are ordered by the fraction of training dataset, ranging from $10^{-4}$ (top-left) to $1.0$ (bottom-right). As the fraction of training data increases, recovered tokens concentrate near zero entropy (high confidence), while unrecovered tokens exhibit a broad distribution at higher entropy.
    }
    \label{fig:additional_recovered_unrecovered_hist_medium}
\end{figure}

\begin{figure}[t]
    \centering
    \includegraphics[width=1.0\linewidth]{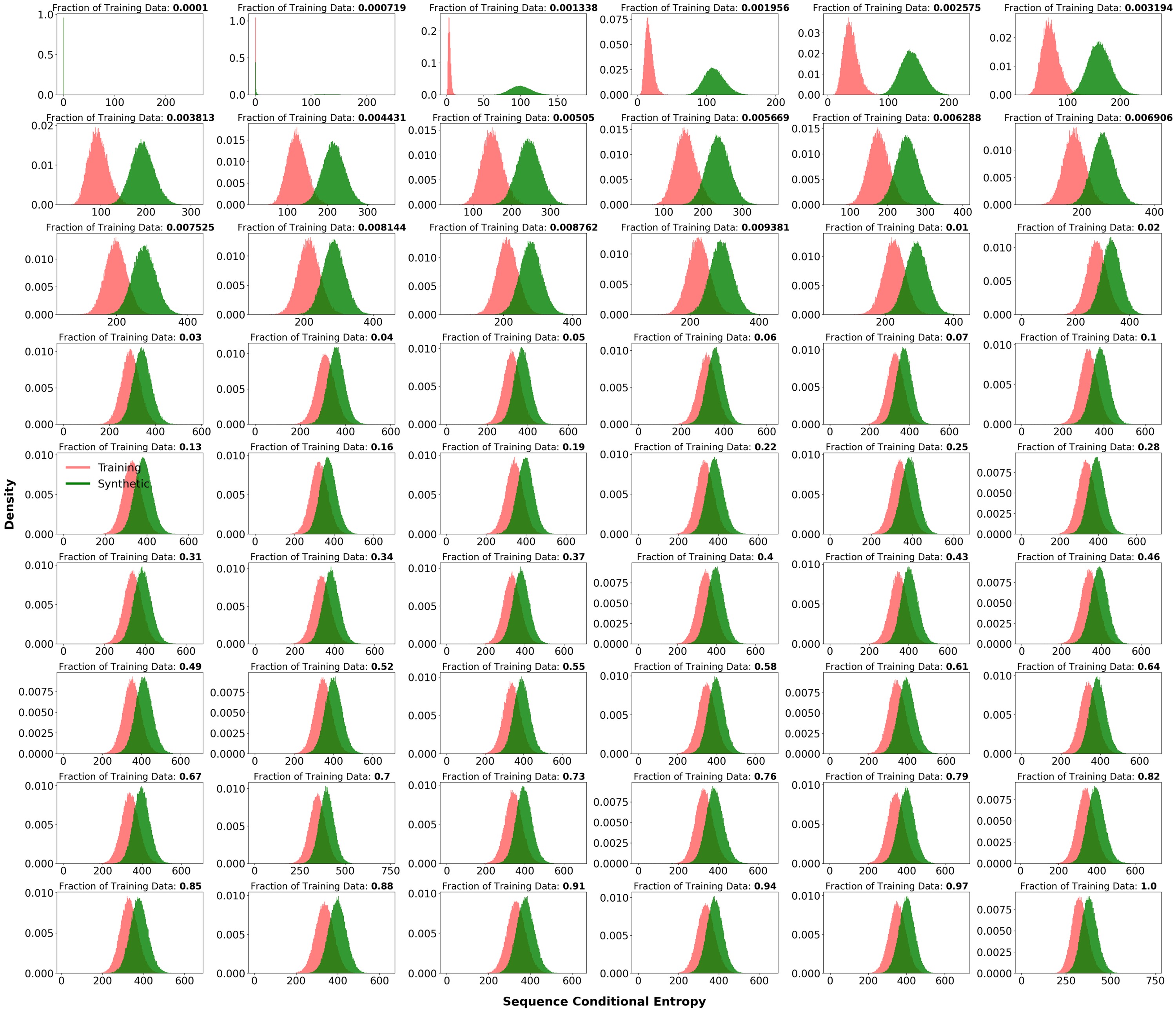}
    \caption{An illustration of the evolution of the density of the average conditional entropy for the probabilities of training and synthetic sequences respectively, computed at $t = 10^{-5}$ using the \textit{Tiny} models, as the training dataset size grows. When the fraction of training set is small, there exists a separation in the average conditional entropies of training and synthetic samples. However, as the training dataset size grows, this separation is reduced and the conditional entropy of synthetic samples becomes similar to that of the training samples.
    }
    \label{fig:additional_train_synth_hist_tiny}
\end{figure}

\begin{figure}[t]
    \centering
    \includegraphics[width=1.0\linewidth]{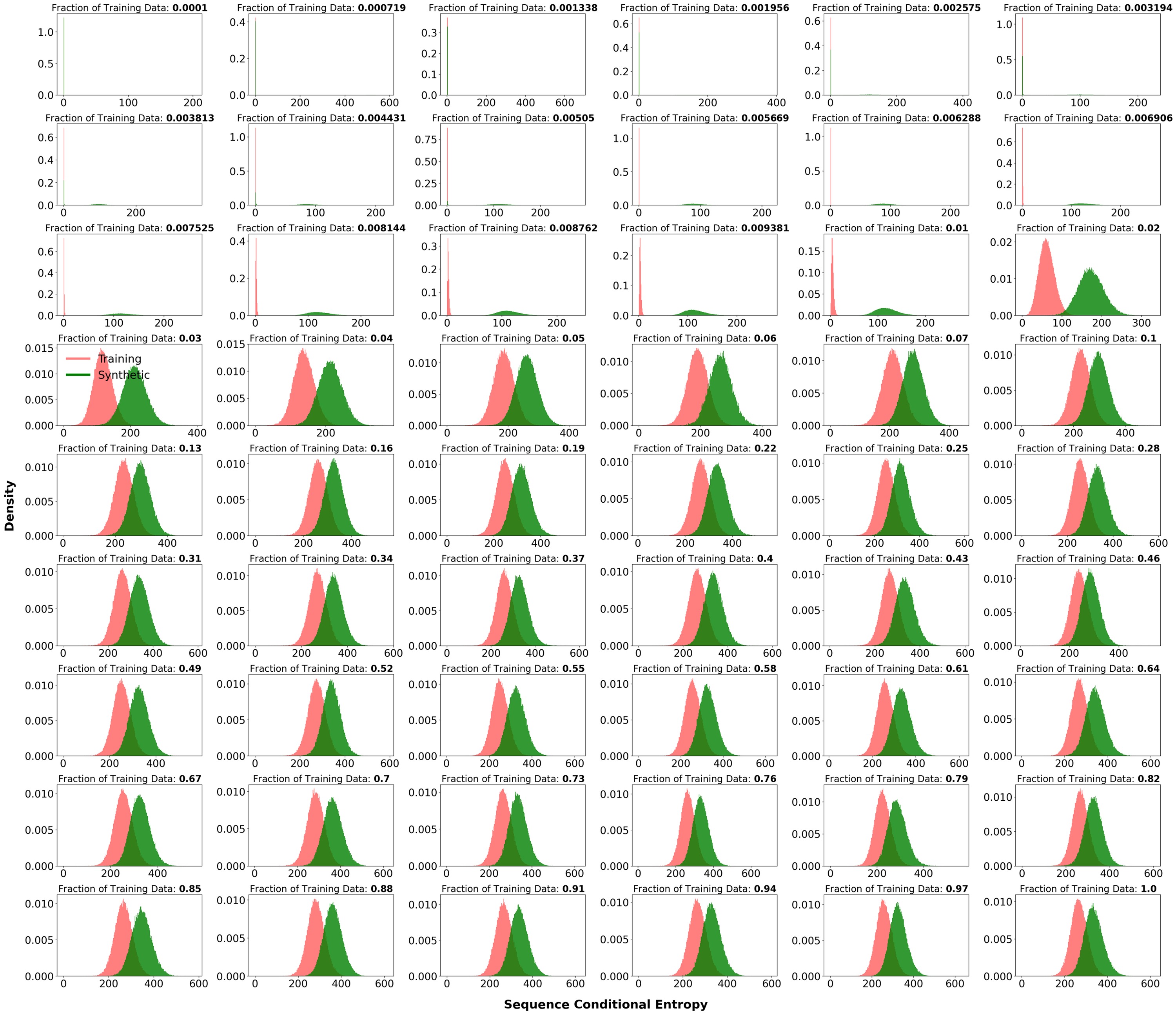}
    \caption{An illustration of the evolution of the density of the average conditional entropy for the probabilities of training and synthetic sequences respectively, computed at $t = 10^{-5}$ using the \textit{Small} models, as the training dataset size grows. When the fraction of training set is small, there exists a separation in the average conditional entropies of training and synthetic samples. However, as the training dataset size grows, this separation is reduced and the conditional entropy of synthetic samples becomes similar to that of the training samples.
    }
    \label{fig:additional_train_synth_hist_small}
\end{figure}

\begin{figure}[t]
    \centering
    \includegraphics[width=1.0\linewidth]{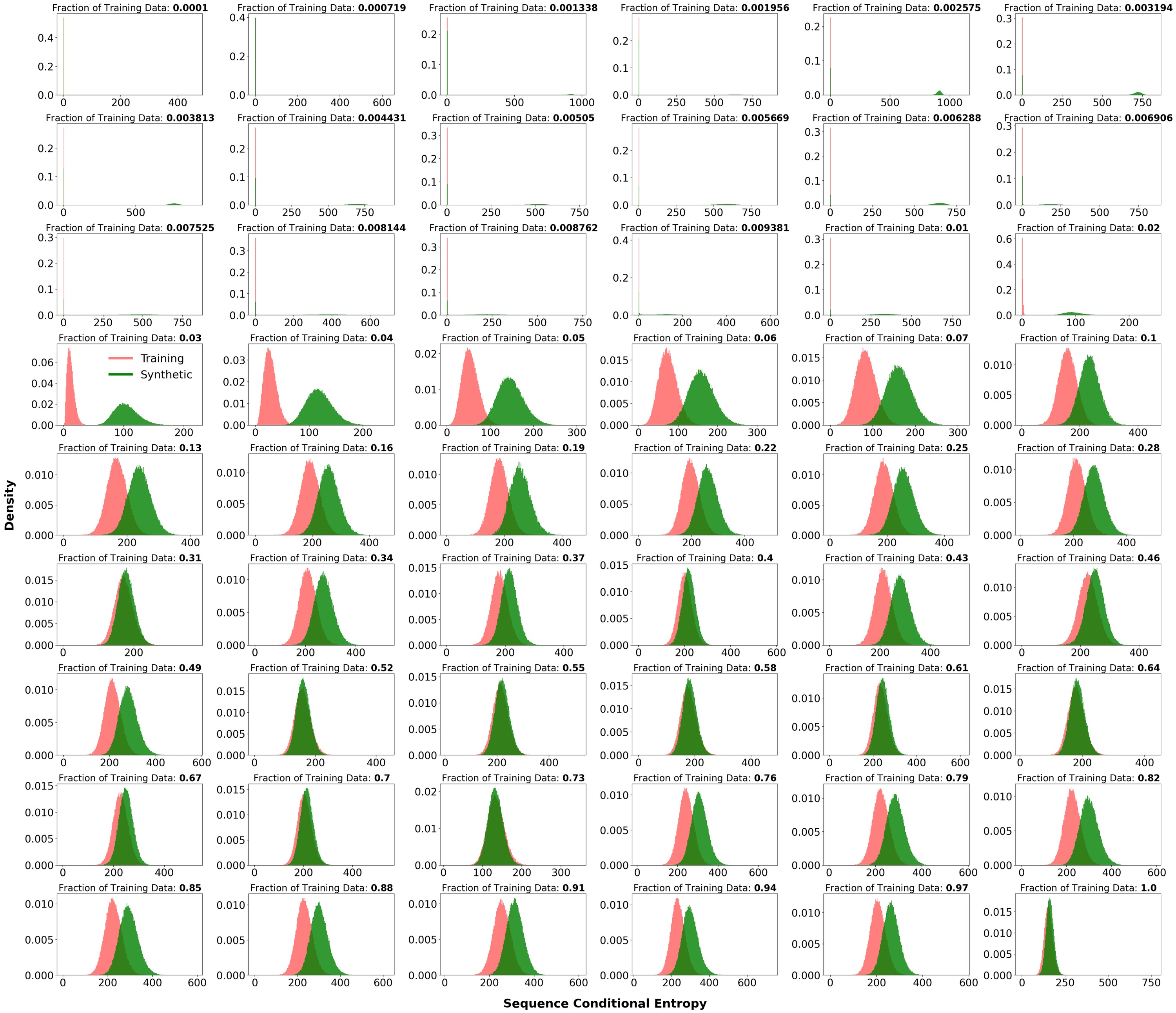}
    \caption{An illustration of the evolution of the density of the average conditional entropy for the probabilities of training and synthetic sequences respectively, computed at $t = 10^{-5}$ using the \textit{Medium} models, as the training dataset size grows. When the fraction of training set is small, there exists a separation in the average conditional entropies of training and synthetic samples. However, as the training dataset size grows, this separation is reduced and the conditional entropy of synthetic samples becomes similar to that of the training samples. Due to larger model size, the UDDM exhibits stronger memorization (and chaotic) behaviors, where the two distributions initially merge at certain fractions of the training dataset and later diverge. Nonetheless, with the full training set, both distributions overlap each other at the end. 
    }
    \label{fig:additional_train_synth_hist_medium}
\end{figure}

\begin{figure}[t]
    \centering
    \includegraphics[width=1\linewidth]{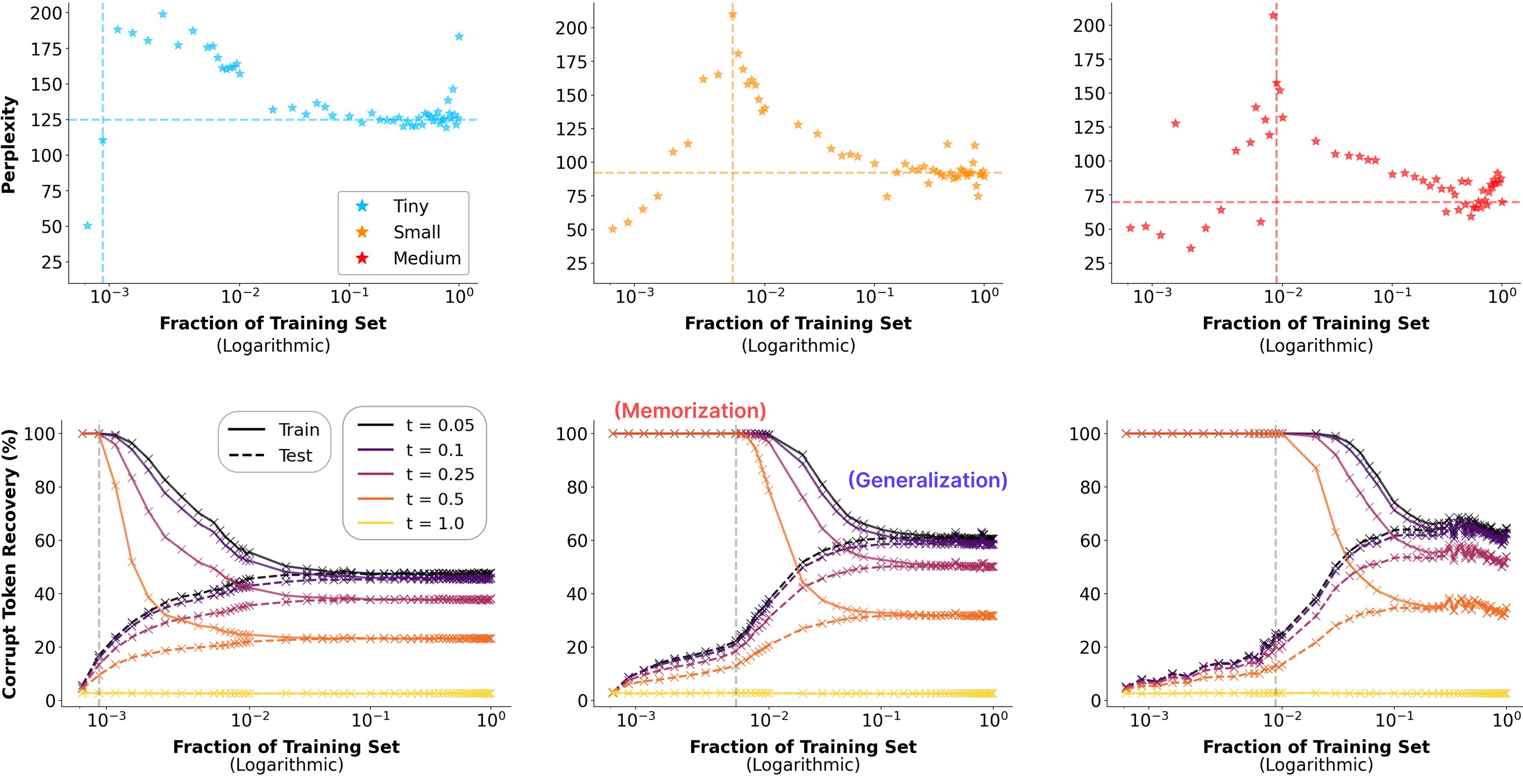}
    \caption{Perplexity trends during the memorization-to-generalization transition together with corrupted token recovery. This figure demonstrates the alignment of phase transitions across perplexity (top row) and the average corrupt token recovery rate (bottom row) for the \textit{Tiny} (left column), \textit{Small} (middle column), and \textit{Medium} (right column) models. The vertical dashed lines roughly represent the critical transitional points where the perplexity (calculated using GPT-2 Large \citep{radford2019language}) aligns with the initial decrease of corrupt token recovery rate for training and the increase of test tokens' recovery rate. This alignment of the two metrics showcases the usefulness of the token recovery rate in determining whether generalization has been reached in UDDMs.
    }
    \label{fig:ppl_combined}
\end{figure}

Figs.~(\ref{fig:additional_recovered_unrecovered_hist_tiny})--(\ref{fig:additional_recovered_unrecovered_hist_medium}) show the token-level conditional entropy of recovered and unrecovered tokens, and Figs.~(\ref{fig:additional_train_synth_hist_tiny})--(\ref{fig:additional_train_synth_hist_medium}) show the sequence-level conditional entropy of training and synthetic sequences, for all fractions of the training set and all model sizes. \Cref{fig:ppl_combined} combines the token recovery rates of \Cref{fig:token_recovery_rate_combined} with the perplexity of \Cref{fig:perplexity}.

%% file: chapters/ch5_conclusion/ch5.tex
\chapter{Conclusion and Future Directions}
\label{chapter:conclusion}

This thesis began with a question that can be asked of every sample a generative model produces, namely whether that sample is recalled from the training data or created anew. We have argued that Associative Memory offers a common language for both answers. Memory recall and creativity (or generalization) are not two separate mechanisms but two regimes of the same dynamics, distinguished by the geometry of the basins of attraction formed around the data during training. In this chapter, we summarize how the preceding chapters address the questions posed in \Cref{sec:intro-questions}, and we then outline the directions that we believe follow most naturally from this perspective.

\section{Summary of Findings}
\label{sec:conclusion-summary}

\noindent \textbf{Generation as a Failure in Memory Recall.}
Chapter~\ref{chapter:dam-and-diffusion} interpreted continuous diffusion models as DenseAMs, in which training writes data into memory and generation attempts to recall it, and showed that their behavior follows the predictions of DenseAM theory. With a small training set, recall succeeds. Each training sample acts as an attractor with its own basin, and the model reproduces what it has seen. As the training set grows beyond the capacity of the model, recall begins to fail. The basins around individual training samples shrink, and the dynamics settle into new attractors that belong to no single training sample. This failure is where generation begins. At the onset of the transition, we found spurious states, samples with sizable basins of attraction that do not correspond to any training sample, as DenseAM theory predicts. Long regarded as a defect of memory, they are, in our interpretation, the first signs of generative capability, and they mark the onset of generalization, although we do not show that they cause it. The ordering of basin volume and energy curvature across memorized, spurious, and generalized samples held across a two-dimensional toy model, several natural image datasets, and multiple architectures, and a small case study of Stable Diffusion found a similar ordering of energy curvature among candidate memorized, spurious, and generalized samples.

\par\addvspace{1em}
\noindent \textbf{Measurement of Memorization and Generalization.}
AM theory also provides quantities that describe memorization and generalization through the geometry of basins of attraction, rather than only through how closely a sample resembles a training data point. In Chapter~\ref{chapter:dam-and-diffusion}, memorized, spurious, and generalized samples were first identified by comparing each generated sample with its nearest neighbors in the training set and in a larger set of generated samples. Once identified, the three sample types were characterized by the volume of their basins of attraction and by the curvature of the energy around them, which follow a largely consistent ordering across datasets and architectures. Because the curvature is computed from the trained model alone, it points toward diagnostics that could flag spurious states, and therefore the onset of generalization, without a nearest-neighbor search over the training set, although developing such a diagnostic remains future work. In language diffusion models, studied in Chapter~\ref{chapter:language-dm-am}, token recovery rates measure how reliably the model returns to a reference sequence after corruption, and therefore require the training and test sequences themselves. Conditional entropy, in contrast, is read directly from the predictions of the model and can be computed for any sequence, including generated ones. It locates a model along the transition, agrees with perplexity measured by an external critic model, and separates recovered tokens, which have near-zero entropy, from unrecovered ones within the same sequence. In both domains, larger models delay the transition and require more data before they generalize.
\par\addvspace{1em}
\noindent \textbf{Associative Memory Can Be Enabled via Conditional Likelihood Maximization.}
Building on prior work showing that maximizing the conditional likelihood of each variable given the others can form basins of attraction without an energy function \citep{d2025pseudo}, Chapter~\ref{chapter:language-dm-am} brought this view to language diffusion models. In a simplified setting of binary variables, this objective yields a Hebbian term that stores the data and a penalty term that concentrates learning on the variables with the smallest classification margins, so that the training data become attractors without symmetric couplings or a global energy function. Uniform-state discrete diffusion models share this conditional-likelihood structure through the cross-entropy terms of their training objective, and empirically they display the retrieval behavior expected of an AM. They undergo a sharp memorization-to-generalization transition governed by the size of the training set, in which basins around training sequences shrink while basins, at least partial ones, form and expand around unseen test sequences, until both are recovered from corruption at the same rate.

\par\addvspace{1em}
\noindent \textbf{Coexistence of Memorizing and Generalizing.}
Taken together, these results suggest an answer to the central question of this thesis. Memorization and generalization appear to be better understood as local properties of individual attractors than as global properties of a model, which is what allows them to coexist. A single model can hold sharp basins around some data points and broader, flatter basins around novel samples at the same time. In images, spurious states occupy the boundary between the two. In language, generated sequences in the generalization regime still contain a fraction of highly stable, low-entropy tokens, which may act as small pieces of recall embedded in an otherwise novel output. The balance between the two is governed largely by the size of the training set relative to the capacity of the model, mirroring the role of the load in classical AMs.

\section{Future Directions}
\label{sec:conclusion-future}

The perspective developed in this thesis treats a generative model as a memory whose geometry can be measured. We see three directions in which this view can be extended, and a fourth in which it becomes a principle for building models rather than only explaining them.

\par\addvspace{1em}
\noindent \textbf{Scaling laws through the lens of capacity.}
Neural scaling laws describe how the loss of a model improves as a power law in the number of parameters, the amount of data, and the available compute \citep{kaplan2020scaling, hoffmann2022training}. These laws are largely empirical, and they say little about what a model does with the data it sees. The capacity of an AM offers a complementary and mechanistic quantity. In both Chapter~\ref{chapter:dam-and-diffusion} and Chapter~\ref{chapter:language-dm-am}, the training set size at which a model leaves the memorization regime grows with the size of the model, although only a few model sizes were studied. A natural next step is to measure this critical size systematically across model sizes, data modalities, and training budgets, and to ask whether it follows a law of its own. Such a law would describe not only how much a model improves with scale but also when it stops reproducing its training data, which matters for privacy and copyright as much as for sample quality. A promising starting point is the recent finding that the implicit bias of gradient-based training produces scaling laws along the entire learning curve, not only at convergence \citep{damico2026implicit}. In a single-layer perceptron trained with the logistic loss, the same margin-maximizing bias that forms basins of attraction in Chapter~\ref{chapter:language-dm-am} gives rise to dynamical scaling laws in norm-based complexity measures. Combined, these laws recover the familiar scaling of the test error at convergence, and they hold for convolutional networks, residual networks, and vision transformers. Because the simplified AM of Chapter~\ref{chapter:language-dm-am} is a collection of such perceptrons, one for each variable, this result offers a path toward scaling laws derived directly from AM theory, relating the growth of the classification margins to the size of the basins and to the location of the transition. Other recent work has connected scaling behavior to the statistics of natural data \citep{cagnetta2026deriving} and to associative memory \citep{niu2024beyond}. Probes that are computed from the model itself, such as energy curvature and conditional entropy, could make the transition point observable during training.

\par\addvspace{1em}
\noindent \textbf{Alternative learning algorithms for energy-based modeling.}
Training an energy-based model by maximum likelihood requires the gradient of the log partition function, an expectation under the model that is intractable in high dimensions. Most existing algorithms are ways of approximating or avoiding this term. Contrastive divergence and its persistent variant estimate it with short Markov chains started at or near the data \citep{hinton2002training, tieleman2008training}, but the resulting gradients are biased, the chains can mix slowly, and training can become unstable \citep{du2020improved, song2021trainebm}. Score matching and its denoising and sliced variants avoid the partition function by matching the gradient of the log density instead \citep{hyvarinen05a, VincentP2011, song2019sliced}, and noise-contrastive estimation recasts density estimation as classification against a known noise distribution \citep{gutmann2010noise}. Diffusion models can be viewed as energy-based models trained in this way, with denoising score matching at many noise levels (\Cref{sec:diffusion-ebm}). Pseudo-likelihood \citep{pseudo-likelihood}, which Chapter~\ref{chapter:language-dm-am} connected to Hebbian learning and margin maximization, is another alternative, since each conditional probability is normalized locally. Each of these objectives shapes the energy landscape differently, and therefore the basins of attraction that determine whether a model memorizes or generalizes, yet they are usually compared only through sample quality or likelihood. In recent work, we showed that for DenseAMs with translation-invariant kernel energies the partition function does not depend on the memories, so maximum likelihood reduces exactly to minimizing the energy of the data \citep{pham2026energy}. This yields a sampling-free training rule, reminiscent of Hebbian learning, that preserves an explicit energy and extends to the latent space of an autoencoder. We are interested in extending such sampling-free rules to broader classes of energy functions, in comparing the objectives above through the geometry of the basins they produce, and in local learning rules, such as equilibrium propagation \citep{scellier2017equilibrium} and competitive Hebbian learning \citep{krotov2019unsupervised}, that train energy-based networks without backpropagating through the retrieval dynamics. An objective that controls where the memorization-to-generalization transition occurs would turn the analysis of this thesis into a design tool.
\par\addvspace{1em}
\noindent \textbf{Test-time scaling and recursive computation.}
Retrieval in an AM is an iterative computation. A state is refined step by step until it settles into an attractor, and running the dynamics for longer is a way of spending more compute at inference. This connects AMs to a growing line of work on test-time scaling, in which a model improves by allocating more computation to each input rather than more parameters \citep{snell2025scaling}. Diffusion models already scale at inference through the number of denoising steps and through search over the initial noise \citep{ma2025inference}. Recursive architectures go further by applying the same block many times, from universal transformers \citep{dehghani2019universal} and deep equilibrium models \citep{bai2019deep} to recurrent-depth language models \citep{geiping2025scaling} and small recursive reasoning networks \citep{wang2025hierarchical, jolicoeur2025less}. Energy-based formulations of this idea treat inference itself as the optimization of a learned energy \citep{du2024learning, gladstone2026energy}. The AM perspective raises questions that, to our knowledge, remain open. In terms of basins, an additional step of computation may pull a state deeper into a memorized attractor, or it may move the state along the flat directions that Chapter~\ref{chapter:language-dm-am} associated with generalization. Spurious states, which sit at the boundary between recall and creation, might be reached or avoided by controlling the length of the dynamics. Understanding these effects would connect the geometry studied here with the way models reason at test time.

\par\addvspace{1em}
\noindent \textbf{Alternative architectures.}
The transformer \citep{Attention2017} is built from four operations, namely attention, a feed-forward network, residual connections, and layer normalization, which are combined in a sequence of independent blocks. The way these operations are combined has been established largely through empirical search, and the principles that should govern it remain unclear. The AM perspective suggests a different starting point, in which an architecture is derived from an energy function chosen for the task, and the forward pass is the dynamics that minimizes this energy. The observation that attention is the update rule of a DenseAM with a softmax activation \citep{Ramsauer2021}, and the construction of hierarchical AMs whose neurons and layers are all described by one energy \citep{krotov2021hierarchical}, are steps in this direction. Architectures of this kind are recursive by construction, since the same dynamics is applied until it converges, and their energy can make the basins of attraction studied in this thesis directly accessible.

The Energy Transformer \citep{et} is an example of this approach. Instead of a stack of feed-forward transformer blocks, it uses a single block that is applied recurrently. Each token is treated as a particle whose state evolves under a global energy with two contributions. The first is an energy-based attention, which aligns the queries of each token with the keys of the tokens it should attend to, so that information flows between tokens. The second is a Hopfield network module, which stores patterns in its weights and pulls each token toward configurations that are consistent with the data. Layer normalization arises as the derivative of a Lagrangian, and the two energy terms act in parallel rather than in sequence. Because every update decreases the energy, the forward pass is a retrieval dynamics whose fixed point is the output of the model. In masked image completion, for instance, the fixed point is a plausible reconstruction of the occluded patches. Requiring a global energy also constrains the architecture in an informative way. The resulting attention contains the conventional attention update together with an additional term that guarantees the decrease of the energy, and it needs no separate value matrix. The Energy Transformer was competitive on image completion, graph anomaly detection, and graph classification, and in the image domain it used far fewer parameters than a standard vision transformer or a weight-sharing variant of it. The results of Chapter~\ref{chapter:language-dm-am} also suggest how such architectures could be trained and probed. Because the energy of the Energy Transformer defines a conditional distribution for each token given the others, the model could be trained by maximizing these conditionals directly, in the spirit of pseudo-likelihood \citep{pseudo-likelihood}, rather than by backpropagating through every step of the energy descent. For discrete tokens, each conditional is normalized over the vocabulary, so no negative sampling would be required. The same conditionals also give the conditional entropy of each token, which could signal when the dynamics has settled into a basin. Studying when such architectures memorize and when they generalize, how their recursive dynamics behave as test-time compute grows, and how they can be trained with the learning algorithms discussed above would help close the loop between explaining generative models and building them.

\section{Closing Remarks}
\label{sec:conclusion-remarks}

Memorization is often treated as a flaw of generative models and generalization as their goal. The view developed in this thesis is that both are expressions of the same mechanism -- the formation of basins of attraction -- and that a model need not be in one regime alone. Associative Memory, an idea that emerged from physics and the concept of energy, provides the language to describe this mechanism, and, as Chapter~\ref{chapter:language-dm-am} and prior work suggest, it does not require an energy function to apply. We hope that this view helps explain why generative models both remember and create, and that it guides the design of models whose balance between the two can be understood and controlled.

%% file: references/references.tex
%
%
\renewcommand{\bibsection}{\specialhead{REFERENCES}}
\setlength{\bibhang}{0.5in}
\begin{singlespace}
\input{references/apa_references.tex}

\end{singlespace}

%% file: references/apa_references.tex

%% file: main.bbl
\begin{thebibliography}{174}
\providecommand{\natexlab}[1]{#1}
\providecommand{\url}[1]{\texttt{#1}}

\bibitem[{Abu-Mostafa \& St. Jacques(1985)Abu-Mostafa \& St. Jacques}]{AbuMostafa1985InformationCO}
Abu-Mostafa, Y.~S., \& St. Jacques, J.-M. (1985). Information capacity of the Hopfield model. \textit{IEEE Transactions on Information Theory, 31}(4), 461--464. \url{https://doi.org/10.1109/TIT.1985.1057069}

\bibitem[{Achilli et~al.(2025)Achilli, Ambrogioni, Lucibello, M{\'e}zard, \& Ventura}]{achilli2025memorization}
Achilli, B., Ambrogioni, L., Lucibello, C., M{\'e}zard, M., \& Ventura, E. (2025). Memorization and generalization in generative diffusion under the manifold hypothesis. \textit{Journal of Statistical Mechanics: Theory and Experiment, 2025}(7), Article 073401. \url{https://doi.org/10.1088/1742-5468/ade136}

\bibitem[{Achilli et~al.(2024)Achilli, Ventura, Silvestri, Pham, Raya, Krotov, Lucibello, \& Ambrogioni}]{achilli2024losing}
Achilli, B., Ventura, E., Silvestri, G., Pham, B., Raya, G., Krotov, D., Lucibello, C., \& Ambrogioni, L. (2024). \textit{Losing dimensions: Geometric memorization in generative diffusion}. arXiv. \url{https://doi.org/10.48550/arXiv.2410.08727}

\bibitem[{Ackley et~al.(1985)Ackley, Hinton, \& Sejnowski}]{AckleyHintonSej1985}
Ackley, D.~H., Hinton, G.~E., \& Sejnowski, T.~J. (1985). A learning algorithm for Boltzmann machines. \textit{Cognitive Science, 9}(1), 147--169. \url{https://doi.org/10.1016/S0364-0213(85)80012-4}

\bibitem[{Agliari et~al.(2020)Agliari, Alemanno, Barra, Centonze, \& Fachechi}]{agliari2020neural}
Agliari, E., Alemanno, F., Barra, A., Centonze, M., \& Fachechi, A. (2020). Neural networks with a redundant representation: Detecting the undetectable. \textit{Physical Review Letters, 124}(2), Article 028301. \url{https://doi.org/10.1103/PhysRevLett.124.028301}

\bibitem[{Agliari \& De Marzo(2020)Agliari \& De Marzo}]{agliari2020tolerance}
Agliari, E., \& De Marzo, G. (2020). Tolerance versus synaptic noise in dense associative memories. \textit{The European Physical Journal Plus, 135}(11), Article 883. \url{https://doi.org/10.1140/epjp/s13360-020-00894-8}

\bibitem[{Albanese et~al.(2022)Albanese, Alemanno, Alessandrelli, \& Barra}]{albanese2022replica}
Albanese, L., Alemanno, F., Alessandrelli, A., \& Barra, A. (2022). Replica symmetry breaking in dense Hebbian neural networks. \textit{Journal of Statistical Physics, 189}(2), Article 24. \url{https://doi.org/10.1007/s10955-022-02966-8}

\bibitem[{Amari(1972)Amari}]{AmariS.I.1972}
Amari, S.-I. (1972). Learning patterns and pattern sequences by self-organizing nets of threshold elements. \textit{IEEE Transactions on Computers, C-21}(11), 1197--1206. \url{https://doi.org/10.1109/T-C.1972.223477}

\bibitem[{Ambrogioni(2024)Ambrogioni}]{ambrogioni2023search}
Ambrogioni, L. (2024). In search of dispersed memories: Generative diffusion models are associative memory networks. \textit{Entropy, 26}(5), Article 381. \url{https://doi.org/10.3390/e26050381}

\bibitem[{Amit et~al.(1985)Amit, Gutfreund, \& Sompolinsky}]{Amit1985StoringIN}
Amit, D.~J., Gutfreund, H., \& Sompolinsky, H. (1985). Storing infinite numbers of patterns in a spin-glass model of neural networks. \textit{Physical Review Letters, 55}(14), 1530--1533. \url{https://doi.org/10.1103/PhysRevLett.55.1530}

\bibitem[{Amit et~al.(1987)Amit, Gutfreund, \& Sompolinsky}]{amit1987statistical}
Amit, D.~J., Gutfreund, H., \& Sompolinsky, H. (1987). Statistical mechanics of neural networks near saturation. \textit{Annals of Physics, 173}(1), 30--67. \url{https://doi.org/10.1016/0003-4916(87)90092-3}

\bibitem[{Anderson(1982)Anderson}]{anderson1982reverse}
Anderson, B.~D.~O. (1982). Reverse-time diffusion equation models. \textit{Stochastic Processes and their Applications, 12}(3), 313--326. \url{https://doi.org/10.1016/0304-4149(82)90051-5}

\bibitem[{Arjovsky et~al.(2017)Arjovsky, Chintala, \& Bottou}]{arjovsky17a}
Arjovsky, M., Chintala, S., \& Bottou, L. (2017). Wasserstein generative adversarial networks. In \textit{Proceedings of the 34th international conference on machine learning} (Vol. 70, pp. 214--223). PMLR.

\bibitem[{Austin et~al.(2021)Austin, Johnson, Ho, Tarlow, \& van den Berg}]{austin2021structured}
Austin, J., Johnson, D.~D., Ho, J., Tarlow, D., \& van den Berg, R. (2021). Structured denoising diffusion models in discrete state-spaces. In \textit{Advances in neural information processing systems} (Vol. 34, pp. 17981--17993). Curran Associates.

\bibitem[{Bai et~al.(2019)Bai, Kolter, \& Koltun}]{bai2019deep}
Bai, S., Kolter, J.~Z., \& Koltun, V. (2019). Deep equilibrium models. In \textit{Advances in neural information processing systems} (Vol. 32). Curran Associates.

\bibitem[{Benedetti et~al.(2022)Benedetti, Ventura, Marinari, Ruocco, \& Zamponi}]{benedetti2022supervised}
Benedetti, M., Ventura, E., Marinari, E., Ruocco, G., \& Zamponi, F. (2022). Supervised perceptron learning vs unsupervised Hebbian unlearning: Approaching optimal memory retrieval in Hopfield-like networks. \textit{The Journal of Chemical Physics, 156}(10), Article 104107. \url{https://doi.org/10.1063/5.0084219}

\bibitem[{Besag(1974)Besag}]{pseudo-likelihood}
Besag, J. (1974). Spatial interaction and the statistical analysis of lattice systems. \textit{Journal of the Royal Statistical Society: Series B (Methodological), 36}(2), 192--236. \url{https://doi.org/10.1111/j.2517-6161.1974.tb00999.x}

\bibitem[{Biroli et~al.(2024)Biroli, Bonnaire, de Bortoli, \& Mézard}]{biroli2024dynamical}
Biroli, G., Bonnaire, T., de Bortoli, V., \& Mézard, M. (2024). Dynamical regimes of diffusion models. \textit{Nature Communications, 15}, Article 9957. \url{https://doi.org/10.1038/s41467-024-54281-3}

\bibitem[{Biroli \& M{\'e}zard(2023)Biroli \& M{\'e}zard}]{biroli2023generative}
Biroli, G., \& M{\'e}zard, M. (2023). Generative diffusion in very large dimensions. \textit{Journal of Statistical Mechanics: Theory and Experiment, 2023}(9), Article 093402. \url{https://doi.org/10.1088/1742-5468/acf8ba}

\bibitem[{Blattmann et~al.(2023)Blattmann, Rombach, Ling, Dockhorn, Kim, Fidler, \& Kreis}]{blattmann2023videoldm}
Blattmann, A., Rombach, R., Ling, H., Dockhorn, T., Kim, S.~W., Fidler, S., \& Kreis, K. (2023). Align your latents: High-resolution video synthesis with latent diffusion models. In \textit{Proceedings of the IEEE/CVF conference on computer vision and pattern recognition} (pp. 22563--22575). IEEE.

\bibitem[{Bonnaire et~al.(2025)Bonnaire, Urfin, Biroli, \& M{\'e}zard}]{bonnaire2025diffusion}
Bonnaire, T., Urfin, R., Biroli, G., \& M{\'e}zard, M. (2025). Why diffusion models don't memorize: The role of implicit dynamical regularization in training. In \textit{Advances in neural information processing systems} (Vol. 38, pp. 141266--141286). Curran Associates. \url{https://doi.org/10.52202/085713-4720}

\bibitem[{Brown et~al.(2020)Brown, Mann, Ryder, Subbiah, Kaplan, Dhariwal, Neelakantan, Shyam, Sastry, Askell, Agarwal, Herbert-Voss, Krueger, Henighan, Child, Ramesh, Ziegler, Wu, Winter, Hesse, Chen, Sigler, Litwin, Gray, Chess, Clark, Berner, McCandlish, Radford, Sutskever, \& Amodei}]{brown2020language}
Brown, T.~B., Mann, B., Ryder, N., Subbiah, M., Kaplan, J., Dhariwal, P., Neelakantan, A., Shyam, P., Sastry, G., Askell, A., Agarwal, S., Herbert-Voss, A., Krueger, G., Henighan, T., Child, R., Ramesh, A., Ziegler, D.~M., Wu, J., Winter, C., \ldots{} Amodei, D. (2020). Language models are few-shot learners. In \textit{Advances in neural information processing systems} (Vol. 33, pp. 1877--1901). Curran Associates.

\bibitem[{Brush(1967)Brush}]{Lenz-Ising-history1967}
Brush, S.~G. (1967). History of the Lenz--Ising model. \textit{Reviews of Modern Physics, 39}(4), 883--893. \url{https://doi.org/10.1103/RevModPhys.39.883}

\bibitem[{Cagnetta et~al.(2026)Cagnetta, Ravent{\'o}s, Ganguli, \& Wyart}]{cagnetta2026deriving}
Cagnetta, F., Ravent{\'o}s, A., Ganguli, S., \& Wyart, M. (2026). Deriving neural scaling laws from the statistics of natural language. In \textit{Proceedings of the 43rd international conference on machine learning} (Vol. 306, pp. 10450--10480). PMLR.

\bibitem[{Campbell et~al.(2024)Campbell, Yim, Barzilay, Rainforth, \& Jaakkola}]{campbell2024generative}
Campbell, A., Yim, J., Barzilay, R., Rainforth, T., \& Jaakkola, T. (2024). Generative flows on discrete state-spaces: Enabling multimodal flows with applications to protein co-design. In \textit{Proceedings of the 41st international conference on machine learning} (Vol. 235, pp. 5453--5512). PMLR.

\bibitem[{Carlini, Hayes, et~al.(2023)Carlini, Hayes, Nasr, Jagielski, Sehwag, Tram\`{e}r, Balle, Ippolito, \& Wallace}]{carlini2023}
Carlini, N., Hayes, J., Nasr, M., Jagielski, M., Sehwag, V., Tram\`{e}r, F., Balle, B., Ippolito, D., \& Wallace, E. (2023). Extracting training data from diffusion models. In \textit{Proceedings of the 32nd USENIX security symposium} (pp. 5253--5270). USENIX Association.

\bibitem[{Carlini, Ippolito, et~al.(2023)Carlini, Ippolito, Jagielski, Lee, Tram{\`e}r, \& Zhang}]{carlini2023quantifying}
Carlini, N., Ippolito, D., Jagielski, M., Lee, K., Tram{\`e}r, F., \& Zhang, C. (2023). Quantifying memorization across neural language models. In \textit{International conference on learning representations}. OpenReview. Retrieved November 6, 2026, from \url{https://openreview.net/forum?id=TatRHT_1cK}

\bibitem[{Carlini et~al.(2021)Carlini, Tram{\`e}r, Wallace, Jagielski, Herbert-Voss, Lee, Roberts, Brown, Song, Erlingsson, Oprea, \& Raffel}]{carlini2021extracting}
Carlini, N., Tram{\`e}r, F., Wallace, E., Jagielski, M., Herbert-Voss, A., Lee, K., Roberts, A., Brown, T., Song, D., Erlingsson, {\'U}., Oprea, A., \& Raffel, C. (2021). Extracting training data from large language models. In \textit{Proceedings of the 30th USENIX security symposium} (pp. 2633--2650). USENIX Association.

\bibitem[{Chelba et~al.(2014)Chelba, Mikolov, Schuster, Ge, Brants, Koehn, \& Robinson}]{chelba2013one}
Chelba, C., Mikolov, T., Schuster, M., Ge, Q., Brants, T., Koehn, P., \& Robinson, T. (2014). One billion word benchmark for measuring progress in statistical language modeling. In \textit{Proceedings of Interspeech 2014} (pp. 2635--2639). International Speech Communication Association. \url{https://doi.org/10.21437/Interspeech.2014-564}

\bibitem[{Chen et~al.(2018)Chen, Rubanova, Bettencourt, \& Duvenaud}]{NeuralODE}
Chen, R.~T.~Q., Rubanova, Y., Bettencourt, J., \& Duvenaud, D. (2018). Neural ordinary differential equations. In \textit{Advances in neural information processing systems} (Vol. 31, pp. 6571--6583). Curran Associates.

\bibitem[{Chen et~al.(2021)Chen, Zhang, Zen, Weiss, Norouzi, \& Chan}]{chen2020wavegrad}
Chen, N., Zhang, Y., Zen, H., Weiss, R.~J., Norouzi, M., \& Chan, W. (2021). WaveGrad: Estimating gradients for waveform generation. In \textit{International conference on learning representations}. OpenReview. Retrieved November 6, 2026, from \url{https://openreview.net/forum?id=NsMLjcFaO8O}

\bibitem[{Cheung et~al.(1987)Cheung, Atlas, \& Marks}]{Cheung87}
Cheung, K.~F., Atlas, L.~E., \& Marks, R.~J. (1987). Synchronous vs asynchronous behavior of Hopfield's CAM neural net. \textit{Applied Optics, 26}(22), 4808--4813. \url{https://doi.org/10.1364/AO.26.004808}

\bibitem[{Cui et~al.(2025)Cui, Pehlevan, \& Lu}]{cui2025precise}
Cui, H., Pehlevan, C., \& Lu, Y.~M. (2025). A solvable model of learning generative diffusion: Theory and insights. In \textit{Advances in neural information processing systems} (Vol. 38, pp. 5253--5296). Curran Associates. \url{https://doi.org/10.52202/085713-0187}

\bibitem[{D'Amico, Bocchi, et~al.(2026)D'Amico, Bocchi, Del Bono, Rossi, \& Negri}]{d2025pseudo}
D'Amico, F., Bocchi, D., Del Bono, L.~M., Rossi, S., \& Negri, M. (2026). Pseudo-likelihood produces associative memories able to generalize, even for asymmetric couplings. \textit{Physica A: Statistical Mechanics and its Applications, 692}, Article 131497. \url{https://doi.org/10.1016/j.physa.2026.131497}

\bibitem[{D'Amico, Bocchi, \& Negri(2026)D'Amico, Bocchi, \& Negri}]{damico2026implicit}
D'Amico, F., Bocchi, D., \& Negri, M. (2026). Implicit bias produces neural scaling laws in learning curves, from perceptrons to deep networks. In \textit{International conference on learning representations}. OpenReview. Retrieved November 6, 2026, from \url{https://openreview.net/forum?id=qBAV2DEvAC}

\bibitem[{Dehghani et~al.(2019)Dehghani, Gouws, Vinyals, Uszkoreit, \& Kaiser}]{dehghani2019universal}
Dehghani, M., Gouws, S., Vinyals, O., Uszkoreit, J., \& Kaiser, {\L}. (2019). Universal transformers. In \textit{International conference on learning representations}. OpenReview. Retrieved November 6, 2026, from \url{https://openreview.net/forum?id=HyzdRiR9Y7}

\bibitem[{Demircigil et~al.(2017)Demircigil, Heusel, L{\"o}we, Upgang, \& Vermet}]{demircigil2017model}
Demircigil, M., Heusel, J., L{\"o}we, M., Upgang, S., \& Vermet, F. (2017). On a model of associative memory with huge storage capacity. \textit{Journal of Statistical Physics, 168}(2), 288--299. \url{https://doi.org/10.1007/s10955-017-1806-y}

\bibitem[{Deng(2012)Deng}]{mnist}
Deng, L. (2012). The MNIST database of handwritten digit images for machine learning research. \textit{IEEE Signal Processing Magazine, 29}(6), 141--142. \url{https://doi.org/10.1109/MSP.2012.2211477}

\bibitem[{Deng et~al.(2020)Deng, Bakhtin, Ott, Szlam, \& Ranzato}]{deng2020residual}
Deng, Y., Bakhtin, A., Ott, M., Szlam, A., \& Ranzato, M. (2020). Residual energy-based models for text generation. In \textit{International conference on learning representations}. OpenReview. Retrieved November 6, 2026, from \url{https://openreview.net/forum?id=B1l4SgHKDH}

\bibitem[{Deng et~al.(2009)Deng, Dong, Socher, Li, Li, \& Fei-Fei}]{deng2009imagenet}
Deng, J., Dong, W., Socher, R., Li, L.-J., Li, K., \& Fei-Fei, L. (2009). ImageNet: A large-scale hierarchical image database. In \textit{Proceedings of the IEEE conference on computer vision and pattern recognition} (pp. 248--255). IEEE. \url{https://doi.org/10.1109/CVPR.2009.5206848}

\bibitem[{Dinh et~al.(2014)Dinh, Krueger, \& Bengio}]{dinh2014nice}
Dinh, L., Krueger, D., \& Bengio, Y. (2014). \textit{NICE: Non-linear independent components estimation}. arXiv. \url{https://doi.org/10.48550/arXiv.1410.8516}

\bibitem[{Dinh et~al.(2017)Dinh, Sohl-Dickstein, \& Bengio}]{dinh2016density}
Dinh, L., Sohl-Dickstein, J., \& Bengio, S. (2017). Density estimation using Real NVP. In \textit{International conference on learning representations}. OpenReview. Retrieved November 6, 2026, from \url{https://openreview.net/forum?id=HkpbnH9lx}

\bibitem[{Dormand \& Prince(1980)Dormand \& Prince}]{DORMAND198019}
Dormand, J.~R., \& Prince, P.~J. (1980). A family of embedded Runge--Kutta formulae. \textit{Journal of Computational and Applied Mathematics, 6}(1), 19--26. \url{https://doi.org/10.1016/0771-050X(80)90013-3}

\bibitem[{Du et~al.(2021)Du, Li, Tenenbaum, \& Mordatch}]{du2020improved}
Du, Y., Li, S., Tenenbaum, J., \& Mordatch, I. (2021). Improved contrastive divergence training of energy-based models. In \textit{Proceedings of the 38th international conference on machine learning} (Vol. 139, pp. 2837--2848). PMLR.

\bibitem[{Du et~al.(2024)Du, Mao, \& Tenenbaum}]{du2024learning}
Du, Y., Mao, J., \& Tenenbaum, J.~B. (2024). Learning iterative reasoning through energy diffusion. In \textit{Proceedings of the 41st international conference on machine learning} (Vol. 235, pp. 11764--11776). PMLR.

\bibitem[{Du \& Mordatch(2019)Du \& Mordatch}]{du2019implicit}
Du, Y., \& Mordatch, I. (2019). Implicit generation and modeling with energy based models. In \textit{Advances in neural information processing systems} (Vol. 32). Curran Associates.

\bibitem[{Eccles(1953)Eccles}]{Eccles1953}
Eccles, J.~C. (1953). \textit{The neurophysiological basis of mind: The principles of neurophysiology}. Clarendon Press.

\bibitem[{Elman(1990)Elman}]{ElmanRNN}
Elman, J.~L. (1990). Finding structure in time. \textit{Cognitive Science, 14}(2), 179--211. \url{https://doi.org/10.1207/s15516709cog1402_1}

\bibitem[{Farn{\'e} et~al.(2026)Farn{\'e}, Boncoraglio, \& Zdeborov{\'a}}]{farne2026rules}
Farn{\'e}, G., Boncoraglio, F., \& Zdeborov{\'a}, L. (2026). \textit{The rules-and-facts model for simultaneous generalization and memorization in neural networks}. arXiv. \url{https://doi.org/10.48550/arXiv.2603.25579}

\bibitem[{Favero et~al.(2026)Favero, Sclocchi, \& Wyart}]{favero2025bigger}
Favero, A., Sclocchi, A., \& Wyart, M. (2026). Bigger isn't always memorizing: Early stopping overparameterized diffusion models. \textit{Transactions on Machine Learning Research}. Retrieved November 6, 2026, from \url{https://openreview.net/forum?id=P6r9OxZ1vm}

\bibitem[{Feldman(2020)Feldman}]{feldman2020does}
Feldman, V. (2020). Does learning require memorization? A short tale about a long tail. In \textit{Proceedings of the 52nd annual ACM SIGACT symposium on theory of computing} (pp. 954--959). Association for Computing Machinery. \url{https://doi.org/10.1145/3357713.3384290}

\bibitem[{Forrest(1988)Forrest}]{forrest1988content}
Forrest, B.~M. (1988). Content-addressability and learning in neural networks. \textit{Journal of Physics A: Mathematical and General, 21}(1), 245--255. \url{https://doi.org/10.1088/0305-4470/21/1/029}

\bibitem[{Gardner(1988)Gardner}]{gardner1988space}
Gardner, E. (1988). The space of interactions in neural network models. \textit{Journal of Physics A: Mathematical and General, 21}(1), 257--270. \url{https://doi.org/10.1088/0305-4470/21/1/030}

\bibitem[{Gardner \& Derrida(1988)Gardner \& Derrida}]{Gardner1988OptimalSP}
Gardner, E., \& Derrida, B. (1988). Optimal storage properties of neural network models. \textit{Journal of Physics A: Mathematical and General, 21}(1), 271--284. \url{https://doi.org/10.1088/0305-4470/21/1/031}

\bibitem[{Gat et~al.(2024)Gat, Remez, Shaul, Kreuk, Chen, Synnaeve, Adi, \& Lipman}]{gat2024discrete}
Gat, I., Remez, T., Shaul, N., Kreuk, F., Chen, R.~T., Synnaeve, G., Adi, Y., \& Lipman, Y. (2024). Discrete flow matching. In \textit{Advances in neural information processing systems} (Vol. 37, pp. 133345--133385). Curran Associates. \url{https://doi.org/10.52202/079017-4239}

\bibitem[{Geiping et~al.(2025)Geiping, McLeish, Jain, Kirchenbauer, Singh, Bartoldson, Kailkhura, Bhatele, \& Goldstein}]{geiping2025scaling}
Geiping, J., McLeish, S., Jain, N., Kirchenbauer, J., Singh, S., Bartoldson, B.~R., Kailkhura, B., Bhatele, A., \& Goldstein, T. (2025). Scaling up test-time compute with latent reasoning: A recurrent depth approach. In \textit{Advances in neural information processing systems} (Vol. 38, pp. 41340--41391). Curran Associates. \url{https://doi.org/10.52202/085713-1380}

\bibitem[{Gelfand \& Smith(1990)Gelfand \& Smith}]{GelfandA1990}
Gelfand, A.~E., \& Smith, A.~F.~M. (1990). Sampling-based approaches to calculating marginal densities. \textit{Journal of the American Statistical Association, 85}(410), 398--409. \url{https://doi.org/10.1080/01621459.1990.10476213}

\bibitem[{Germain et~al.(2015)Germain, Gregor, Murray, \& Larochelle}]{germain15}
Germain, M., Gregor, K., Murray, I., \& Larochelle, H. (2015). MADE: Masked autoencoder for distribution estimation. In \textit{Proceedings of the 32nd international conference on machine learning} (Vol. 37, pp. 881--889). PMLR.

\bibitem[{Gladstone et~al.(2026)Gladstone, Nanduru, Islam, Han, Ha, Chadha, Du, Ji, Li, \& Iqbal}]{gladstone2026energy}
Gladstone, A., Nanduru, G., Islam, M.~M., Han, P., Ha, H., Chadha, A., Du, Y., Ji, H., Li, J., \& Iqbal, T. (2026). Energy-based transformers are scalable learners and thinkers. In \textit{International conference on learning representations}. OpenReview. Retrieved November 6, 2026, from \url{https://openreview.net/forum?id=ZBj3Qp1bYg}

\bibitem[{Gradshteyn \& Ryzhik(2014)Gradshteyn \& Ryzhik}]{gradshteyn2014table}
Gradshteyn, I.~S., \& Ryzhik, I.~M. (2014). \textit{Table of integrals, series, and products} (8th ed.). Academic Press.

\bibitem[{Graves(2013)Graves}]{graves2013generating}
Graves, A. (2013). \textit{Generating sequences with recurrent neural networks}. arXiv. \url{https://doi.org/10.48550/arXiv.1308.0850}

\bibitem[{Gu et~al.(2025)Gu, Du, Pang, Li, Lin, \& Wang}]{gu2023memorization}
Gu, X., Du, C., Pang, T., Li, C., Lin, M., \& Wang, Y. (2025). On memorization in diffusion models. \textit{Transactions on Machine Learning Research}. Retrieved November 6, 2026, from \url{https://openreview.net/forum?id=D3DBqvSDbj}

\bibitem[{Gutmann \& Hyv{\"a}rinen(2010)Gutmann \& Hyv{\"a}rinen}]{gutmann2010noise}
Gutmann, M., \& Hyv{\"a}rinen, A. (2010). Noise-contrastive estimation: A new estimation principle for unnormalized statistical models. In \textit{Proceedings of the thirteenth international conference on artificial intelligence and statistics} (Vol. 9, pp. 297--304). PMLR.

\bibitem[{He et~al.(2016)He, Zhang, Ren, \& Sun}]{he2016deep}
He, K., Zhang, X., Ren, S., \& Sun, J. (2016). Deep residual learning for image recognition. In \textit{Proceedings of the IEEE conference on computer vision and pattern recognition} (pp. 770--778). IEEE. \url{https://doi.org/10.1109/CVPR.2016.90}

\bibitem[{Hebb(1949)Hebb}]{Hebb1949}
Hebb, D.~O. (1949). \textit{The organization of behavior: A neuropsychological theory}. Wiley.

\bibitem[{Hertz et~al.(1991)Hertz, Krogh, \& Palmer}]{HertzJ-Intro-NeuralComp}
Hertz, J., Krogh, A., \& Palmer, R.~G. (1991). \textit{Introduction to the theory of neural computation}. Addison-Wesley.

\bibitem[{Heusel et~al.(2017)Heusel, Ramsauer, Unterthiner, Nessler, \& Hochreiter}]{heusel2017gans}
Heusel, M., Ramsauer, H., Unterthiner, T., Nessler, B., \& Hochreiter, S. (2017). GANs trained by a two time-scale update rule converge to a local Nash equilibrium. In \textit{Advances in neural information processing systems} (Vol. 30, pp. 6626--6637). Curran Associates.

\bibitem[{Hinton(2002)Hinton}]{hinton2002training}
Hinton, G.~E. (2002). Training products of experts by minimizing contrastive divergence. \textit{Neural Computation, 14}(8), 1771--1800. \url{https://doi.org/10.1162/089976602760128018}

\bibitem[{Ho et~al.(2020)Ho, Jain, \& Abbeel}]{ho2020denoising}
Ho, J., Jain, A., \& Abbeel, P. (2020). Denoising diffusion probabilistic models. In \textit{Advances in neural information processing systems} (Vol. 33, pp. 6840--6851). Curran Associates.

\bibitem[{Ho et~al.(2022)Ho, Salimans, Gritsenko, Chan, Norouzi, \& Fleet}]{ho2022video}
Ho, J., Salimans, T., Gritsenko, A., Chan, W., Norouzi, M., \& Fleet, D.~J. (2022). Video diffusion models. In \textit{Advances in neural information processing systems} (Vol. 35, pp. 8633--8646). Curran Associates. \url{https://doi.org/10.52202/068431-0628}

\bibitem[{Hochreiter \& Schmidhuber(1997)Hochreiter \& Schmidhuber}]{LSTM}
Hochreiter, S., \& Schmidhuber, J. (1997). Long short-term memory. \textit{Neural Computation, 9}(8), 1735--1780. \url{https://doi.org/10.1162/neco.1997.9.8.1735}

\bibitem[{Hoffmann et~al.(2022)Hoffmann, Borgeaud, Mensch, Buchatskaya, Cai, Rutherford, de Las Casas, Hendricks, Welbl, Clark, Hennigan, Noland, Millican, van den Driessche, Damoc, Guy, Osindero, Simonyan, Elsen, Rae, Vinyals, \& Sifre}]{hoffmann2022training}
Hoffmann, J., Borgeaud, S., Mensch, A., Buchatskaya, E., Cai, T., Rutherford, E., de Las Casas, D., Hendricks, L.~A., Welbl, J., Clark, A., Hennigan, T., Noland, E., Millican, K., van den Driessche, G., Damoc, B., Guy, A., Osindero, S., Simonyan, K., Elsen, E., \ldots{} Sifre, L. (2022). An empirical analysis of compute-optimal large language model training. In \textit{Advances in neural information processing systems} (Vol. 35, pp. 30016--30030). Curran Associates. \url{https://doi.org/10.52202/068431-2176}

\bibitem[{Hoover, Liang, et~al.(2023)Hoover, Liang, Pham, Panda, Strobelt, Chau, Zaki, \& Krotov}]{et}
Hoover, B., Liang, Y., Pham, B., Panda, R., Strobelt, H., Chau, D.~H., Zaki, M., \& Krotov, D. (2023). Energy transformer. In \textit{Advances in neural information processing systems} (Vol. 36, pp. 27532--27559). Curran Associates. \url{https://doi.org/10.52202/075280-1197}

\bibitem[{Hoover, Strobelt, et~al.(2023)Hoover, Strobelt, Krotov, Hoffman, Kira, \& Chau}]{hoover2023memory}
Hoover, B., Strobelt, H., Krotov, D., Hoffman, J., Kira, Z., \& Chau, D.~H. (2023). \textit{Memory in plain sight: A survey of the uncanny resemblances between diffusion models and associative memories}. arXiv. \url{https://doi.org/10.48550/arXiv.2309.16750}

\bibitem[{Hopfield(1982)Hopfield}]{Hopfield1982}
Hopfield, J.~J. (1982). Neural networks and physical systems with emergent collective computational abilities. \textit{Proceedings of the National Academy of Sciences, 79}(8), 2554--2558. \url{https://doi.org/10.1073/pnas.79.8.2554}

\bibitem[{Hopfield(1984)Hopfield}]{Hopfield1984}
Hopfield, J.~J. (1984). Neurons with graded response have collective computational properties like those of two-state neurons. \textit{Proceedings of the National Academy of Sciences, 81}(10), 3088--3092. \url{https://doi.org/10.1073/pnas.81.10.3088}

\bibitem[{Hopfield et~al.(1983)Hopfield, Feinstein, \& Palmer}]{Hopfield1983UnlearningHA}
Hopfield, J.~J., Feinstein, D.~I., \& Palmer, R.~G. (1983). `Unlearning' has a stabilizing effect in collective memories. \textit{Nature, 304}(5922), 158--159. \url{https://doi.org/10.1038/304158a0}

\bibitem[{Hutchinson(1990)Hutchinson}]{Hutchinson1990}
Hutchinson, M.~F. (1990). A stochastic estimator of the trace of the influence matrix for Laplacian smoothing splines. \textit{Communications in Statistics -- Simulation and Computation, 19}(2), 433--450. \url{https://doi.org/10.1080/03610919008812866}

\bibitem[{Hyv{{\"a}}rinen(2005)Hyv{{\"a}}rinen}]{hyvarinen05a}
Hyv{{\"a}}rinen, A. (2005). Estimation of non-normalized statistical models by score matching. \textit{Journal of Machine Learning Research, 6}(24), 695--709.

\bibitem[{Jeon et~al.(2025)Jeon, Kim, \& No}]{jeon2024understanding}
Jeon, D., Kim, D., \& No, A. (2025). Understanding and mitigating memorization in generative models via sharpness of probability landscapes. In \textit{Proceedings of the 42nd international conference on machine learning} (Vol. 267, pp. 27091--27112). PMLR.

\bibitem[{Ji et~al.(2023)Ji, Lee, Frieske, Yu, Su, Xu, Ishii, Bang, Madotto, \& Fung}]{ji2023survey}
Ji, Z., Lee, N., Frieske, R., Yu, T., Su, D., Xu, Y., Ishii, E., Bang, Y.~J., Madotto, A., \& Fung, P. (2023). Survey of hallucination in natural language generation. \textit{ACM Computing Surveys, 55}(12), Article 248. \url{https://doi.org/10.1145/3571730}

\bibitem[{Jolicoeur-Martineau(2025)Jolicoeur-Martineau}]{jolicoeur2025less}
Jolicoeur-Martineau, A. (2025). \textit{Less is more: Recursive reasoning with tiny networks}. arXiv. \url{https://doi.org/10.48550/arXiv.2510.04871}

\bibitem[{Jordan(1997)Jordan}]{JordanRNN}
Jordan, M.~I. (1997). Serial order: A parallel distributed processing approach. In J.~W. Donahoe \& V.~P. Dorsel (Eds.), \textit{Neural-network models of cognition} (Vol. 121, pp. 471--495). North-Holland. \url{https://doi.org/10.1016/S0166-4115(97)80111-2}

\bibitem[{Kadkhodaie et~al.(2024)Kadkhodaie, Guth, Simoncelli, \& Mallat}]{kadkhodaie2023generalization}
Kadkhodaie, Z., Guth, F., Simoncelli, E.~P., \& Mallat, S. (2024). Generalization in diffusion models arises from geometry-adaptive harmonic representations. In \textit{International conference on learning representations}. OpenReview. Retrieved November 6, 2026, from \url{https://openreview.net/forum?id=ANvmVS2Yr0}

\bibitem[{Kalaj et~al.(2025)Kalaj, Lauditi, Perugini, Lucibello, Malatesta, \& Negri}]{kalaj2024random}
Kalaj, S., Lauditi, C., Perugini, G., Lucibello, C., Malatesta, E.~M., \& Negri, M. (2025). Random features Hopfield networks generalize retrieval to previously unseen examples. \textit{Physica A: Statistical Mechanics and its Applications, 678}, Article 130946.

\bibitem[{Kamb \& Ganguli(2025)Kamb \& Ganguli}]{kamb2024analytic}
Kamb, M., \& Ganguli, S. (2025). An analytic theory of creativity in convolutional diffusion models. In \textit{Proceedings of the 42nd international conference on machine learning} (Vol. 267, pp. 28795--28831). PMLR.

\bibitem[{Kaplan et~al.(2020)Kaplan, McCandlish, Henighan, Brown, Chess, Child, Gray, Radford, Wu, \& Amodei}]{kaplan2020scaling}
Kaplan, J., McCandlish, S., Henighan, T., Brown, T.~B., Chess, B., Child, R., Gray, S., Radford, A., Wu, J., \& Amodei, D. (2020). \textit{Scaling laws for neural language models}. arXiv. \url{https://doi.org/10.48550/arXiv.2001.08361}

\bibitem[{Kingma \& Ba(2015)Kingma \& Ba}]{Adam}
Kingma, D.~P., \& Ba, J. (2015). Adam: A method for stochastic optimization. In \textit{International conference on learning representations}. Retrieved November 6, 2026, from \url{https://arxiv.org/abs/1412.6980}

\bibitem[{Kingma \& Dhariwal(2018)Kingma \& Dhariwal}]{kingma2018glow}
Kingma, D.~P., \& Dhariwal, P. (2018). Glow: Generative flow with invertible 1$\times$1 convolutions. In \textit{Advances in neural information processing systems} (Vol. 31, pp. 10215--10224). Curran Associates.

\bibitem[{Kingma \& Welling(2014)Kingma \& Welling}]{kingmaVAE}
Kingma, D.~P., \& Welling, M. (2014). \textit{Auto-encoding variational Bayes}. arXiv. \url{https://doi.org/10.48550/arXiv.1312.6114}

\bibitem[{Kingma \& Welling(2019)Kingma \& Welling}]{kingmaIntroVAE2019}
Kingma, D.~P., \& Welling, M. (2019). An introduction to variational autoencoders. \textit{Foundations and Trends in Machine Learning, 12}(4), 307--392. \url{https://doi.org/10.1561/2200000056}

\bibitem[{Kloeden \& Platen(1992)Kloeden \& Platen}]{kloeden1992stochastic}
Kloeden, P.~E., \& Platen, E. (1992). \textit{Numerical solution of stochastic differential equations}. Springer. \url{https://doi.org/10.1007/978-3-662-12616-5}

\bibitem[{Kojima et~al.(2022)Kojima, Gu, Reid, Matsuo, \& Iwasawa}]{kojima2022large}
Kojima, T., Gu, S.~S., Reid, M., Matsuo, Y., \& Iwasawa, Y. (2022). Large language models are zero-shot reasoners. In \textit{Advances in neural information processing systems} (Vol. 35, pp. 22199--22213). Curran Associates. \url{https://doi.org/10.52202/068431-1613}

\bibitem[{Kong et~al.(2021)Kong, Ping, Huang, Zhao, \& Catanzaro}]{kong2020diffwave}
Kong, Z., Ping, W., Huang, J., Zhao, K., \& Catanzaro, B. (2021). DiffWave: A versatile diffusion model for audio synthesis. In \textit{International conference on learning representations}. OpenReview. Retrieved November 6, 2026, from \url{https://openreview.net/forum?id=a-xFK8Ymz5J}

\bibitem[{Kozachkov et~al.(2025)Kozachkov, Slotine, \& Krotov}]{kozachkov2025neuron}
Kozachkov, L., Slotine, J.-J., \& Krotov, D. (2025). Neuron--astrocyte associative memory. \textit{Proceedings of the National Academy of Sciences, 122}(21), Article e2417788122. \url{https://doi.org/10.1073/pnas.2417788122}

\bibitem[{Krizhevsky(2009)Krizhevsky}]{cifar10}
Krizhevsky, A. (2009). \textit{Learning multiple layers of features from tiny images} [Technical report]. University of Toronto.

\bibitem[{Krizhevsky et~al.(2012)Krizhevsky, Sutskever, \& Hinton}]{AlexNet}
Krizhevsky, A., Sutskever, I., \& Hinton, G.~E. (2012). ImageNet classification with deep convolutional neural networks. In \textit{Advances in neural information processing systems} (Vol. 25, pp. 1097--1105). Curran Associates.

\bibitem[{Krotov(2021)Krotov}]{krotov2021hierarchical}
Krotov, D. (2021). \textit{Hierarchical associative memory}. arXiv. \url{https://doi.org/10.48550/arXiv.2107.06446}

\bibitem[{Krotov(2023)Krotov}]{krotov2023new}
Krotov, D. (2023). A new frontier for Hopfield networks. \textit{Nature Reviews Physics, 5}, 366--367. \url{https://doi.org/10.1038/s42254-023-00595-y}

\bibitem[{Krotov et~al.(2025)Krotov, Hoover, Ram, \& Pham}]{krotov2025modern}
Krotov, D., Hoover, B., Ram, P., \& Pham, B. (2025, July). \textit{Modern methods in associative memory} [Tutorial]. 42nd International Conference on Machine Learning, Vancouver, Canada.

\bibitem[{Krotov \& Hopfield(2016)Krotov \& Hopfield}]{DAM2016}
Krotov, D., \& Hopfield, J.~J. (2016). Dense associative memory for pattern recognition. In \textit{Advances in neural information processing systems} (Vol. 29, pp. 1172--1180). Curran Associates.

\bibitem[{Krotov \& Hopfield(2018)Krotov \& Hopfield}]{krotov2018dense}
Krotov, D., \& Hopfield, J. (2018). Dense associative memory is robust to adversarial inputs. \textit{Neural Computation, 30}(12), 3151--3167. \url{https://doi.org/10.1162/neco_a_01143}

\bibitem[{Krotov \& Hopfield(2019)Krotov \& Hopfield}]{krotov2019unsupervised}
Krotov, D., \& Hopfield, J.~J. (2019). Unsupervised learning by competing hidden units. \textit{Proceedings of the National Academy of Sciences, 116}(16), 7723--7731. \url{https://doi.org/10.1073/pnas.1820458116}

\bibitem[{Krotov \& Hopfield(2021)Krotov \& Hopfield}]{krotov2021large}
Krotov, D., \& Hopfield, J.~J. (2021). Large associative memory problem in neurobiology and machine learning. In \textit{International conference on learning representations}. OpenReview. Retrieved November 6, 2026, from \url{https://openreview.net/forum?id=X4y_10OX-hX}

\bibitem[{Kullback \& Leibler(1951)Kullback \& Leibler}]{kullback1951information}
Kullback, S., \& Leibler, R.~A. (1951). On information and sufficiency. \textit{The Annals of Mathematical Statistics, 22}(1), 79--86. \url{https://doi.org/10.1214/aoms/1177729694}

\bibitem[{LeCun et~al.(2006)LeCun, Chopra, Hadsell, Ranzato, \& Huang}]{LeCun2006ATO}
LeCun, Y., Chopra, S., Hadsell, R., Ranzato, M., \& Huang, F.~J. (2006). A tutorial on energy-based learning. In G. Bak{\i}r, T. Hofmann, B. Sch{\"o}lkopf, A.~J. Smola, B. Taskar, \& S.~V.~N. Vishwanathan (Eds.), \textit{Predicting structured data} (pp. 191--246). MIT Press.

\bibitem[{Li \& Chen(2024)Li \& Chen}]{li2024critical}
Li, M., \& Chen, S. (2024). Critical windows: Non-asymptotic theory for feature emergence in diffusion models. In \textit{Proceedings of the 41st international conference on machine learning} (Vol. 235, pp. 27474--27498). PMLR.

\bibitem[{Li et~al.(2023)Li, Li, Zhang, \& Bian}]{li2024generalization}
Li, P., Li, Z., Zhang, H., \& Bian, J. (2023). On the generalization properties of diffusion models. In \textit{Advances in neural information processing systems} (Vol. 36, pp. 2097--2127). Curran Associates. \url{https://doi.org/10.52202/075280-0101}

\bibitem[{Liu et~al.(2021)Liu, Yan, Oztekin, \& Ji}]{liu2021graphebm}
Liu, M., Yan, K., Oztekin, B., \& Ji, S. (2021). \textit{GraphEBM: Molecular graph generation with energy-based models}. arXiv. \url{https://doi.org/10.48550/arXiv.2102.00546}

\bibitem[{Lyapunov(1992)Lyapunov}]{lyapunov1992general}
Lyapunov, A.~M. (1992). The general problem of the stability of motion. \textit{International Journal of Control, 55}(3), 531--773. \url{https://doi.org/10.1080/00207179208934253}

\bibitem[{Ma et~al.(2025)Ma, Tong, Jia, Hu, Su, Zhang, Yang, Li, Jaakkola, Jia, \& Xie}]{ma2025inference}
Ma, N., Tong, S., Jia, H., Hu, H., Su, Y.-C., Zhang, M., Yang, X., Li, Y., Jaakkola, T., Jia, X., \& Xie, S. (2025). Scaling inference time compute for diffusion models. In \textit{Proceedings of the IEEE/CVF conference on computer vision and pattern recognition} (pp. 2523--2534). IEEE.

\bibitem[{Meehan et~al.(2020)Meehan, Chaudhuri, \& Dasgupta}]{meehan2020non}
Meehan, C., Chaudhuri, K., \& Dasgupta, S. (2020). A three sample hypothesis test for evaluating generative models. In \textit{Proceedings of the twenty third international conference on artificial intelligence and statistics} (Vol. 108, pp. 3546--3556). PMLR.

\bibitem[{Millidge et~al.(2022)Millidge, Salvatori, Song, Lukasiewicz, \& Bogacz}]{millidge2022universal}
Millidge, B., Salvatori, T., Song, Y., Lukasiewicz, T., \& Bogacz, R. (2022). Universal Hopfield networks: A general framework for single-shot associative memory models. In \textit{Proceedings of the 39th international conference on machine learning} (Vol. 162, pp. 15561--15583). PMLR.

\bibitem[{Montanari et~al.(2024)Montanari, Zhong, \& Zhou}]{Montanari2024NegativePerceptron}
Montanari, A., Zhong, Y., \& Zhou, K. (2024). Tractability from overparametrization: The example of the negative perceptron. \textit{Probability Theory and Related Fields, 188}(3--4), 805--910. \url{https://doi.org/10.1007/s00440-023-01248-y}

\bibitem[{Ngiam et~al.(2011)Ngiam, Chen, Koh, \& Ng}]{ngiam2011}
Ngiam, J., Chen, Z., Koh, P.~W., \& Ng, A.~Y. (2011). Learning deep energy models. In \textit{Proceedings of the 28th international conference on machine learning} (pp. 1105--1112). Omnipress.

\bibitem[{Niu et~al.(2024)Niu, Bai, Deng, \& Han}]{niu2024beyond}
Niu, X., Bai, B., Deng, L., \& Han, W. (2024). \textit{Beyond scaling laws: Understanding transformer performance with associative memory}. arXiv. \url{https://doi.org/10.48550/arXiv.2405.08707}

\bibitem[{{\O}ksendal(1992){\O}ksendal}]{OksendalSDE}
{\O}ksendal, B. (1992). \textit{Stochastic differential equations: An introduction with applications} (3rd ed.). Springer.

\bibitem[{Parisi(1981)Parisi}]{parisiLangevin}
Parisi, G. (1981). Correlation functions and computer simulations. \textit{Nuclear Physics B, 180}(3), 378--384. \url{https://doi.org/10.1016/0550-3213(81)90056-0}

\bibitem[{Peebles \& Xie(2023)Peebles \& Xie}]{peebles2023scalable}
Peebles, W., \& Xie, S. (2023). Scalable diffusion models with transformers. In \textit{Proceedings of the IEEE/CVF international conference on computer vision} (pp. 4195--4205). IEEE.

\bibitem[{Pham, Hoover, et~al.(2026)Pham, Hoover, Krotov, \& Ram}]{pham2026energy}
Pham, B., Hoover, B., Krotov, D., \& Ram, P. (2026, April). \textit{Energy minimization for training dense associative memory} [Paper presentation]. ICLR 2026 Workshop on New Frontiers in Associative Memories, Rio de Janeiro, Brazil. Retrieved November 6, 2026, from \url{https://openreview.net/forum?id=d7PjRsguov}

\bibitem[{Pham, Raya, et~al.(2026)Pham, Raya, Negri, Zaki, Ambrogioni, \& Krotov}]{pham2025memorization}
Pham, B., Raya, G., Negri, M., Zaki, M.~J., Ambrogioni, L., \& Krotov, D. (2026). Emergence of diffusion models from associative memory through memorization and generalization. \textit{Nature Communications}. \url{https://doi.org/10.1038/s41467-026-78727-y}

\bibitem[{Radford et~al.(2019)Radford, Wu, Child, Luan, Amodei, \& Sutskever}]{radford2019language}
Radford, A., Wu, J., Child, R., Luan, D., Amodei, D., \& Sutskever, I. (2019). \textit{Language models are unsupervised multitask learners} [Technical report]. OpenAI. Retrieved November 6, 2026, from \url{https://cdn.openai.com/better-language-models/language_models_are_unsupervised_multitask_learners.pdf}

\bibitem[{Ramachandran et~al.(2017)Ramachandran, Zoph, \& Le}]{Swish}
Ramachandran, P., Zoph, B., \& Le, Q.~V. (2017). \textit{Searching for activation functions}. arXiv. \url{https://doi.org/10.48550/arXiv.1710.05941}

\bibitem[{Ramsauer et~al.(2021)Ramsauer, Sch{\"a}fl, Lehner, Seidl, Widrich, Gruber, Holzleitner, Adler, Kreil, Kopp, Klambauer, Brandstetter, \& Hochreiter}]{Ramsauer2021}
Ramsauer, H., Sch{\"a}fl, B., Lehner, J., Seidl, P., Widrich, M., Gruber, L., Holzleitner, M., Adler, T., Kreil, D., Kopp, M.~K., Klambauer, G., Brandstetter, J., \& Hochreiter, S. (2021). Hopfield networks is all you need. In \textit{International conference on learning representations}. OpenReview. Retrieved November 6, 2026, from \url{https://openreview.net/forum?id=tL89RnzIiCd}

\bibitem[{Raya \& Ambrogioni(2023)Raya \& Ambrogioni}]{raya2024spontaneous}
Raya, G., \& Ambrogioni, L. (2023). Spontaneous symmetry breaking in generative diffusion models. In \textit{Advances in neural information processing systems} (Vol. 36, pp. 66377--66389). Curran Associates. \url{https://doi.org/10.52202/075280-2897}

\bibitem[{Rezende \& Mohamed(2015)Rezende \& Mohamed}]{pmlr-v37-rezende15}
Rezende, D., \& Mohamed, S. (2015). Variational inference with normalizing flows. In \textit{Proceedings of the 32nd international conference on machine learning} (Vol. 37, pp. 1530--1538). PMLR.

\bibitem[{Rezende et~al.(2014)Rezende, Mohamed, \& Wierstra}]{rezende14}
Rezende, D.~J., Mohamed, S., \& Wierstra, D. (2014). Stochastic backpropagation and approximate inference in deep generative models. In \textit{Proceedings of the 31st international conference on machine learning} (Vol. 32, pp. 1278--1286). PMLR.

\bibitem[{Rombach et~al.(2022)Rombach, Blattmann, Lorenz, Esser, \& Ommer}]{rombach2022high}
Rombach, R., Blattmann, A., Lorenz, D., Esser, P., \& Ommer, B. (2022). High-resolution image synthesis with latent diffusion models. In \textit{Proceedings of the IEEE/CVF conference on computer vision and pattern recognition} (pp. 10684--10695). IEEE.

\bibitem[{Rosenblatt(1958)Rosenblatt}]{Rosenblatt1958ThePA}
Rosenblatt, F. (1958). The perceptron: A probabilistic model for information storage and organization in the brain. \textit{Psychological Review, 65}(6), 386--408. \url{https://doi.org/10.1037/h0042519}

\bibitem[{Ross et~al.(2025)Ross, Kamkari, Wu, Hosseinzadeh, Liu, Stein, Cresswell, \& Loaiza-Ganem}]{ross2024geometric}
Ross, B.~L., Kamkari, H., Wu, T., Hosseinzadeh, R., Liu, Z., Stein, G., Cresswell, J.~C., \& Loaiza-Ganem, G. (2025). A geometric framework for understanding memorization in generative models. In \textit{International conference on learning representations}. OpenReview. Retrieved November 6, 2026, from \url{https://openreview.net/forum?id=aZ1gNJu8wO}

\bibitem[{Saha et~al.(2023)Saha, Krotov, Zaki, \& Ram}]{sahaCLAM}
Saha, B., Krotov, D., Zaki, M.~J., \& Ram, P. (2023). End-to-end differentiable clustering with associative memories. In \textit{Proceedings of the 40th international conference on machine learning} (Vol. 202, pp. 29649--29670). PMLR.

\bibitem[{Sahoo et~al.(2024)Sahoo, Arriola, Schiff, Gokaslan, Marroquin, Chiu, Rush, \& Kuleshov}]{sahoo2024simple}
Sahoo, S., Arriola, M., Schiff, Y., Gokaslan, A., Marroquin, E., Chiu, J., Rush, A., \& Kuleshov, V. (2024). Simple and effective masked diffusion language models. In \textit{Advances in neural information processing systems} (Vol. 37, pp. 130136--130184). Curran Associates. \url{https://doi.org/10.52202/079017-4135}

\bibitem[{Sahoo et~al.(2025)Sahoo, Deschenaux, Gokaslan, Wang, Chiu, \& Kuleshov}]{sahoo2025duality}
Sahoo, S.~S., Deschenaux, J., Gokaslan, A., Wang, G., Chiu, J.~T., \& Kuleshov, V. (2025). The diffusion duality. In \textit{Proceedings of the 42nd international conference on machine learning} (Vol. 267, pp. 52584--52619). PMLR.

\bibitem[{Salimans et~al.(2017)Salimans, Karpathy, Chen, \& Kingma}]{salimans2017pixelcnn++}
Salimans, T., Karpathy, A., Chen, X., \& Kingma, D.~P. (2017). PixelCNN++: Improving the PixelCNN with discretized logistic mixture likelihood and other modifications. In \textit{International conference on learning representations}. OpenReview. Retrieved November 6, 2026, from \url{https://openreview.net/forum?id=BJrFC6ceg}

\bibitem[{Scellier \& Bengio(2017)Scellier \& Bengio}]{scellier2017equilibrium}
Scellier, B., \& Bengio, Y. (2017). Equilibrium propagation: Bridging the gap between energy-based models and backpropagation. \textit{Frontiers in Computational Neuroscience, 11}, Article 24. \url{https://doi.org/10.3389/fncom.2017.00024}

\bibitem[{Schacter \& Addis(2007)Schacter \& Addis}]{schacter2007cognitive}
Schacter, D.~L., \& Addis, D.~R. (2007). The cognitive neuroscience of constructive memory: Remembering the past and imagining the future. \textit{Philosophical Transactions of the Royal Society B: Biological Sciences, 362}(1481), 773--786. \url{https://doi.org/10.1098/rstb.2007.2087}

\bibitem[{Schuhmann et~al.(2022)Schuhmann, Beaumont, Vencu, Gordon, Wightman, Cherti, Coombes, Katta, Mullis, Wortsman, Schramowski, Kundurthy, Crowson, Schmidt, Kaczmarczyk, \& Jitsev}]{schuhmann2022laion}
Schuhmann, C., Beaumont, R., Vencu, R., Gordon, C., Wightman, R., Cherti, M., Coombes, T., Katta, A., Mullis, C., Wortsman, M., Schramowski, P., Kundurthy, S., Crowson, K., Schmidt, L., Kaczmarczyk, R., \& Jitsev, J. (2022). LAION-5B: An open large-scale dataset for training next generation image-text models. In \textit{Advances in neural information processing systems} (Vol. 35, pp. 25278--25294). Curran Associates. \url{https://doi.org/10.52202/068431-1833}

\bibitem[{Sebah \& Gourdon(2002)Sebah \& Gourdon}]{sebah2002introductionGammaFunction}
Sebah, P., \& Gourdon, X. (2002). \textit{Introduction to the gamma function}. Numbers, constants and computation. Retrieved November 6, 2026, from \url{http://numbers.computation.free.fr/Constants/constants.html}

\bibitem[{Simonyan \& Zisserman(2015)Simonyan \& Zisserman}]{VGGsimonyan2014very}
Simonyan, K., \& Zisserman, A. (2015). \textit{Very deep convolutional networks for large-scale image recognition}. arXiv. \url{https://doi.org/10.48550/arXiv.1409.1556}

\bibitem[{Snell et~al.(2025)Snell, Lee, Xu, \& Kumar}]{snell2025scaling}
Snell, C., Lee, J., Xu, K., \& Kumar, A. (2025). \textit{Scaling LLM test-time compute optimally can be more effective than scaling model parameters}. arXiv. \url{https://doi.org/10.48550/arXiv.2408.03314}

\bibitem[{Sohl-Dickstein et~al.(2015)Sohl-Dickstein, Weiss, Maheswaranathan, \& Ganguli}]{sohl-dickstein15}
Sohl-Dickstein, J., Weiss, E., Maheswaranathan, N., \& Ganguli, S. (2015). Deep unsupervised learning using nonequilibrium thermodynamics. In \textit{Proceedings of the 32nd international conference on machine learning} (Vol. 37, pp. 2256--2265). PMLR.

\bibitem[{Somepalli et~al.(2023a)Somepalli, Singla, Goldblum, Geiping, \& Goldstein}]{somepalli2023diffusion}
Somepalli, G., Singla, V., Goldblum, M., Geiping, J., \& Goldstein, T. (2023a). Diffusion art or digital forgery? Investigating data replication in diffusion models. In \textit{Proceedings of the IEEE/CVF conference on computer vision and pattern recognition} (pp. 6048--6058). IEEE.

\bibitem[{Somepalli et~al.(2023b)Somepalli, Singla, Goldblum, Geiping, \& Goldstein}]{somepalli2023understanding}
Somepalli, G., Singla, V., Goldblum, M., Geiping, J., \& Goldstein, T. (2023b). Understanding and mitigating copying in diffusion models. In \textit{Advances in neural information processing systems} (Vol. 36, pp. 47783--47803). Curran Associates. \url{https://doi.org/10.52202/075280-2071}

\bibitem[{Song \& Ermon(2019)Song \& Ermon}]{song2019generative}
Song, Y., \& Ermon, S. (2019). Generative modeling by estimating gradients of the data distribution. In \textit{Advances in neural information processing systems} (Vol. 32). Curran Associates.

\bibitem[{Song et~al.(2019)Song, Garg, Shi, \& Ermon}]{song2019sliced}
Song, Y., Garg, S., Shi, J., \& Ermon, S. (2019). Sliced score matching: A scalable approach to density and score estimation. In \textit{Proceedings of the thirty-fifth conference on uncertainty in artificial intelligence} (Vol. 115, pp. 574--584). PMLR.

\bibitem[{Song \& Kingma(2021)Song \& Kingma}]{song2021trainebm}
Song, Y., \& Kingma, D.~P. (2021). \textit{How to train your energy-based models}. arXiv. \url{https://doi.org/10.48550/arXiv.2101.03288}

\bibitem[{Song, Meng, \& Ermon(2021)Song, Meng, \& Ermon}]{song2020denoising}
Song, J., Meng, C., \& Ermon, S. (2021). Denoising diffusion implicit models. In \textit{International conference on learning representations}. OpenReview. Retrieved November 6, 2026, from \url{https://openreview.net/forum?id=St1giarCHLP}

\bibitem[{Song, Sohl-Dickstein, et~al.(2021)Song, Sohl-Dickstein, Kingma, Kumar, Ermon, \& Poole}]{song2021scorebased}
Song, Y., Sohl-Dickstein, J., Kingma, D.~P., Kumar, A., Ermon, S., \& Poole, B. (2021). Score-based generative modeling through stochastic differential equations. In \textit{International conference on learning representations}. OpenReview. Retrieved November 6, 2026, from \url{https://openreview.net/forum?id=PxTIG12RRHS}

\bibitem[{Soudry et~al.(2018)Soudry, Hoffer, Nacson, Gunasekar, \& Srebro}]{Soudry2018}
Soudry, D., Hoffer, E., Nacson, M.~S., Gunasekar, S., \& Srebro, N. (2018). The implicit bias of gradient descent on separable data. \textit{Journal of Machine Learning Research, 19}(70), 1--57.

\bibitem[{Stanczuk et~al.(2024)Stanczuk, Batzolis, Deveney, \& Sch{\"o}nlieb}]{stanczuk2022your}
Stanczuk, J.~P., Batzolis, G., Deveney, T., \& Sch{\"o}nlieb, C.-B. (2024). Diffusion models encode the intrinsic dimension of data manifolds. In \textit{Proceedings of the 41st international conference on machine learning} (Vol. 235, pp. 46412--46440). PMLR.

\bibitem[{Tian et~al.(2024)Tian, Jiang, Yuan, Peng, \& Wang}]{tian2024visual}
Tian, K., Jiang, Y., Yuan, Z., Peng, B., \& Wang, L. (2024). Visual autoregressive modeling: Scalable image generation via next-scale prediction. In \textit{Advances in neural information processing systems} (Vol. 37, pp. 84839--84865). Curran Associates. \url{https://doi.org/10.52202/079017-2694}

\bibitem[{Tieleman(2008)Tieleman}]{tieleman2008training}
Tieleman, T. (2008). Training restricted Boltzmann machines using approximations to the likelihood gradient. In \textit{Proceedings of the 25th international conference on machine learning} (pp. 1064--1071). Association for Computing Machinery. \url{https://doi.org/10.1145/1390156.1390290}

\bibitem[{van den Burg \& Williams(2021)van den Burg \& Williams}]{burg2021on}
van den Burg, G.~J.~J., \& Williams, C.~K.~I. (2021). On memorization in probabilistic deep generative models. In \textit{Advances in neural information processing systems} (Vol. 34, pp. 27916--27928). Curran Associates.

\bibitem[{van den Oord, Kalchbrenner, et~al.(2016)van den Oord, Kalchbrenner, Espeholt, Kavukcuoglu, Vinyals, \& Graves}]{van2016conditional}
van den Oord, A., Kalchbrenner, N., Espeholt, L., Kavukcuoglu, K., Vinyals, O., \& Graves, A. (2016). Conditional image generation with PixelCNN decoders. In \textit{Advances in neural information processing systems} (Vol. 29, pp. 4790--4798). Curran Associates.

\bibitem[{van den Oord, Kalchbrenner, \& Kavukcuoglu(2016)van den Oord, Kalchbrenner, \& Kavukcuoglu}]{van2016pixel}
van den Oord, A., Kalchbrenner, N., \& Kavukcuoglu, K. (2016). Pixel recurrent neural networks. In \textit{Proceedings of the 33rd international conference on machine learning} (Vol. 48, pp. 1747--1756). PMLR.

\bibitem[{Vastola(2025)Vastola}]{vastola2025generalization}
Vastola, J.~J. (2025). Generalization through variance: How noise shapes inductive biases in diffusion models. In \textit{International conference on learning representations}. OpenReview. Retrieved November 6, 2026, from \url{https://openreview.net/forum?id=7lUdo8Vuqa}

\bibitem[{Vaswani et~al.(2017)Vaswani, Shazeer, Parmar, Uszkoreit, Jones, Gomez, Kaiser, \& Polosukhin}]{Attention2017}
Vaswani, A., Shazeer, N., Parmar, N., Uszkoreit, J., Jones, L., Gomez, A.~N., Kaiser, {\L}., \& Polosukhin, I. (2017). Attention is all you need. In \textit{Advances in neural information processing systems} (Vol. 30, pp. 5998--6008). Curran Associates.

\bibitem[{Ventura et~al.(2025)Ventura, Achilli, Silvestri, Lucibello, \& Ambrogioni}]{ventura2024manifolds}
Ventura, E., Achilli, B., Silvestri, G., Lucibello, C., \& Ambrogioni, L. (2025). Manifolds, random matrices and spectral gaps: The geometric phases of generative diffusion. In \textit{International conference on learning representations}. OpenReview. Retrieved November 6, 2026, from \url{https://openreview.net/forum?id=KlN00vQEY2}

\bibitem[{Vincent(2011)Vincent}]{VincentP2011}
Vincent, P. (2011). A connection between score matching and denoising autoencoders. \textit{Neural Computation, 23}(7), 1661--1674. \url{https://doi.org/10.1162/NECO_a_00142}

\bibitem[{Virtanen et~al.(2020)Virtanen, Gommers, Oliphant, Haberland, Reddy, Cournapeau, Burovski, Peterson, Weckesser, Bright, van der Walt, Brett, Wilson, Millman, Mayorov, Nelson, Jones, Kern, Larson, Carey, Polat, Feng, Moore, VanderPlas, Laxalde, Perktold, Cimrman, Henriksen, Quintero, Harris, Archibald, Ribeiro, Pedregosa, van Mulbregt, \& SciPy 1.0 Contributors}]{SciPy}
Virtanen, P., Gommers, R., Oliphant, T.~E., Haberland, M., Reddy, T., Cournapeau, D., Burovski, E., Peterson, P., Weckesser, W., Bright, J., van der Walt, S.~J., Brett, M., Wilson, J., Millman, K.~J., Mayorov, N., Nelson, A.~R.~J., Jones, E., Kern, R., Larson, E., \ldots{} SciPy 1.0 Contributors (2020). SciPy 1.0: Fundamental algorithms for scientific computing in Python. \textit{Nature Methods, 17}(3), 261--272. \url{https://doi.org/10.1038/s41592-019-0686-2}

\bibitem[{von Platen et~al.(2022)von Platen, Patil, Lozhkov, Cuenca, Lambert, Rasul, Davaadorj, Nair, Paul, Berman, Xu, Liu, \& Wolf}]{von-platen-etal-2022-diffusers}
von Platen, P., Patil, S., Lozhkov, A., Cuenca, P., Lambert, N., Rasul, K., Davaadorj, M., Nair, D., Paul, S., Berman, W., Xu, Y., Liu, S., \& Wolf, T. (2022). \textit{Diffusers: State-of-the-art diffusion models} [Computer software]. GitHub. Retrieved November 6, 2026, from \url{https://github.com/huggingface/diffusers}

\bibitem[{Wang et~al.(2025)Wang, Li, Sun, Chen, Liu, Wu, Lu, Song, \& Abbasi Yadkori}]{wang2025hierarchical}
Wang, G., Li, J., Sun, Y., Chen, X., Liu, C., Wu, Y., Lu, M., Song, S., \& Abbasi Yadkori, Y. (2025). \textit{Hierarchical reasoning model}. arXiv. \url{https://doi.org/10.48550/arXiv.2506.21734}

\bibitem[{Webster(2023)Webster}]{webster2023reproducible}
Webster, R. (2023). \textit{A reproducible extraction of training images from diffusion models}. arXiv. \url{https://doi.org/10.48550/arXiv.2305.08694}

\bibitem[{Webster et~al.(2023)Webster, Rabin, Simon, \& Jurie}]{webster2023duplication}
Webster, R., Rabin, J., Simon, L., \& Jurie, F. (2023). \textit{On the de-duplication of LAION-2B}. arXiv. \url{https://doi.org/10.48550/arXiv.2303.12733}

\bibitem[{Welling \& Teh(2011)Welling \& Teh}]{WellingLangevin}
Welling, M., \& Teh, Y.~W. (2011). Bayesian learning via stochastic gradient Langevin dynamics. In \textit{Proceedings of the 28th international conference on machine learning} (pp. 681--688). Omnipress.

\bibitem[{Wen et~al.(2024)Wen, Liu, Chen, \& Lyu}]{wen2024detecting}
Wen, Y., Liu, Y., Chen, C., \& Lyu, L. (2024). Detecting, explaining, and mitigating memorization in diffusion models. In \textit{International conference on learning representations}. OpenReview. Retrieved November 6, 2026, from \url{https://openreview.net/forum?id=84n3UwkH7b}

\bibitem[{Weng(2018)Weng}]{weng2018flow}
Weng, L. (2018, October 13). \textit{Flow-based deep generative models}. Lil'Log. Retrieved November 6, 2026, from \url{https://lilianweng.github.io/posts/2018-10-13-flow-models/}

\bibitem[{Xiao et~al.(2017)Xiao, Rasul, \& Vollgraf}]{fmnist}
Xiao, H., Rasul, K., \& Vollgraf, R. (2017). \textit{Fashion-MNIST: A novel image dataset for benchmarking machine learning algorithms}. arXiv. \url{https://doi.org/10.48550/arXiv.1708.07747}

\bibitem[{Yoon et~al.(2023)Yoon, Choi, Kwon, \& Ryu}]{yoon2023diffusion}
Yoon, T., Choi, J.~Y., Kwon, S., \& Ryu, E.~K. (2023, July). \textit{Diffusion probabilistic models generalize when they fail to memorize} [Paper presentation]. ICML 2023 Workshop on Structured Probabilistic Inference \& Generative Modeling, Honolulu, HI, United States.

\bibitem[{Yu et~al.(2015)Yu, Seff, Zhang, Song, Funkhouser, \& Xiao}]{lsun}
Yu, F., Seff, A., Zhang, Y., Song, S., Funkhouser, T., \& Xiao, J. (2015). \textit{LSUN: Construction of a large-scale image dataset using deep learning with humans in the loop}. arXiv. \url{https://doi.org/10.48550/arXiv.1506.03365}

\bibitem[{Yu et~al.(2022)Yu, Xu, Koh, Luong, Baid, Wang, Vasudevan, Ku, Yang, Ayan, Hutchinson, Han, Parekh, Li, Zhang, Baldridge, \& Wu}]{yu2022scaling}
Yu, J., Xu, Y., Koh, J.~Y., Luong, T., Baid, G., Wang, Z., Vasudevan, V., Ku, A., Yang, Y., Ayan, B.~K., Hutchinson, B., Han, W., Parekh, Z., Li, X., Zhang, H., Baldridge, J., \& Wu, Y. (2022). Scaling autoregressive models for content-rich text-to-image generation. \textit{Transactions on Machine Learning Research}. Retrieved November 6, 2026, from \url{https://openreview.net/forum?id=AFDcYJKhND}

\bibitem[{Yuille \& Rangarajan(2001)Yuille \& Rangarajan}]{CCP-Yuilee}
Yuille, A.~L., \& Rangarajan, A. (2001). The concave-convex procedure (CCCP). In \textit{Advances in neural information processing systems} (Vol. 14, pp. 1033--1040). MIT Press.

\bibitem[{Zhang et~al.(2017)Zhang, Bengio, Hardt, Recht, \& Vinyals}]{zhang2017understanding}
Zhang, C., Bengio, S., Hardt, M., Recht, B., \& Vinyals, O. (2017). Understanding deep learning requires rethinking generalization. In \textit{International conference on learning representations}. OpenReview. Retrieved November 6, 2026, from \url{https://openreview.net/forum?id=Sy8gdB9xx}

\bibitem[{Zhang et~al.(2018)Zhang, Isola, Efros, Shechtman, \& Wang}]{lpips}
Zhang, R., Isola, P., Efros, A.~A., Shechtman, E., \& Wang, O. (2018). The unreasonable effectiveness of deep features as a perceptual metric. In \textit{Proceedings of the IEEE conference on computer vision and pattern recognition} (pp. 586--595). IEEE.

\end{thebibliography}
